\documentclass{article}

\usepackage{scalefnt,letltxmacro}
\LetLtxMacro{\oldtextsc}{\textsc}
\renewcommand{\textsc}[1]{\oldtextsc{\scalefont{1.10}#1}}
\usepackage[acronym,nowarn]{glossaries}
\setacronymstyle{long-sc-short}

\glsdisablehyper
\makeglossaries
\usepackage{xspace}
\usepackage{float}
\usepackage{physics}
\usepackage[normalem]{ulem}

\usepackage{enumitem}

\usepackage{amssymb}
\usepackage{mathtools}
\usepackage{amsfonts}
\usepackage{amsmath}
\usepackage{amsthm,thmtools}
\usepackage{booktabs}
\usepackage[]{microtype}

\usepackage{hyperref}

\usepackage[capitalize,nameinlink,]{cleveref} 
\crefname{section}{\S}{\S\S}
\Crefname{section}{\S}{\S\S}

\usepackage[linesnumbered,ruled,vlined]{algorithm2e}

\SetCommentSty{mycommfont}
\SetKwInput{KwInput}{Input}
\SetKwInput{KwOutput}{Output}

\makeatletter
\@ifundefined{c@rownum}{%
	\let\c@rownum\rownum
}{}
\@ifundefined{therownum}{%
	\def\therownum{\@arabic\rownum}%
}{}
\makeatother

\makeatletter
\newcommand*{\addFileDependency}[1]{
	\typeout{(#1)}
	\@addtofilelist{#1}
	\IfFileExists{#1}{}{\typeout{No file #1.}}
}
\makeatother

\usepackage[font=footnotesize,labelfont=bf,tableposition=top]{caption}
\usepackage{tikz}
\usepackage{graphicx}
\usepackage[]{subcaption}   

\usepackage{pgfplots}
\pgfplotsset{compat=1.6}
\usepgfplotslibrary{groupplots}
\tikzstyle{every picture}+=[font=\sffamily]
\tikzstyle{optimized} = [circle,fill=white,draw=black, dashed,inner sep=1pt, minimum size=20pt, font=\fontsize{10}{10}\selectfont, node distance=1]
\pgfkeys{/pgf/number format/.cd,1000 sep={}}
\pgfplotsset{
	tick label style = {font=\sffamily},
	every axis label/.append style={font=\sffamily},
	typeset ticklabels with strut,
}
\pgfplotsset{every axis/.append style={
			every x tick label/.append style={font=\fontsize{6pt}{6pt}\sffamily, yshift=.5ex,},
			every y tick label/.append style={font=\fontsize{6pt}{6pt}\sffamily, xshift=.5ex},
			every y label/.append style={xshift=10ex, font=\sffamily},
			every x label/.append style={yshift=3ex, font=\sffamily},
			every title/.append style={font=\sffamily}
		},
}
\pgfplotsset{
	xticklabel={$\mathsf{\pgfmathprintnumber{\tick}}$},
	yticklabel={$\mathsf{\pgfmathprintnumber{\tick}}$},
}
\pgfplotsset{every axis title/.append style={yshift=-1ex}}
\newlength\figureheight
\newlength\figurewidth
\usepgfplotslibrary{external}
\tikzexternaldisable

\usepackage{url}

\usepackage{subcaption}

\usepackage{tikz}
\usepackage{graphicx}
\usepackage{xcolor}
\usepackage{amsmath}
\usepackage{amssymb}
\usepackage{bm}
\usepackage{amsthm} 
\usepackage{nicefrac}

\declaretheorem[name=Proposition]{proposition}

\usepackage{adjustbox}
\usepackage{multirow}
\usepackage{mathtools}

\usepackage{xspace}

\usepackage{balance}

\newacronym{MAP}{map}{maximum-a-posteriori}
\newacronym{MLE}{mle}{maximum likelihood estimation}
\newacronym{MNLL}{mnll}{mean negative loglikelihood}
\newacronym{NLL}{nll}{negative loglikelihood}
\newacronym{LL}{ll}{log-likelihood}
\newacronym{RMSE}{rmse}{root mean square error}
\newacronym{ECE}{ece}{expected calibration error}
\newacronym{SNR}{snr}{signal-to-noise ratio}
\newacronym{FID}{fid}{Fr\'echet Inception Distance}
\newacronym{FAD}{fad}{Fr\'echet Audio Distance}
\newacronym{FMD}{fmd}{Fr\'echet Modality Distance}
\newacronym{BPD}{bpd}{bit per dimension}
\newacronym{NFE}{nfe}{neural function evaluations}

\newacronym{AE}{ae}{autoencoder}
\newacronym{WAE}{wae}{Wasserstein Autoencoder}
\newacronym{VAE}{vae}{Variational Autoencoder}
\newacronym{BAE}{bae}{Bayesian autoencoder}
\newacronym{CDF}{cdf}{cumulative density function}
\newacronym{GAN}{gan}{Generative Adversarial Network}
\newacronym{DPGMM}{dpgmm}{Dirichlet process Gaussian mixture model}

\newacronym{MC}{mc}{Monte Carlo}
\newacronym{SDE}{sde}{Stochastic Differential Equation}
\newacronym{CNF}{cnf}{Continuous Normaxlizing Flow}
\newacronym{ODE}{ode}{Ordinary Differential Equation}
\newacronym{MCMC}{mcmc}{Markov chain Monte Carlo}
\newacronym{HMC}{hmc}{Hamiltonian Monte Carlo}
\newacronym{MH}{mh}{Metropolis-Hastings}
\newacronym{NUTS}{nuts}{no-u-turn sampler}
\newacronym{SGHMC}{sghmc}{stochastic gradient Hamiltonian Monte Carlo}

\newacronym{DGP}{dgp}{deep Gaussian process} 
\newacronym{GPLVM}{gplvm}{Gaussian process latent variable model}
\newacronym{DPMM}{dpmm}{Dirichlet Process Mixture Model}

\newacronym{VFE}{vfe}{variational free energy}

\newacronym[firstplural=Gaussian Processes]{GP}{gp}{Gaussian Process}

\newacronym{VI}{vi}{variational inference}

\newacronym{PDE}{pde}{Partial Differential Equation}
\newacronym{ELBO}{elbo}{evidence lower bound}
\newacronym{NELBO}{nelbo}{negative evidence lower bound}
\newacronym{ELL}{ell}{expected log likelihood}
\newacronym{KL}{kl}{Kullback-Leibler}
\newacronym{AUC}{auc}{area under the curve}

\newacronym[firstplural=Bayesian neural networks]{BNN}{bnn}{Bayesian neural network}
\newacronym[firstplural=deep neural networks]{DNN}{dnn}{deep neural network}
\newacronym[]{CNN}{cnn}{convolutional neural network}
\newacronym{MLP}{mlp}{multilayer perceptron}
\newacronym{NN}{nn}{neural network}
\newacronym{RELU}{ReLU}{rectified linear unit}

\newacronym{NF}{nf}{normalizing flow}

\newacronym{RBF}{rbf}{radial basis function}
\newacronym{ARD}{ard}{automatic relevance determination}

\newacronym{RKHS}{rkhs}{reproducing kernel Hilbert space}

\newacronym{OT}{ot}{optimal transport}
\newacronym{WD}{wd}{Wasserstein distance}
\newacronym{SWD}{swd}{sliced-Wasserstein distance}
\newacronym{DSWD}{dswd}{distributional sliced-Wasserstein distance}

\newacronym{MLD}{mld}{Multi-modal Latent Diffusion}
\newacronym{MLD Inpaint}{mld in-paint}{Multi-modal Latent Diffusion with In-painting}
\newacronym{MLD Uni}{mld uni}{Multi-modal Latent Diffusion UniDiffuser }
\newacronym{MOPOE}{mopoe}{Mixture of Product of Experts}
\newacronym{MVAE}{mvae}{Product of Experts}
\newacronym{MMVAE}{mmvae}{Mixture of Expert}
\newacronym{NEXUS}{nexus}{Hierarchical Genertive Model}

\newacronym{MMVAEplus}{MMVAE+}{}

\newacronym{MVTCAE}{mvtcae}{Multi-view Total Correlation Autoencoder}
\newacronym{CLIP-S}{clip-s}{CLIP-Score}
\newacronym{MHD}{mhd}{The Multimodal Handwritten Digits data-set}
\newacronym{VPSDE}{vpsde}{Variance preserving SDE}
\newacronym{EMA}{ema}{Exponential moving average}
\newacronym{MSE}{mse}{Mean square error}

\newacronym{PID}{pid}{Partial Information Decomposition }
\newacronym{MI}{mi}{Mutual Information }
\newacronym{PI}{pi}{Partial Information}
\newacronym{O-information}{o-information}{O-information }

\newacronym{SOI}{s$\Omega$i}{ Score-based O-Information estimation }

\newacronym{S-information}{s-information}{S-information }
\newacronym{PED}{ped}{Partial Entropy Decomposition}

\newacronym{TC}{tc}{Total Correlation}

\newacronym{DTC}{dtc}{Dual Total Correlation}

\newacronym{MINE}{mine}{MINE}
\newacronym{MINDE}{minde}{MINDE}
\newacronym{NWJ}{nwj}{NWJ}
\newacronym{InfoNCE}{infonce}{InfoNCE}
\newacronym{CLUB}{club}{CLUB}

\newcommand{\g}{\,|\,}

\DeclareRobustCommand{\KL}[2]{\ensuremath{\textsc{kl}\left[#1\;\|\;#2\right]}}
\DeclarePairedDelimiterX{\infdivx}[2]{[}{]}{%
#1\;\delimsize\|\;#2%
}

\newcommand\indep{\protect\mathpalette{\protect\independenT}{\perp}}
\def\independenT#1#2{\mathrel{\rlap{$#1#2$}\mkern2mu{#1#2}}}

\newcommand{\cD}{\mathcal{D}}

\newcommand{\cS}{\mathcal{S}}

\newcommand{\cI}{\mathcal{I}}
\newcommand{\cT}{\mathcal{T}}

\newcommand{\cH}{\mathcal{H}}

\newcommand{\E}{\mathbb{E}}

\newcommand{\defeq}{\stackrel{\text{\tiny def}}{=}}

\DeclareMathAlphabet{\pazocal}{OMS}{zplm}{m}{n}

\usepackage[accepted]{icml2024}

\begin{document}

\twocolumn[

\icmltitle{S$\Omega$I: Score-based O-INFORMATION Estimation}



\icmlsetsymbol{equal}{*}

\begin{icmlauthorlist}
\icmlauthor{Mustapha Bounoua}{comp,un}
\icmlauthor{Giulio Franzese}{un}
\icmlauthor{Pietro Michiardi}{un}
\end{icmlauthorlist}

\icmlaffiliation{comp}{Ampere Software Technology, France}
\icmlaffiliation{un}{Department of Data Science, Eurecom, France}

\icmlcorrespondingauthor{}{mustapha.bounoua@eurecom.fr}

\icmlkeywords{Machine Learning, Mutual information, Score based models, Diffusion models}

\vskip 0.3in
]



 \printAffiliationsAndNotice{}  
\begin{abstract}

The analysis of scientific data and complex multivariate systems requires information quantities that capture relationships among multiple random variables. Recently, new information-theoretic measures have been developed to overcome the shortcomings of classical ones, such as mutual information, that are restricted to considering pairwise interactions. Among them, the concept of information synergy and redundancy is crucial for understanding the high-order dependencies between variables. One of the most prominent and versatile measures based on this concept is \acrshort{O-information}, which provides a clear and scalable way to quantify the synergy-redundancy balance in multivariate systems. However, its practical application is limited to simplified cases. In this work, we introduce \acrshort{SOI}, which allows to compute \acrshort{O-information} without restrictive assumptions   about the system while leveraging a unique model. Our experiments validate our approach on synthetic data, and demonstrate the effectiveness of \acrshort{SOI} in the context of a real-world use case.
\end{abstract}

\section{Introduction}

\gls{MI} is a fundamental measure which allows investigation of the non-linear dependence between random variables \citep{shannon1948mathematical, mackay2003information}. Despite its success in various domains, classical \gls{MI} suffers from limitations when analyzing systems composed by more than two variables. This constitutes an important limitation, considering that many scientific endeavors aim at an accurate statistical characterization of systems which are composed of many random variables. Examples includes neuroscience \citep{latham2005synergy,ganmor2011sparse,gat1998synergy}, climate models \citep{runge2019inferring}, econometrics \citep{dosi2019more}, and machine learning \cite{e19090474}, to name a few.

A recent attempt to overcome such limitations, and to extend the applicability of information-theoretic tools to multivariate systems, is represented by \gls{PID} \cite{williams2010nonnegative}. The key idea behind such method is the \textit{decomposition} of the overall \gls{MI} between a set of source variables and a given target variable into non-negative constituents. In particular, \gls{PID} quantifies how much of the total information about the target variable is encoded redundantly, synergistically or uniquely into given subsets of variables. 
\textit{Redundancy} quantifies information that is shared between subsets of the partition, \textit{synergy} describes the additional information that is endowed to all subsets observed jointly but that is not available from individual constituents of the partition, and \textit{uniqueness} quantifies the information that is lost when a given subset is not observed, removing the amount of redundant and synergistic information associated to that subset. 
The \gls{PID} method requires partitioning the source system into all its possible subsets and computing the information decomposition of all constituents with respect to the target variable. 

Despite its elegance, this measure is not without drawbacks. Indeed, there is no consensus on the best way to define and compute \gls{PID}, and several variants have emerged, including~\citep{Adam14}, who reformulate synergy and redundancy for Gaussian systems (but that has been judged as poorly motivated by~\citep{venkatesh2023gaussian}),~\cite{Finn2019GeneralisedMO}, who use the algebraic structure of information sharing,~\cite{Ay2019InformationDB}, who rely on cooperative game theory, \cite{Rosas2020AnOI}, who build on concepts related to data privacy and disclosure, \cite{Kolchinsky19}, who use set theory, \cite{Enk2023PoolingPD}, who deal with scalability issues by pooling probabilities, \cite{gutknecht2023babel}, who use a mereological formulation, and \citep{makkeh2021introducing,ehrlich2023partial}, who advocate for methods based on the exclusions of probability mass. 
Nevertheless, the main limitations of \gls{PID} persist in all variants. Indeed, computational complexity grows extremely fast, precisely as the Dedekind number of variables (which is more than $10^{31}$ for 9 variables). Moreover, \gls{PID} computation relies on a partition of the system into a set of sources and a unique target. This can be an artificial distinction which limits usability and interpretability of the results. This latter problem is partially addressed in~\cite{Varley2023PartialED}, who introduce \gls{PED}.

Motivated by these limitations,~\cite{Rosas2019QuantifyingHI} introduce the concept of \acrshort{O-information}, a measure which captures the synergy-redundancy dominance in multivariate systems. In contrast to \gls{PID}, this measure does not require the system to be partitioned into sources and a target, and gracefully scales in the number of its  components~\cite{martinez2022integrated}. Furthermore, recent extensions such as \acrshort{O-information} locality \cite{Scagliarini2021QuantifyingHI} and gradient computation \cite{Scagliarini23} allow a fine-grained analysis of system behavior. 
However, \acrshort{O-information} measures are accessible only in restricted scenarios. Indeed, existing methods rely on estimation techniques that requires either i) discrete distributions (or binning of continuous ones) or ii) Gaussian distributions. In this work, we show that such limitations can be lifted by using and extending recent methods to estimate \gls{MI}~\cite{franzese2023minde, kong2023interpretable}. 

Our work is organized as follows: \Cref{prem} introduces the high-dimensional interaction measures which we investigate in this work, while \Cref{method} proposes \gls{SOI}, our novel methodology which allows scalable and flexible \acrshort{O-information} estimation. \Cref{experiment} validates experimentally our proposed method, where we report a series of compelling results on various synthetic systems, for which ground truth values are known and accessible analytically. 
Furthermore, we consider a realistic endeavor by revisiting previous studies~\cite{venkatesh2023gaussian} that focus on the analysis of brain activity in mice. Our method allows lifting previous limiting assumptions, and allow synergy-redundancy characterizations that are compatible with observations made by domain experts.
Finally, we summarize our findings in \Cref{conclusions}.

\section{High dimensional interaction measures}
\label{prem}
Consider the continuous \textbf{multivariate} random variable $X=\{X^{1},\dots,X^{N}\} \sim p(x^1,\dots,x^N)$. We indicate the collection of all but the $i_{th}$ random variable with the symbol $X^{\setminus i}\defeq \{X^1,..,X^{i-1},X^{i+1},..,X^N \}$. When necessary, we indicate marginal and conditional distributions by properly specifying the arguments of the distribution, e.g. $X^i\sim p(x^i)$ or $X^{\setminus i}\g {X^i}\sim p(x^1,\dots,x^{i-1},x^{i},\dots, x^M \g x^i)$.

A central quantity in this work is the Shannon entropy associated to a given random variable $\cH(X)\defeq \mathbb{E} \left[  - \log p (X)  \right]$~\cite{cover1999elements}. 
Considering the case of bi-variate (i.e. $N=2$) random variable $X$, entropy and conditional entropy allow computation of the mutual information (\gls{MI}) flow $\mathcal{I}$ between the two random variables $X^1,X^2$: 
$\mathcal{I}(X^1; X^2) = \cH(X^1) - \cH(X^1|X^2)$, where $\cH(X^1|X^2)= \mathbb{E} \left[  - \log p (X^1\g X^2)  \right]$. Importantly, such quantity can also be expressed as the \gls{KL} divergence \citep{cover1999elements} between the joint and the product of marginal distributions: $\mathcal{I}(X^1; X^2)=\KL{p(x^1,x^2)}{p(x^1)p(x^2)}$. For the case of $N=3$, it is possible to define the \gls{MI} as $\mathcal{I}(X^1;X^2;X^3) = \mathcal{I}(X^1; X^2) - \mathcal{I}(X^1;X^2|X^3)$, where $\mathcal{I}(X^1;X^2|X^3)=\cH(X^1\g X^3)-\cH({X^1 \g X^2,X^3})$. This quantity, also known as co-information or interaction information, can counter-intuitively result in a negative value, and measures the difference between synergistic and redundant interactions \cite{Rosas2019QuantifyingHI}. 

Since, for $N>3$, interaction information becomes difficult to grasp~\cite{williams2010nonnegative, Rosas2019QuantifyingHI}, our goal in this work is to consider extensions to \gls{MI}, while preserving interpretability. In particular, a measure of the interaction strengths in a system with $N>3$ can be obtained by studying the summand mutual information between one variable and the rest of the system:
\begin{equation}\label{eq:def_s}
    \cS(X) \defeq \sum\limits_{i=1}^N \cI(X^i;X^{\setminus i}).
\end{equation}

This quantity, named \acrshort{S-information}, can be decomposed into the redundant and synergistic components of the considered multivariate system. 
In particular, since $X^{\setminus i}=\{X^{<i},X^{>i}\}$, where $X^{<i}=\{X^1,\dots,X^{i-1}\}$ and $X^{>i}=\{X^{i+1},\dots,X^{N}\}$ (with $X^{>N}=\emptyset$), we can use the conditional mutual information laws \cite{cover1999elements} and rewrite $\cS(X)$ as:
\begin{equation}
\cS(X)=\sum\limits_{i=1}^N \cI(X^i;X^{>i})+\sum\limits_{i=1}^N \cI(X^i;X^{<i}\g X^{>i}).
\end{equation}
The two \textbf{positive} series which constitute $\cS(X)$ are equivalent to the \gls{TC} \cite{sun1975linear} and the \gls{DTC} \cite{SUNHAN198026} denoted by $\cT(.)$ and $\cD(.)$ respectively.
Then, $\cS(X)=\cT(X)+\cD(X)$, where (proof in \Cref{proof1})
\begin{align}
& \cT(X) = \sum_{i=1}^{N} \cH(X^i)- \cH(X),\label{eq:tc} \\
& \cD( X) = \cH(X) - \sum_{i=1}^{N} \cH(X^i|X^{\setminus i}). \label{eq:dtc}
\end{align}
\gls{TC} is high in cases where, for each variable $X^i$, at least one of its ``\textit{children}'' (variables in $ X^{>i}$) carries information about it. Importantly, the number of children conveying information (whether 1, 2, or $ N-1 $) is irrelevant. Since $\cT(X)$ is permutation invariant, a high value implies that for every ordering of the variables, and hence for all possible combinations of children of a given variable, the summand mutual information between variables and their children remains high. This intuition, which suggests \textit{redundancy}, can similarly be obtained by considering the entropic formulation. Indeed, whenever a system is composed of perfectly independent variables ($X^i\perp X^j,i\neq j$) $\cH(X)=\sum_{i=1}^{N} \cH(X^i)$ and consequently $\cT(X)=0$. On the other hand, a \textit{copy} system ($X^i=X^j,\forall i,j$) achieves infinite $\cT(X)$, as $\cH(X)=-\infty$, since the support of the joint distribution is on a lower than $N-$dimensional space. \gls{TC} also admits a representation in terms of \gls{KL} divergences, $\cT(X) = \KL{p(x)}{\prod\limits_{i=1}^N p(x^i)}$, which we will exploit later in our proposed methodology.   

Similar considerations can be carried out for the \gls{DTC}. Consider a single \gls{MI} term $ I(X^i; X^{<i} | X^{>i}) $. The focus of this conditioning is about quantifying how much \textbf{additional} information the variables $X^{<i}$ carry about $X^i$ if we are also given access $ X^{>i}$. Whenever the variables are independent or redundant (the copy system), this value is identically zero. However, whenever the aid of the \textit{extra} measurements unlocks new bits of information, which suggests a \textit{synergistic} scenario, its value is positive. 

Having recognized that $\cS(X)$ in a multivariate system can be decomposed into measures of redundancy $\cT(X)$ and synergy $\cD(X)$, 
we can introduce a new information theoretic measure which quantifies the difference between the two behaviours. This quantity, named \acrshort{O-information} 
\citep{Rosas2019QuantifyingHI}, is defined as 
\begin{equation}\label{eq:oinfo}
  \Omega(X)=\cT(X)-\cD(X).
\end{equation}

In summary, while \acrshort{S-information} only quantifies the strength of interactions in a system, \acrshort{O-information} also determines the \textit{nature} of these interactions, being them redundant or synergistic. Intuitively, a redundancy-dominated system is the most parsimonious explanation --- in an Occam's razor sense --- whenever $\Omega(X)>0$. Conversely, a negative value $\Omega(X)<0$ is associated with a synergy-dominated system. \acrshort{O-information} is a natural generalization of \gls{MI} for more than 3 variables: indeed, it is equal to the co-information for $N=3$, and is a measure which preserves interpretability for any positive $N$.

One important property of \acrshort{O-information} is that it gracefully scales with the number of random variables composing a system, as opposed to, e.g. the \gls{PID} measure, which has much worse scalability. 
 
Since \acrshort{O-information} measures the \textit{overall} information dynamics among variables, recent work focus on ways to study the \textit{individual} influence of variables to the high-order interactions, and capture the interaction structure of a multivariate system~\cite{Scagliarini23}. 
The first order difference, called the \textit{gradient} of \acrshort{O-information}, captures how much \acrshort{O-information} changes when adding or removing a given system variable $i$:
\begin{equation}
    \partial_i \Omega( X) =  \Omega( X) - \Omega( X^{\setminus i}).
\end{equation}
A positive value implies that $X^i$ provides redundant information to the system, while a negative one suggests that its interaction with other variables is mainly synergistic.

\section{Score-based \acrshort{O-information} estimation}
\label{method}
\acrshort{O-information} and its gradient represent extremely useful information theoretic measures to study multivariate systems. However, as it is clear from \Cref{eq:tc,eq:dtc,eq:oinfo}, their estimation requires access to entropies, conditional entropies and \gls{KL} divergence measures. 
When strict assumptions about the distribution of variables composing the system are possible, such as discrete or Gaussian distributions, existing implementations of \acrshort{O-information} estimators have been used successfully in a number of application domains~\citep{Varley2022MultivariateIT,Sparacino2023StatisticalAT,Stramaglia_Sebastiano,Chiarion2023ConnectivityAI}. 
However, in more realistic cases where such assumptions are not valid, there currently does not exist a method to estimate the constituents of \acrshort{O-information} in a reliable and scalable manner. In this work, we present the first methodology allowing estimation of \acrshort{O-information} for more general scenarios. Our method unfolds according to the observation that  all quantities of interest can be expressed in terms of \gls{KL} divergences, and relies on a technique to estimate such divergences which scales gracefully with the system size. Our key ingredient is the score function associated to data distributions \cite{vincent2011,song2019} and the method we present leverages recent advances in the field of \gls{MI} estimation~\citep{franzese2023minde,kong2023interpretable}.

\subsection{Score-based divergence estimation}\label{sec:sde_est}

Consider the generic multivariate random variable $X$ with associated distribution $p(x)$. Provided that certain minimal regularity assumptions are met \cite{vincent2011}, it is always possible to associate the distribution $p(x)$ to its \textit{score function}, defined as the gradient of its logarithm, $\nabla\log p(x)$. 

Recently, the community has showed tremendous interest \cite{song2019,song2021a} in a generalization of such concept, which involves computing the score function of a \textit{noised} version of the variable $X$, due to the possibility of adopting such concept for generative modelling purposes. Accordingly, in this work we define a noised version of the variable $X$ with corresponding intensity indexed by $t\in[0,\infty)$. Then, the new variable is constructed as $X_t=X+\sqrt{2t} W$, where $W$ is a Gaussian random vector with the same dimension of $X$, zero mean and identity covariance matrix. 

This new random variable can be associated to its \textit{time-varying} score function $\nabla\log p_{t}(x)$. In particular the analytic expression of $p_{t}(x)$ can be obtained as the solution of the \gls{PDE} $\frac{\dd p_{t}(x)}{\dd t}=\Delta p_{t}(x)$, with initial conditions given by $p_{0}(x)=p(x)$.

Next, we consider the \gls{KL} divergence between two generic distributions and define how it can be computed using score functions, a result which we will use later for computing \acrshort{O-information}. 
\begin{proposition}\label{prop:kl_est}
    ~\cite{franzese2023minde,kong2023interpretable} The \gls{KL} divergence between two generic distributions $p(x)$ and $q(x)$, defined as
    \begin{equation*}
        \KL{p(x)}{q(x)}=\int p(x)\log\frac{p(x)}{q(x)}\dd x, 
    \end{equation*}
    can be computed considering the time-varying score functions $\nabla \log(p_t)$ and $\nabla \log(q_t)$, according to the following expression:
    \begin{equation*}\label{eq:klest}
        \KL{p(x)}{q(x)}=\int p_{t}(x)\norm{\nabla \log(\frac{p_{t}(x)}{q_{t}(x)})}^2\dd x\dd t.
    \end{equation*}
\end{proposition}

\noindent \textbf{Proof sketch.} To avoid clutter, we drop the dependence on $x$ of the distributions. Let's define $r_t\defeq \int p_{t}\log\frac{p_{t}}{q_t}\dd x$.

Since it holds that $r_\infty-\KL{p}{q}=\int_{0}^\infty \frac{\dd r_t}{\dd t} \dd t$, we need  
$$\int \frac{\dd r_t}{\dd t}\dd t=\int \frac{\dd p_{t}}{\dd t}\log(\frac{p_{t}}{q_t})+p_{t}\frac{\dd}{\dd t}\log(\frac{p_{t}}{q_t})\dd x\dd t.$$

Note that  $\int p_{t}\frac{\dd}{\dd t}\log(\frac{p_{t}}{q_t})\dd x\dd t=\int \frac{\dd}{\dd t} p_{t}-\frac{p_{t}}{q_t}\Delta q_{t} \dd x\dd t$, and $\int \frac{\dd}{\dd t} p_{t} \dd x\dd t=0$ ( See \cref{proof_prop1} for detailed proof). Then, the expression above can be rewritten as $\int p_{t}\Delta \log(\frac{p_{t}}{q_t})-\frac{p_{t}}{q_t}\Delta q_t \dd x\dd t$. Integrating by parts we obtain $\int -\nabla p_{t}\nabla \log(\frac{p_{t}}{q_t})+\nabla(\frac{p_{t}}{q_t})\nabla q_t  \dd x\dd t $. Since $\nabla p_{t}=p_{t}\nabla \log p_{t}$ and $\nabla(\frac{p_{t}}{q_t})\nabla q_t=p_{t}\nabla \log q_t\nabla \log(\frac{p_{t}}{q_t})$, and $r_{\infty}=0$~\cite{franzese2022,villani2009optimal,collet2008logarithmic}, the proposition follows. $\qedsymbol$

The result in \Cref{prop:kl_est} allows, in principle, the exact computation of \gls{KL} divergences, provided knowledge of the score functions $\nabla \log{p_{t}},\nabla \log{q_{t}}$. Such knowledge is however out of reach in practical cases, which is why in this work we consider a \textit{parametric} approximation of such vector fields, leading to a \gls{KL} divergence \textit{estimator}. In particular, we leverage the methodology considered in \cite{song2019,song2021a} where the parametric score $s_t$ is obtained by minimizing the so called \textit{denoising score-matching} loss 
\begin{equation*}\label{eq:loss}
    \int p(x)p_{0t}(\tilde{x}\g x)\norm{s_t(\tilde{x})-\nabla\log(p_{0t}(\tilde{x}\g x))}^2\dd x\dd \tilde{x} \dd t,
\end{equation*}

where $p_{0t}(\tilde{x}\g x)$ is the conditional distribution of the noised random variable \textit{given} initial conditions $X=x$, i.e. $p_t(\tilde{x})=\int p_{0t}(\tilde{x}\g x)p(x)\dd x$. Note that $p_{0t}$ has known Gaussian distribution with mean $x$ and variance $2t$. This allows, together with the knowledge of the score functions, the implementation of an estimator for the \gls{KL} divergence.

Informally, learning the score can be understood as learning to \textit{denoise} the variable $X_t$ to obtain $X$. Indeed, the score functions have analytic expression $\nabla \log{p_{t}(x)}=\frac{\E[X\g X_t=x]-x}{2t}$, where the only unknown is $\E[X\g X_t=x]$. An alternative, but equivalent parametrization of the problem, consists in estimating the noise $W$, given $X_t$.
We use this approach in our work since is considered to be more stable numerically~\cite{ho2020}. In practice, the VP-SDE \cite{song2021a} framework is adopted as the noising process. With such a schedule varying between $[0,T]$, it's valid to assume that $X_T$ is practically indistinguishable from pure noise (More details in \Cref{apdx:detail} ).

\subsection{Estimating \acrshort{O-information}}

Armed with \Cref{prop:kl_est}, we can leverage score functions to estimate the information-theoretic quantities introduced in \Cref{prem}. Here we consider an extension of the simple noising process described in \Cref{sec:sde_est}, where we allow i) noising of only certain subsets of the variables or ii) deletion of subset of variables.
In practice, the first case corresponds to learning to denoise a portion of the variables, given auxiliary information about the other (noiseless) variables, e.g. to learn $\E[X^i\g X^i_t=\tilde{x}^i, X^{\setminus i}=x^{\setminus i}]$. Instead, the second case amounts to denoising problems akin to $\E[X^i\g X^i_t=\tilde{x}^i]$. In our implementation, we follow the approach proposed in \cite{bounoua2023multimodal} (See \Cref{apdx:detail}).
Next, we use such an intuition to derive a series of propositions that pave the way to \acrshort{O-information} computation.

In what follows, we use the compact notation $\begin{bmatrix} (\cdot)^i\end{bmatrix}_{i=1}^{N}$, to indicate a concatenation of $N$ elements in a column vector.

\begin{proposition}\label{prop:Test}
    Given a multivariate random variable $X=\{X^{1},\dots,X^{N}\} \sim p(x^1,\dots,x^N)$, and its corresponding noised version, the Total Correlation $\cT(X)$ is equal to:     
    \begin{equation*} \label{tc_est}
      \int \frac{1}{4t^2}\E\norm{\E[X\g X_t]-
      \begin{bmatrix}
          \E[X^i\g X^i_t]
      \end{bmatrix}_{i=1}^{N}}^2\dd t.   
    \end{equation*}        
\end{proposition}
\noindent \textbf{Proof Sketch.}
Recall that $\cT(X)=\KL{p(x)}{\prod\limits_{i=1}^N p(x^i)}$. Then, by virtue of \Cref{prop:kl_est}, we have that $\cT(X)$ equals
\begin{equation*}\label{eq:tc_est}
    \int p_t(x)\norm{\nabla\log p_{t}(x)-\begin{bmatrix}\frac{\partial}{\partial x^i}\log p_{t}(x^i)\end{bmatrix}_{i=1}^{N}}^2 \dd x \dd t. 
\end{equation*}

The terms $\frac{\partial}{\partial x^i}\log p_{t}(x^i)$ correspond to $\nicefrac{1}{2t}(\E[X^i\g X^i_t=x^i]-x^i)$. Then, the proposition follows. $\qedsymbol$

\begin{proposition}\label{prop:Sest}
    Given a multivariate random variable $X=\{X^{1},\dots,X^{N}\} \sim p(x^1,\dots,x^N)$, and its corresponding noised version, the \acrshort{S-information} $\cS(X)$ is equal to:
    \begin{equation*}
    \int \frac{1}{4t^2}\E\norm{
        \begin{bmatrix}
          \E[X^i\g X^i_t]
        \end{bmatrix}_{i=1}^{N}
          -
        \begin{bmatrix}
          \E[X^i\g X^i_t, X^{\setminus i}]
        \end{bmatrix}_{i=1}^{N}}^2\dd t.
    \end{equation*}
\end{proposition}
\noindent \textbf{Proof Sketch.} In light of \Cref{eq:def_s}, it holds that
\begin{equation*}
    \cS(X)=\sum_{i=1}^N \int p(x^{\setminus i})\KL{p(x^i\g x^{\setminus i})}{p(x^i)}\dd x^{\setminus i},
\end{equation*}
where the $i_{th}$ \gls{KL} term of the sum is equal to (\Cref{prop:kl_est})
\begin{flalign*}
& \int p(x^i\g x^{\setminus i})p_{0t}(\tilde{x}^i\g x^i) \\
& \norm{\frac{\partial}{\partial \tilde{x}^i} \log(\frac{p_{t}(\tilde{x}^i)}{\hat{p}_{0t}(\tilde{x}^i\g x^{\setminus i})})}^2\dd \tilde{x}^i\dd x^i\dd t.    
\end{flalign*}
Now, we can move the terms $p(x^{\setminus i})$ inside the \gls{KL} computation integrals and write the sum of the norms as the norm of a vector, which allows computing \acrshort{S-information} as
\begin{flalign*}\label{eq:s_est}
    &\cS(X)=
    \int p(x)p_{0t}(\tilde{x}\g x) \nonumber\\
    &\norm{
    \begin{bmatrix}
        \frac{\partial}{\partial \tilde{x}^i}\log p_{t}(\tilde{x}^i)
    \end{bmatrix}_{i=1}^{N}
    -
    \begin{bmatrix}\frac{\partial}{\partial \tilde{x}^i}\log p_{t}(\tilde{x}^i\g x^{\setminus i})
    \end{bmatrix}_{i=1}^{N}}^2 \dd \tilde{x} \dd x \dd t,
\end{flalign*}
where ${p}_{t}(\tilde{x}^i\g x^{\setminus i})=\int {p}_{0t}(\tilde{x}^i\g x^{i}){p}({x}^i\g x^{\setminus i}) \dd x^i$. 

Finally, the proposition follows since we can interpret the elements inside the square norm in terms of denoisers, with $\frac{\partial}{\partial \tilde{x}^i}\log p_{t}(\tilde{x}^i\g x^{\setminus i})=\nicefrac{1}{2t}(\E[X^i\g X^i_t=\tilde{x}^i, X^{\setminus i}=x^{\setminus i}]-\tilde{x}^i)$$\qedsymbol$

\begin{proposition}\label{prop:Dest}
    Given a multivariate random variable $X=\{X^{1},\dots,X^{N}\} \sim p(x^1,\dots,x^N)$, and its corresponding noised version, the Dual Total Correlation $\cD(X)$ equals:
    \begin{equation*}\label{dtc_est}
        \int \frac{1}{4t^2}\E\norm{\E[X\g X_t]-
        \begin{bmatrix}
            \E[X^i\g X^i_t, X^{\setminus i}]
        \end{bmatrix}_{i=1}^{N}}^2\dd t 
    \end{equation*}
\end{proposition}
\noindent \textbf{Proof Sketch.}
The starting point to obtain \gls{DTC} is to recall that $\cD(X)=\cS(X)-\cT(X)$. Then, it is sufficient to expand the square norms of $\cS(X)$ and $\cT(X)$ and combine the different terms, to state that $\cD(X)$ equals:
\begin{flalign*}\label{eq:dtc_est}
&\int p(x)p_{0t}(\tilde{x}\g x) \\
&\norm{\nabla\log p_{t}(\tilde{x})-
\begin{bmatrix}
    \frac{\partial}{\partial \tilde{x}^i}\log p_{t}(\tilde{x}^i\g x^{\setminus i})
\end{bmatrix}_{i=1}^{N}}^2 \dd \tilde{x} \dd x \dd t.
\end{flalign*}
This can be proven considering that  i) $\E\left[\E[X^i\g X_t]]\E[X^i\g X^i_t]\right]=\E[(\E[X^i\g X^i_t])^2]$ ii) $\E\left[\E[X^i\g X^i_t]\E[X^i\g X^i_t, X^{\setminus i}] \right]=\E\left[(\E[X^i\g X^i_t])^2\right]$ and iii) $\E\left[\E[X^i\g X_t]]\E[X^i\g X^i_t,X^{\setminus i}]\right]=\E[(\E[X^i\g X^i_t])^2]$.
Then, the proposition follows. $\qedsymbol$

Finally, to estimate \acrshort{O-information}, it is sufficient to combine \Cref{prop:Test} and \Cref{prop:Dest}, and apply \Cref{eq:oinfo}. In practical terms, our method requires access to \textit{denoisers} for the three following scenarios: i) given $X_t$ estimate $X$ ii) given $X_t^i$ estimate $X^i$ iii) given $X_t^i$ and $X^{\setminus i}$ estimate $X^i$. To achieve this, we extend the methodology proposed in \cite{bounoua2023multimodal}, and amortize the three different scenarios with a \textit{unique denoising network}, which takes as input the concatenation of noised and clean variables and outputs the corresponding estimates (see \Cref{apdx:detail}). Additionally, the estimation of the gradients of O-information requires approximating additional denoising score functions to access \Cref{grad_eq} (More details in \Cref{apdx_grad}).

\section{Experimental validation}

\label{experiment}

We evaluate our method according to two strategies. First, we focus on a synthetic setup that allows analytic computation of \acrshort{O-information} and full control on system scale. Then, we consider real data collected in a study of brain activity in mice, to demonstrate how \gls{SOI} unlocks new avenues in the application of information measures in real systems without the need for restrictive assumptions.

\subsection{Synthetic benchmark}
\label{syntheticgaussian}

We consider a canonical Gaussian system, whereby we control the number of variables describing the system $N$, the dimension of each variable (\textbf{Dim}), the inter-dependencies between variables describing how they interact, and the strength of interaction (More details in \Cref{apdx:detail} ). Inspired by \cite{czyz2023beyond}, we consider more challenging distribution going beyond the Gaussian setting (Please refer to \Cref{apdx:non_norm} ). No other neural estimator capable of estimating \acrshort{O-information} was explored in the literature. Next, we construct an original baseline that relies on neural estimation of \gls{MI} to access the \gls{MI} decomposition of \acrshort{O-information}.

\paragraph{Baseline.} Recent work~\cite{bai2023estimating} describes a method to compute \gls{TC} by leveraging a decomposition into pairwise \gls{MI} terms. Clearly, \gls{DTC} can also be decomposed into \gls{MI} terms. Therefore, we extend~\cite{bai2023estimating} such that it can be used as a baseline to compute \acrshort{O-information}. The main limitation of such a baseline is poor scalability: it requires training an individual model for each \gls{MI} term in which \gls{TC} and \gls{DTC} are decomposed in. We adopt the linear-decomposition method~\cite{bai2023estimating}, which results in $2(N-1)$ \gls{MI} terms (see \Cref{apdx:exp}), and propose four variants to estimate \gls{MI} based on \cite{belghazi2018mine,nguyen2007neurips,oord2019representation,cheng2020}. We label this baseline approach according to the \gls{MI} estimators: \acrshort{MINE} ,\acrshort{NWJ}, \acrshort{InfoNCE}, and \acrshort{CLUB}.
 
\paragraph{Experimental protocol}
For each experiment, we use $100k$ samples for training the various neural estimators, and $10k$ samples at inference time, to estimate \acrshort{O-information}. For our method \gls{SOI}, we use the \textsc{vp}-\textsc{sde} formulation~\cite{song2021a} and learn a \textit{unique} denoising network to estimate the various score terms. The denoiser is a simple, stacked \gls{MLP} with skip connections, adapted to the input dimension. We apply importance sampling~\citep{huang2021variational,song2021a} at both training and inference time. Finally, we use 10-sample Monte Carlo estimates for computing integrals. More details about the implementation are included in \Cref{apdx:exp}. For the baseline variants, for each \gls{MI} term we use an \gls{MLP} that is sufficiently expressive given the data dimension. All results are averaged over 5 seeds.
Additional results are included in \Cref{additional}.

Our experiments unfold according to three inter-dependency scenarios, for systems characterized by either redundancy, synergy or a mix of both interactions.

\paragraph{Redundancy benchmark.}
We consider $R = \mathcal{N}(0,\mathbb{I})$ as the redundant information component in the system. All system variables are of the form $X = \{X^1,\dots,X^N\} = \{R + \epsilon_i,\dots, R + \epsilon_N\}$, where $\epsilon_i \sim \mathcal{N}(0,\sigma \mathbb{I})$ are mutually independent random noise samples with standard deviation $\sigma$. We use $\sigma$ to modulate the redundancy level: higher noise levels decrease the strength of redundant interaction, and this has an impact on the value of \acrshort{O-information}.

\begin{figure}[h]

\centering
\begin{subfigure}{0.3\textwidth}
         \centering
    \includegraphics[page=1,width=\linewidth]{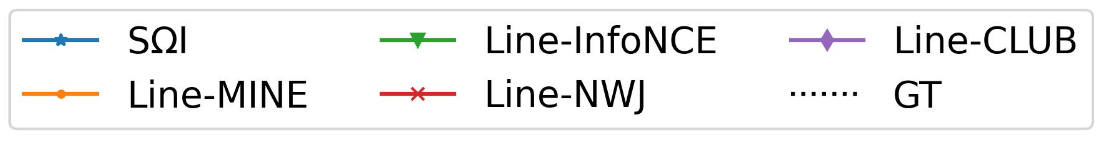}
    
     \end{subfigure}
     
     \begin{subfigure}{0.22\textwidth}
         \centering
    \includegraphics[page=1,width=\linewidth]{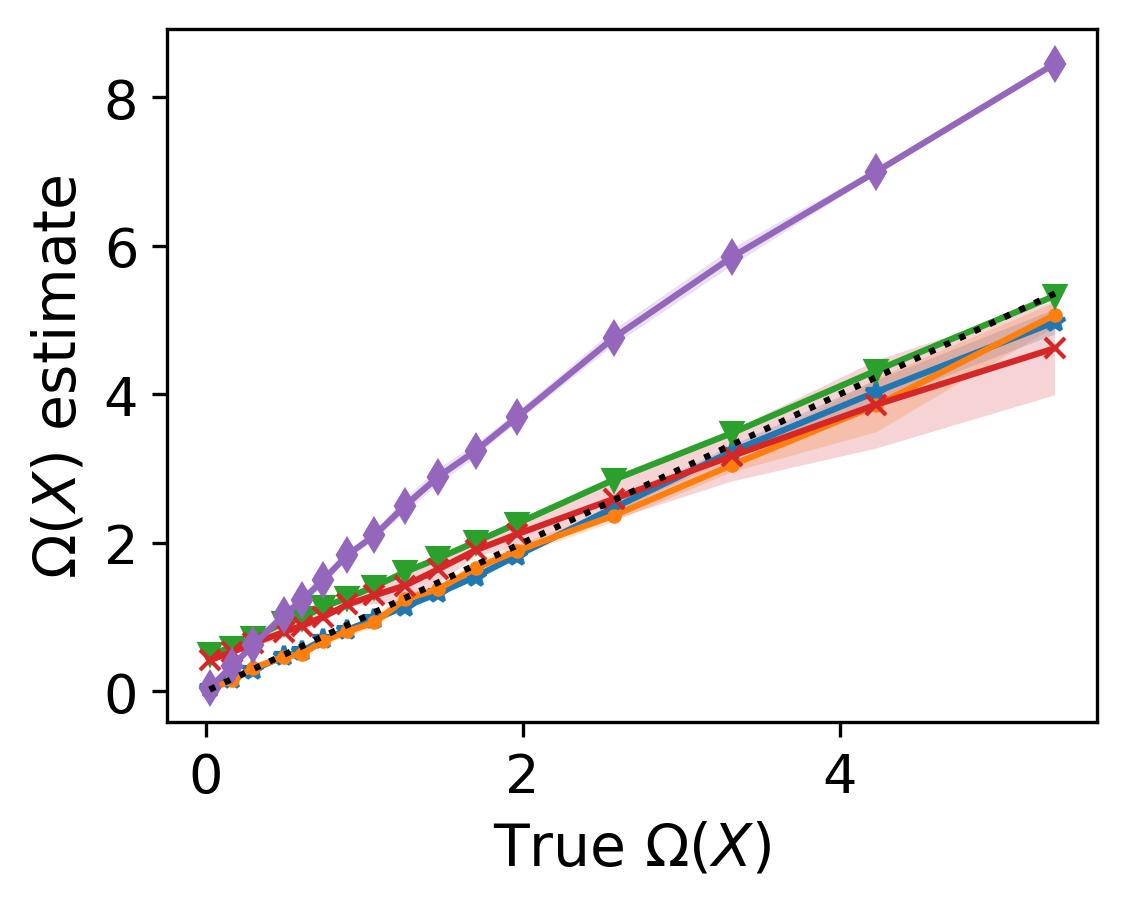}
         \caption{Dim = 5}
     \end{subfigure}
     \begin{subfigure}{0.22\textwidth}
         \centering
         \includegraphics[page=1,width=\linewidth]{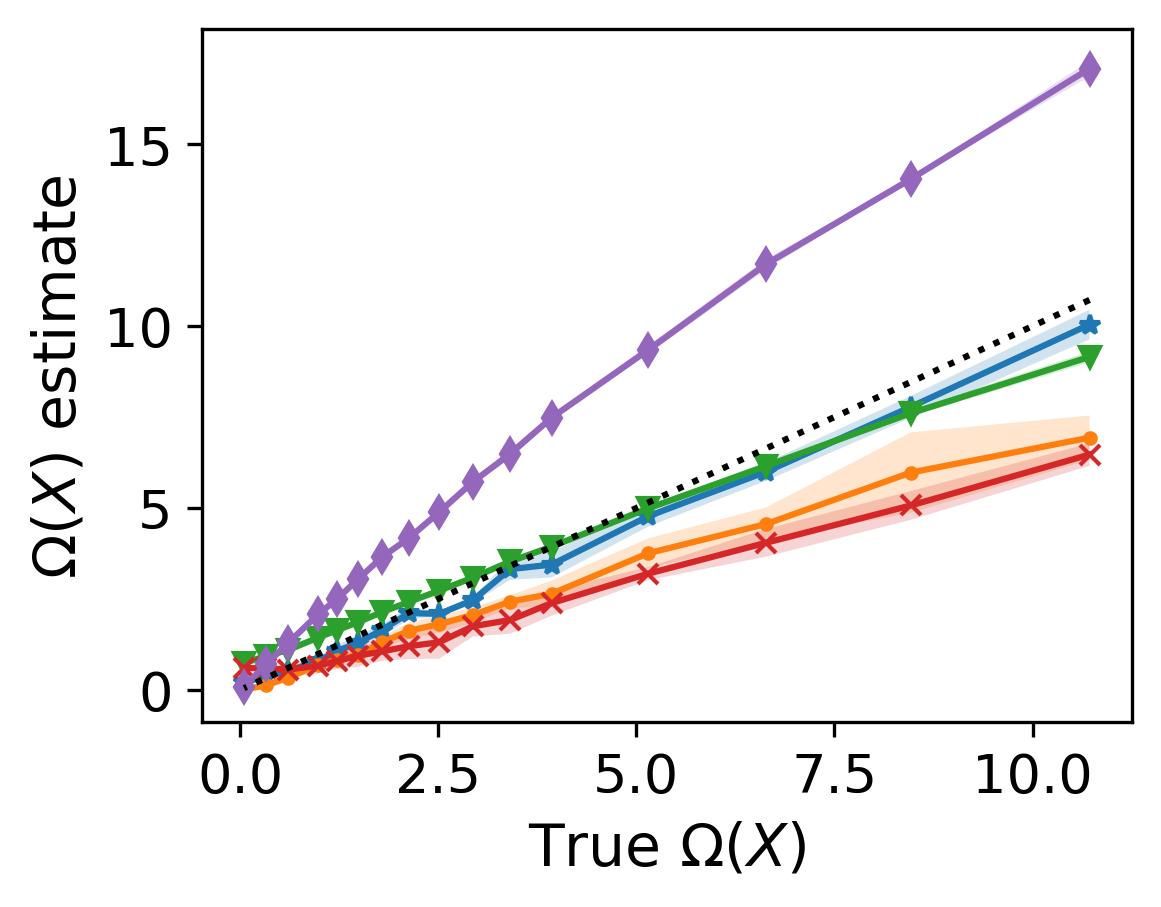}
         \caption{Dim = 10}
     \end{subfigure}
     
      \begin{subfigure}{0.22\textwidth}
         \centering
         \includegraphics[page=1,width=\linewidth]{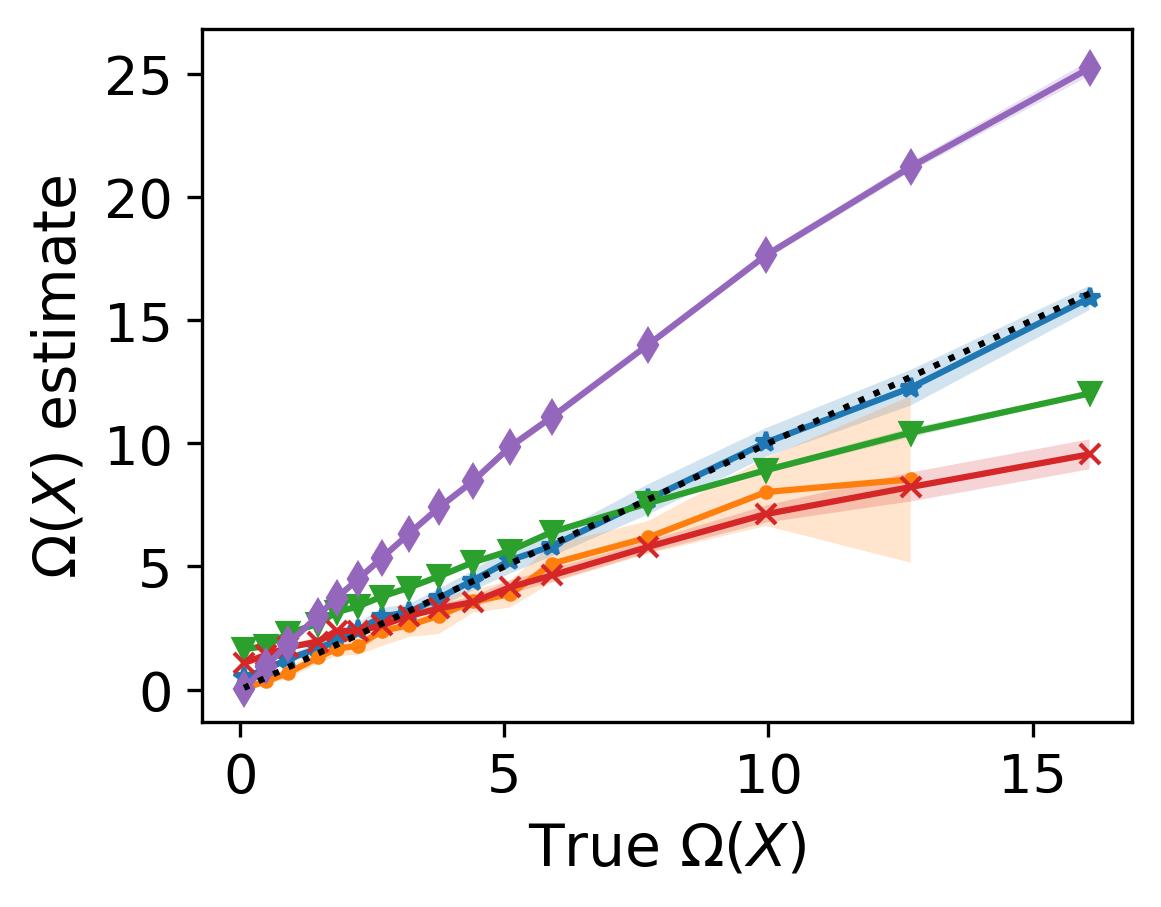}
         \caption{Dim = 15}
     \end{subfigure}
      \begin{subfigure}{0.22\textwidth}
         \centering
         \includegraphics[page=1,width=\linewidth]{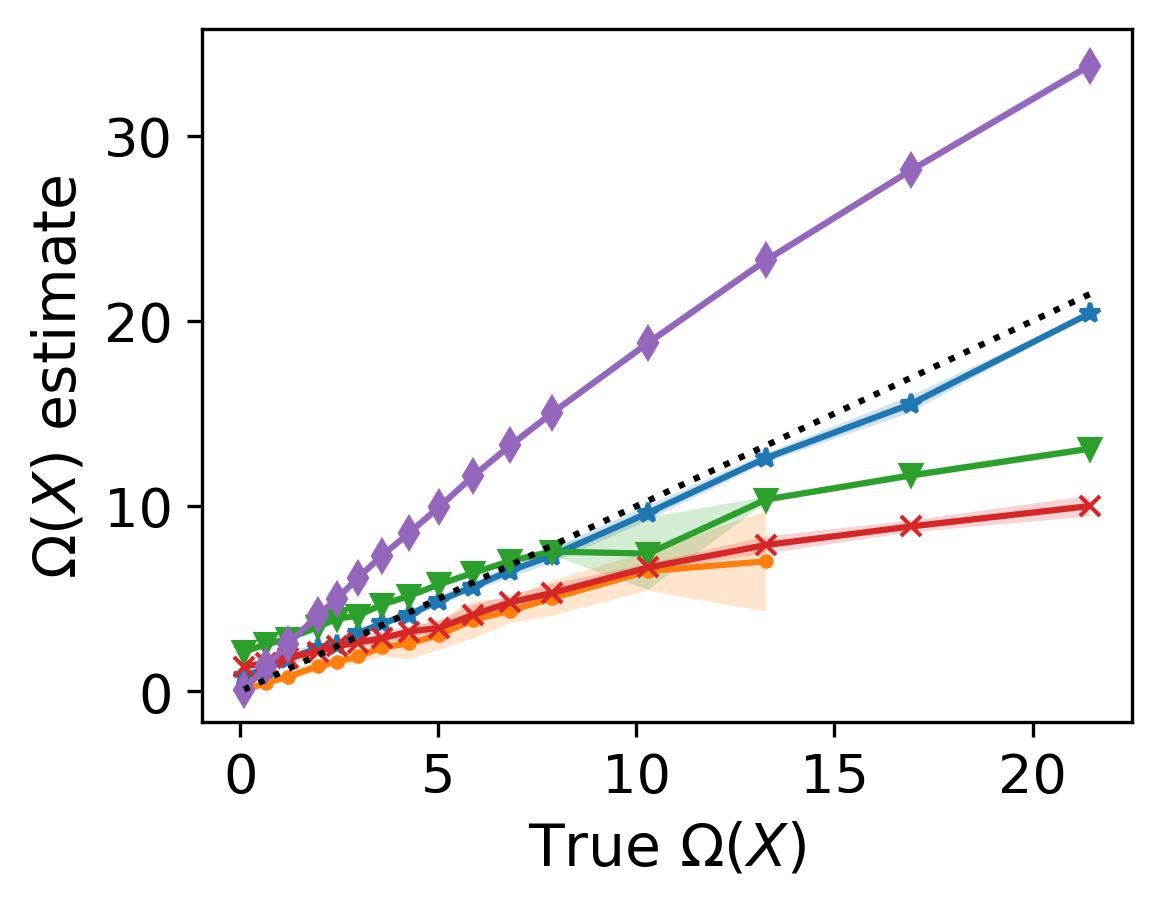}
         \caption{Dim = 20}
     \end{subfigure}
     \caption{Redundant system with $N=$10 variables, organized into subsets of sizes $\{3,3,4\}$ and increasing interaction strength.}
     \label{red_fig_10}
\end{figure}

Next, we discuss results for a system with $N=10$ variables, organized as 3 redundant subsystem, each defined as described above. \Cref{red_fig_10} illustrates, for various variable dimension, ranging from 5 to 20 dimensional Gaussians, the ground-truth and the estimated \acrshort{O-information}, for \gls{SOI} and the various baselines.
In this scenario, \gls{SOI} and baseline competitors produce fairly accurate \acrshort{O-information} estimates, when the dimensionality of each random variable is small. When the dimension of systems variables grows, however, the performance of the baseline methods degrades considerably. This is due to the inherent limitations of the pairwise neural \gls{MI} estimators, that struggle with high dimensional data~\cite{czyz2023beyond}. Instead, the performance of \gls{SOI} remains stable when increasing variable dimension, and \acrshort{O-information} estimates are accurate, even when interaction strength is high.

\paragraph{Synergy benchmark.} 
In this case, we synthesize synergistic inter-dependency among system variables by considering the following setup. For simplicity, consider three random variables that behave as follows:
\begin{align*}
    & X^1 \sim \mathcal{N}(0, \mathbb{I}) , \quad X^2 = X^1 +  S \\
    & X^3 =  S + \epsilon ,\quad  \epsilon \sim \mathcal{N}(0,\sigma) \text{  and  } X^1 \indep X^3
\end{align*}
with $  S \sim \mathcal{N}(0, \mathbb{I}) $.

When $\sigma = 0 $, the synergy emerges through the Markov chain $\{X^2,X^3\} - X^1 $, $ \{X^1,X^3\} - X^2 $ and $\{X^1,X^2\} - X^3 $, since no element alone is sufficient to recover the remaining variables. We modulate $\sigma$ to achieve different synergistic strengths. More generally, we simulate $N$ synergistic variables as: $X^1 \sim \mathcal{N}(0, \mathbb{I})$, $X^2 = X^1 +  S_1 + ... + S_{N-2}$ and $X^i = S_{i-2} + \epsilon_{i-2} \forall i \in \{3,..,N\}$.

\begin{figure}[h!]

\centering
\begin{subfigure}{0.3\textwidth}
         \centering
    \includegraphics[page=1,width=\linewidth]{assets/figures/exp_red/legend.PNG}
    
     \end{subfigure}
     
     \begin{subfigure}{0.22\textwidth}
         \centering
         \includegraphics[page=1,width=\linewidth]{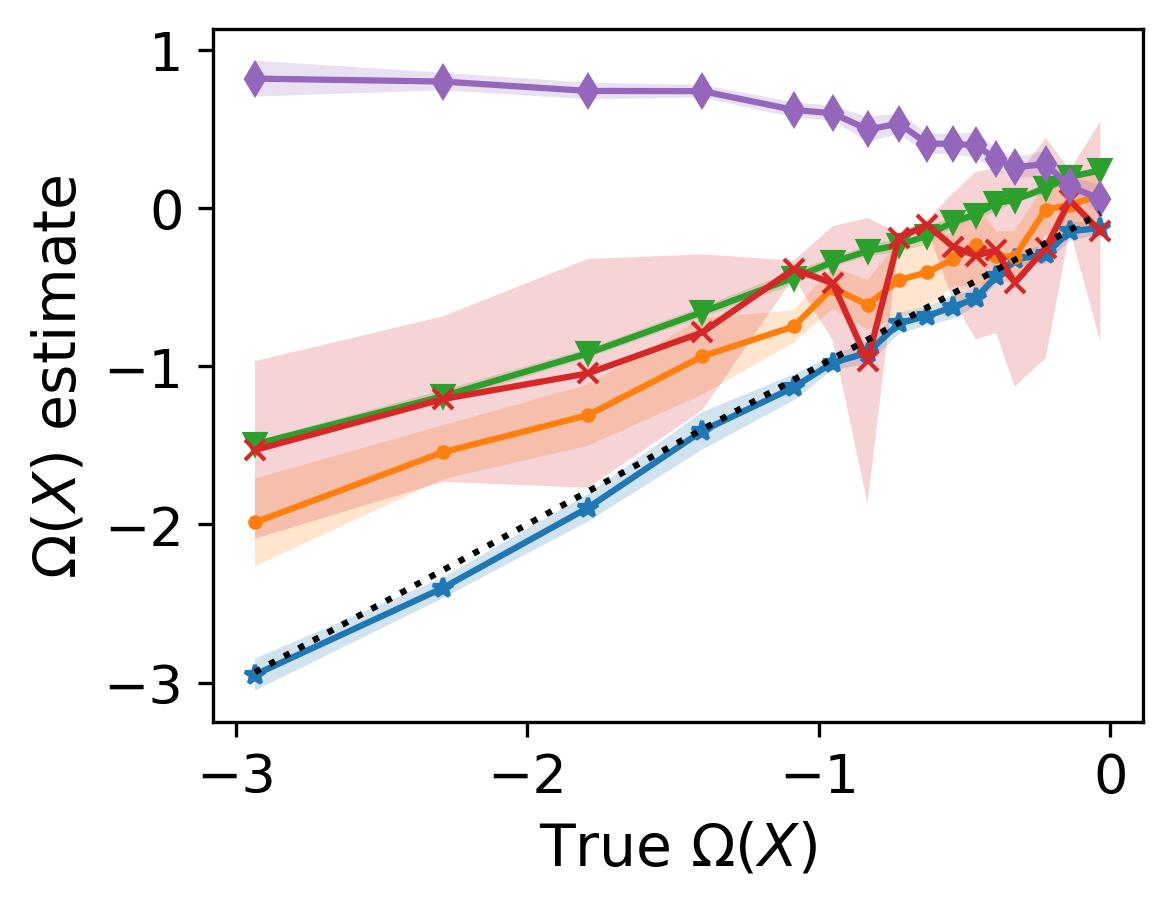}
         \caption{Dim = 5}
     \end{subfigure}
     \begin{subfigure}{0.22\textwidth}
         \centering
         \includegraphics[page=1,width=\linewidth]{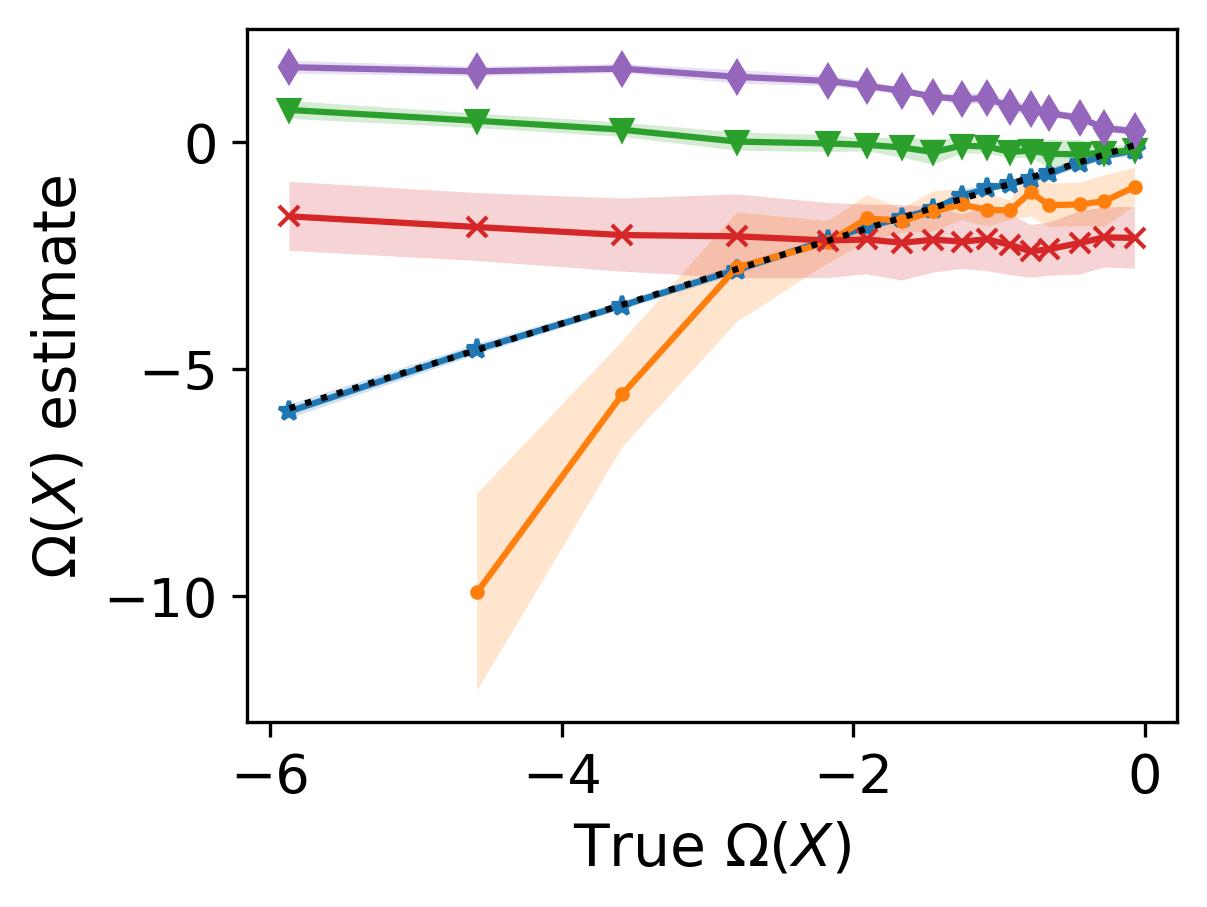}
         \caption{Dim = 10}
     \end{subfigure}
     
      \begin{subfigure}{0.22\textwidth}
         \centering
         \includegraphics[page=1,width=\linewidth]{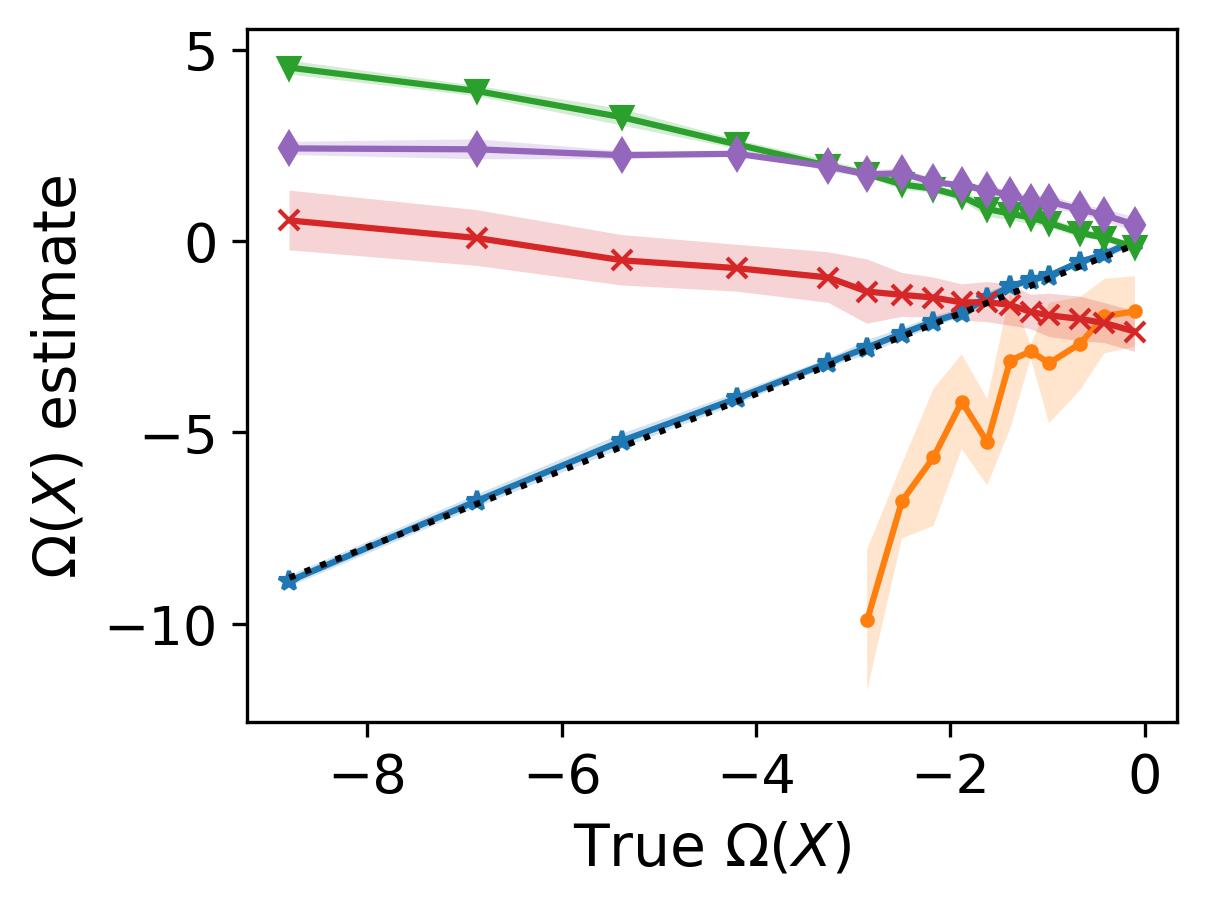}
         \caption{Dim = 15}
     \end{subfigure}
      \begin{subfigure}{0.22\textwidth}
         \centering
         \includegraphics[page=1,width=\linewidth]{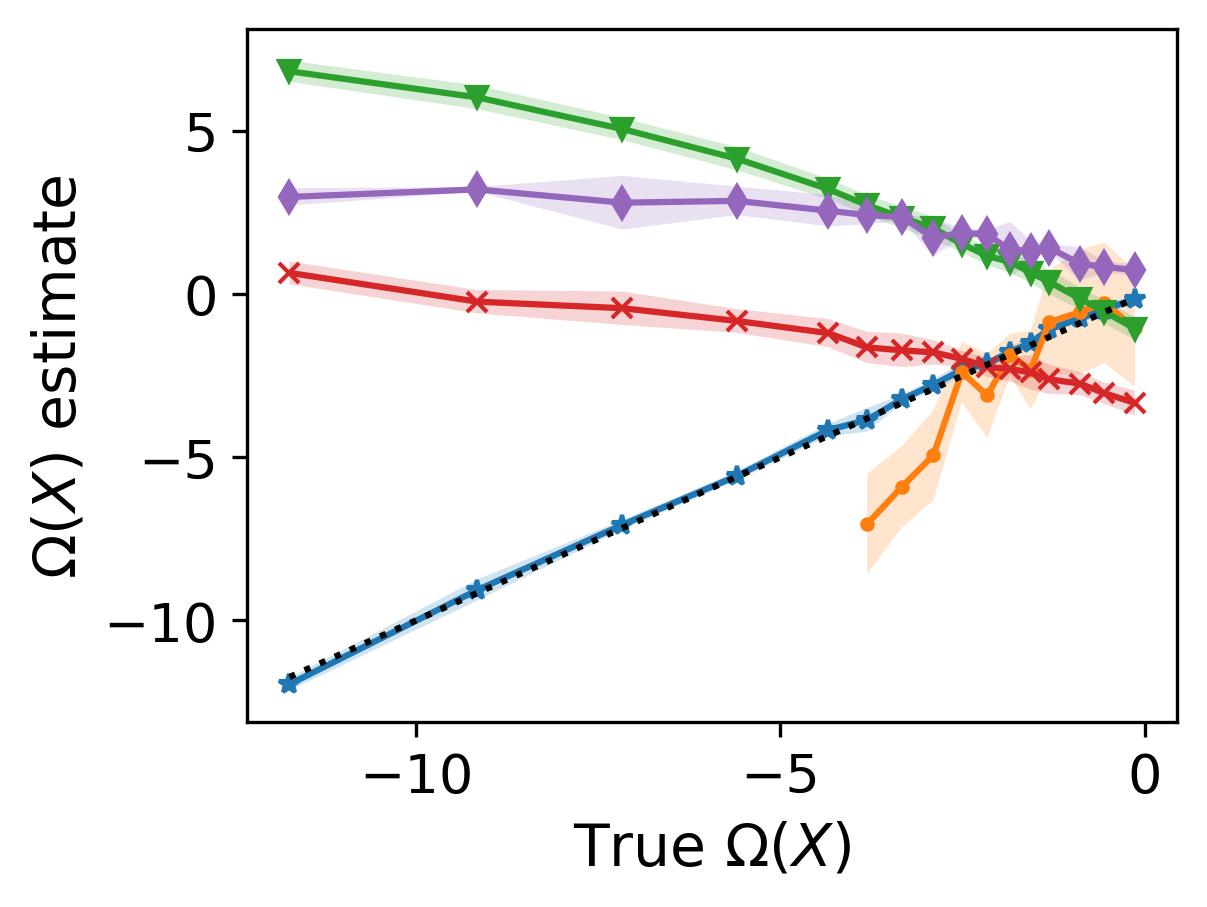}
         \caption{Dim = 20}
     \end{subfigure}
          \caption{Synergistic system with $N=$10 variables, organized into subsets of sizes $\{3,3,4\}$ and increasing interaction strength.}
          \label{syn_fig_10}
\end{figure}

Results in \cref{syn_fig_10} show that \gls{SOI} achieves consistent results in all scenarios, whereas the baselines behave poorly. Indeed, a synergy-only setting is challenging, as it's dominated by high \gls{DTC} values required to capture high-order interactions, on which the baselines based on pairwise \gls{MI} estimator fail. 

\paragraph{Mixed benchmark.} 
In general, systems components are characterized by a mix of redundant and synergistic interactions. Then, we synthesize such a system by creating subgroups dominated by redundancy and synergy, respectively, following the procedures defined above. 

\begin{figure}[h]

\centering
\begin{subfigure}{0.3\textwidth}
         \centering
    \includegraphics[page=1,width=\linewidth]{assets/figures/exp_red/legend.PNG}
    
     \end{subfigure}
     
     \begin{subfigure}{0.22\textwidth}
         \centering
         \includegraphics[page=1,width=\linewidth]{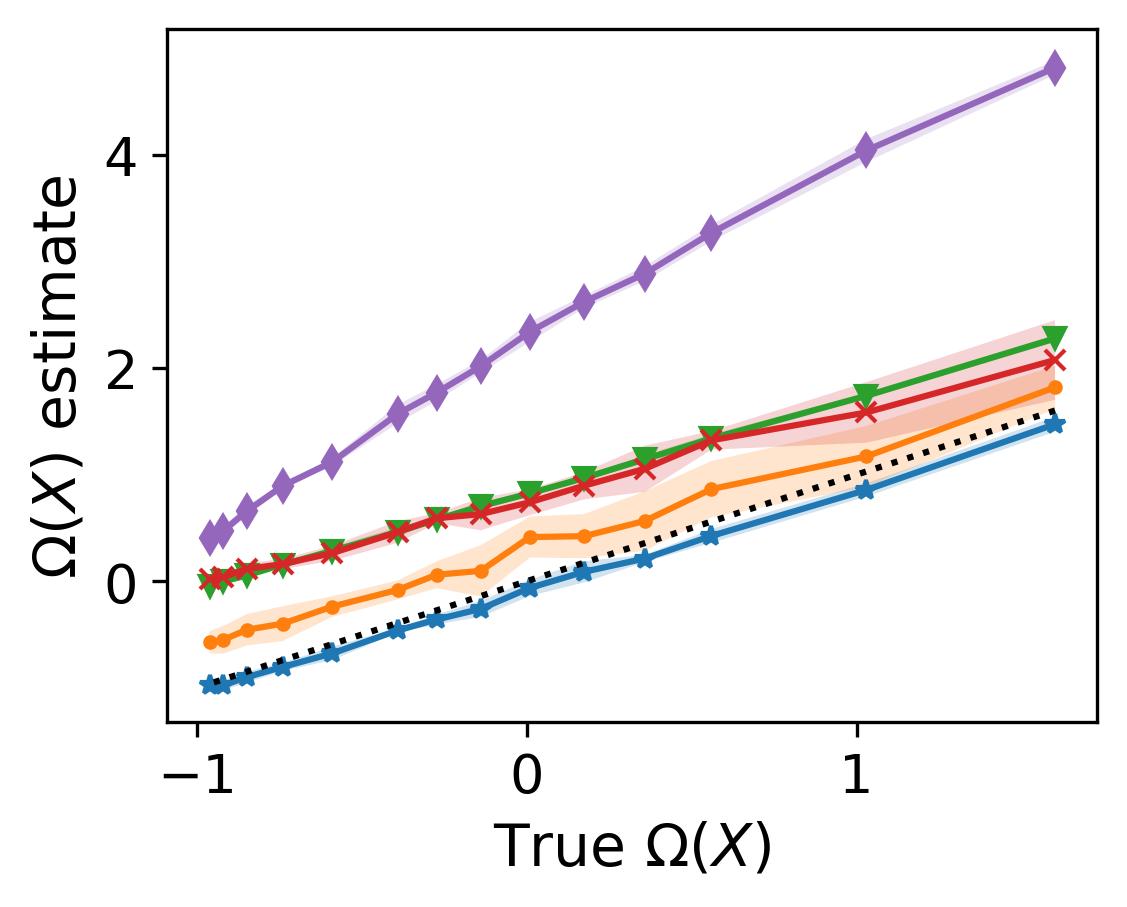}
         \caption{Dim = 5}
     \end{subfigure}
     \begin{subfigure}{0.22\textwidth}
         \centering
         \includegraphics[page=1,width=\linewidth]{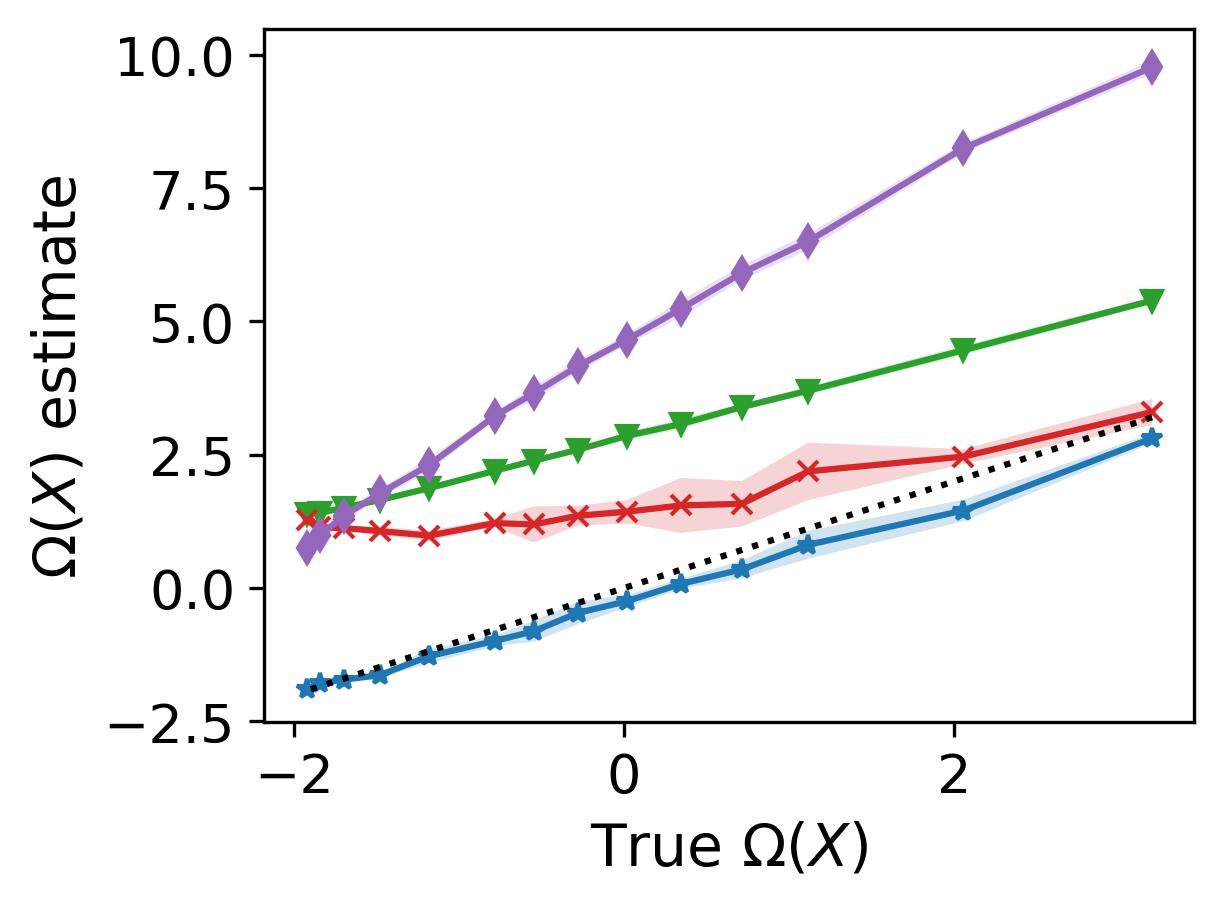}
         \caption{Dim = 10}
     \end{subfigure}
     
      \begin{subfigure}{0.22\textwidth}
         \centering
         \includegraphics[page=1,width=\linewidth]{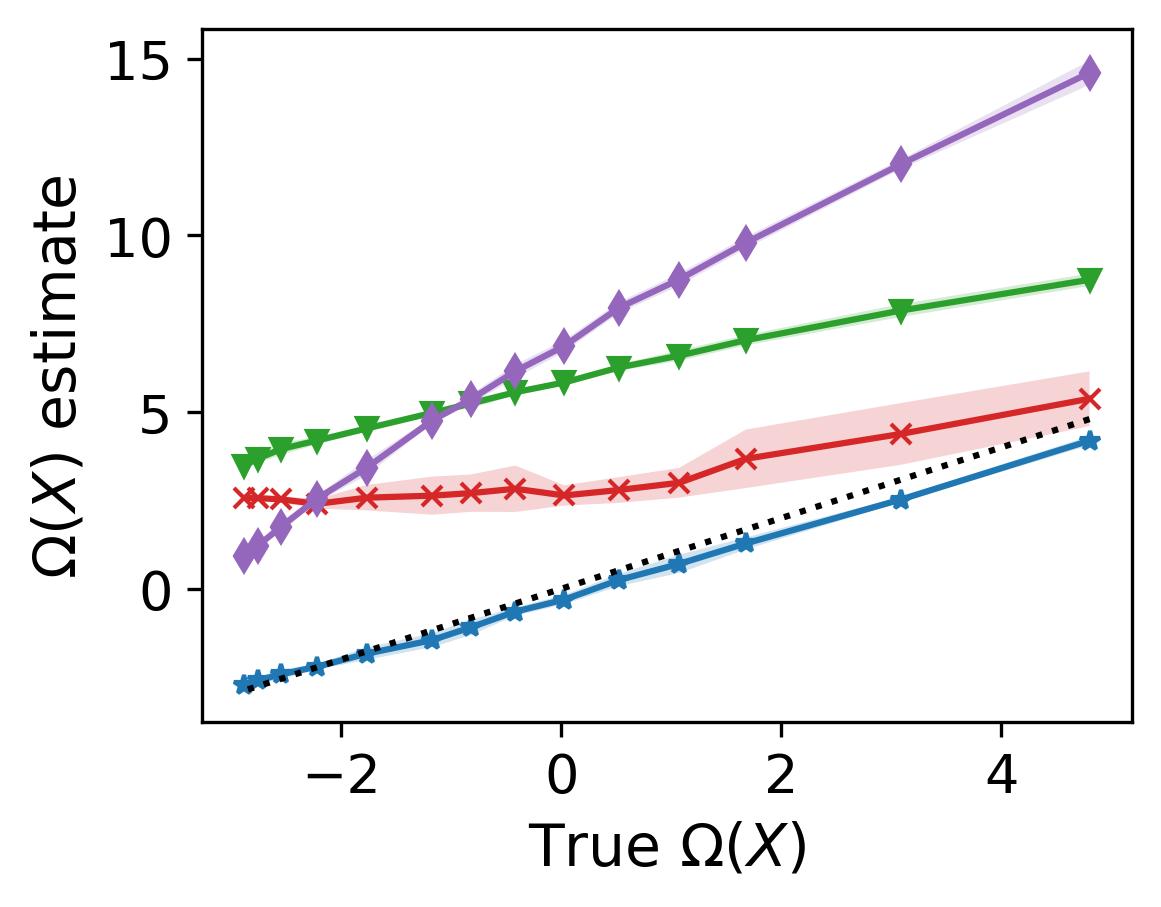}
         \caption{Dim = 15}
     \end{subfigure}
      \begin{subfigure}{0.22\textwidth}
         \centering
         \includegraphics[page=1,width=\linewidth]{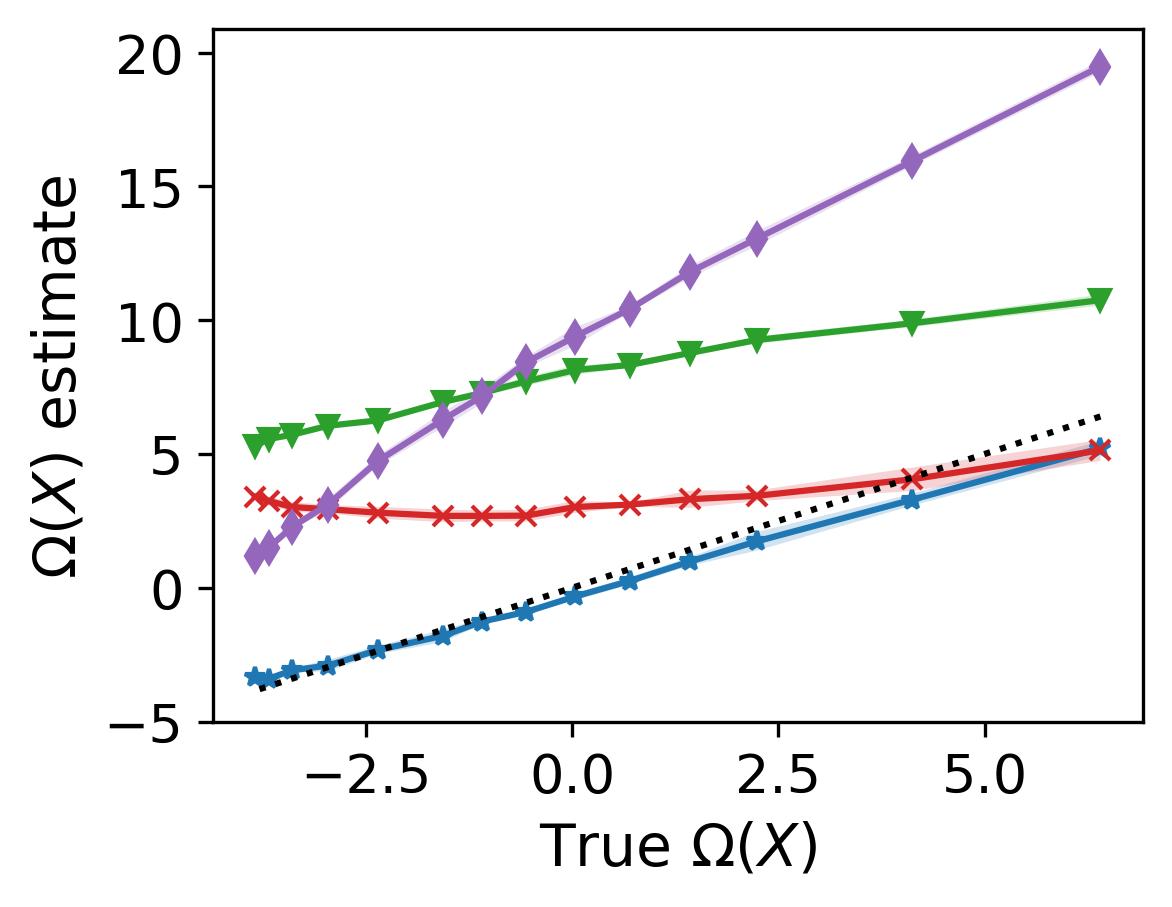}
         \caption{Dim = 20}
     \end{subfigure}
      \caption{Mixed-interaction system with $N=$10 variables, organized into 2 redundancy-dominant subsets of size $\{3,4\}$ variables and one synergy-dominant subset with $3$ variables. \acrshort{O-information} is modulated by fixing the synergy inter-dependency and increasing the redundancy.}
      \label{mix_fig_10}
\end{figure}

Results in \cref{mix_fig_10}, demonstrate that our method \gls{SOI} stands out as the best estimator in this challenging scenario. Baseline methods produce poor estimates, especially when the synergistic interaction is dominant. Note that \gls{SOI} reports a negative \acrshort{O-information} whenever the system is synergy-dominant and also succeeds in capturing interaction strengths, when the system equilibrium changes in favor redundant interactions, by estimating correctly a positive \acrshort{O-information}.

\paragraph{Discussion.} 
We attribute the superior performance of \gls{SOI}, compared to the baselines, to several factors. Score-based estimators have shown to be extremely successful in fitting complex distributions, for example in the context of generative modeling \cite{song2020, song2021a}. Moreover, our technique relies on \Cref{prop:kl_est}, whereby the difference of score functions has been shown to produce an accurate estimate of \gls{KL} divergences, due to canceling effects of estimation errors \cite{franzese2023minde}. 
Note also that the baselines we adopt in our work use \gls{MI} estimators that produce a bound only. Moreover, using individual models to estimate several \gls{MI} terms can naturally suffer from cumulative bias, which is avoided in our case by amortizing computation with a unique neural network.

\begin{figure} [h]

     \begin{subfigure}{0.22\textwidth}
         \centering
         \includegraphics[page=1,width=\linewidth]{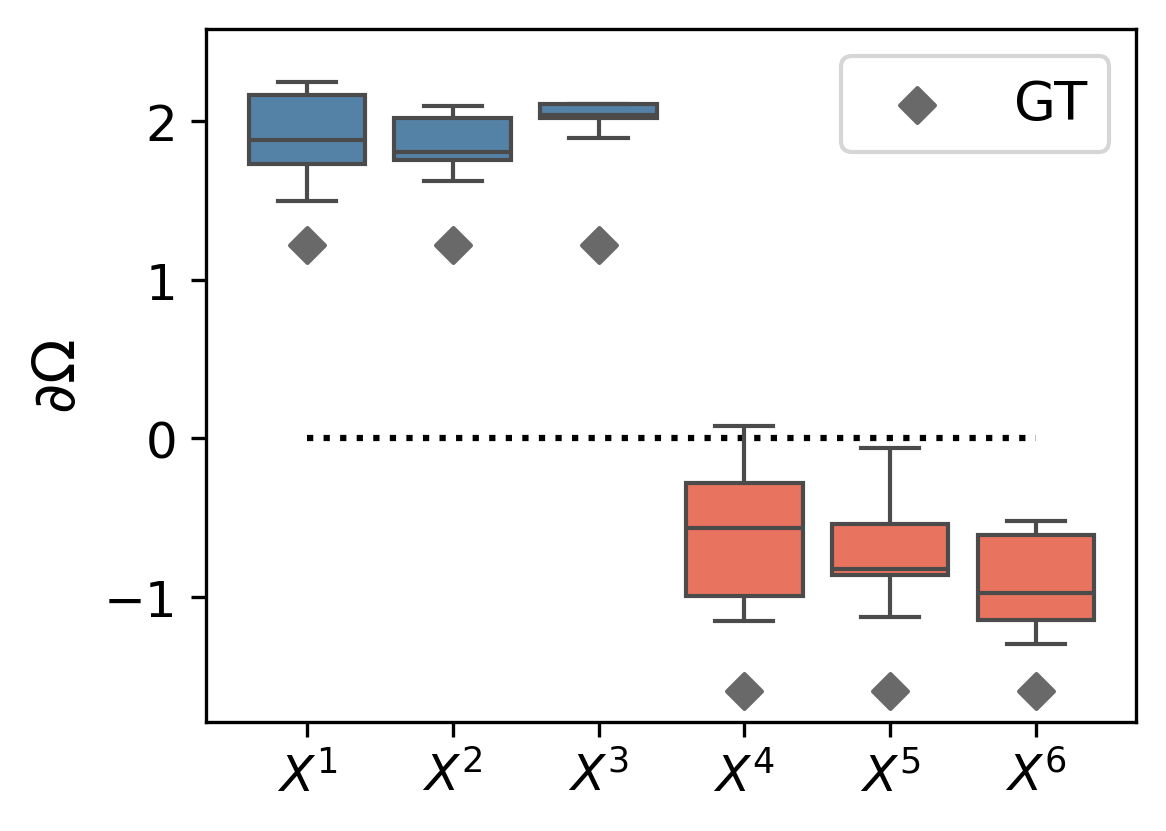}
         \caption{Dim = 5}
     \end{subfigure}
      \begin{subfigure}{0.22\textwidth}
         \centering
         \includegraphics[page=1,width=\linewidth]{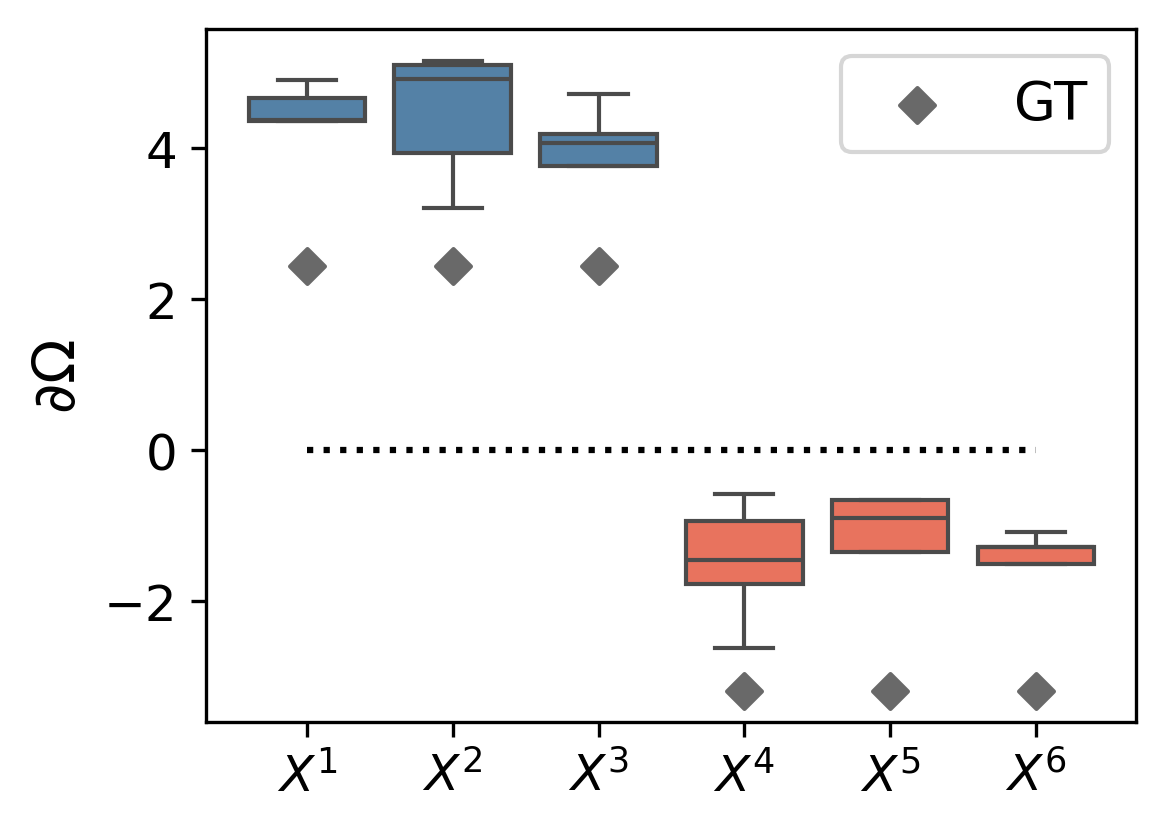}
         \caption{Dim = 10}
     \end{subfigure}
     
      \begin{subfigure}{0.22\textwidth}
         \centering
         \includegraphics[page=1,width=\linewidth]{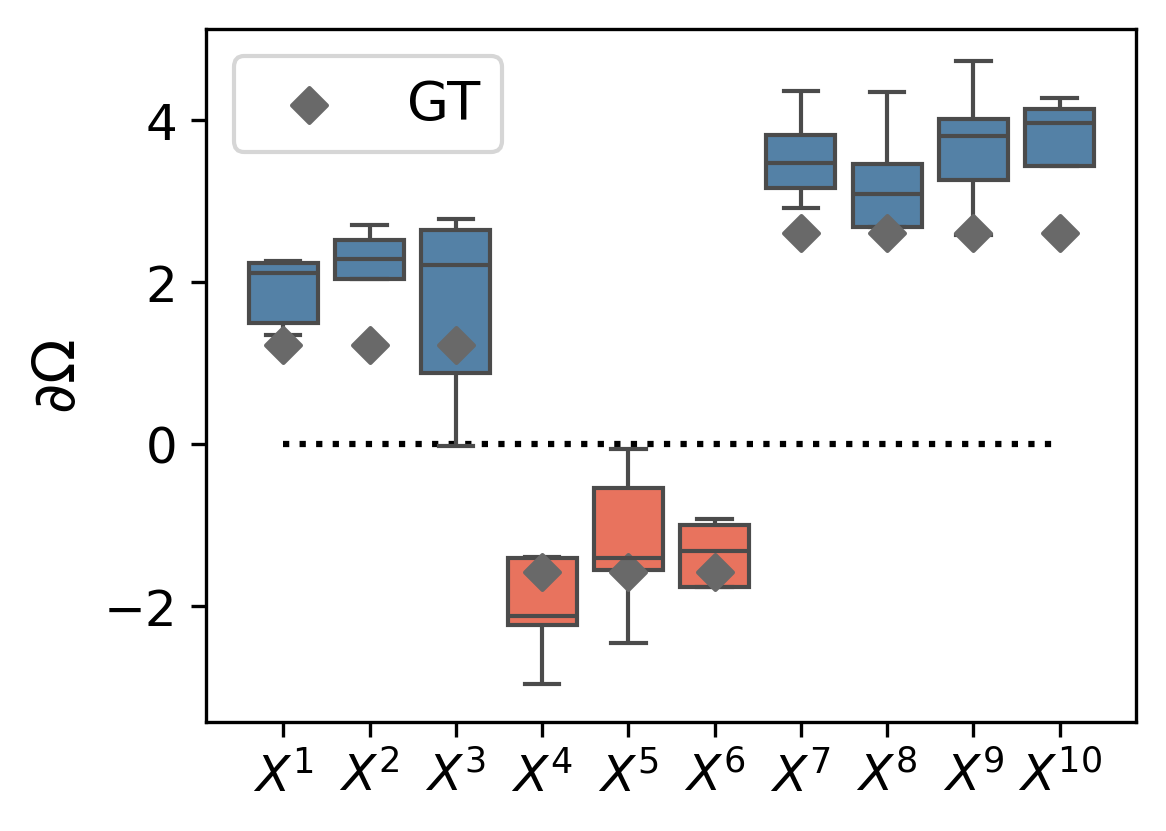}
         \caption{Dim = 5}
     \end{subfigure}
      \begin{subfigure}{0.22\textwidth}
         \centering
         \includegraphics[page=1,width=\linewidth]{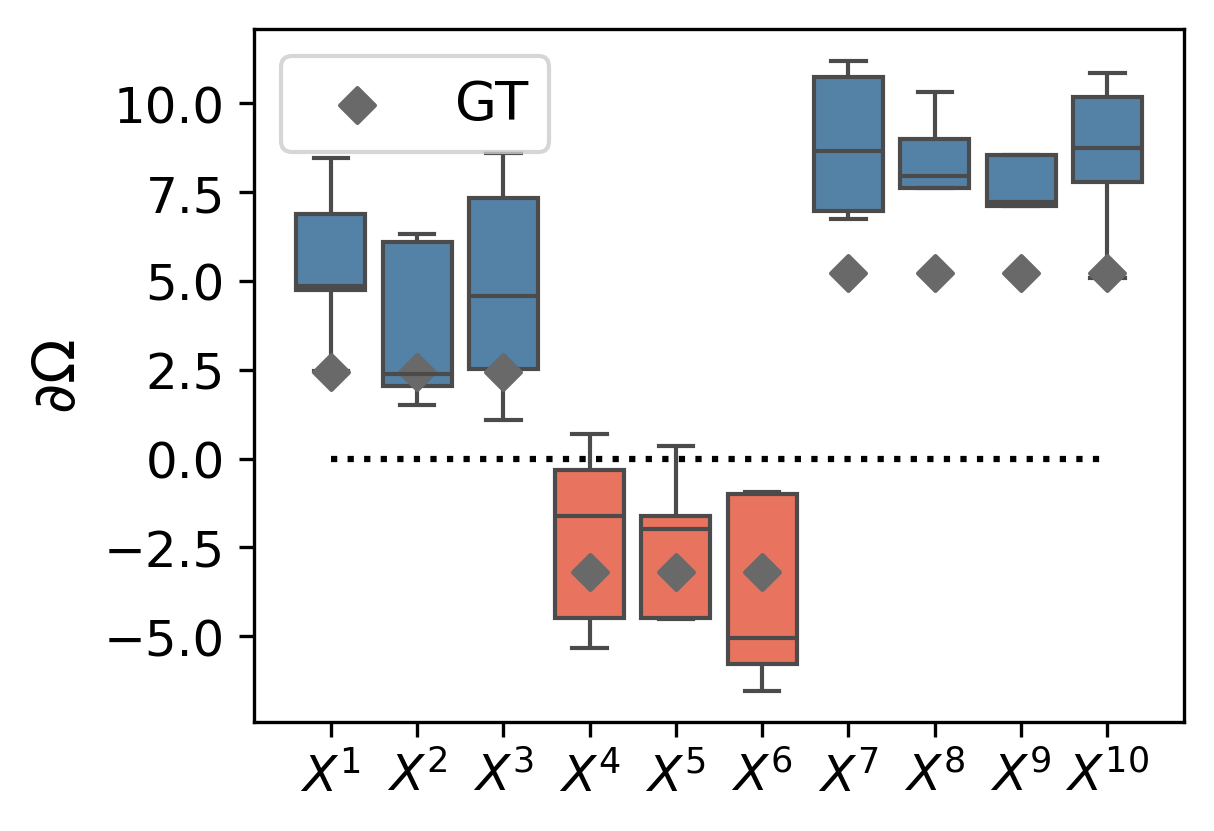}
         \caption{Dim = 10}
     \end{subfigure}

      \caption{Gradient of \acrshort{O-information} for the mixed benchmark, for a system of $N=$6 variables, and a system of $N=$10 variables, and different dimension of variables. 
      }
      \label{grad_oinf}
\end{figure}

\paragraph{Gradient of O-information}
While \acrshort{O-information} provides global information about dominance of either synergy or redundancy, the contribution of individual variables to either effects is not available. Next, we rely on the gradient of \acrshort{O-information} to study individual system components, as introduced in \Cref{prem}. Indeed, our method \gls{SOI} can be easily extended to output such gradients, by estimating additional score functions, as described in \Cref{apdx:detail}.
In \Cref{grad_oinf}, we illustrate gradients of \acrshort{O-information} applied to the mixed benchmark scenario discussed above. While \acrshort{O-information} of the whole system can be positive due to the redundancy strength of some subgroup of variables, we notice that three variables report a negative gradient, which is indicative of their synergistic interaction. In \Cref{grad_oinf}, ground truth gradient values are showed using a diamond marker. Our estimator, despite suffering from some bias, correctly attributes the role and interaction type of each system constituent.

\subsection{Application to a real system}
\label{sec:vbn}
Multivariate analysis is a powerful tool for the field of neuroscience, as it allows scientists to analyze activity patterns of different brain regions. Understanding how the brain processes and transmits information during different stimulus requires analysing the underlying inter-dependencies between different brain regions. To show that \gls{SOI} is an effective tool also in practical use cases, we now consider the Visual Behavior project, which used the Allen Brain Observatory to collect a highly standardized dataset consisting of recordings of neural activity in mice that have learned to perform a visually guided task~\cite{allen-inst}.

A visual change-detection task experiment was conducted on 80 mice using six neuropixels probes tasked to report the activity of different regions of the visual cortex. During the recordings, a set of 8 natural scenes were presented in 250 ms flashes, at intervals of 750 ms. The same image was shown during several flashes before a change to a new image. The mouse had to perform an action to receive a water reward when the image changed. Ultimately, the purpose of this experiment is to investigate how the different brain region of the mice react to different types of stimulus, such as detecting a new image (\textit{change}) or not (\textit{no change}).

In this work, we follow the prepossessing procedure described by \cite{venkatesh2023gaussian}, where in each experimental session, good quality units from each area are chosen (See \Cref{apdx:exp}). 
For each trial, the recorded spikes are binned in 50 ms intervals, starting from the stimulus flash. 
We consider two types of flashes: \textit{change} and \textit{no change}. For both cases, \gls{SOI} is used for each time bin to estimate \gls{O-information}. The reported estimation is done using 10 Monte Carlo integration steps and averaged over multiple seeds.  
We first consider three visual cortex regions \textsc{VISp}, \textsc{VISl} and \textsc{VISal}, as done in~\cite{venkatesh2023gaussian}. We then extend the experiment to six brain regions by including \textsc{VISrl}, \textsc{VISam} and \textsc{VISpm}. 

\begin{figure} [h]

\centering
     \begin{subfigure}{0.38\textwidth}
         \centering
         \includegraphics[page=1,width=\linewidth]{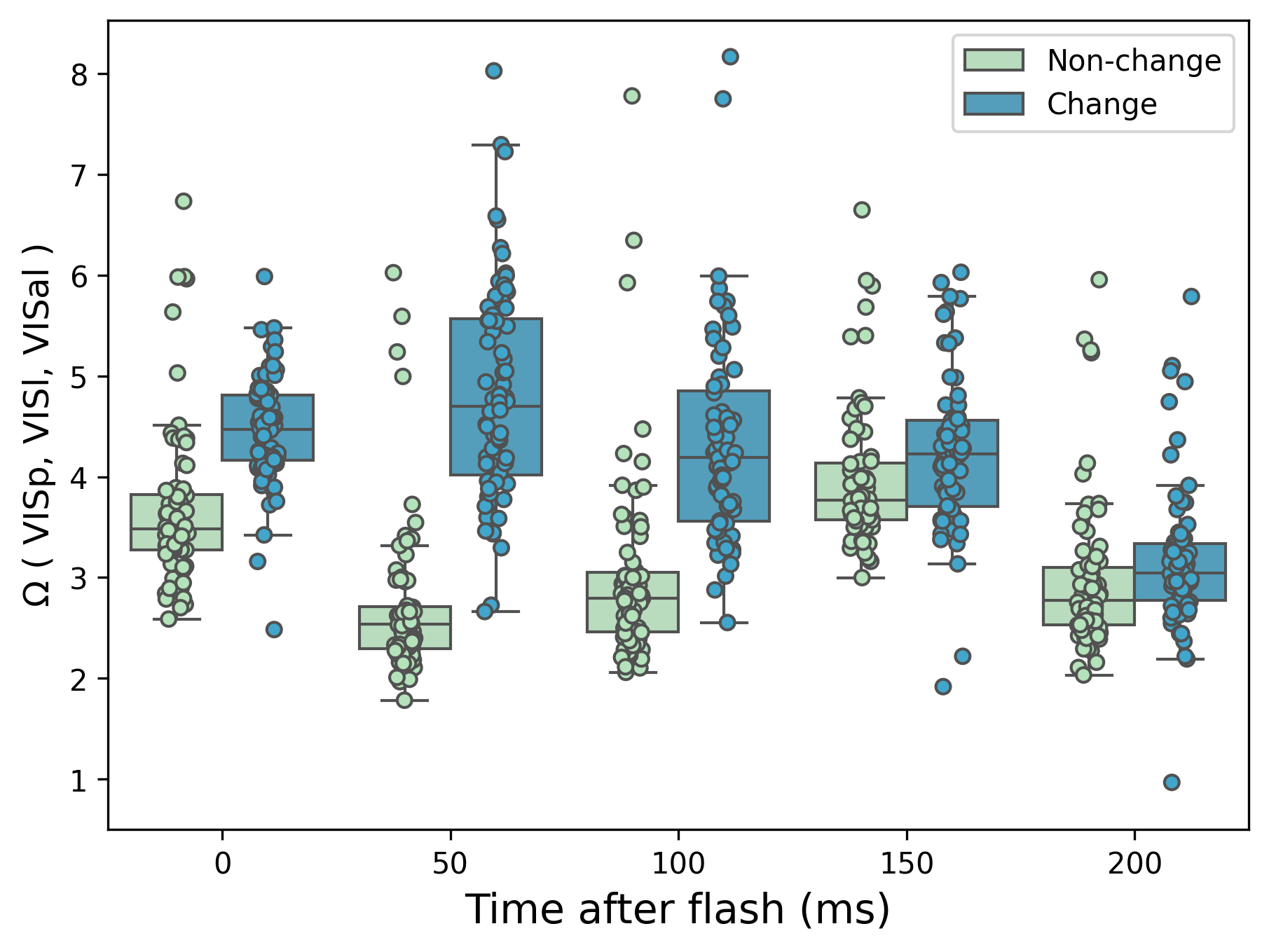}
         \caption{3 areas}
     \end{subfigure}
      \begin{subfigure}{0.38\textwidth}
         \centering
         \includegraphics[page=1,width=\linewidth]{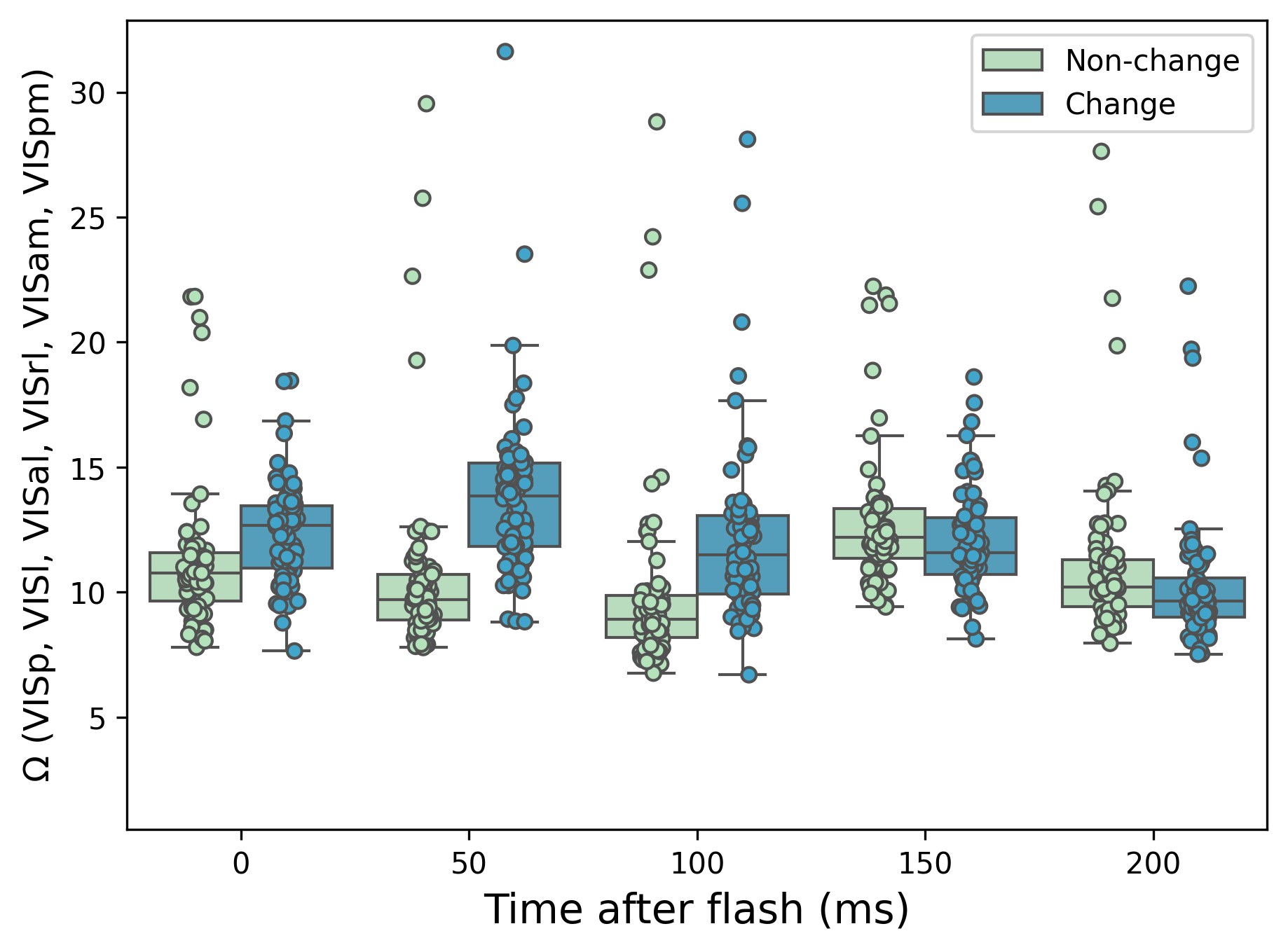}
         \caption{6 areas}
     \end{subfigure}
   
      \caption{\acrshort{O-information} estimate in the visual cortex region activity after two types of stimulus flash across 72 trial sessions. Top: Analysis using three brain region areas, Bottom: Extended analysis using six brain region areas. The step size is set to $2ms$ which results in 25 dimensional data for each bin per area. Different step sizes led to the same behavior (see \Cref{additional}).
      }
      \label{vbn}
\end{figure}

We show our results in \Cref{vbn}, where the distribution of \acrshort{O-information} values are reported as box-plots for each bin. We remark that values of \gls{O-information} are higher in cases of \textit{change} stimulus, and lower for the \textit{no change} stimulus.
This suggests that higher amount of redundant information in the visual cortex regions is transmitted in case of a flash with new scene. Interestingly, when considering six areas of the visual cortex, our observations remain valid, suggesting that the measured behaviour is common to these other brain areas as well. Our results are aligned with~\cite{venkatesh2023gaussian}. However, prior work rely on the \gls{PID} measure, which requires the brain regions to be artificially organized into two areas and a target variable, due to scalability issues affecting \gls{PID}. Our work confirms that \gls{SOI} does not have such a limitation and allows a single estimation procedure to obtain the same conclusions.

\balance

\section{Conclusion}\label{conclusions}
We addressed the problem of analyzing multivariate systems, whereby the essence of complexity does not only lie in the nature of the individual system components, but also in the structure of their inter-dependencies. Indeed, the analysis of high-order interaction among variables has emerged as an important tool to deepen our understanding of such complex systems, with application domains including machine learning, neuroscience, climate modeling, and many more.

Recently, the scientific community has spent considerable effort on extending information theory to allow the study of complex, multivariate systems according to notions of uniqueness, redundancy and synergy. While no consensus exists yet, on a information measure that can fully and reliably characterize high-order interactions, in this work we focused on \acrshort{O-information}, which has desirable properties such as interpretability and scalability in number of variables. The current state of the art is however at a roadblock. The existing techniques rely on strong assumptions on the data distribution. Additionally, we explore an exhaustive use of the neural \gls{MI} estimators to access the \gls{O-information} which resulted in sub-optimal performance and scalability issues. Then, the endeavour of our work was to present a method to lift such limitations, and endow practitioners and scientists with a flexible and reliable tool to study complex systems associated to natural phenomena.

In this paper, we proposed \gls{SOI}, a novel technique that leverages recent neural estimators of mutual information and uses score functions of joint and conditional distributions to compute divergences. We showed that \gls{SOI} can compute \acrshort{O-information} by training a unique parametric model, which is efficient and flexible. We validated our technique with a comprehensive experimental protocol, both in synthetic and realistic settings. We demonstrated that \gls{SOI} is accurate and robust across different system configurations and complexities. We also applied \gls{SOI} to a case study of mice brain activity, where we obtained plausible and interpretable results, and showcased the scalability of \gls{SOI} to handle larger systems than previously possible.
We believe that our work contributes to a substantial advancement of information measures computation and their applications to real-world, complex systems.

\balance

\section*{Acknowledgment} Pietro Michiardi was partially funded by project MUSE-COM$^2$ - AI-enabled MUltimodal SEmantic COMmunications and COMputing, in the Machine Learning-based Communication Systems, towards Wireless AI (WAI), Call 2022, ChistERA.

\section*{Impact Statement}
This paper presents work to improve current methods to compute information measures of complex systems, modeled as ensembles of multiple random variables.
Such information measures have been recently brought to the attention of the scientific community, for their potential in explaining the high-order interactions between systems part, and specifically to understand information redundancy, uniqueness and synergy. Applications of such measures range from multi-modal machine learning, neuroscience, climate modeling and many more.
There are many potential societal consequences of our work, none which we feel must be specifically highlighted here.

\bibliography{paper}

\begin{thebibliography}{52}
\providecommand{\natexlab}[1]{#1}
\providecommand{\url}[1]{\texttt{#1}}
\expandafter\ifx\csname urlstyle\endcsname\relax
  \providecommand{\doi}[1]{doi: #1}\else
  \providecommand{\doi}{doi: \begingroup \urlstyle{rm}\Url}\fi

\bibitem[Allen-Institute(2022)]{allen-inst}
Allen-Institute.
\newblock Visual behavior neuropixels dataset overview.
\newblock 2022.
\newblock URL \url{https://portal.brain-map.org/explore/circuits/visual-behavior-neuropixels}.

\bibitem[Ay et~al.(2019)Ay, Polani, and Virgo]{Ay2019InformationDB}
Ay, N., Polani, D., and Virgo, N.
\newblock Information decomposition based on cooperative game theory.
\newblock \emph{ArXiv}, abs/1910.05979, 2019.
\newblock URL \url{https://api.semanticscholar.org/CorpusID:204512236}.

\bibitem[Bai et~al.(2023)Bai, Cheng, Hao, Henao, and Carin]{bai2023estimating}
Bai, K., Cheng, P., Hao, W., Henao, R., and Carin, L.
\newblock Estimating total correlation with mutual information estimators.
\newblock In \emph{International Conference on Artificial Intelligence and Statistics}, pp.\  2147--2164. PMLR, 2023.

\bibitem[Barrett(2014)]{Adam14}
Barrett, A.~B.
\newblock An exploration of synergistic and redundant information sharing in static and dynamical gaussian systems.
\newblock \emph{CoRR}, abs/1411.2832, 2014.
\newblock URL \url{http://arxiv.org/abs/1411.2832}.

\bibitem[Belghazi et~al.(2018)Belghazi, Baratin, Rajeshwar, Ozair, Bengio, Courville, and Hjelm]{belghazi2018mine}
Belghazi, M.~I., Baratin, A., Rajeshwar, S., Ozair, S., Bengio, Y., Courville, A., and Hjelm, D.
\newblock Mutual information neural estimation.
\newblock In \emph{Proceedings of the 35th International Conference on Machine Learning}, 2018.

\bibitem[Bounoua et~al.(2024)Bounoua, Franzese, and Michiardi]{bounoua2023multimodal}
Bounoua, M., Franzese, G., and Michiardi, P.
\newblock Multi-modal latent diffusion.
\newblock \emph{Entropy}, 26\penalty0 (4), 2024.
\newblock ISSN 1099-4300.
\newblock \doi{10.3390/e26040320}.
\newblock URL \url{https://www.mdpi.com/1099-4300/26/4/320}.

\bibitem[Cheng et~al.(2020)Cheng, Hao, Dai, Liu, Gan, and Carin]{cheng2020}
Cheng, P., Hao, W., Dai, S., Liu, J., Gan, Z., and Carin, L.
\newblock Club: A contrastive log-ratio upper bound of mutual information.
\newblock In \emph{International conference on machine learning}, pp.\  1779--1788. PMLR, 2020.

\bibitem[Chiarion et~al.(2023)Chiarion, Sparacino, Antonacci, Faes, and Mesin]{Chiarion2023ConnectivityAI}
Chiarion, G., Sparacino, L., Antonacci, Y., Faes, L., and Mesin, L.
\newblock Connectivity analysis in eeg data: A tutorial review of the state of the art and emerging trends.
\newblock \emph{Bioengineering}, 10\penalty0 (3), 2023.
\newblock ISSN 2306-5354.
\newblock \doi{10.3390/bioengineering10030372}.
\newblock URL \url{https://www.mdpi.com/2306-5354/10/3/372}.

\bibitem[Collet \& Malrieu(2008)Collet and Malrieu]{collet2008logarithmic}
Collet, J.-F. and Malrieu, F.
\newblock Logarithmic sobolev inequalities for inhomogeneous markov semigroups.
\newblock \emph{ESAIM: Probability and Statistics}, 12:\penalty0 492--504, 2008.

\bibitem[Cover et~al.(1991)Cover, Thomas, et~al.]{cover1999elements}
Cover, T.~M., Thomas, J.~A., et~al.
\newblock Entropy, relative entropy and mutual information.
\newblock \emph{Elements of information theory}, 2\penalty0 (1):\penalty0 12--13, 1991.

\bibitem[Czy{\.z} et~al.(2023)Czy{\.z}, Grabowski, Vogt, Beerenwinkel, and Marx]{czyz2023beyond}
Czy{\.z}, P., Grabowski, F., Vogt, J.~E., Beerenwinkel, N., and Marx, A.
\newblock Beyond normal: On the evaluation of mutual information estimators.
\newblock \emph{Advances in Neural Information Processing Systems}, 2023.

\bibitem[Dosi \& Roventini(2019)Dosi and Roventini]{dosi2019more}
Dosi, G. and Roventini, A.
\newblock More is different... and complex! the case for agent-based macroeconomics.
\newblock \emph{Journal of Evolutionary Economics}, 29:\penalty0 1--37, 2019.

\bibitem[Ehrlich et~al.(2023)Ehrlich, Schick-Poland, Makkeh, Lanfermann, Wollstadt, and Wibral]{ehrlich2023partial}
Ehrlich, D.~A., Schick-Poland, K., Makkeh, A., Lanfermann, F., Wollstadt, P., and Wibral, M.
\newblock Partial information decomposition for continuous variables based on shared exclusions: Analytical formulation and estimation.
\newblock \emph{arXiv preprint arXiv:2311.06373}, 2023.

\bibitem[Finn \& Lizier(2020)Finn and Lizier]{Finn2019GeneralisedMO}
Finn, C. and Lizier, J.~T.
\newblock Generalised measures of multivariate information content.
\newblock \emph{Entropy}, 22\penalty0 (2), 2020.
\newblock ISSN 1099-4300.
\newblock \doi{10.3390/e22020216}.
\newblock URL \url{https://www.mdpi.com/1099-4300/22/2/216}.

\bibitem[Franzese et~al.(2023)Franzese, Rossi, Yang, Finamore, Rossi, Filippone, and Michiardi]{franzese2022}
Franzese, G., Rossi, S., Yang, L., Finamore, A., Rossi, D., Filippone, M., and Michiardi, P.
\newblock How much is enough? a study on diffusion times in score-based generative models.
\newblock \emph{Entropy}, 2023.

\bibitem[Franzese et~al.(2024)Franzese, BOUNOUA, and Michiardi]{franzese2023minde}
Franzese, G., BOUNOUA, M., and Michiardi, P.
\newblock {MINDE}: Mutual information neural diffusion estimation.
\newblock In \emph{The Twelfth International Conference on Learning Representations}, 2024.
\newblock URL \url{https://openreview.net/forum?id=0kWd8SJq8d}.

\bibitem[Ganmor et~al.(2011)Ganmor, Segev, and Schneidman]{ganmor2011sparse}
Ganmor, E., Segev, R., and Schneidman, E.
\newblock Sparse low-order interaction network underlies a highly correlated and learnable neural population code.
\newblock \emph{Proceedings of the National Academy of sciences}, 108\penalty0 (23):\penalty0 9679--9684, 2011.

\bibitem[Gat \& Tishby(1998)Gat and Tishby]{gat1998synergy}
Gat, I. and Tishby, N.
\newblock Synergy and redundancy among brain cells of behaving monkeys.
\newblock \emph{Advances in neural information processing systems}, 11, 1998.

\bibitem[Gutknecht et~al.(2023)Gutknecht, Makkeh, and Wibral]{gutknecht2023babel}
Gutknecht, A.~J., Makkeh, A., and Wibral, M.
\newblock From babel to boole: The logical organization of information decompositions.
\newblock \emph{ArXiv}, abs/2306.00734, 2023.

\bibitem[Ho et~al.(2020)Ho, Jain, and Abbeel]{ho2020}
Ho, J., Jain, A., and Abbeel, P.
\newblock Denoising diffusion probabilistic models.
\newblock In Larochelle, H., Ranzato, M., Hadsell, R., Balcan, M., and Lin, H. (eds.), \emph{Advances in Neural Information Processing Systems}, volume~33, pp.\  6840--6851. Curran Associates, Inc., 2020.

\bibitem[Huang et~al.(2021)Huang, Lim, and Courville]{huang2021variational}
Huang, C.-W., Lim, J.~H., and Courville, A.~C.
\newblock A variational perspective on diffusion-based generative models and score matching.
\newblock \emph{Advances in Neural Information Processing Systems}, 34:\penalty0 22863--22876, 2021.

\bibitem[Kingma \& Ba(2015)Kingma and Ba]{kingma2014adam}
Kingma, D. and Ba, J.
\newblock Adam: A method for stochastic optimization.
\newblock In \emph{International Conference on Learning Representations (ICLR)}, 2015.

\bibitem[Kolchinsky(2019)]{Kolchinsky19}
Kolchinsky, A.
\newblock A novel approach to multivariate redundancy and synergy.
\newblock \emph{CoRR}, abs/1908.08642, 2019.
\newblock URL \url{http://arxiv.org/abs/1908.08642}.

\bibitem[Kong et~al.(2024)Kong, Liu, Li, Yogatama, and Steeg]{kong2023interpretable}
Kong, X., Liu, O., Li, H., Yogatama, D., and Steeg, G.~V.
\newblock Interpretable diffusion via information decomposition.
\newblock In \emph{The Twelfth International Conference on Learning Representations}, 2024.
\newblock URL \url{https://openreview.net/forum?id=X6tNkN6ate}.

\bibitem[Latham \& Nirenberg(2005)Latham and Nirenberg]{latham2005synergy}
Latham, P.~E. and Nirenberg, S.
\newblock Synergy, redundancy, and independence in population codes, revisited.
\newblock \emph{Journal of Neuroscience}, 25\penalty0 (21):\penalty0 5195--5206, 2005.

\bibitem[MacKay(2003)]{mackay2003information}
MacKay, D.~J.
\newblock \emph{Information theory, inference and learning algorithms}.
\newblock Cambridge university press, 2003.

\bibitem[Makkeh et~al.(2021)Makkeh, Gutknecht, and Wibral]{makkeh2021introducing}
Makkeh, A., Gutknecht, A.~J., and Wibral, M.
\newblock Introducing a differentiable measure of pointwise shared information.
\newblock \emph{Physical Review E}, 103\penalty0 (3):\penalty0 032149, 2021.

\bibitem[Martinez~Mediano(2022)]{martinez2022integrated}
Martinez~Mediano, P.~A.
\newblock Integrated information theory in complex neural systems.
\newblock 2022.

\bibitem[Nguyen et~al.(2007)Nguyen, Wainwright, and Jordan]{nguyen2007neurips}
Nguyen, X., Wainwright, M.~J., and Jordan, M.
\newblock Estimating divergence functionals and the likelihood ratio by penalized convex risk minimization.
\newblock In \emph{Advances in Neural Information Processing Systems}, 2007.

\bibitem[Oord et~al.(2018)Oord, Li, and Vinyals]{oord2019representation}
Oord, A. v.~d., Li, Y., and Vinyals, O.
\newblock Representation learning with contrastive predictive coding.
\newblock \emph{Advances in neural information processing systems}, 2018.

\bibitem[Peebles \& Xie(2023)Peebles and Xie]{peebles2023scalable}
Peebles, W. and Xie, S.
\newblock Scalable diffusion models with transformers.
\newblock In \emph{Proceedings of the IEEE/CVF International Conference on Computer Vision}, pp.\  4195--4205, 2023.

\bibitem[Rosas et~al.(2019)Rosas, Mediano, Gastpar, and Jensen]{Rosas2019QuantifyingHI}
Rosas, F.~E., Mediano, P. A.~M., Gastpar, M., and Jensen, H.~J.
\newblock Quantifying high-order interdependencies via multivariate extensions of the mutual information.
\newblock \emph{Physical review. E}, 100 3-1:\penalty0 032305, 2019.
\newblock URL \url{https://api.semanticscholar.org/CorpusID:67855406}.

\bibitem[Rosas et~al.(2020)Rosas, Mediano, Rassouli, and Barrett]{Rosas2020AnOI}
Rosas, F.~E., Mediano, P. A.~M., Rassouli, B., and Barrett, A.
\newblock An operational information decomposition via synergistic disclosure.
\newblock \emph{Journal of Physics A: Mathematical and Theoretical}, 53, 2020.
\newblock URL \url{https://api.semanticscholar.org/CorpusID:210932609}.

\bibitem[Runge et~al.(2019)Runge, Bathiany, Bollt, Camps-Valls, Coumou, Deyle, Glymour, Kretschmer, Mahecha, Mu{\~n}oz-Mar{\'\i}, et~al.]{runge2019inferring}
Runge, J., Bathiany, S., Bollt, E., Camps-Valls, G., Coumou, D., Deyle, E., Glymour, C., Kretschmer, M., Mahecha, M.~D., Mu{\~n}oz-Mar{\'\i}, J., et~al.
\newblock Inferring causation from time series in earth system sciences.
\newblock \emph{Nature communications}, 10\penalty0 (1):\penalty0 2553, 2019.

\bibitem[Scagliarini et~al.(2021)Scagliarini, Marinazzo, Guo, Stramaglia, and Rosas]{Scagliarini2021QuantifyingHI}
Scagliarini, T., Marinazzo, D., Guo, Y., Stramaglia, S., and Rosas, F.~E.
\newblock Quantifying high-order interdependencies on individual patterns via the local o-information: Theory and applications to music analysis.
\newblock \emph{Physical Review Research}, 2021.
\newblock URL \url{https://api.semanticscholar.org/CorpusID:237303787}.

\bibitem[Scagliarini et~al.(2023)Scagliarini, Nuzzi, Antonacci, Faes, Rosas, Marinazzo, and Stramaglia]{Scagliarini23}
Scagliarini, T., Nuzzi, D., Antonacci, Y., Faes, L., Rosas, F., Marinazzo, D., and Stramaglia, S.
\newblock Gradients of o-information: Low-order descriptors of high-order dependencies.
\newblock \emph{Physical Review Research}, 5, 01 2023.
\newblock \doi{10.1103/PhysRevResearch.5.013025}.

\bibitem[Shannon(1948)]{shannon1948mathematical}
Shannon, C.~E.
\newblock A mathematical theory of communication.
\newblock \emph{The Bell system technical journal}, 27\penalty0 (3):\penalty0 379--423, 1948.

\bibitem[Song \& Ermon(2019)Song and Ermon]{song2019}
Song, Y. and Ermon, S.
\newblock Generative modeling by estimating gradients of the data distribution.
\newblock In Wallach, H., Larochelle, H., Beygelzimer, A., d'~Alch\'{e}-Buc, F., Fox, E., and Garnett, R. (eds.), \emph{Advances in Neural Information Processing Systems}, volume~32. Curran Associates, Inc., 2019.

\bibitem[Song \& Ermon(2020)Song and Ermon]{song2020}
Song, Y. and Ermon, S.
\newblock Improved techniques for training score-based generative models.
\newblock In Larochelle, H., Ranzato, M., Hadsell, R., Balcan, M., and Lin, H. (eds.), \emph{Advances in Neural Information Processing Systems}, volume~33, pp.\  12438--12448. Curran Associates, Inc., 2020.

\bibitem[Song et~al.(2021)Song, Sohl-Dickstein, Kingma, Kumar, Ermon, and Poole]{song2021a}
Song, Y., Sohl-Dickstein, J., Kingma, D.~P., Kumar, A., Ermon, S., and Poole, B.
\newblock Score-based generative modeling through stochastic differential equations.
\newblock In \emph{International Conference on Learning Representations}, 2021.

\bibitem[Sparacino et~al.(2023)Sparacino, Faes, Mijatovi{\'c}, Parla, Re, Miraglia, de~Ville~de Goyet, and Sparacia]{Sparacino2023StatisticalAT}
Sparacino, L., Faes, L., Mijatovi{\'c}, G., Parla, G., Re, V.~L., Miraglia, R., de~Ville~de Goyet, J., and Sparacia, G.
\newblock Statistical approaches to identify pairwise and high-order brain functional connectivity signatures on a single-subject basis.
\newblock \emph{Life}, 13, 2023.
\newblock URL \url{https://api.semanticscholar.org/CorpusID:264314627}.

\bibitem[Stramaglia et~al.(2021)Stramaglia, Scagliarini, Daniels, and Marinazzo]{Stramaglia_Sebastiano}
Stramaglia, S., Scagliarini, T., Daniels, B.~C., and Marinazzo, D.
\newblock Quantifying dynamical high-order interdependencies from the o-information: An application to neural spiking dynamics.
\newblock \emph{Frontiers in Physiology}, 11, 2021.
\newblock ISSN 1664-042X.
\newblock \doi{10.3389/fphys.2020.595736}.
\newblock URL \url{https://www.frontiersin.org/articles/10.3389/fphys.2020.595736}.

\bibitem[Sun(1975)]{sun1975linear}
Sun, T.
\newblock Linear dependence structure of the entropy space.
\newblock \emph{Inf Control}, 29\penalty0 (4):\penalty0 337--68, 1975.

\bibitem[{Sun Han}(1980)]{SUNHAN198026}
{Sun Han}, T.
\newblock Multiple mutual informations and multiple interactions in frequency data.
\newblock \emph{Information and Control}, 46\penalty0 (1):\penalty0 26--45, 1980.
\newblock ISSN 0019-9958.
\newblock \doi{https://doi.org/10.1016/S0019-9958(80)90478-7}.
\newblock URL \url{https://www.sciencedirect.com/science/article/pii/S0019995880904787}.

\bibitem[Tax et~al.(2017)Tax, Mediano, and Shanahan]{e19090474}
Tax, T.~M., Mediano, P.~A., and Shanahan, M.
\newblock The partial information decomposition of generative neural network models.
\newblock \emph{Entropy}, 19\penalty0 (9), 2017.
\newblock ISSN 1099-4300.
\newblock \doi{10.3390/e19090474}.
\newblock URL \url{https://www.mdpi.com/1099-4300/19/9/474}.

\bibitem[van Enk(2023)]{Enk2023PoolingPD}
van Enk, S.~J.
\newblock Pooling probability distributions and partial information decomposition.
\newblock \emph{Physical review. E}, 107 5-1:\penalty0 054133, 2023.
\newblock URL \url{https://api.semanticscholar.org/CorpusID:256615444}.

\bibitem[Varley et~al.(2022)Varley, Pope, Faskowitz, and Sporns]{Varley2022MultivariateIT}
Varley, T.~F., Pope, M., Faskowitz, J., and Sporns, O.
\newblock Multivariate information theory uncovers synergistic subsystems of the human cerebral cortex.
\newblock \emph{Communications Biology}, 6, 2022.
\newblock URL \url{https://api.semanticscholar.org/CorpusID:249642639}.

\bibitem[Varley et~al.(2023)Varley, Pope, Puxeddu, Faskowitz, and Sporns]{Varley2023PartialED}
Varley, T.~F., Pope, M., Puxeddu, M.~G., Faskowitz, J., and Sporns, O.
\newblock Partial entropy decomposition reveals higher-order information structures in human brain activity.
\newblock \emph{Proceedings of the National Academy of Sciences of the United States of America}, 120, 2023.
\newblock URL \url{https://api.semanticscholar.org/CorpusID:255825886}.

\bibitem[Venkatesh et~al.(2023)Venkatesh, Bennett, Gale, Ramirez, Heller, Durand, Olsen, and Mihalas]{venkatesh2023gaussian}
Venkatesh, P., Bennett, C., Gale, S., Ramirez, T.~K., Heller, G., Durand, S., Olsen, S.~R., and Mihalas, S.
\newblock Gaussian partial information decomposition: Bias correction and application to high-dimensional data.
\newblock In \emph{Thirty-seventh Conference on Neural Information Processing Systems}, 2023.
\newblock URL \url{https://openreview.net/forum?id=1PnSOKQKvq}.

\bibitem[Villani(2009)]{villani2009optimal}
Villani, C.
\newblock \emph{Optimal transport: old and new}, volume 338.
\newblock Springer, 2009.

\bibitem[Vincent(2011)]{vincent2011}
Vincent, P.
\newblock A connection between score matching and denoising autoencoders.
\newblock \emph{Neural Computation}, 23\penalty0 (7):\penalty0 1661--1674, 2011.

\bibitem[Williams \& Beer(2010)Williams and Beer]{williams2010nonnegative}
Williams, P.~L. and Beer, R.~D.
\newblock Nonnegative decomposition of multivariate information, 2010.

\end{thebibliography}
\bibliographystyle{icml2024}

\newpage
\appendix
\onecolumn
\label{apx:detail}

\section*{Score-based \acrshort{O-information} Estimation  --- Supplementary material}

\section{Proofs}

\label{proof1}

\subsection{Detailed proof of \Cref{prop:kl_est}}
\label{proof_prop1}
Here we provide  the full proof for \Cref{prop:kl_est} (to avoid unnecessary complications, we assume the 1-d case, the vector proof is identical). Starting from the equation :

$$ C= \int \frac{d p_t}{ d t}\log (\frac{p_t}{q_t} )+p_t \frac{d}{dt} \log (\frac{p_t}{q_t} ) dx dt $$

Concerning the first part of the integral:

$$ \int \frac{d p_t}{ d t}\log (\frac{p_t}{q_t} ) dx dt = \int \Delta(p_t) \log (\frac{p_t}{q_t} ) dx dt = \int p_t \Delta( \log (\frac{p_t}{q_t} ) ) dx dt,$$

Where the first equality is simply due to $\frac{d p_t}{ d t}=\Delta p_t$, and the second is obtained by properties of the adjoint of the $\Delta$ operator. In particular, we need to perform a double application of integration by parts, where we should remember that densities $p_t$, $q_t$ are equal to zero at infinite values of $x$ and that $ \Delta=\nabla \nabla $ .

Focusing on the second part of the integral:

$$\int p_t \frac{d}{dt} \log (\frac{p_t}{q_t} ) dx dt =\int p_t ( \frac{d\log p_t}{dt} - \frac{d\log q_t}{dt}) dx dt = \int p_t ( \frac{ \frac{d p_t}{dt} }{p_t} - \frac{ \frac{d q_t}{dt} }{q_t} ) dx dt $$

The first summand $p_t \frac{ \frac{d p_t}{dt} }{p_t}$ simplifies to $\frac{d p_t}{dt}$.

Since $ \int \frac{d p_t}{dt} dx dt = \int \frac{d }{dt}(\int p_t dx ) dt=\int \frac{d }{dt}(1 ) dt=0 $, this term is cancelled.

The second is transformed as :

$ p_t \frac{ \frac{d q_t}{dt} }{q_t}= \frac{p_t}{q_t} \frac{d q_t}{dt}= \frac{p_t}{q_t} \Delta q_t $ where again we leveraged $\frac{d q_t}{dt} =\Delta q_t $.

Consequently, we obtain:

$$ C= \int p_t \Delta \log (\frac{p_t}{q_t} ) - \frac{p_t}{q_t} \Delta q_t dx dt $$

We apply one step of integration by parts on both $\Delta$ operators and obtain :

$$ \int -\nabla p_t \nabla \log (\frac{p_t}{q_t} ) + \nabla( \frac{p_t}{q_t}) \nabla q_t dx dt $$

The remaining missing clarification in the sketch proof of \Cref{prop:kl_est} is that :

\begin{flalign*}  
&\nabla( \frac{p_t}{q_t}) \nabla (q_t) = \frac{\nabla(p_t) q_t-\nabla(q_t) p_t }{q^2_t } \nabla (q_t)= 
\\&\frac{\nabla (p_t)}{q_t}\nabla (q_t)- p_t (\frac{\nabla q_t}{q_t} )^2=\nabla p_t \nabla(\log(q_t)) - p_t (\nabla(\log q_t))^2=
\\&p_t \nabla (\log p_t) \nabla(\log(q_t)) - p_t (\nabla(\log q_t))^2= p_t \nabla (\log q_t) ( \nabla (\log p_t) -\nabla (\log q_t) )=p_t \nabla (\log q_t) ( \nabla(\log \frac{p_t}{q_t}) ) 
\end{flalign*}

\subsection{\gls{TC} and \gls{DTC} equivalences}
We here prove the equivalences about \gls{TC} and \gls{DTC}. Starting from \gls{TC} :

\begin{flalign*}
\sum_{i=1}^{N} \cH(X^i)- \cH(X)=\sum_{i=1}^{N} \cH(X^i)- \sum_{i=1}^{N} \cH(X^i \g X^{>i})=\sum\limits_{i=1}^{N-1}\cI(X^i;X^{>i}) = \cT(X) 
\end{flalign*}

Concerning \gls{DTC}
\begin{flalign*}
& \cH(X) - \sum_{i=1}^{N} \cH(X^i|X^{\setminus i})=\cH(X^1)+\cH(X^{\setminus 1}\g X^1)-\cH(X^1|X^{\setminus 1})  - \sum_{i=2}^{N} \cH(X^i|X^{\setminus i})=\\&\cI(X^1;X^{\setminus 1})+\cH(X^{\setminus 1}\g X^1)  - \sum_{i=2}^{N} \cH(X^i|X^{\setminus i})=\\&\cI(X^1;X^{\setminus 1})+\cH(X^2\g X^1)+\cH(X^{\setminus 1,2}\g X^1,X^2)-\cH(X^2|X^{\setminus 2})- \sum_{i=3}^{N} \cH(X^i|X^{\setminus i})=\\&
\cI(X^1;X^{\setminus 1})+\cI(X^2;X^{>2}|X^{1})+\cH(X^{\setminus 1,2}\g X^1,X^2)- \sum_{i=3}^{N} \cH(X^i|X^{\setminus i})=\dots\\&\sum\limits_{i=1}^{N-1} \cI(X^i;X^{>i}\g X^{<i})=\cD( X)
\end{flalign*}

Where for the last equality it suffices to consider trivial reordering arguments, $\sum\limits_{i=2}^{N} \cI(X^i;X^{<i}\g X^{>i})=\sum\limits_{i=1}^{N-1} \cI(X^i;X^{>i}\g X^{<i})$.

\section{Details of \gls{SOI}}
\label{apdx:detail}
In the section we provide additional implementation details about \gls{SOI}.

\subsection{Computing \gls{O-information}}

In \Cref{sec:sde_est}, we presented how \gls{TC} and \gls{DTC} can be estimated using denoising score functions. Our estimators requires different score functions which can be obtained by learning different denoisers. More particularly, \gls{TC}  requires the joint denoiser $ \E[X\g X_t]$ and the marginals $ \E[X^i\g X^i_t]$ for $i \in \{1,\dots,N \} $. \gls{DTC} estimation is obtained using the joint and the following conditional terms $ \E[X^i \g X^i_t,X^{\setminus i} ] $ for $i \in \{1,\dots,N \} $. 
Our formulation in \Cref{sec:sde_est} is general and can be applied to a wide range of denoising score learning techniques. For the implementation of \gls{SOI}, we adopt VP-\gls{SDE} framework \cite{song2019}. The latter  perturbs the data using an \gls{SDE} parameterized by a drift $f_t$ and a diffusion coefficient $g_t$.

\paragraph{Muti-variate denoising score network.}

 We extend the work from \cite{bounoua2023multimodal} to amortize the learning of all the required terms using a \textbf{unique} denoising score network.  The denoising score network $\epsilon_\theta$ accepts as input the concatenation of the variables each perturbed at different times. The second input is a vector of size $N$ which describes the state of each variable and allows a parametrization of different denoising score functions.

The joint term corresponds to the case where all the variables are perturbed with the same intensity $t$ and all the elements of the vector $\tau= [t,\dots,t]$ are set equivalently to $t$.
The conditional terms correspond to the case where only the conditioned variable $i$ is perturbed with intensity $t$ whereas the remaining conditioning variables $\setminus i_{\text{th}} $ are kept unperturbed at $t=0$. Consequently the parameter describing this case is of the form $[0,\dots, t,\dots,0]$. 

While \cite{bounoua2023multimodal} framework is not able to learn the marginal denoising score, it's possible via an additional parameterization to include this configuration.  This corresponds to the case where the marginal variable $i$ is perturbed with intensity $t$ while all the other variables are made uninformative. The non marginal variables $\setminus i_{\text{th}} $ are replaced with pure noise corresponding to a maximal perturbation  at $t=T$.   Consequently the parameter describing this case is of the form $[T,\dots, t,\dots,T]$. 

\paragraph{Training.}
 \label{train}
The training is carried out through a randomized procedure. At each training step, we select randomly a set of the denoising score functions required for the \gls{O-information} estimation (joint, conditional or marginals). These denoising scores function are learned by the unique network following Algorithm \ref{algo:soi_training}. In total, estimating \gls{O-information} requires calling  $2N +1$ denoising score functions which we learn using a unique denoising network.

\begin{algorithm}[H]
\DontPrintSemicolon
\SetAlgoLined
\SetNoFillComment
\LinesNotNumbered

\caption{\gls{SOI} Training step}
\KwData{ $ X = \{X^i\}_{i=1}^N $ } 
$t \sim \mathcal{U}[0,T]$  \tcp*{Importance sampling schemes \citep{huang2021variational,song2021a} can be adopted to reduce variance}
\uIf{ Joint }{
\hfill \\
$X_t \sim p_t$  \tcp*{ Obtain noisy version of all the variables using \acrshort{VPSDE} \cite{song2019} with drift $f_t$ and  diffusion coefficient $g_t$.} 
$ s_t(X_t) = \epsilon_\theta([X_t^1,\dots, X_t^N ], \tau = [t,\dots,t,\dots,t] ) $ 
\hfill \\
\textbf{Return} $ \nabla_\theta  \norm{  s_t(X_t) - \nabla \log{p_{t}(X_t|X)} }$  \tcp*{Denoising score matching of all the variables} 
}
\uIf{Conditional }{ 
\hfill \\
$X^i_t \sim p_t $   \tcp*{Obtain noisy version of the variable $i$ while the remaining variables are kept unperturbed at ($t=0$)}  
$ s_t(X^i_t |X^{\setminus i} ) = \epsilon_\theta \left([X^1,\dots,X^{i-1},X^i_t,X^{i+1},\dots, X^N ], \tau = [0,\dots,t,\dots,0] \right) $ 
\hfill \\

\textbf{Return} $ \nabla_\theta \norm{  s_t(X^i_t|X^{\setminus i} ) - \nabla \log{p_{t}(X^i_t|X^i)} }$ \tcp*{Denoising score matching of the conditioning variable $i$ } 
}
\uIf{Marginal }{

\hfill \\
$X^i_t \sim p_t$   \\
$X^{\setminus i}_T \gets p_T = \mathcal{N}(0, \mathbb{I})$  \tcp*{Obtain noisy version of the variable $i$ while the remaining variables are replaced with pure noise ($t=T$).}
$ s_t(X^i_t) = \epsilon_\theta([X_T^1,\dots,X_T^{i-1},X_t^i,X_T^{i+1},\dots, X_T^N ], \tau = [T,\dots,t,\dots,T] ) $ 

\hfill \\
\textbf{Return} $ \nabla_\theta \norm{  s_t(X^i_t) - \nabla \log{p_{t}(X^i_t|X^i)} }$  \tcp*{Denoising score matching of the marginal variable $i$ } 
}
\label{algo:soi_training}
\end{algorithm}

\paragraph{Inference.}

Once all the denoising score functions are learned, it's possible to estimate \gls{TC} and \gls{DTC} via a Monte Carlo estimation of the integral over $t$ in \Cref{prop:Test} and \Cref{prop:Dest} . 
The outer integration w.r.t. to the time instant is  possible by sampling $t\sim\mathcal{U}(0,T)$, and then using the estimation $\int_0^T(\cdot) \dd t=T\E_{t\sim\mathcal{U}(0,T)}[(\cdot)]$. In practice we adopt 10 steps for the  computation of the expectation. The procedure to estimate \gls{O-information} is described in  algorithm \ref{soi:inference}. First,  samples from $x \sim p(x) $ are considered, then sampling the time $t \sim \mathcal{U}[0,T]$. A perturbed version of the variables $X_t$ is computed using the \gls{VPSDE}.
The joint, conditional and marginal denoising scores are computed leveraging the unique denoising score network. This is possible by choosing different perturbation times and manipulating the vector $\tau$ as described earlier. Computing the difference of the denoising scores functions (see \Cref{prop:Test} and \Cref{prop:Test} ) allows the computation of \gls{TC} and \gls{DTC} respectively. Please note that it is possible to implement importance sampling schemes to reduce the variance, along the lines of what described by \citet{huang2021variational}.

\begin{algorithm}[H]
\DontPrintSemicolon
\SetAlgoLined
\SetNoFillComment
\LinesNotNumbered 
\caption{\gls{SOI} inference time }\label{soi:inference}
\KwData{ $ X = \{X^i\}_{i=1}^N  $ } 
$t \sim \mathcal{U}[0,T]$  \tcp*{Importance sampling scheme can also be adopted}
$X_t \sim p_t$   \tcp*{ Obtain the noisy version of all the variables using \acrshort{VPSDE} \cite{song2019} with drift $f_t$ and  diffusion coefficient $g_t$.} 
\hfill \\
$ s_t(X_t) \gets \epsilon_\theta([X_t^1,\dots, X_t^N ], \tau = [t,\dots,t,\dots,t] ) $ \tcp{Compute the joint score}
\hfill \\
\For{i = 1 \textbf{to} N \tcp{Compute the conditional and marginal terms} }{
\hfill \\
$ s_t(X^i_t | X^{\setminus i} ) \gets \epsilon_\theta \left([X^1,\dots,X^{i-1},X^i_t,X^{i+1},\dots, X^N ], \tau = [0,\dots,t,\dots,0] \right) $ 
\hfill \\

$ s_t(X^i_t) \gets \epsilon_\theta([X_T^1,\dots,X_T^{i-1},X_t^i,X_T^{i+1},\dots, X_T^N ], \tau = [T,\dots,t,\dots,T] ) $\tcp*{Similarly to Algorithm \ref{algo:soi_training} the non marginal variables are replaced with pure noise $X_T^{\setminus i} \sim \mathcal{N}(0,\mathbb{I} ) $
}
}
\hfill \\ 
$ \hat{\cT}(X) \gets \frac{g^2_t}{2} \norm{ s_t(X_t) - \left[ s_t(X^i_t) \right]_{i=1}^{N} }^2  $  \tcp{See \Cref{tc_est}}
$ \hat{\cD}(X) \gets \frac{g^2_t}{2} \norm{ s_t(X_t) - \left[ s_t(X^i_t | X^{\setminus i} ) \right]_{i=1}^{N} }^2  $ 
\tcp{See \Cref{dtc_est}}
$ \hat{\Omega}(X) \gets \hat{\cT}(X)  - \hat{\cD}(X) $
\hfill \\ 
\textbf{Return} $\hat{\Omega}(X)$
\end{algorithm}

\subsection{Computing gradient of \gls{O-information}}
\label{apdx_grad}
To compute the gradient of \gls{O-information}
recall that $\partial_i \Omega( X) =  \Omega( X) - \Omega( X^{\setminus i})$. The first order gradient of \gls{O-information} requires the estimation of \gls{O-information} of all the subsystems of size $N-1$. 

\begin{align}
    & \Omega( X^{\setminus i}) = \cT(X^{\setminus i}) - \cD(X^{\setminus i\textbf{}}) \\
    & = \sum_{j=1, j\neq i}^{N} \cH(X^j) - \cH(X^{\setminus i}) \\
    & - ( \cH(X^{\setminus i}) - \sum_{j=1, j \neq i }^{N} \cH(X^j|X^{\setminus \{i,j\} }) ) \label{grad_eq}    
\end{align}

It's possible to use an alternative formulation to estimate the gradient of \gls{O-information} based on \gls{MI} terms:

\begin{align}
    & \partial_i \Omega( X) = (2-N) \cI(X^i,X^{\setminus i}) + \sum_{j=1, j \neq i }^{N} \cI(X^i,X^{\setminus \{i,j\} })\\
    & = (2-N)  \left[ \cH(X^i) - \cH(X^i|X^{\setminus i}) \right] + \sum_{j=1, j \neq i }^{N} \cH(X^i) - \cH(X^i|X^{\setminus \{i,j\} })
   \label{grad_eq_2}    
\end{align}

Many denoising score functions in \Cref{grad_eq} were also used to estimate the global \gls{O-information}. To learn the additional necessary terms to compute $\Omega( X^{\setminus i}) $, the randomized set of scores adopted during the training step (see \Cref{train}) is extended to account for the new requirements. Please note that we still use a unique denoising network that considers all the terms necessary to compute \gls{O-information} and its gradient. A large number of learned denoising score functions is a potential reason for the bias observed in our experiment \Cref{grad_oinf}. A highly flexible architecture capable of fitting large number of scores may be needed to infer gradient of \gls{O-information}.

\section{Experimental settings }
\label{apdx:exp}
\subsection{Canonical multivariate Gaussian system}

In this section we provide additional details about the construction of the synthetic benchmark \Cref{syntheticgaussian}. 
\paragraph{Redundancy benchmark.}

All the variable of the system are composed of a redundant component and unique information specific to each variable.

We modulate the redundant inter-dependency strength by setting different values for $\sigma $. We consider a standardized system where all the variables mean is $0$ and standard deviation equal to $\mathbb{I}$. This results in the following covariance matrix:
\begin{equation}
    \begin{bmatrix}
         \mathbb{I}           & \rho  \mathbb{I}       & \vdots & \rho  \mathbb{I} \\
        \rho   \mathbb{I} &       \mathbb{I} & \dots &\rho \mathbb{I} \\
        \vdots                & \vdots          & \ddots           & \rho  \mathbb{I}   \\
        \rho \mathbb{I} & \rho  \mathbb{I}     & \dots &  \mathbb{I} \\
    \end{bmatrix}
\end{equation}

With $\rho = \frac{1}{1+\sigma^2}$ which modulates the interactions strength in the system.

\paragraph{Synergy benchmark.}
We consider a standardized system where all the variables mean is $0$ and standard deviation equal to $\mathbb{I}$. This 
results in the following covariance matrix :
\begin{equation}
    \begin{bmatrix}
        \mathbb{I}           & \frac{1}{\sqrt{N-1}} \mathbb{I} &  0  & \dots & 0  \\
        \frac{1}{\sqrt{N-1}} \mathbb{I} &  \mathbb{I} &  \frac{\rho}{\sqrt{N-1}} \mathbb{I}& \dots    & \frac{\rho}{\sqrt{N-1}} \mathbb{I}\\
        0  &  \frac{\rho}{\sqrt{N-1}} \mathbb{I} &  \mathbb{I} & \dots              &        0  \\
       0   & \vdots & 0  & \ddots   &   0 \\
         0   & \frac{\rho}{\sqrt{N-1}} & 0 &\dots &   \mathbb{I}  \\
    \end{bmatrix}
\end{equation}
Where $\rho = \frac{1}{\sqrt{1+\sigma^2}}$   modulates the interactions strength in the system.

\paragraph{Mixed benchmark.}

The covariance matrix is easy to obtain as the mixed benchmark is made of independent subsystems.

\paragraph{Ground Truth.}

Having access to the covariance matrix of the system, computing entropy in close form for Gaussian distribution is possible. For $X \sim \mathcal{N}(\mu,\sigma)$ : 

\begin{equation}
    \mathcal{H} (X) = \frac{1}{2}  \log (2 \pi \sigma^2)  + \frac{1}{2} \ 
\end{equation}

For a multivariate Gaussian distribution $X^d \sim \mathcal{N}_d(\mu, \Sigma )$ :

\begin{equation}
    \mathcal{H} (X) = \frac{D}{2}  (1+ \log(2\pi) ) + \frac{1}{2} \log det(\Sigma) \ 
\end{equation}

\subsection{\gls{SOI} implementation details }

We provide code-base for \gls{SOI} implementation at \footnote{\url{https://github.com/MustaphaBounoua/soi} }. The training of \gls{SOI} is carried out using \textit{Adam optimizer} \citep{kingma2014adam}. We use Exponential moving average (EMA) with a momentum parameter $m = 0.999$. Importance sampling \cite{huang2021variational} (\footnote{\url{https://github.com/CW-Huang/sdeflow-light}}) at train and test-time. The hyper-parameters are presented in \Cref{table:soi}. To estimate the gradient of \gls{O-information} (\Cref{grad_oinf}) the model width is double the one presented in \Cref{table:soi} to account for the additional necessary terms to learn. Concerning the experiments in \Cref{vbn} , we use the same architecture used for the canonical examples and follow the same procedure to choose the model capacity( see \Cref{table:soi} for the hyper-parameters details).

\renewcommand{\tabcolsep}{2.0pt}

\begin{table}[H]
\caption{\gls{SOI} network training details. $Dim$ of the task correspond the sum of the dimensions of all variables of the system. For the neural data application we report the number of training iterations (.,.) corresponding the "change" case and "No change" case. The number of iteration used for the "No change" is higher since the dataset contains more "no change" flashes compared to "change" flashes.}
\centering
\begin{tabular}{ccccccccc}
\toprule
  & Width &Time embed & Batch size & Lr & Iterations & Number of params  \\
\midrule 
($Dim \leq 50$)  & 128 &128& 256 &  1e-2 & 195k& 320k \\
($Dim \leq 100$)  & 192 & 192 & 256 &  1e-2 & 195k& 747k\\ 
($Dim \ge 100$)  & 256 & 256 & 256 &  1e-2 & 
195k &1003k \\
\midrule
Neural application  &  &&  &  & &   \\
($Dim \leq 30$) & 128 &128& 256 &  1e-2 & (100k,160k)& 320k \\
($Dim \leq 75$) & 192 &192& 256 &  1e-2 & (100k,160k)& 737k \\
($Dim \leq 150$) & 256 &256& 256 &  1e-2 & (100k,160k)& 1300k \\
($Dim \ge 150$) & 384 &384& 256 &  1e-2 & (100k,160k)& 3000k \\
\bottomrule
\end{tabular}
\label{table:soi}
\end{table}

\subsection{Baselines}

\cite{bai2023estimating} decomposes \gls{TC} into $N-1$ \gls{MI} terms which are estimated using pairwise neural \gls{MI} estimator. Similarly by leveraging \Cref{eq:base} \gls{DTC} can also be retrieved by estimating $N-1$ additional \gls{MI} terms.

\begin{align}
    & \cT(X) =\sum\limits_{i=1}^{N-1} \cI(X^i;X^{>i}) \\
    & \cD(X) = \cS(X) - \cT(X) = \sum\limits_{i=1}^{N} \cI(X^i;X^{\setminus i}) -\cT(X)     \\ \label{eq:base}
    & \cD(X) = \sum\limits_{i=2}^{N} \cI(X^i;X^{\setminus i}) - \sum\limits_{i=2}^{N-1} \cI(X^i;X^{>i}) 
\end{align}

Our implementation in based on the the official codebase \footnote{\url{https://github.com/Linear95/TC-estimation}} of \cite{bai2023estimating}.
We use the same architecture and hyper parameters from \cite{bai2023estimating}: $\text{LR} =1e-3 $, $\text{Batch size} =64 $. We use an \gls{MLP} architecture for all the variants of the baseline with 3 linear layers with varying width. For each \gls{MI} term, the capacity of the neural network is aligned to the input dimension. \textit{Adam optimizer} \citep{kingma2014adam} is used for training. We increase the width of the hidden layer to accommodate the data dimension. For the variant of the baseline implemented with \acrshort{MINE}, we used smaller layer size as large capacity led to divergence during training. To ensure the best performance, we train each \gls{MI} estimator model for $80k$ steps for a number of variables $N=10$ and $40k$ for number of variables $N=6$. In the different experiments, we reported  the performance results averaged over 5 seeds and dropped the baseline in case of divergence during training.

\paragraph{Limitations of the baseline in computing gradients of \gls{O-information}}

It’s possible to leverage the decomposition of \cite{bai2023estimating}, using the  compact gradient of O-information formulation \Cref{grad_oinf_tx}.

This will require $N$ \gls{MI} term for each  $\partial_i \Omega(X)$. Consequently to compute all the terms, it’s required to train $N * N$   pairwise MI models.  While it's possible to leverage some \gls{MI} terms, if already estimated for the computation of O-information, the overall complexity remains of order $\mathcal{O}(N^2)$.

This naturally raises a scalability problem in training a large number of neural estimator models. Moreover, as the number of MI terms increases, this approach is likely to suffer from cumulative errors observed when estimating O-information.

To compute the gradient of O-information with \gls{SOI}, we are instead required to approximate an additional number of denoising score functions. However, our method \gls{SOI} amortizes the training costs : we use a unique score network to approximate all the required score functions. 

\subsection{The Visual Behavior Neuropixels}

Hereafter we describe the different pre-processing steps applied on the Visual Behavior Neuropixels in \Cref{sec:vbn}. We follow  the same procedure described by \cite{venkatesh2023gaussian}. The selected mice are  the ones with both
familiar and novel sessions and a minimum number of 20 units in each of the six brain regions: \textsc{VISp}, \textsc{VISl}, \textsc{VISal} ,\textsc{VISrl}, \textsc{VISam} and \textsc{VISpm}. Only the units of good quality are kept. The selection criteria was based on an SNR at least 1, and with fewer than 1 inter-spike interval violations. The non-change flashes correspond to the ones where the image does not change and happen between 4 and 10 flashes after the trial start. Trials corresponding to a change are naturally the ones when the image has changed. Only flashes that occurred while the animal was engaged ( based on the reward information)  is kept, while 
the ones corresponding to an omission, or after an omission, and flashes during which the animal licked, were all removed. 

\begin{figure} [H]

\centering
     \begin{subfigure}{0.35\textwidth}
         \centering

         \includegraphics[page=1,width=\linewidth]{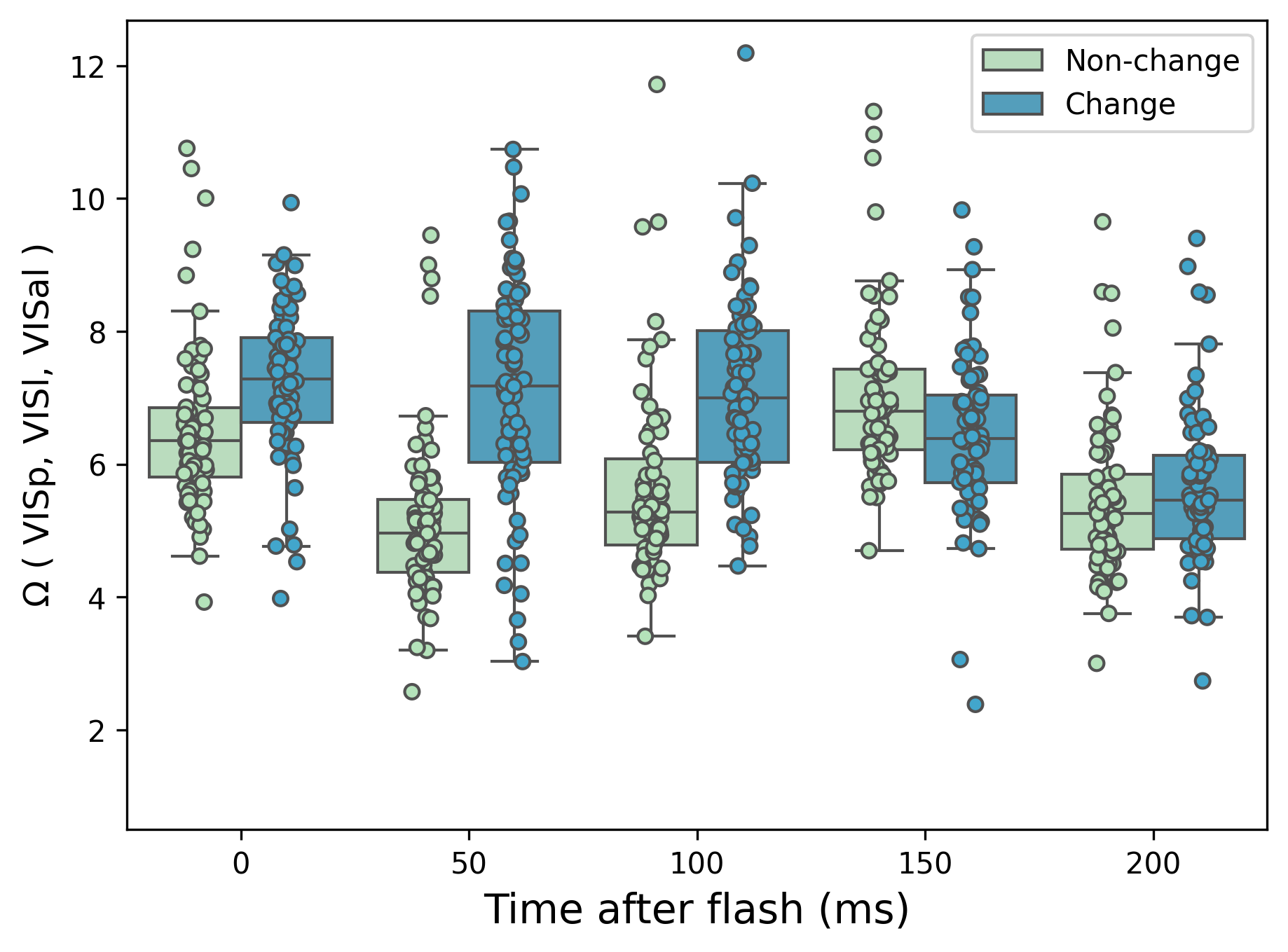}
         \caption{3 areas}
     \end{subfigure}
      \begin{subfigure}{0.35\textwidth}
         \centering
 \includegraphics[page=1,width=\linewidth]{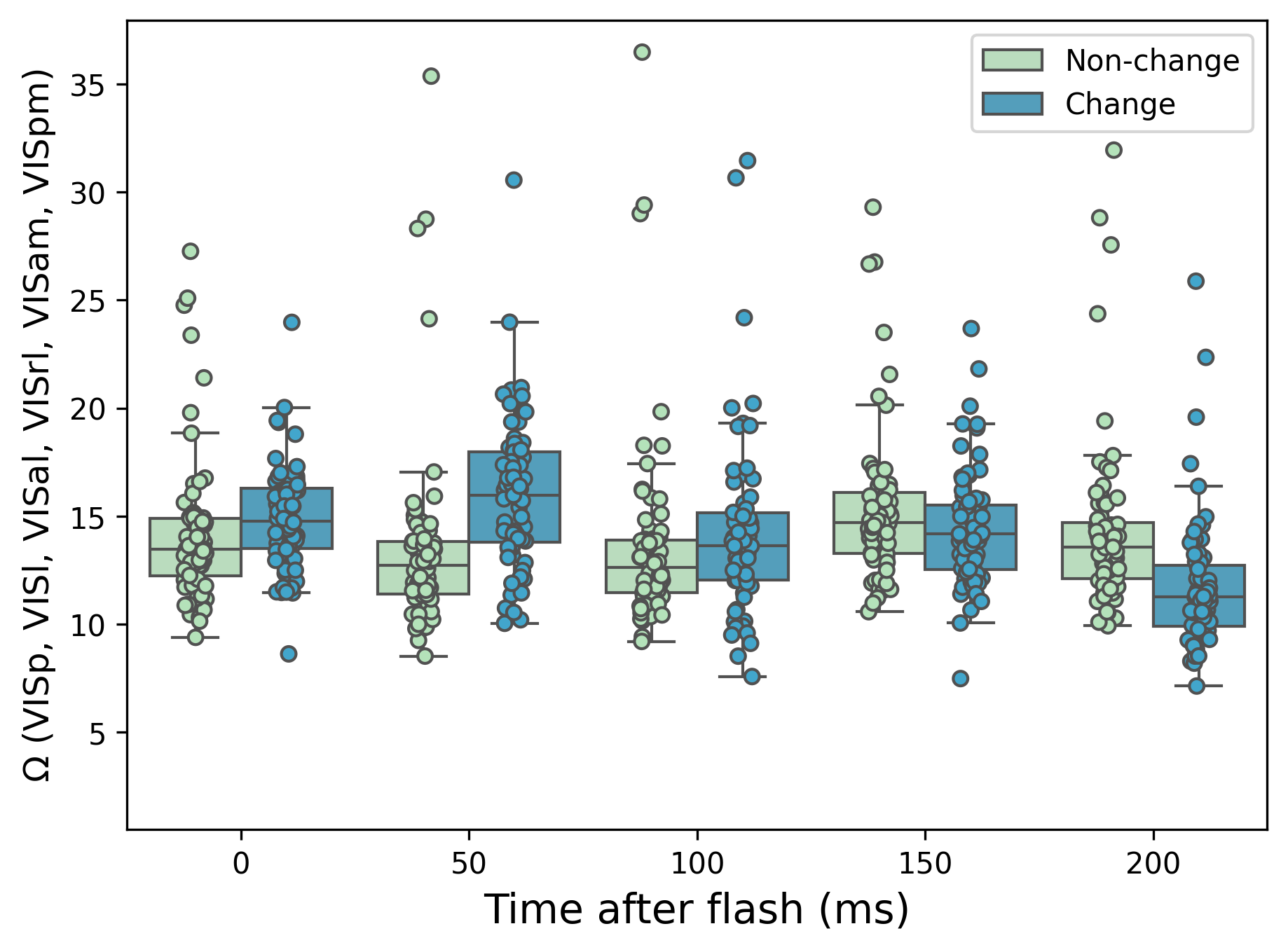}
         \caption{6 areas}
     \end{subfigure}

      \begin{subfigure}{0.35\textwidth}
         \centering

         \includegraphics[page=1,width=\linewidth]{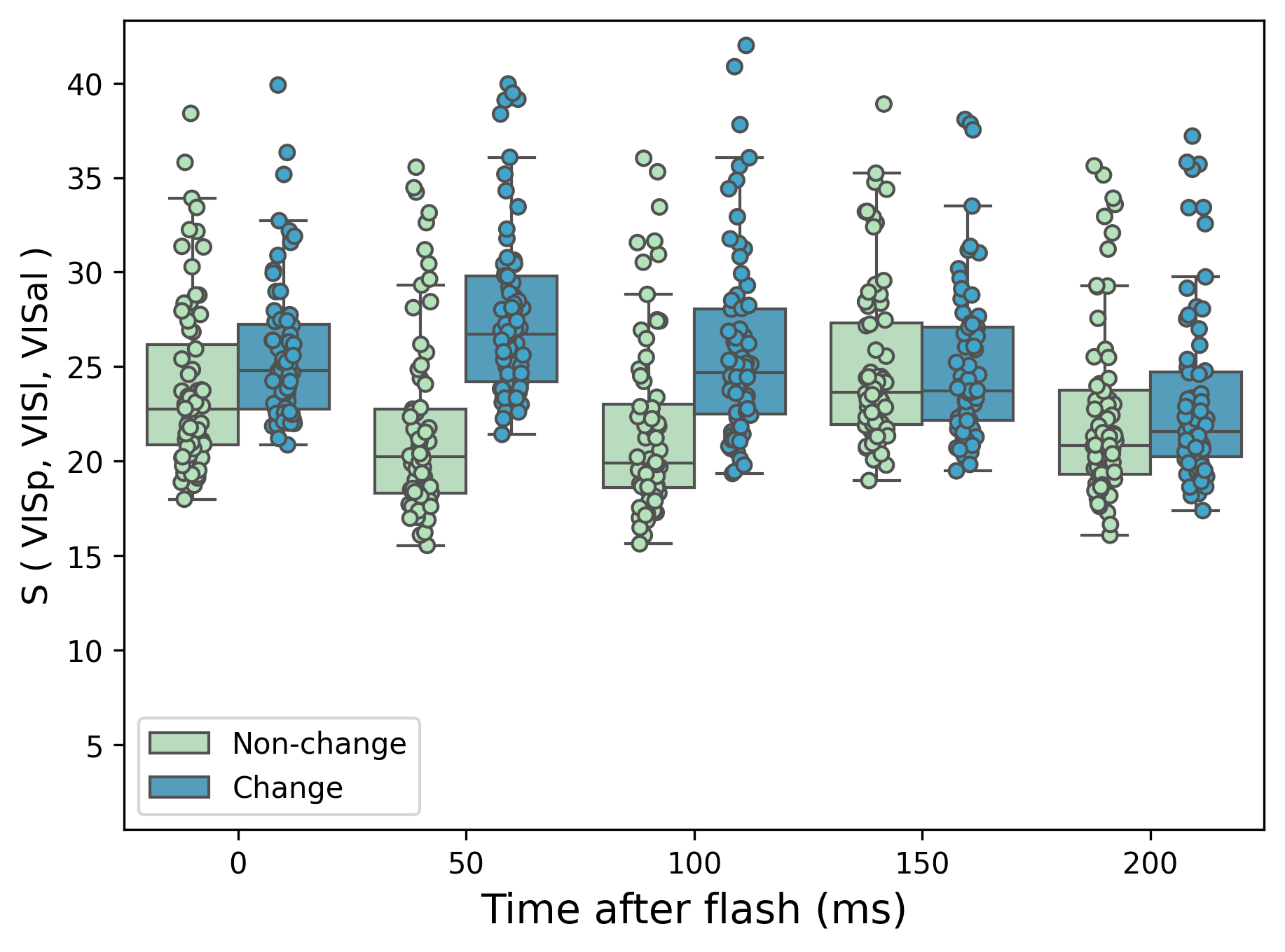}
         \caption{3 areas}
     \end{subfigure}
      \begin{subfigure}{0.35\textwidth}
         \centering
 \includegraphics[page=1,width=\linewidth]{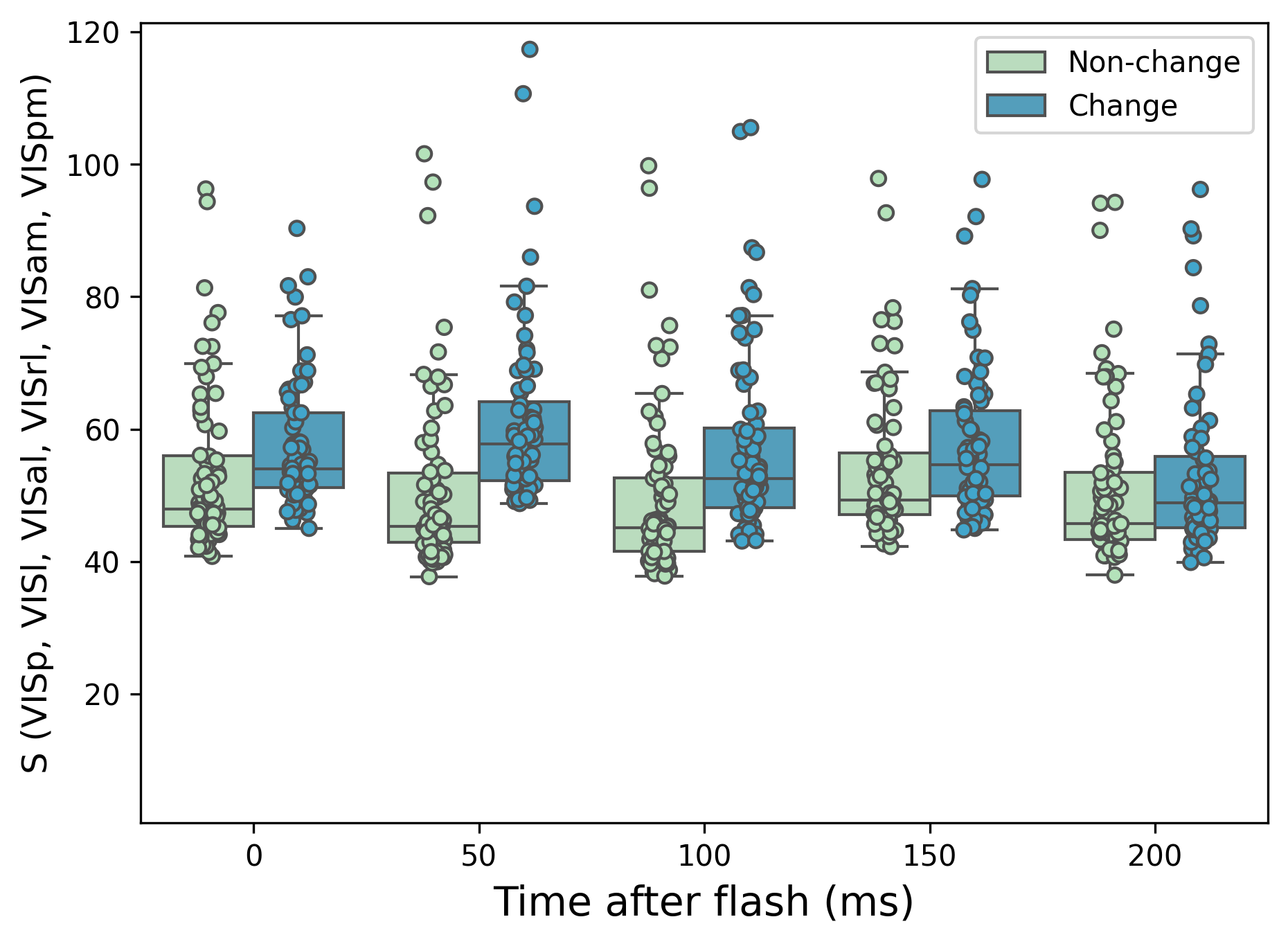}
         \caption{6 areas}
     \end{subfigure}
      \caption{\acrshort{O-information} and \acrshort{S-information} estimate in the visual cortex region activity after two types of stimulus flash across 72 trial sessions. \textbf{Left}: Analysis using three brain region areas, \textbf{Right}: Extended analysis using six brain region areas. The step size is set to $1ms$ which results in \textbf{50} dimensional data for each bin per area.
      }
      \label{vbn_50}
\end{figure}

\begin{figure} [h]

\centering
      \begin{subfigure}{0.35\textwidth}
         \centering
         \includegraphics[page=1,width=\linewidth]{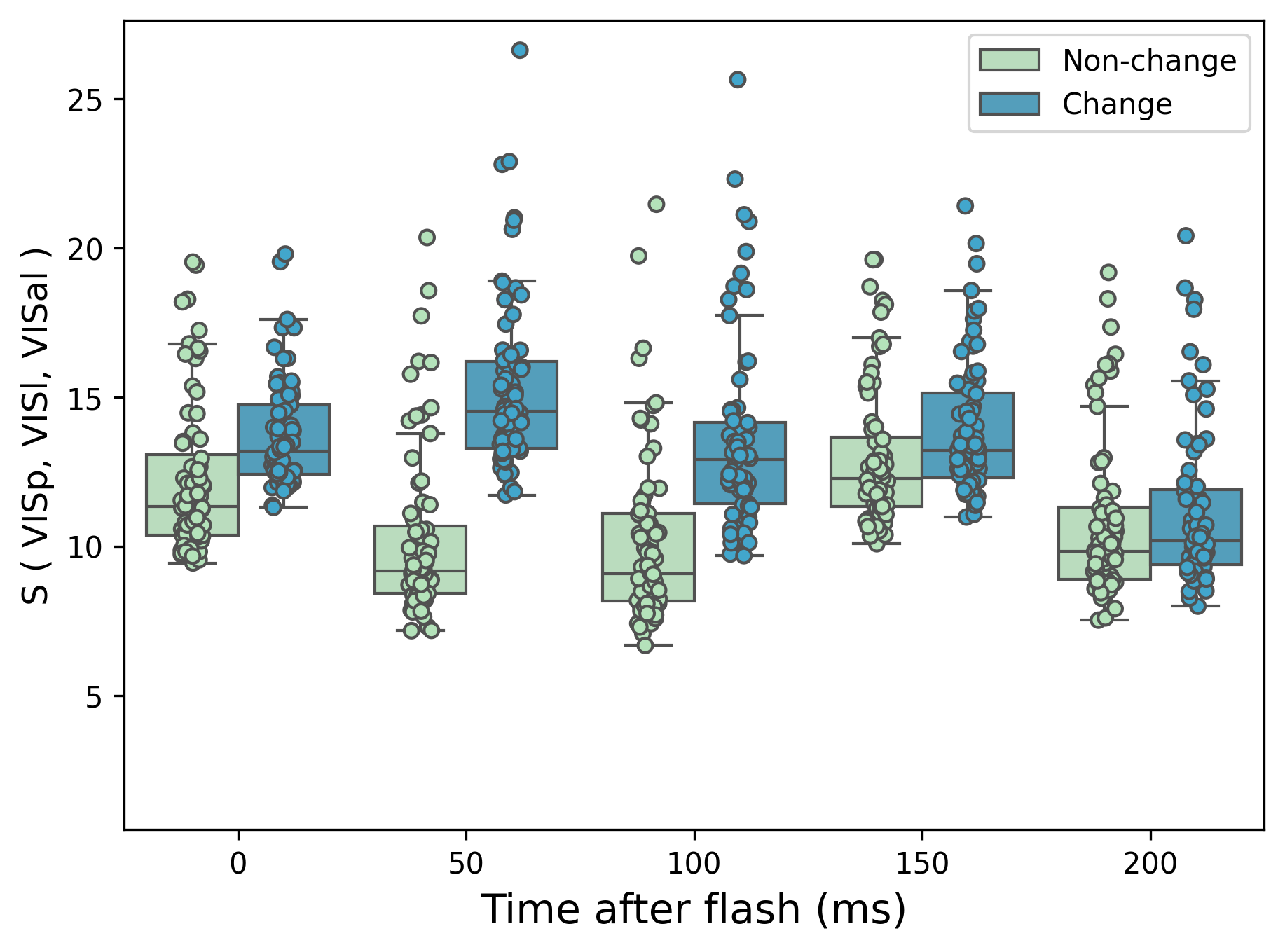}
         \caption{3 areas}
     \end{subfigure}
      \begin{subfigure}{0.35\textwidth}
         \centering
 \includegraphics[page=1,width=\linewidth]{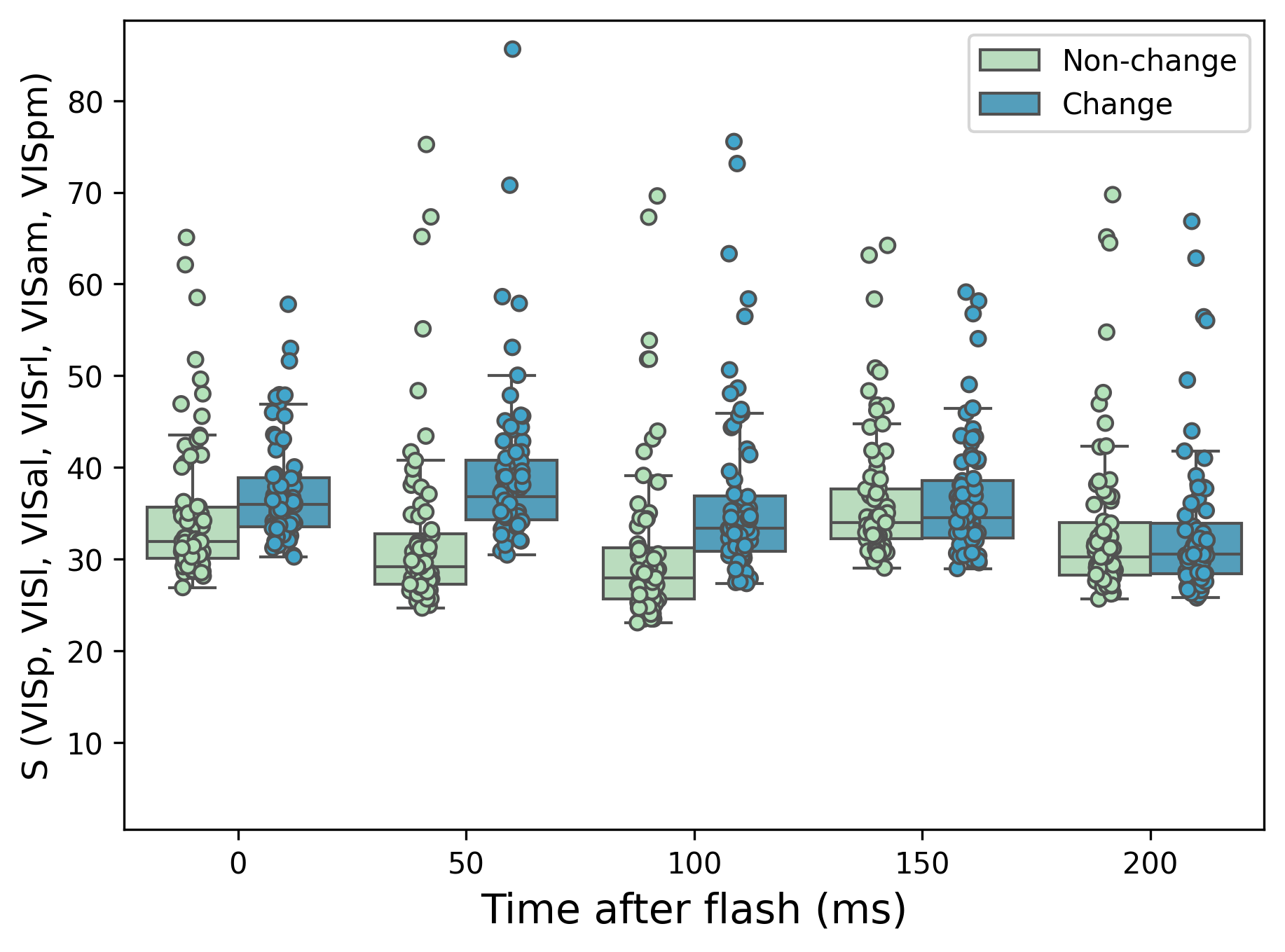}
         \caption{6 areas}
     \end{subfigure}
      \caption{\acrshort{S-information} estimate in the visual cortex region activity after two types of stimulus flash across 72 trial sessions. \textbf{Left}: Analysis using three brain region areas, \textbf{Right}: Extended analysis using six brain region areas. The step size is set to $2ms$ which results in \textbf{25} dimensional data for each bin per area.
      }
      \label{vbn_25}
\end{figure}

\begin{figure} [h]

\centering
     \begin{subfigure}{0.35\textwidth}
         \centering
         \includegraphics[page=1,width=\linewidth]{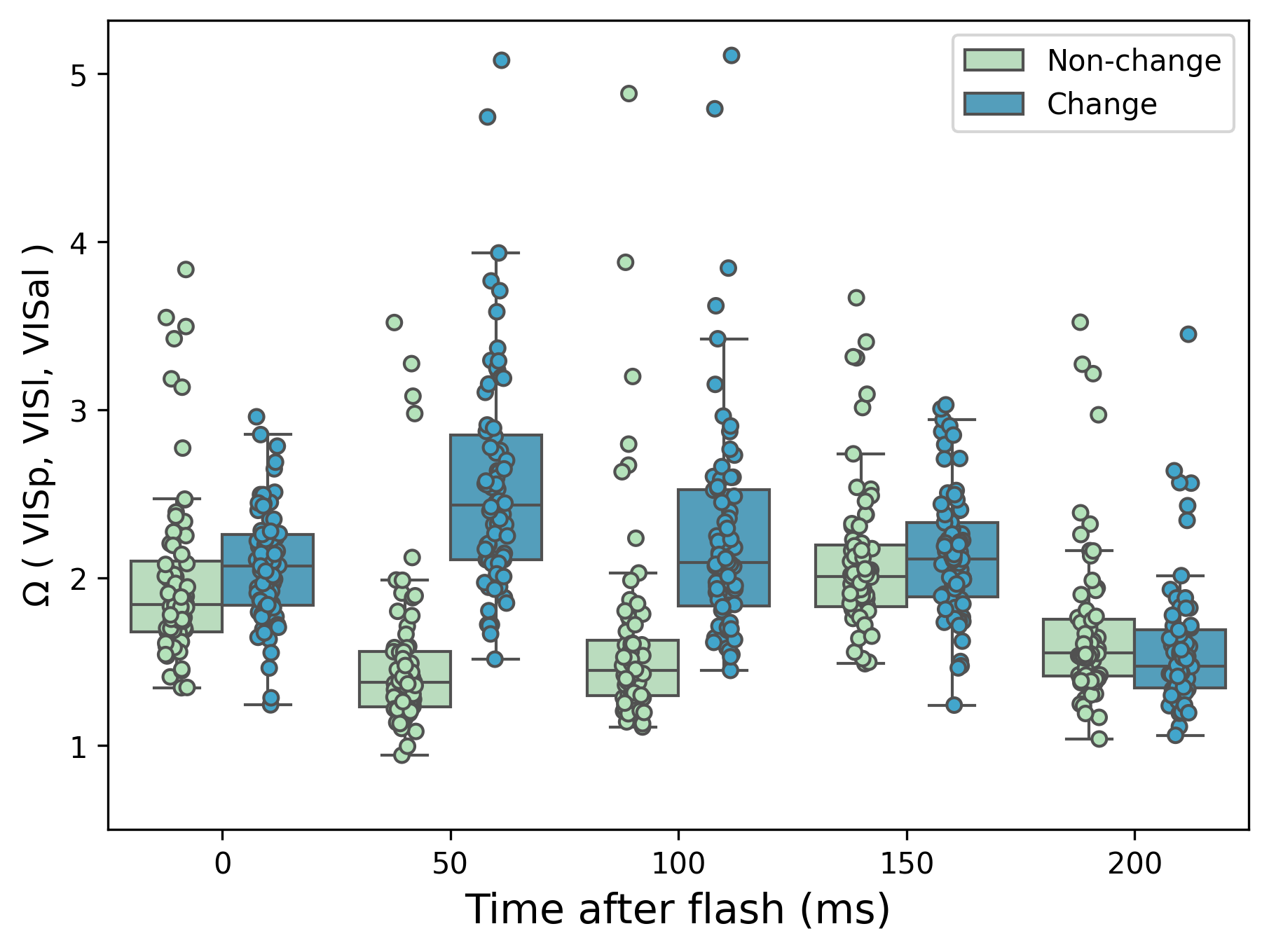}
         \caption{3 areas}
     \end{subfigure}
      \begin{subfigure}{0.35\textwidth}
         \centering
 \includegraphics[page=1,width=\linewidth]{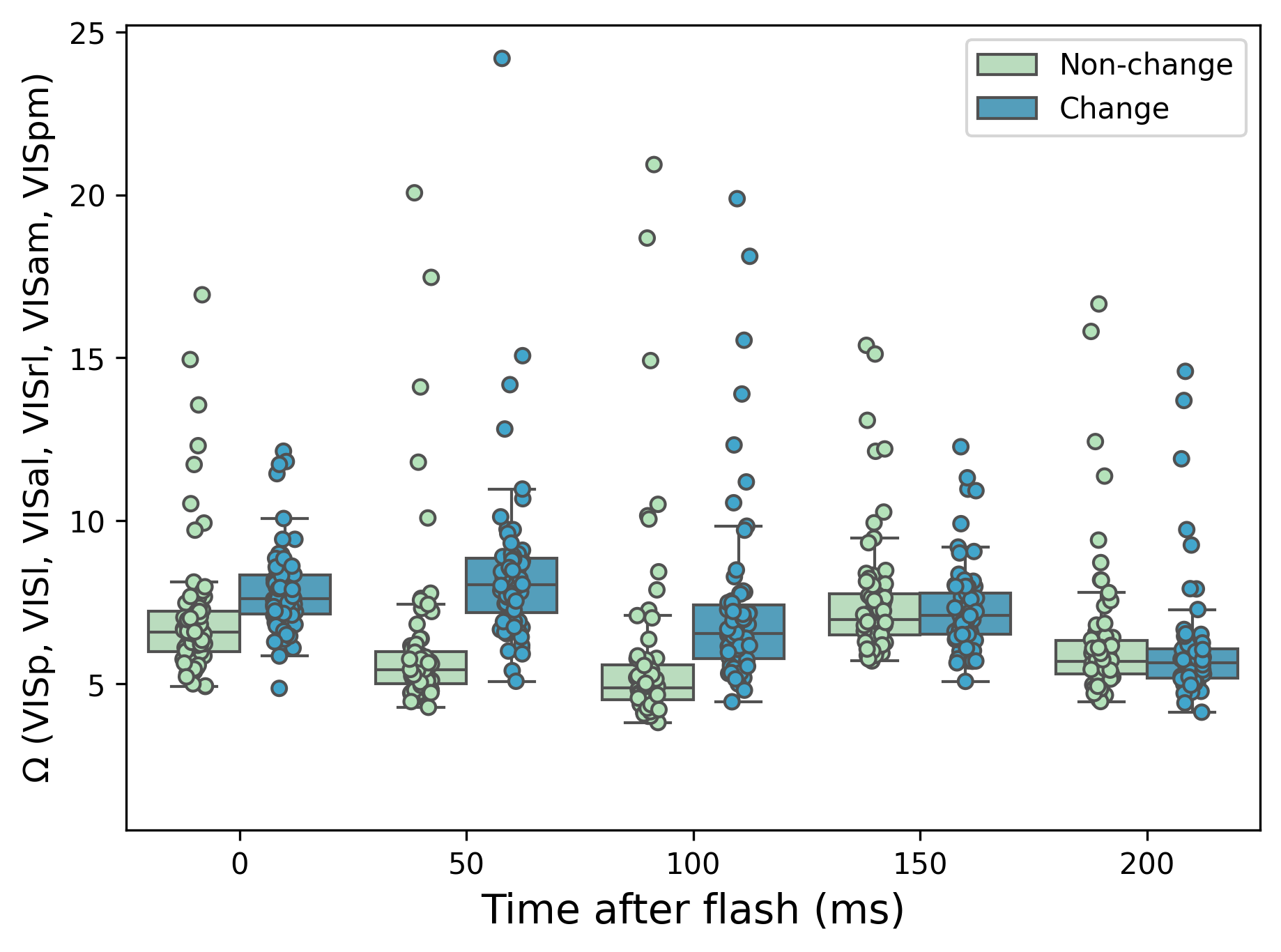}
         \caption{6 areas}
     \end{subfigure}

      \begin{subfigure}{0.35\textwidth}
         \centering
         \includegraphics[page=1,width=\linewidth]{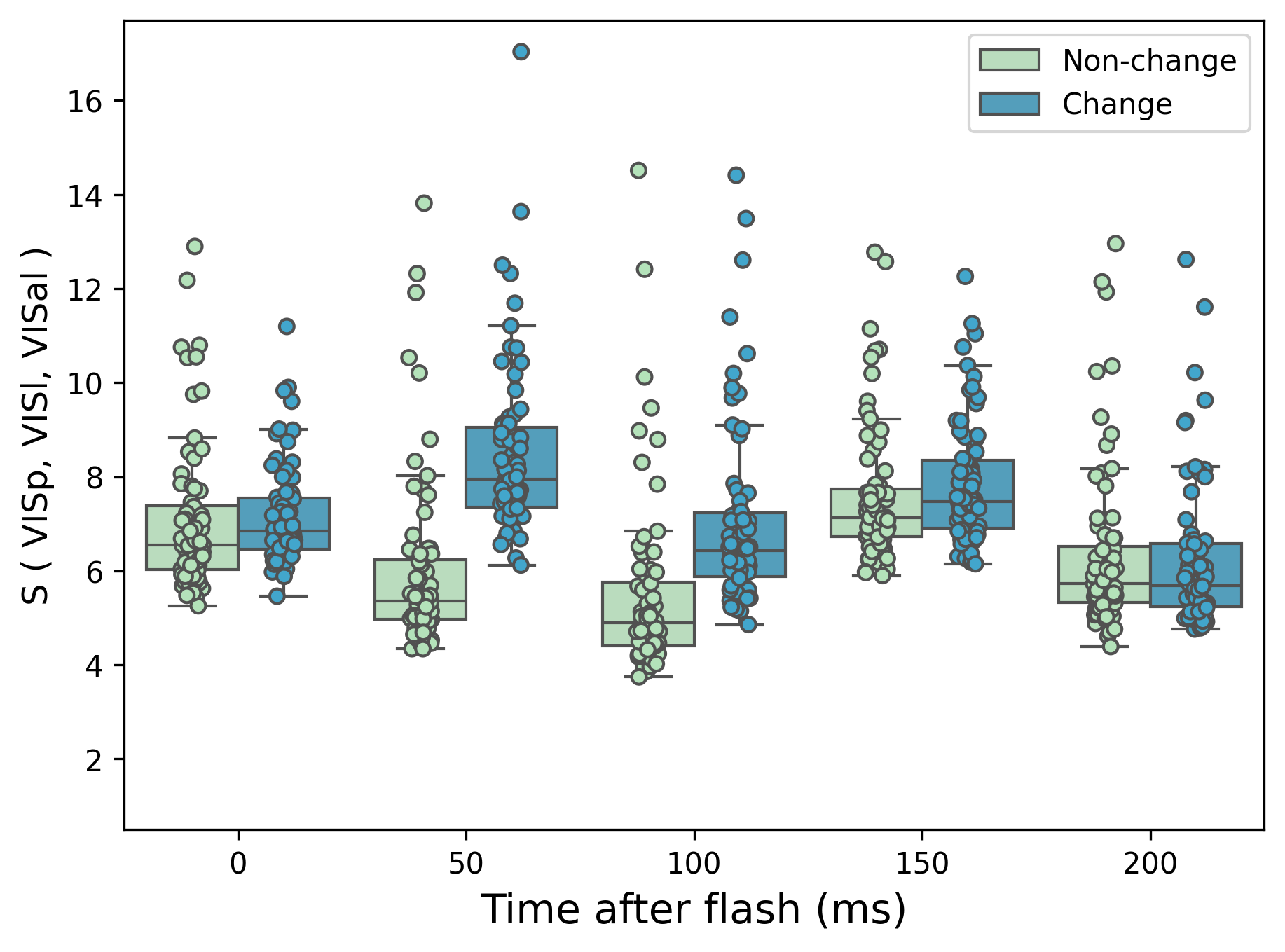}
         \caption{3 areas}
     \end{subfigure}
      \begin{subfigure}{0.35\textwidth}
         \centering
 \includegraphics[page=1,width=\linewidth]{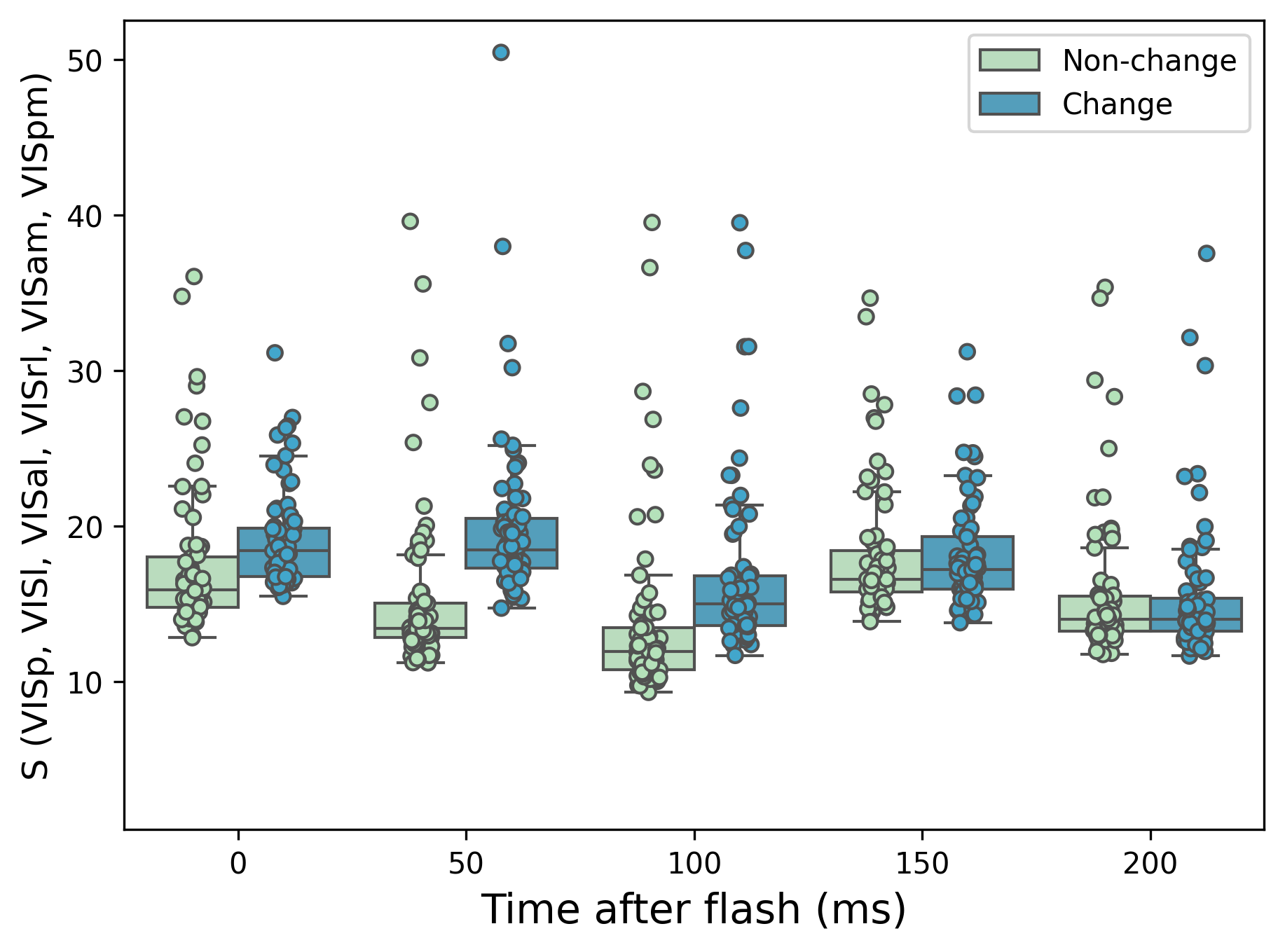}
         \caption{6 areas}
     \end{subfigure}
      \caption{\acrshort{O-information} and \acrshort{S-information} estimate in the visual cortex region activity after two types of stimulus flash across 72 trial sessions. \textbf{Left}: Analysis using three brain region areas, \textbf{Right}: Extended analysis using six brain region areas. The step size is set to $5ms$ which results in \textbf{10} dimensional data for each bin per area.
      }
      \label{vbn_10}
\end{figure}

The trials were aligned to the start of each stimulus flash, and the 250ms recordings were divided into 5 bins of 50 ms duration averaged over the units of the same region. We use different step sizes to count the spikes which resulted in different dimensional representation but resulted in the same intuition (See \Cref{vbn_10},\Cref{vbn_25} and \Cref{vbn_50}). Please note that unlike \cite{venkatesh2023gaussian}, we don't use PCA to reduce the dimension of the data, and count the number of spikes per unit by averaging the activity over the units of the same region indexed by time.

\section{A transformer based \gls{SOI}}
\label{apdx:tx_exp}
Throughout our experimental campaign as referenced in \Cref{experiment}, we employed an \gls{MLP} structure enhanced with skip connections. While this setup reliably estimated \gls{O-information}, it produced perfectible gradient of \gls{O-information} estimation. We address this shortcoming by integrating a more robust architecture capable of scaling with an increased number of denoising score functions. Our approach is based on the latest developments in denoising score matching, incorporating a transformer-based model.

Our method is simple: we adopt the architecture from \cite{peebles2023scalable} to learn the denoising score functions, treating each modality as a distinct token, while substituting any non-marginal modality with a NULL token (a token with zero value). A transformer block is employed to learn the conditional signal, which is subsequently merged with the temporal signal. This conditioning employs the adaLN-Zero configuration. Our model consists of 4 Blocks, each with 6 attention heads, and the width of the transformer's linear layers is scaled according to the dimension size of the benchmark.The training follows a randomized approach akin to that detailed in \Cref{apx:detail} eliminating the need for a multi-time vector. To compute gradient of \gls{O-information}, we utilize the formulation presented in \Cref{grad_eq_2}.

\begin{figure} [h]

     \begin{subfigure}{0.22\textwidth}
         \centering
         \includegraphics[page=1,width=\linewidth]{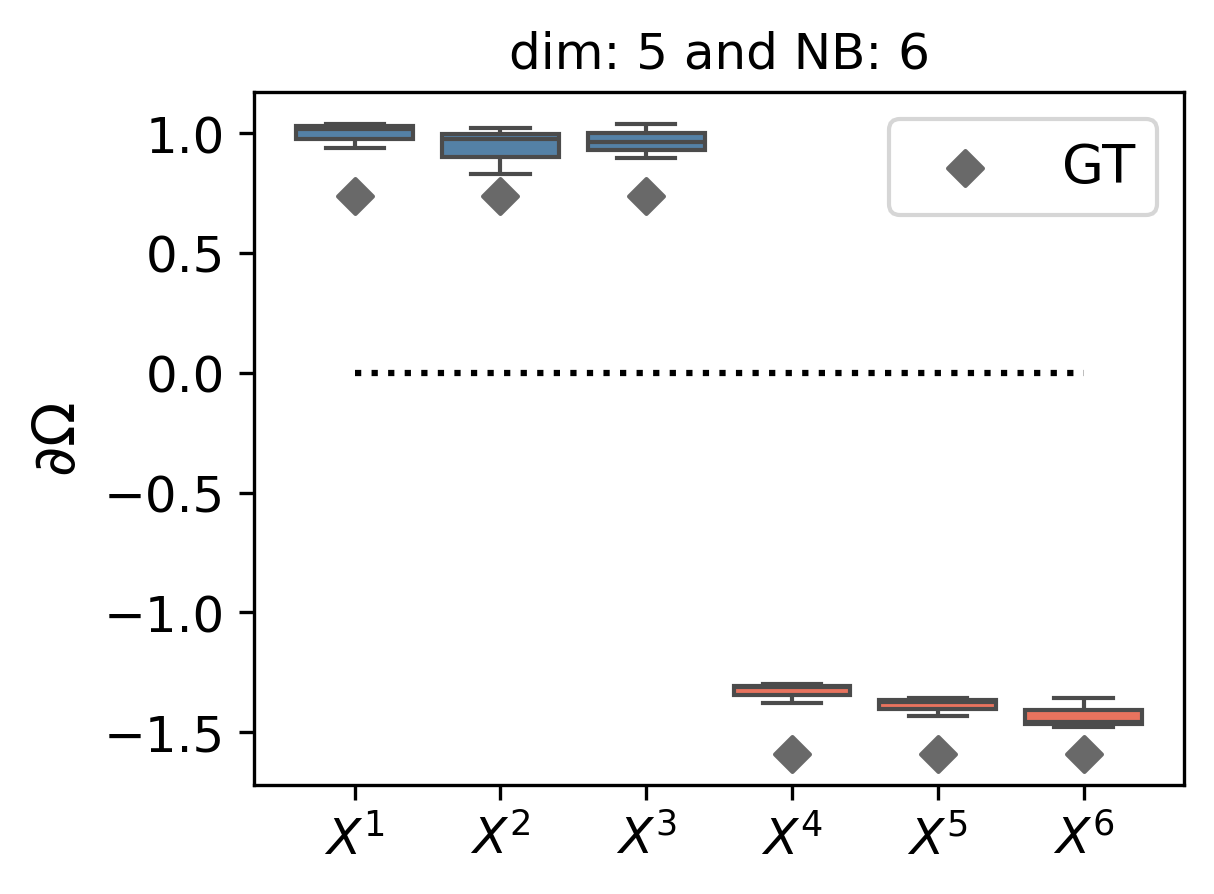}
         \caption{Dim = 5}
     \end{subfigure}
      \begin{subfigure}{0.22\textwidth}
         \centering
         \includegraphics[page=1,width=\linewidth]{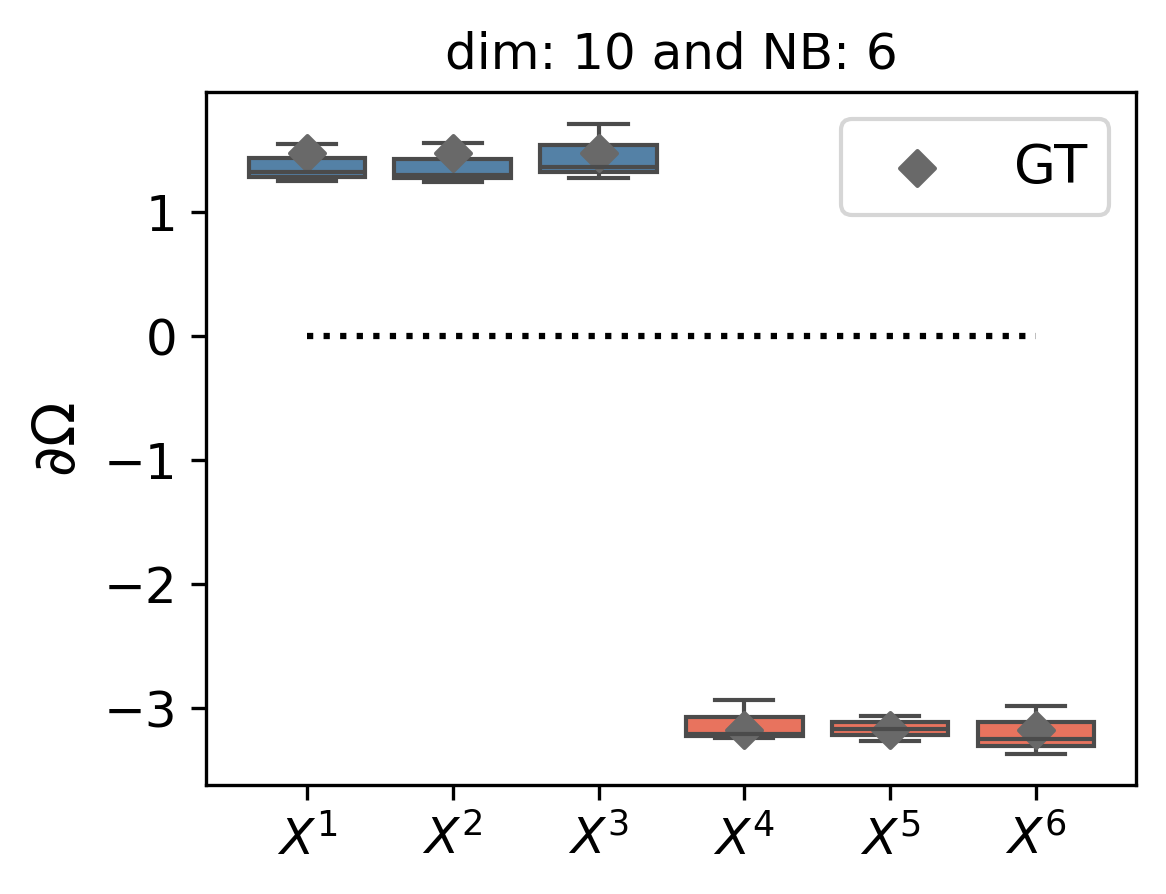}
         \caption{Dim = 10}
     \end{subfigure}
   \begin{subfigure}{0.22\textwidth}
         \centering
         \includegraphics[page=1,width=\linewidth]{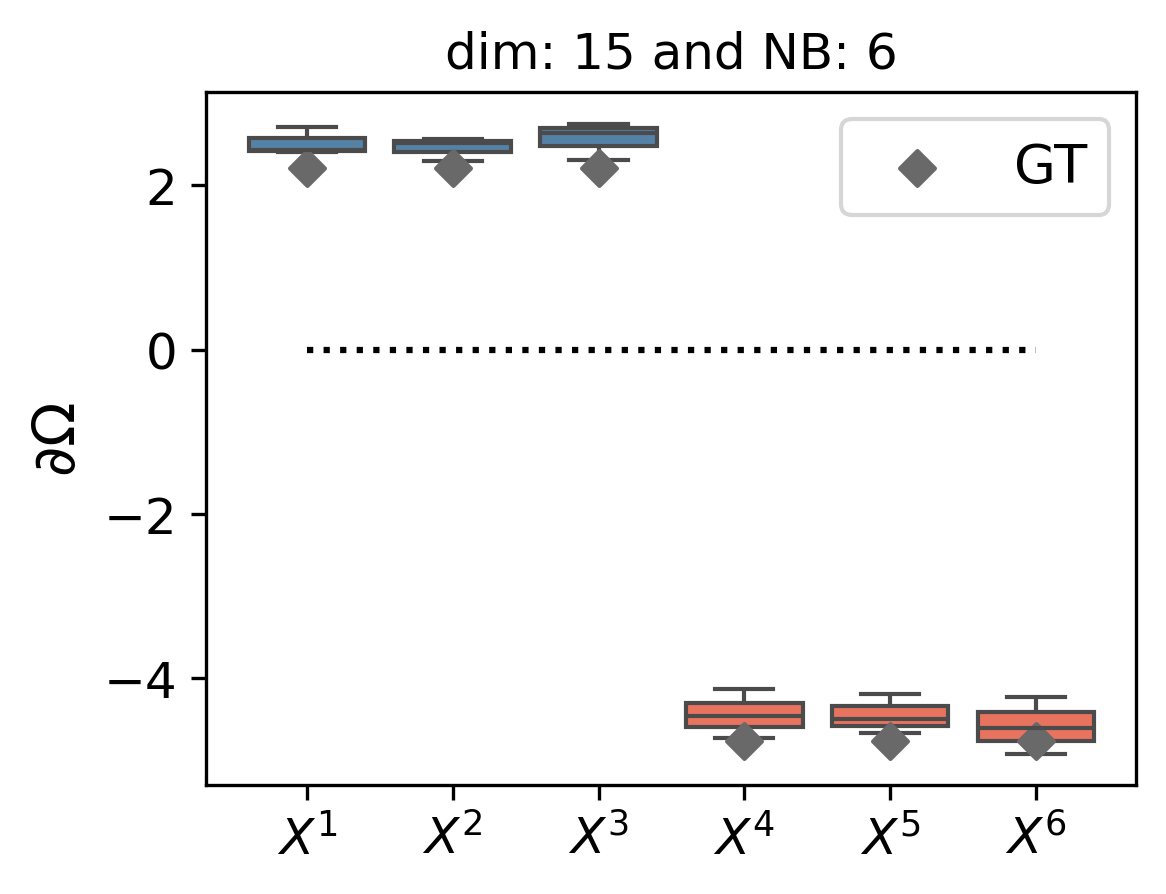}
         \caption{Dim = 15}
     \end{subfigure}
      \begin{subfigure}{0.22\textwidth}
         \centering
         \includegraphics[page=1,width=\linewidth]{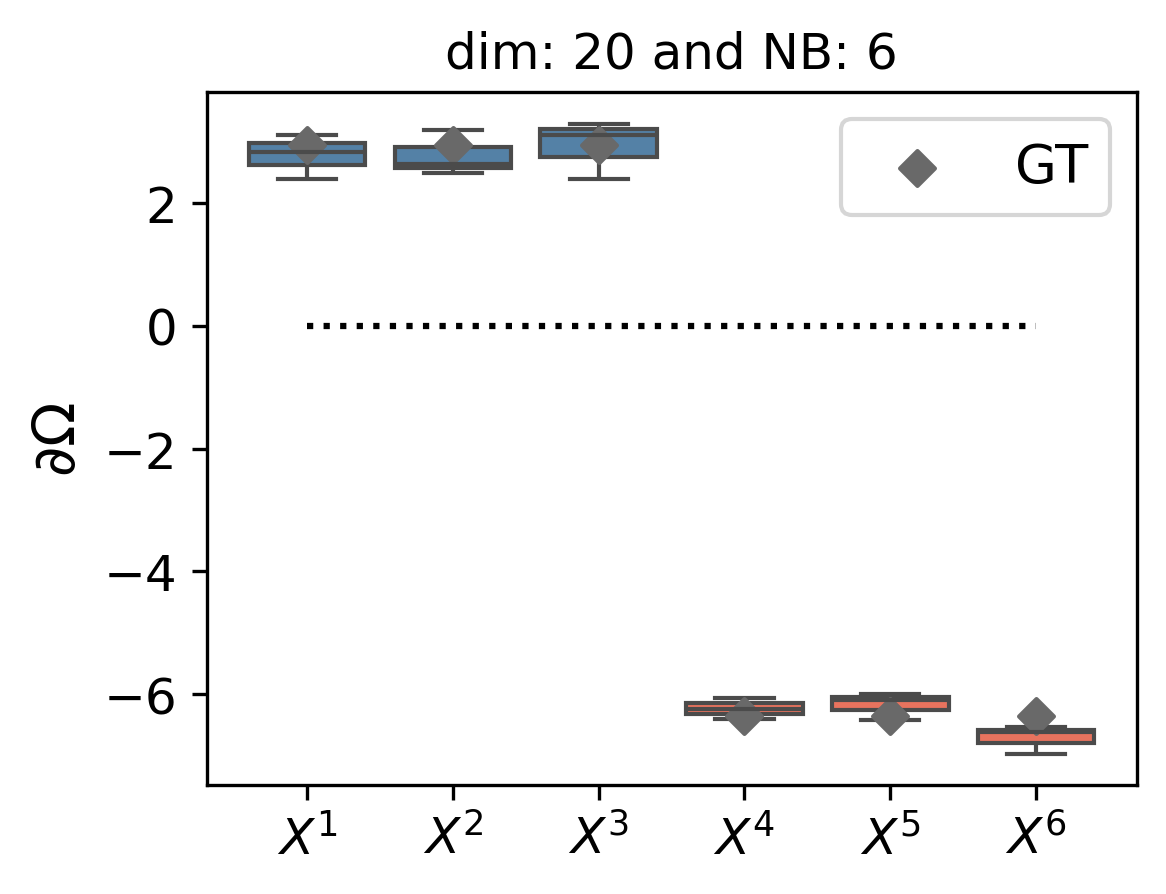}
         \caption{Dim = 20}
     \end{subfigure}
     
         \begin{subfigure}{0.22\textwidth}
         \centering
         \includegraphics[page=1,width=\linewidth]{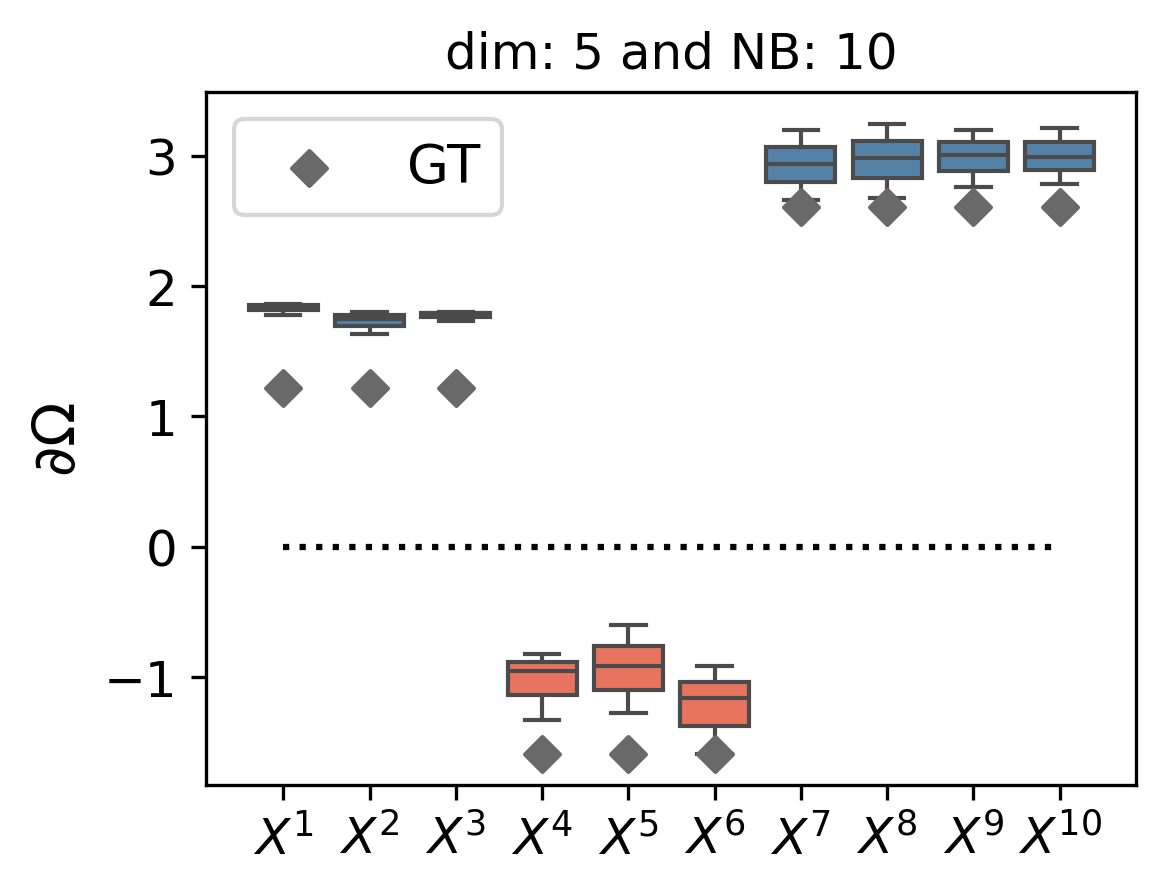}
         \caption{Dim = 5}
     \end{subfigure}
      \begin{subfigure}{0.22\textwidth}
         \centering
         \includegraphics[page=1,width=\linewidth]{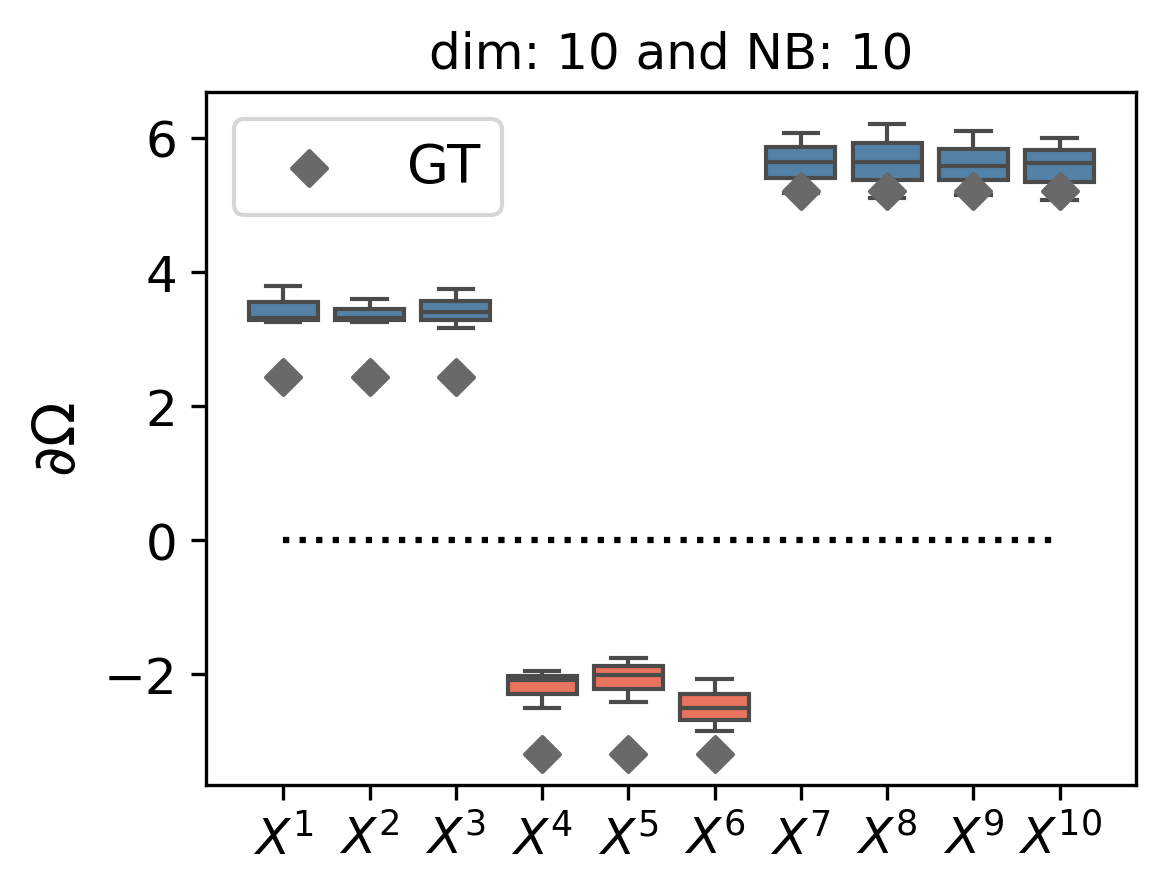}
         \caption{Dim = 10}
     \end{subfigure}
   \begin{subfigure}{0.22\textwidth}
         \centering
         \includegraphics[page=1,width=\linewidth]{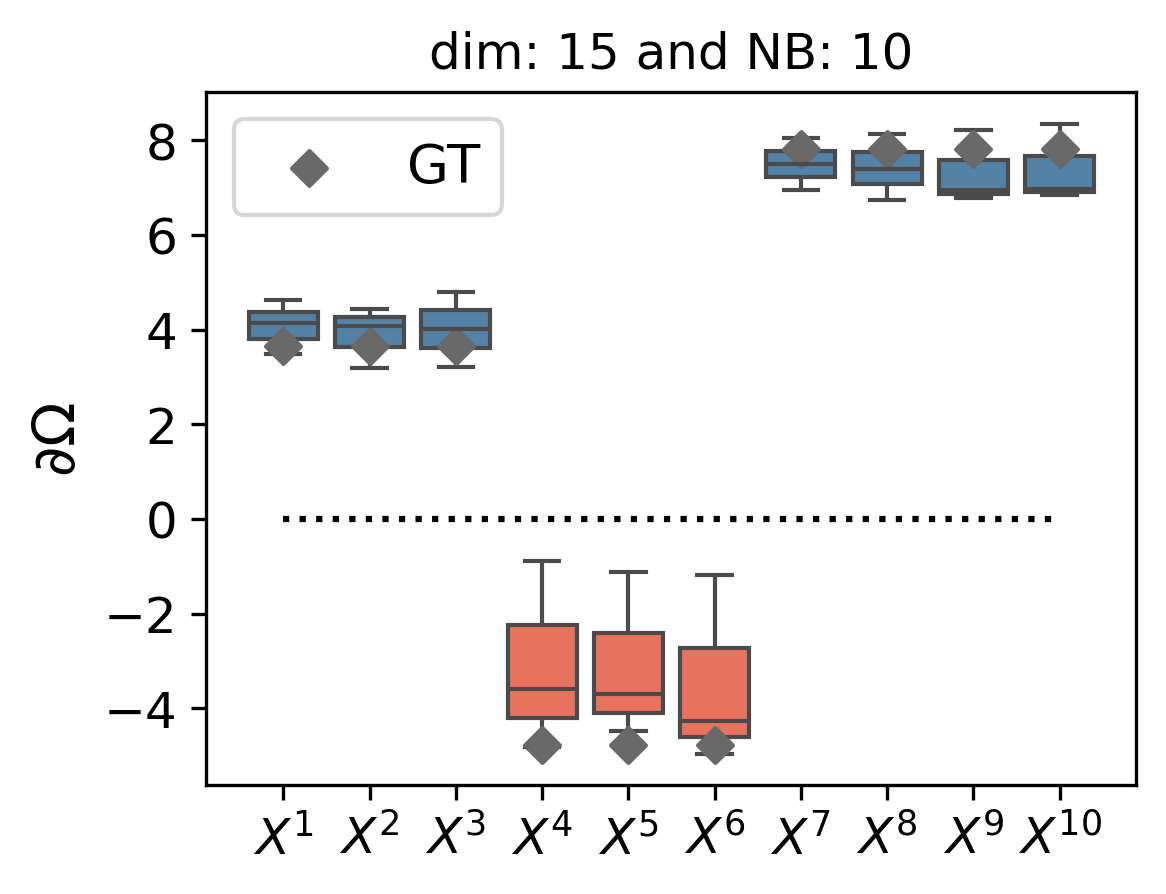}
         \caption{Dim = 15}
     \end{subfigure}
      \begin{subfigure}{0.22\textwidth}
         \centering
         \includegraphics[page=1,width=\linewidth]{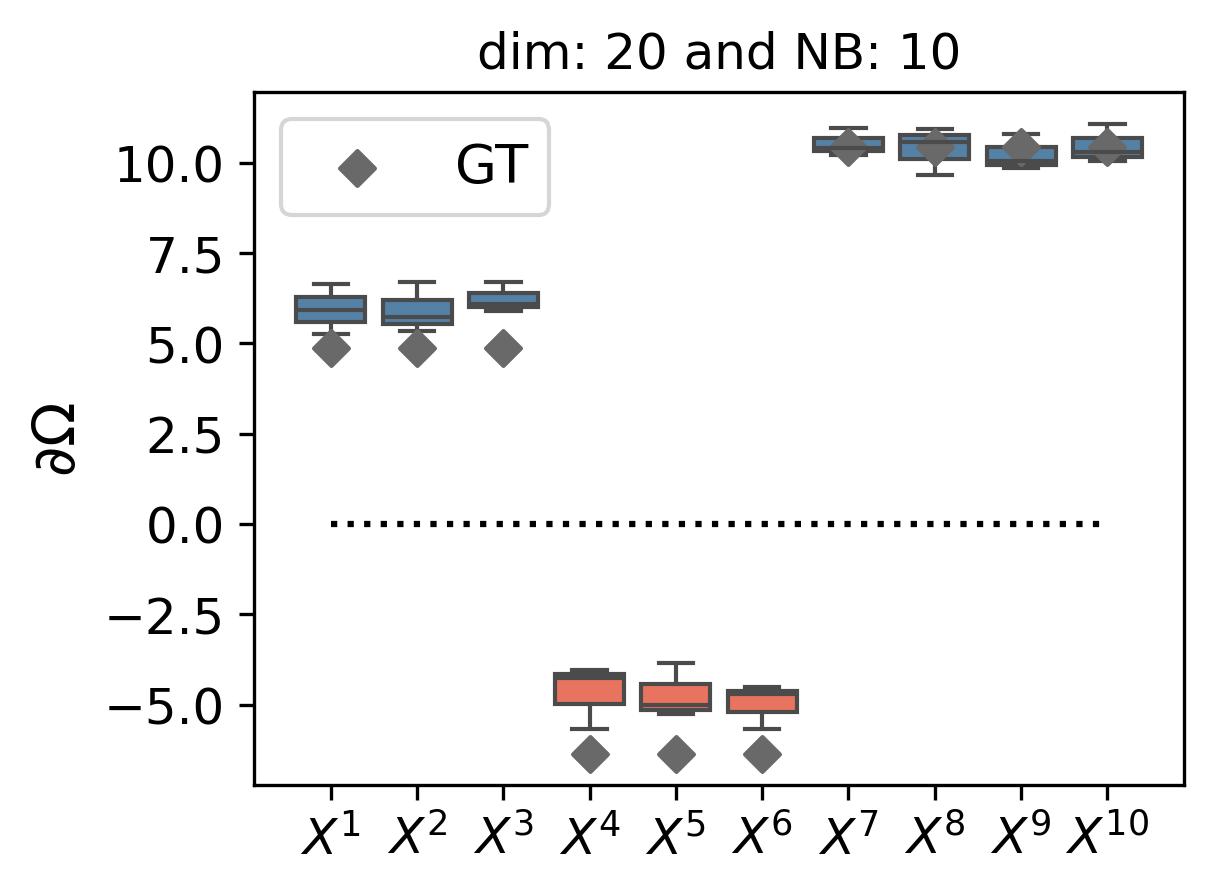}
         \caption{Dim = 20}
     \end{subfigure}

      \caption{Gradient of \acrshort{O-information} using a \textbf{transformer based architecture} for the mixed benchmark, for a system of 6 variables, and a system of 10 variables, and different dimension of variables.
      }
      \label{grad_oinf_tx}
\end{figure}

The results presented in \Cref{grad_oinf_tx} demonstrate the ability of \gls{SOI} to accurately estimate the gradients of \gls{O-information}, provided that the denoising network has sufficient capacity to approximate all the denoising score functions.

\section{Beyond Normal Benchmarks}
\label{apdx:non_norm}
In this section, we evaluate \gls{SOI} and alternatives across more challenging distributions. To construct such settings we apply \gls{MI}-invariant transformations to the benchmarks established in Section \Cref{experiment}. Since \gls{TC} and \gls{DTC} can be written in terms of \gls{MI} terms, the in-variance of \gls{O-information} to \gls{MI} invariant transformations is self-evident. 

\begin{figure} [h]
\centering
\begin{subfigure}{0.3\textwidth}
         \centering
\includegraphics[page=1,width=\linewidth]{assets/figures/exp_red/legend.PNG}
     \end{subfigure}

     \begin{subfigure}{0.24\textwidth}
         \centering

         \includegraphics[page=1,width=\linewidth]{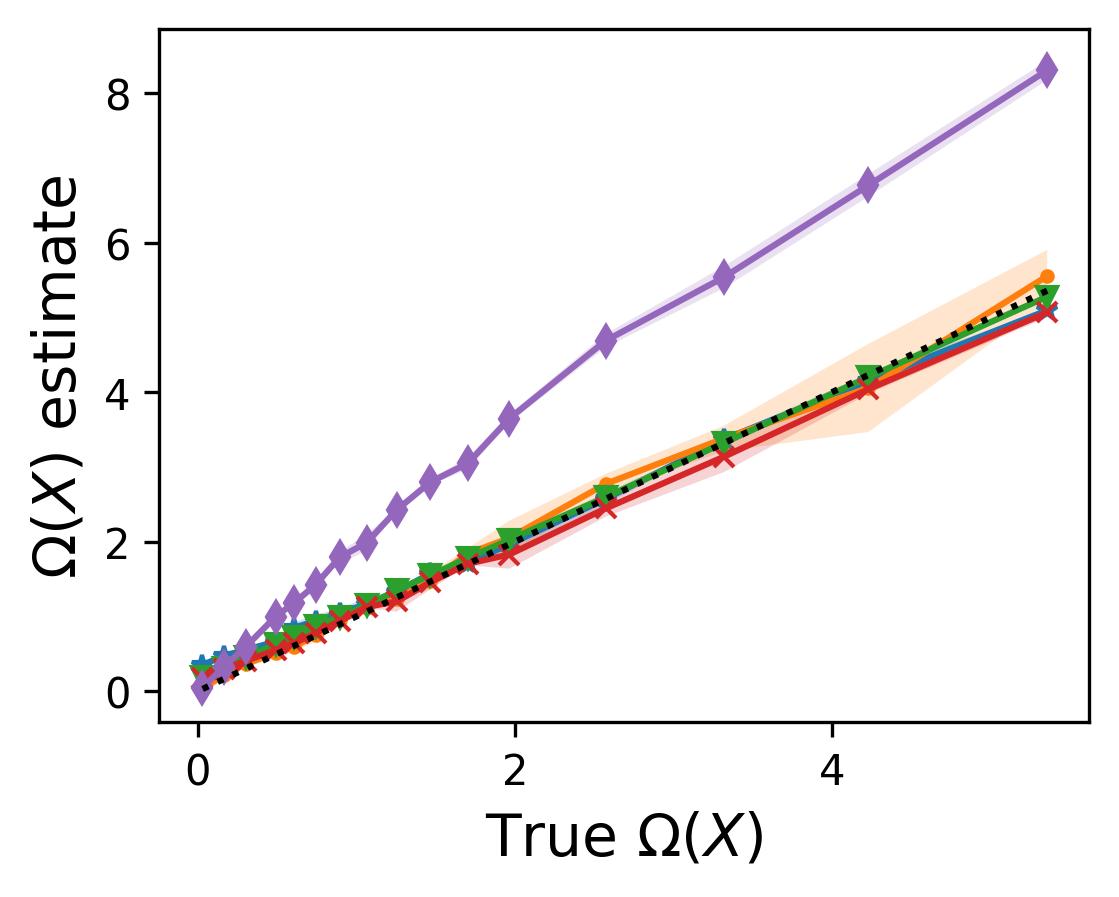}
         \caption{Dim=5}
     \end{subfigure}
      \begin{subfigure}{0.24\textwidth}
         \centering
  
         \includegraphics[page=1,width=\linewidth]{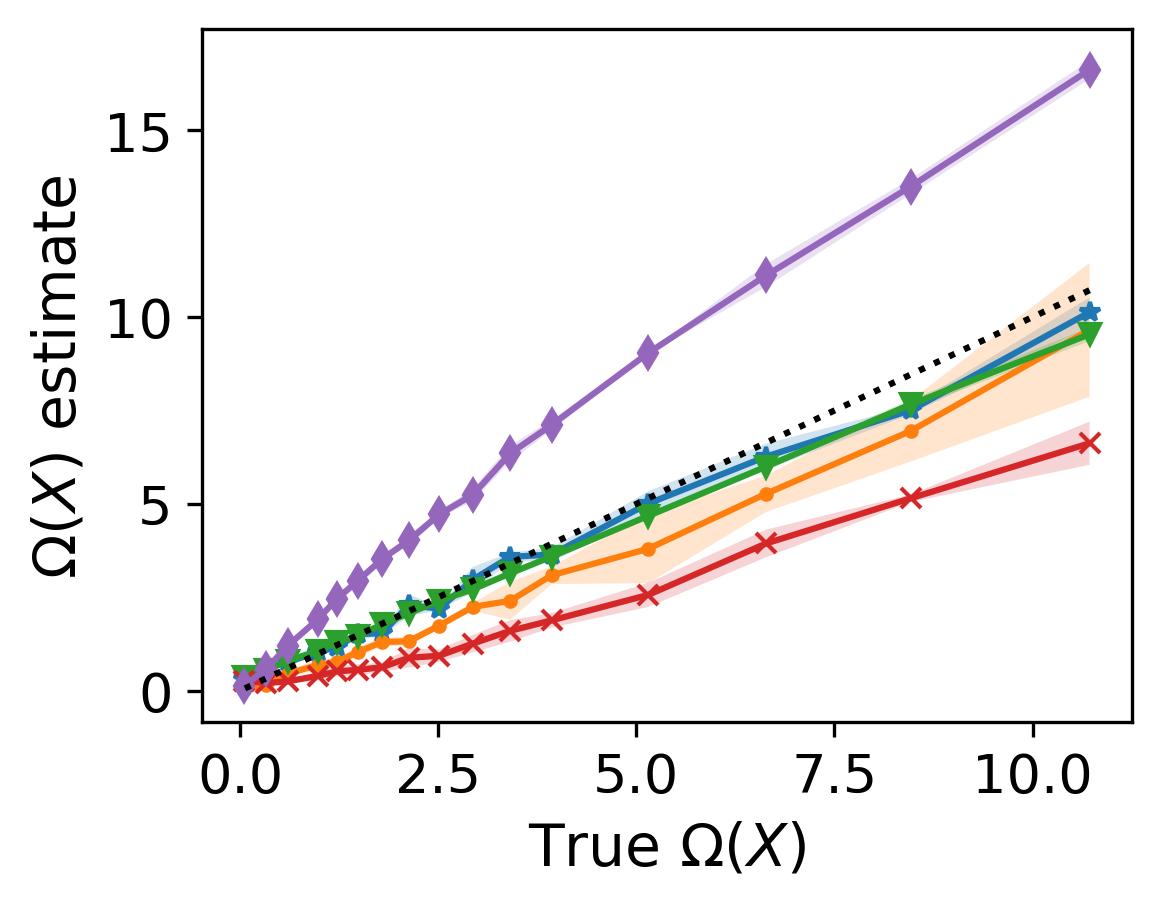}
         \caption{Dim=10}
     \end{subfigure}
     \begin{subfigure}{0.24\textwidth}
         \centering
    
         \includegraphics[page=1,width=\linewidth]{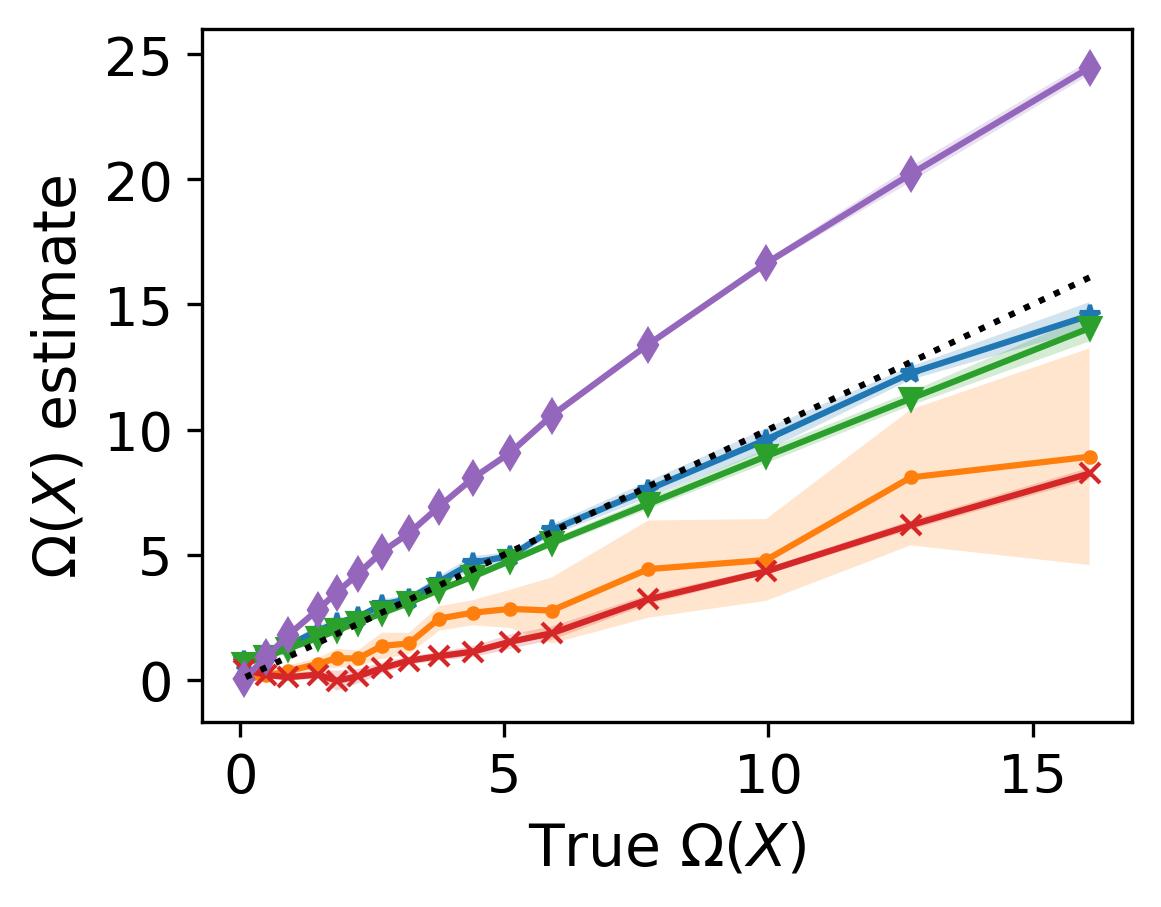}
         \caption{Dim=15}
     \end{subfigure}
         \begin{subfigure}{0.24\textwidth}
         \centering

         \includegraphics[page=1,width=\linewidth]{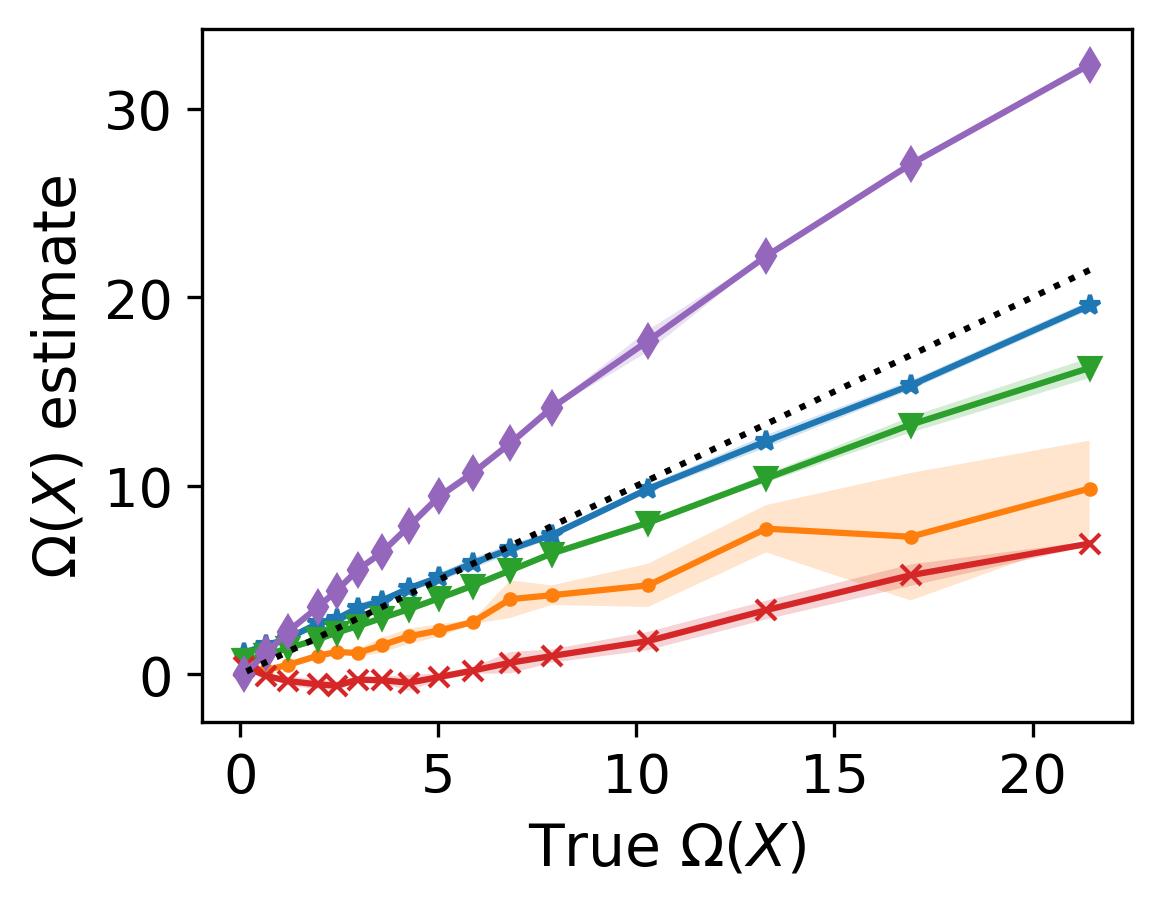}
         \caption{Dim=20}
     \end{subfigure}
      \caption{ Redundant system with 10 variables, organized into subsets of sizes $\{3,3,4\}$ and increasing interaction strength. A \textbf{half-cube} transformation is applied on-top of the multi-normal distribution
      }
       \label{hc_red}

\end{figure}

\begin{figure} [h]
\centering
\begin{subfigure}{0.3\textwidth}
         \centering
\includegraphics[page=1,width=\linewidth]{assets/figures/exp_red/legend.PNG}
     \end{subfigure}

     \begin{subfigure}{0.24\textwidth}
         \centering

         \includegraphics[page=1,width=\linewidth]{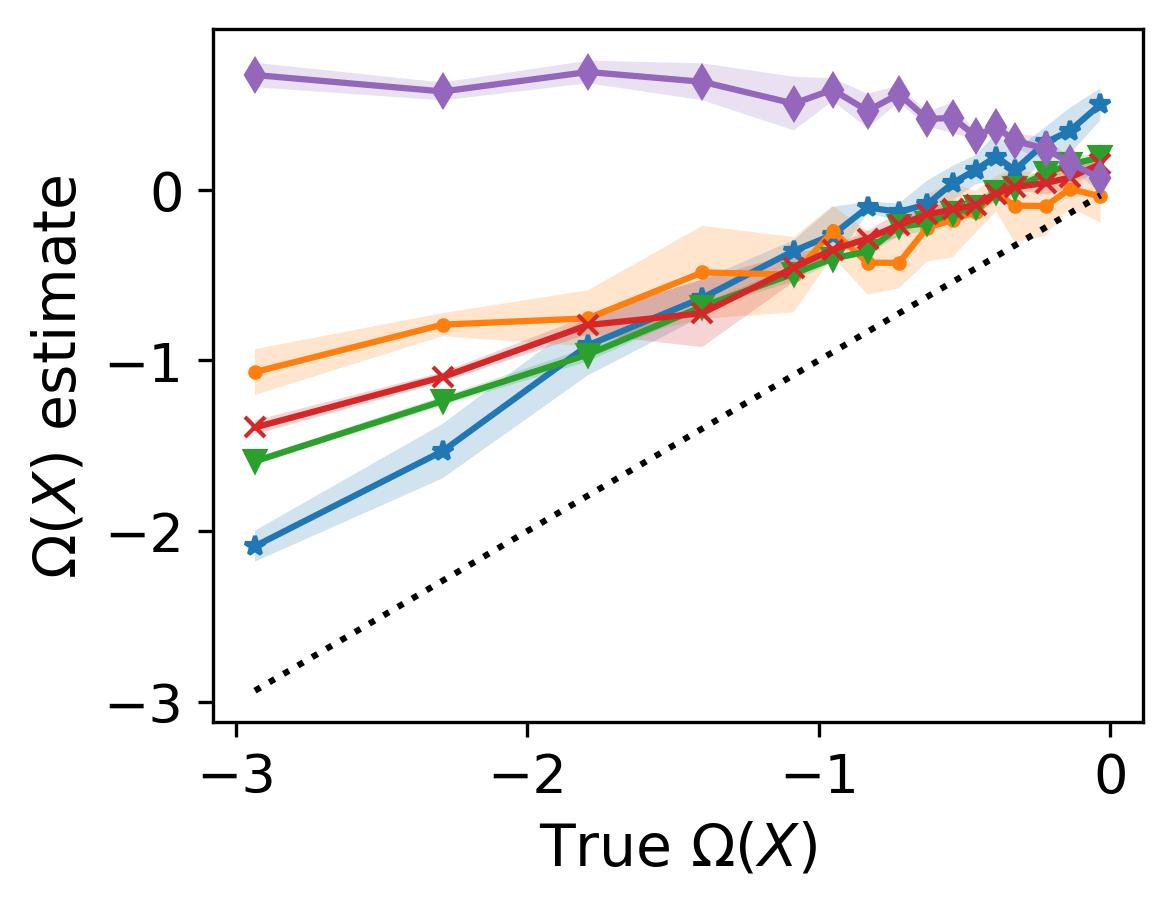}
         \caption{Dim=5}
     \end{subfigure}
      \begin{subfigure}{0.24\textwidth}
         \centering
  
         \includegraphics[page=1,width=\linewidth]{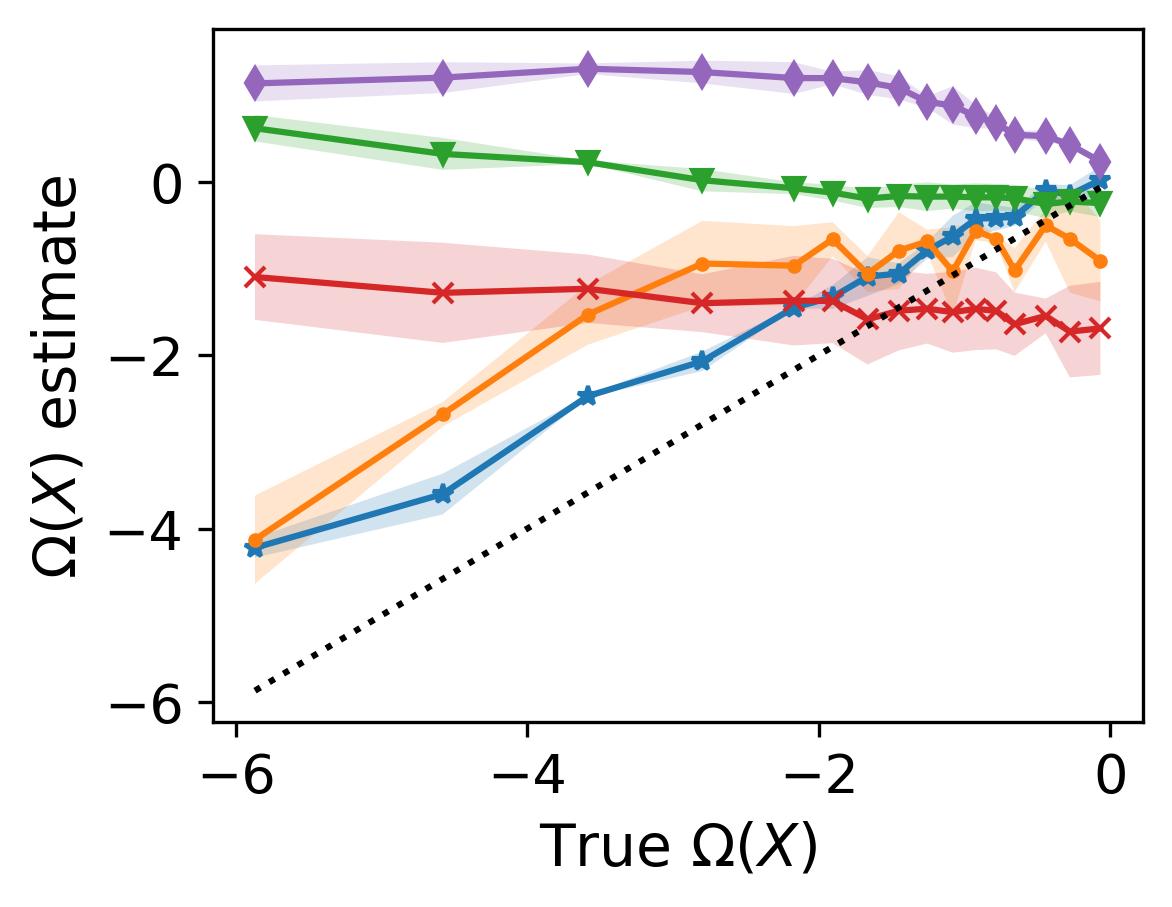}
         \caption{Dim=10}
     \end{subfigure}
     \begin{subfigure}{0.24\textwidth}
         \centering
    
         \includegraphics[page=1,width=\linewidth]{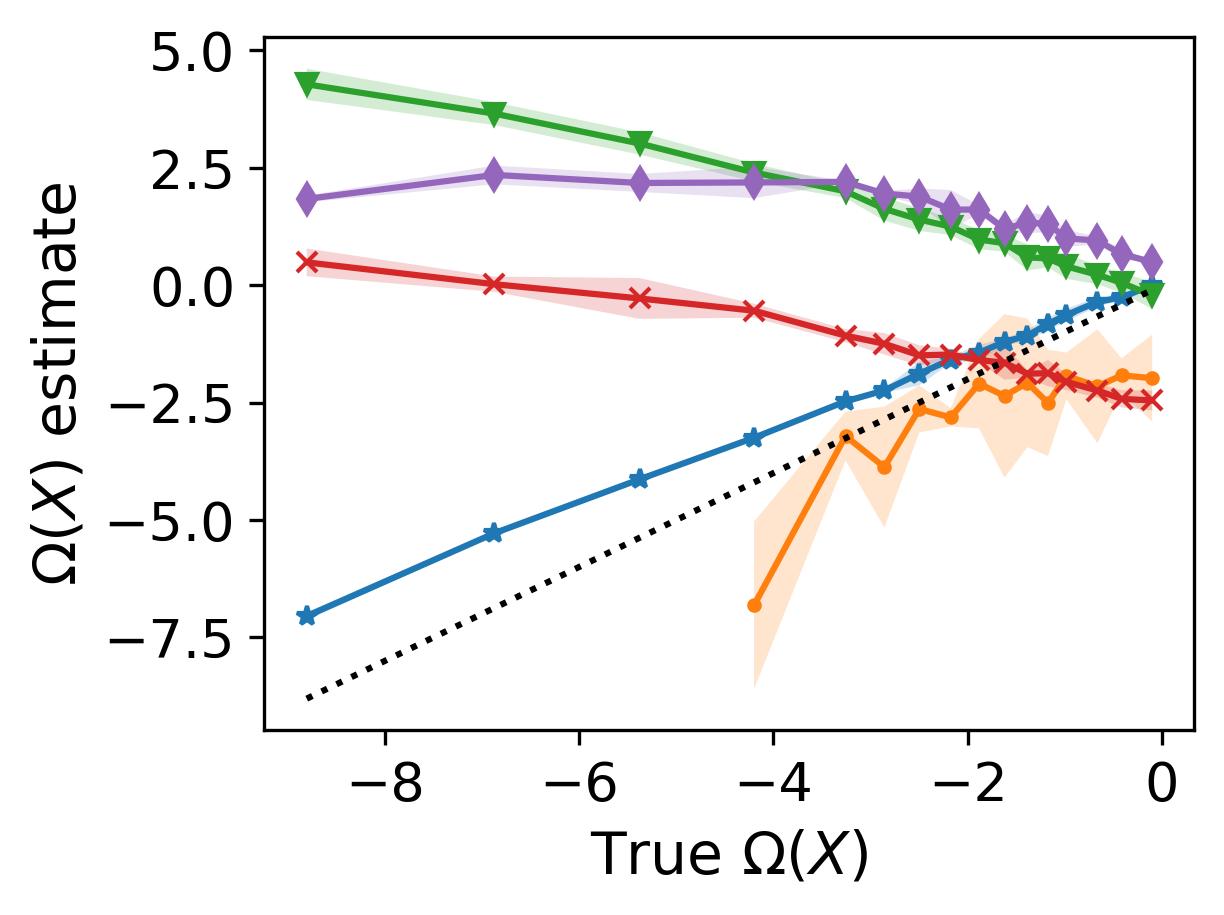}
         \caption{Dim=15}
     \end{subfigure}
         \begin{subfigure}{0.24\textwidth}
         \centering

         \includegraphics[page=1,width=\linewidth]{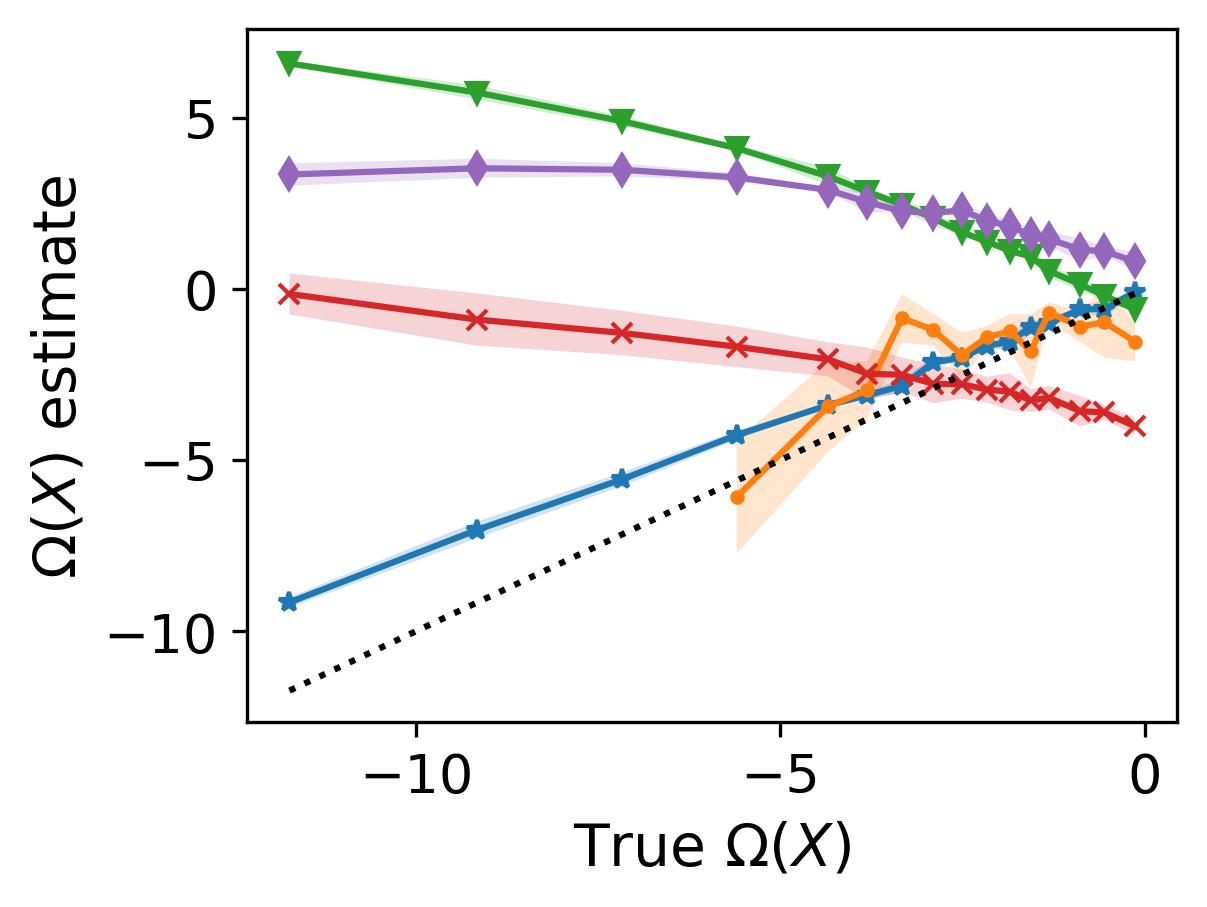}
         \caption{Dim=20}
     \end{subfigure}
      \caption{ Synergistic system with 10 variables, organized into subsets of sizes $\{3,3,4\}$ and increasing interaction strength. A \textbf{half-cube} transformation is applied on-top of the multi-normal distribution.
      }
      \label{hc_syn}

\end{figure}

\begin{figure} [H]
\centering
\begin{subfigure}{0.3\textwidth}
         \centering
\includegraphics[page=1,width=\linewidth]{assets/figures/exp_red/legend.PNG}
     \end{subfigure}

     \begin{subfigure}{0.24\textwidth}
         \centering

         \includegraphics[page=1,width=\linewidth]{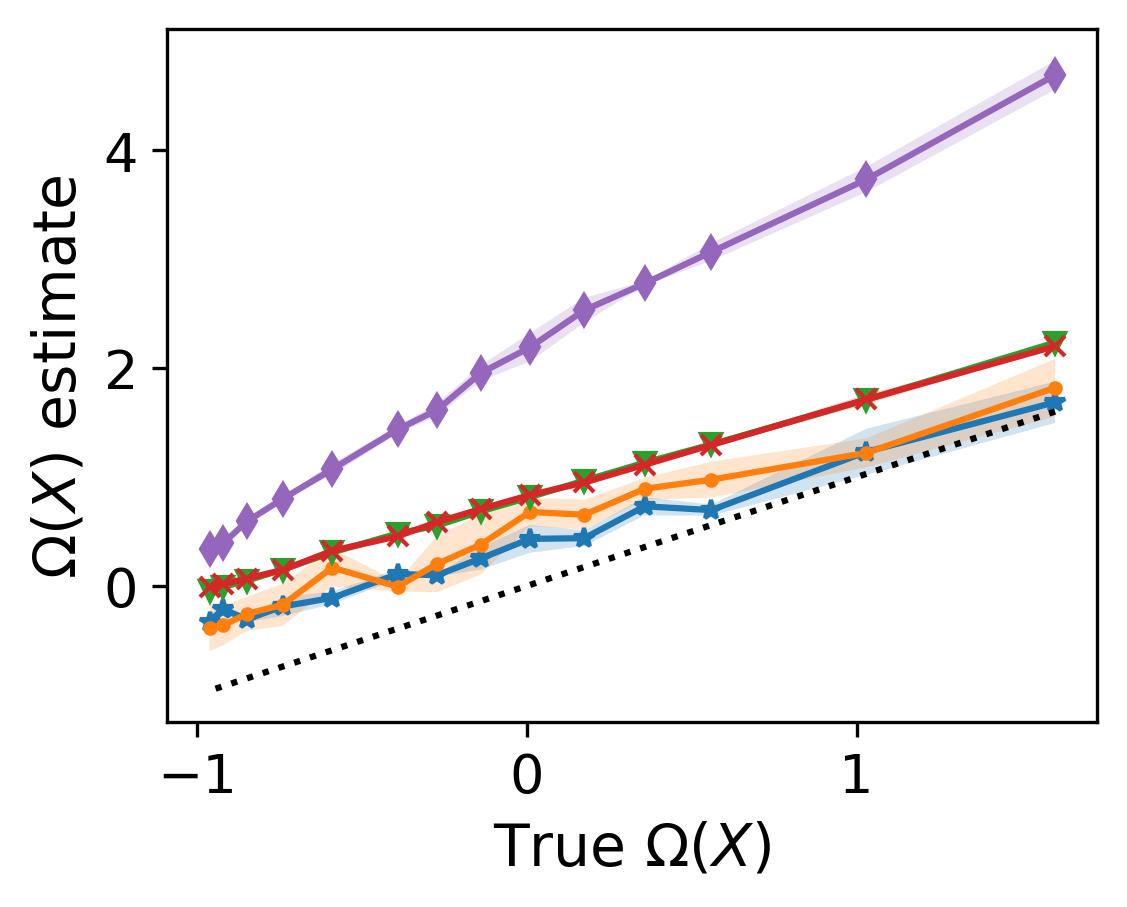}
         \caption{Dim=5}
     \end{subfigure}
      \begin{subfigure}{0.24\textwidth}
         \centering
  
         \includegraphics[page=1,width=\linewidth]{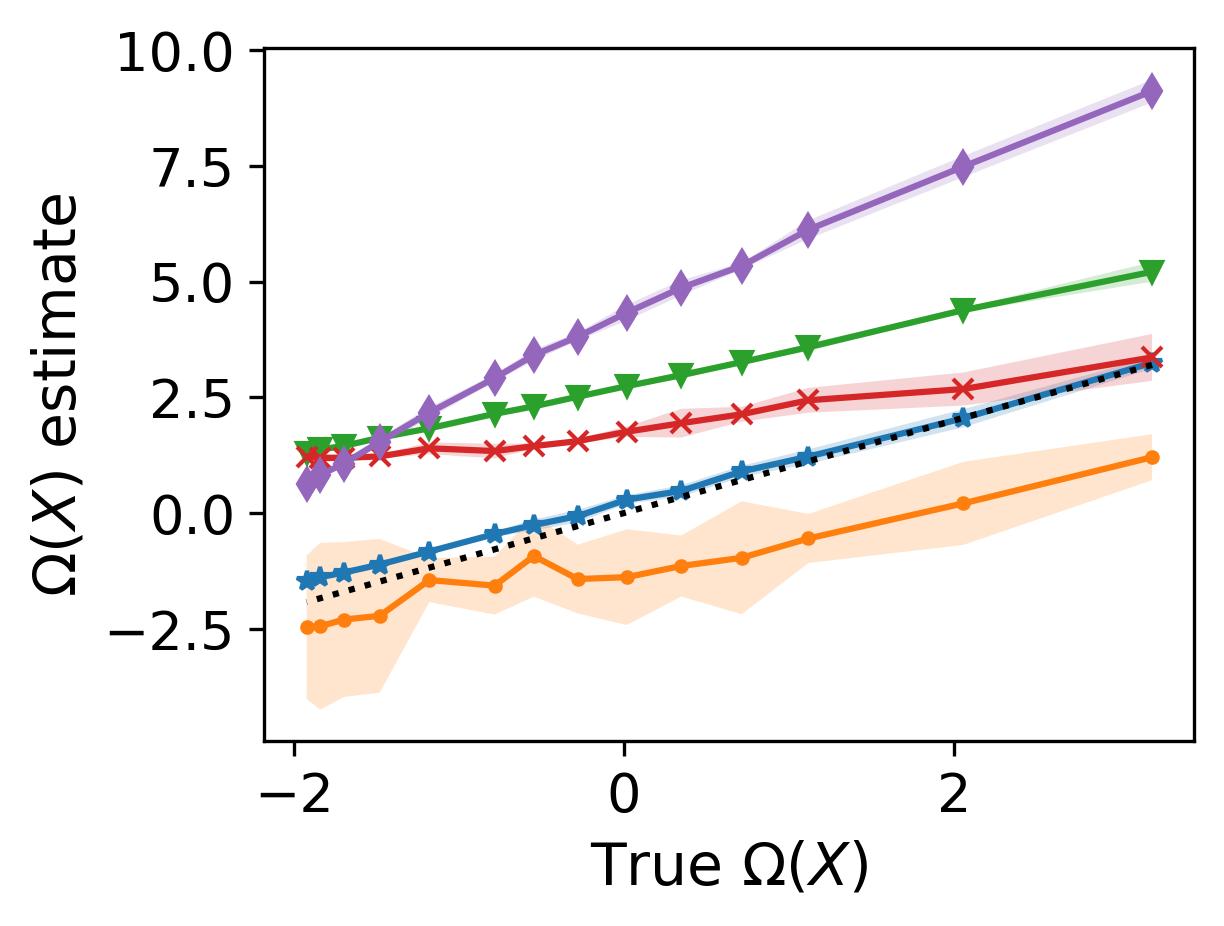}
         \caption{Dim=10}
     \end{subfigure}
     \begin{subfigure}{0.24\textwidth}
         \centering
    
         \includegraphics[page=1,width=\linewidth]{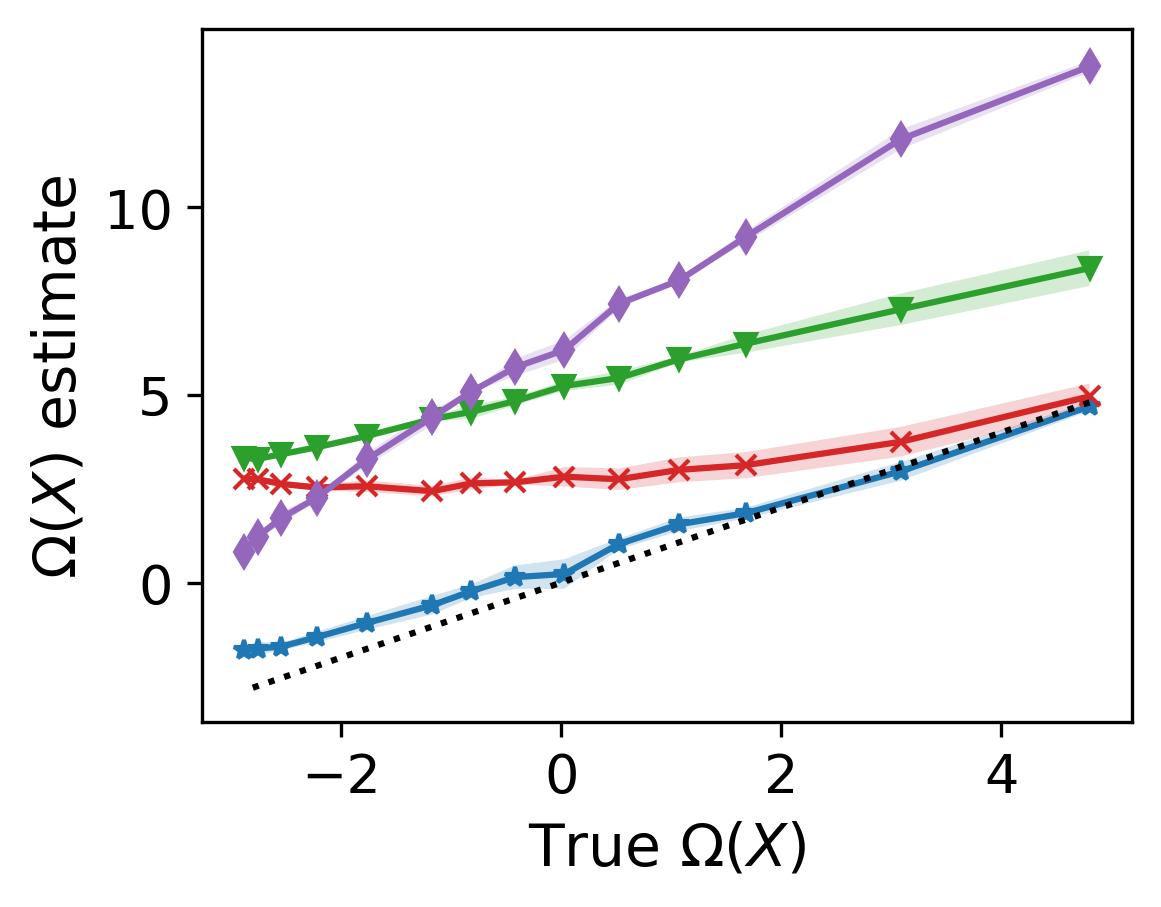}
         \caption{Dim=15}
     \end{subfigure}
         \begin{subfigure}{0.24\textwidth}
         \centering

         \includegraphics[page=1,width=\linewidth]{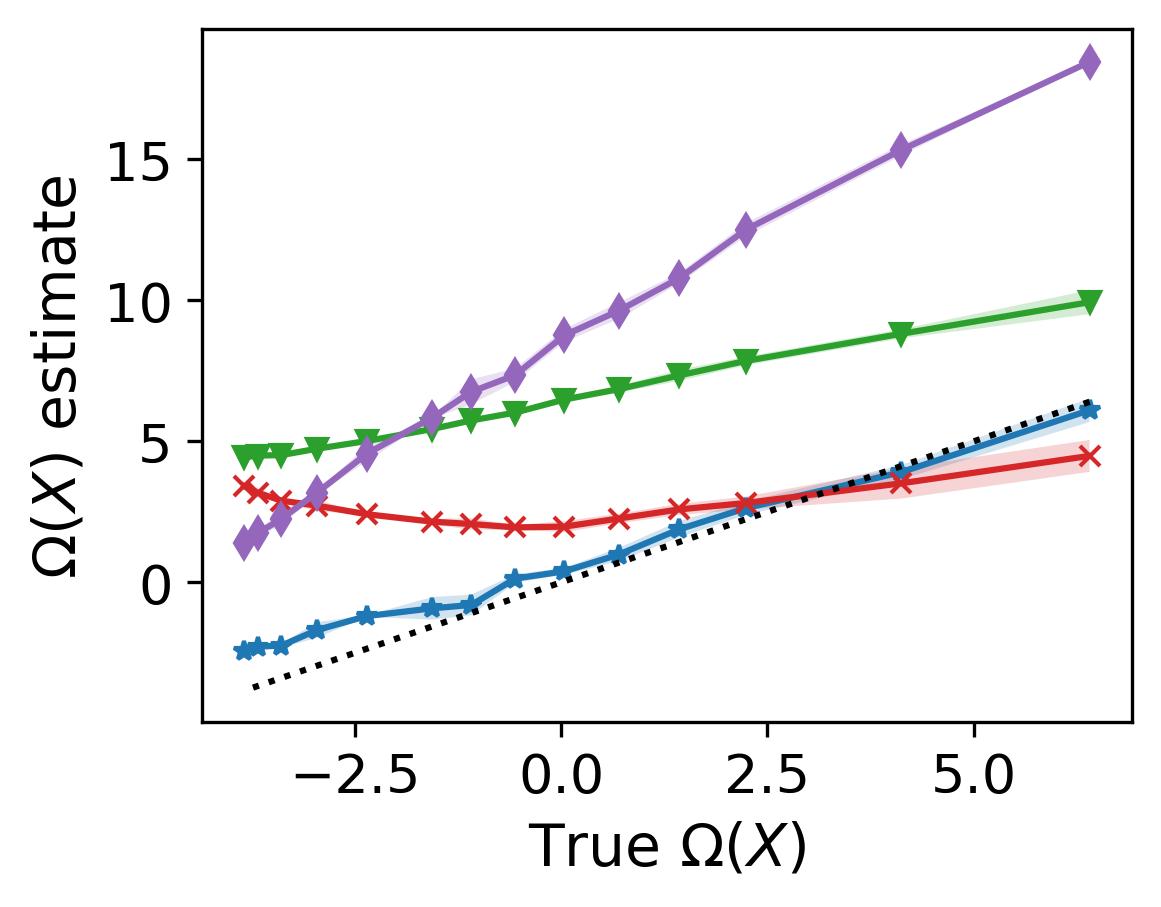}
         \caption{Dim=20}
     \end{subfigure}
      \caption{Mixed-interaction system with 10 variables, organized into 2 redundancy-dominant subsets of size $\{3,4\}$ variables and one synergy-dominant subset with $3$ variables. \acrshort{O-information} is modulated by fixing the synergy inter-dependency and increasing the redundancy. A \textbf{half-cube} transformation is applied on-top of the multivariate-normal distribution.
      }
      \label{hc_mix}

\end{figure}

\paragraph{Half-cube}  $ x \rightarrow x\sqrt{|x|}$  is recognized as an \gls{MI} invariant transformation, which serves to lengthen the tail of the distribution. Addressing the long tail distribution poses a significant challenge for neural MI estimators, as highlighted in recent studies by\citep{franzese2023minde,czyz2023beyond}. In\Cref{hc_red},\Cref{hc_syn} and \Cref{hc_mix}, we showcase the performance outcomes of \gls{SOI} and other baselines on half-Cube transformed benchmarks that exhibit similar interactions as detailed in \Cref{experiment}. Our approach stands out by delivering superior performance. Notably, the synergistic transformed benchmark emerges as the most demanding scenario: competitors suffer particularly with high-dimensional variables, while \gls{SOI} shows bias, especially in cases of high synergistic interactions, indicated by very low \gls{O-information} values.

\begin{figure} [h]
\centering
\begin{subfigure}{0.3\textwidth}
         \centering
\includegraphics[page=1,width=\linewidth]{assets/figures/exp_red/legend.PNG}
     \end{subfigure}

     \begin{subfigure}{0.24\textwidth}
         \centering

         \includegraphics[page=1,width=\linewidth]{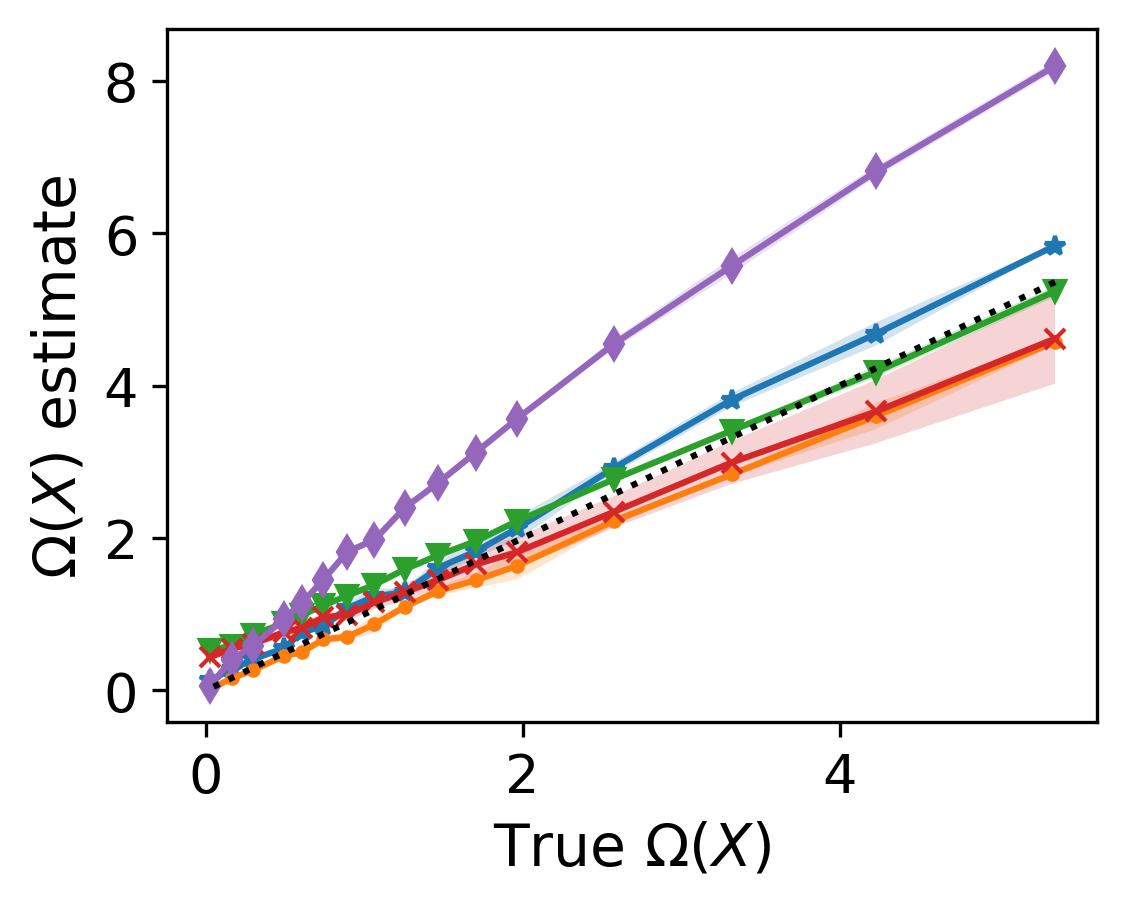}
         \caption{Dim=5}
     \end{subfigure}
      \begin{subfigure}{0.24\textwidth}
         \centering
  
         \includegraphics[page=1,width=\linewidth]{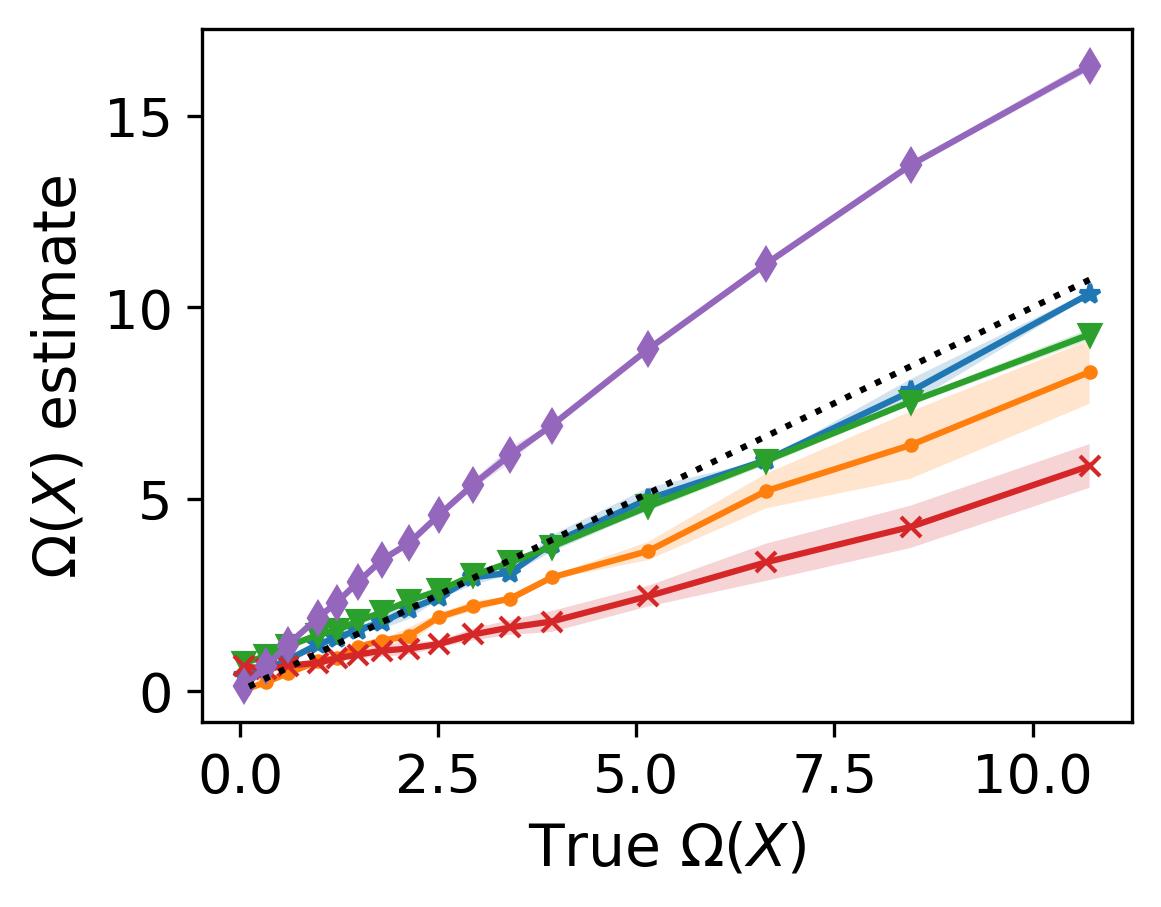}
         \caption{Dim=10}
     \end{subfigure}
     \begin{subfigure}{0.24\textwidth}
         \centering
    
         \includegraphics[page=1,width=\linewidth]{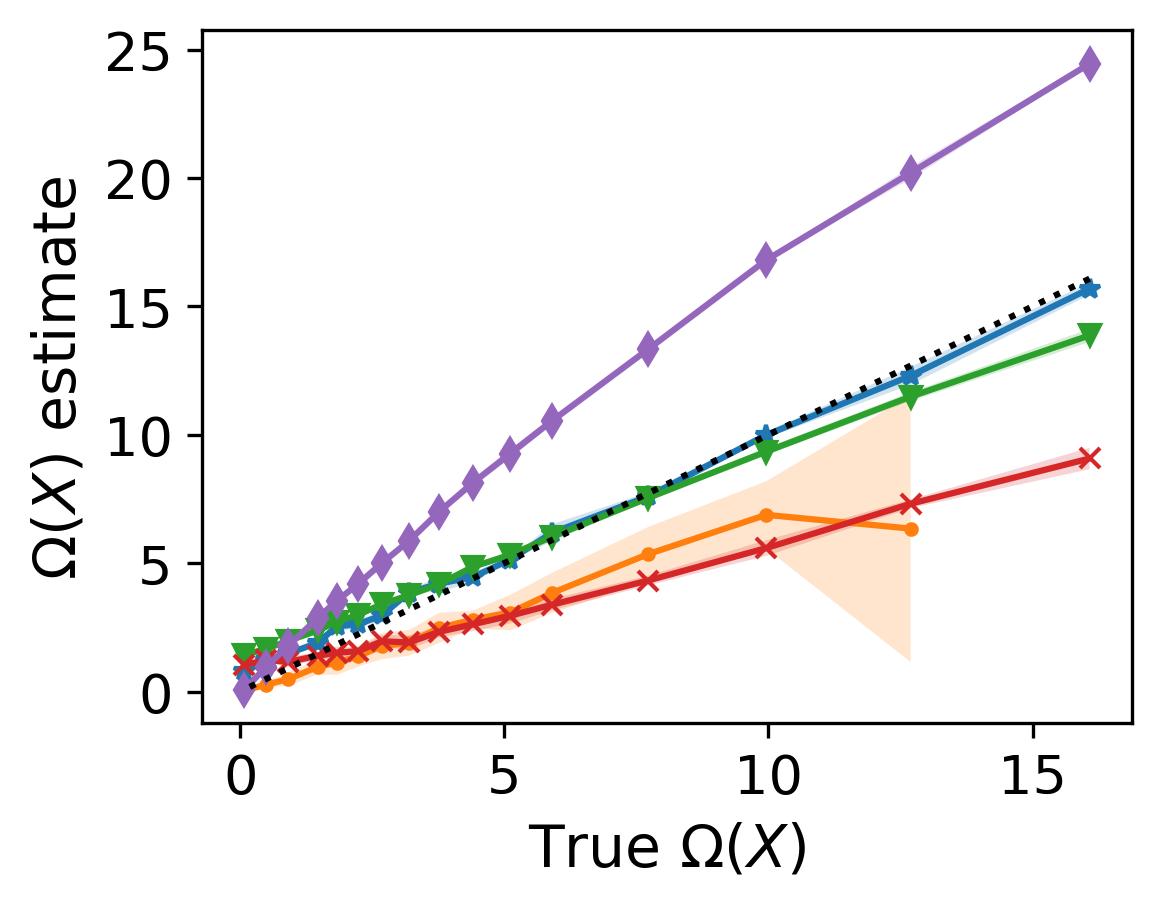}
         \caption{Dim=15}
     \end{subfigure}
         \begin{subfigure}{0.24\textwidth}
         \centering

         \includegraphics[page=1,width=\linewidth]{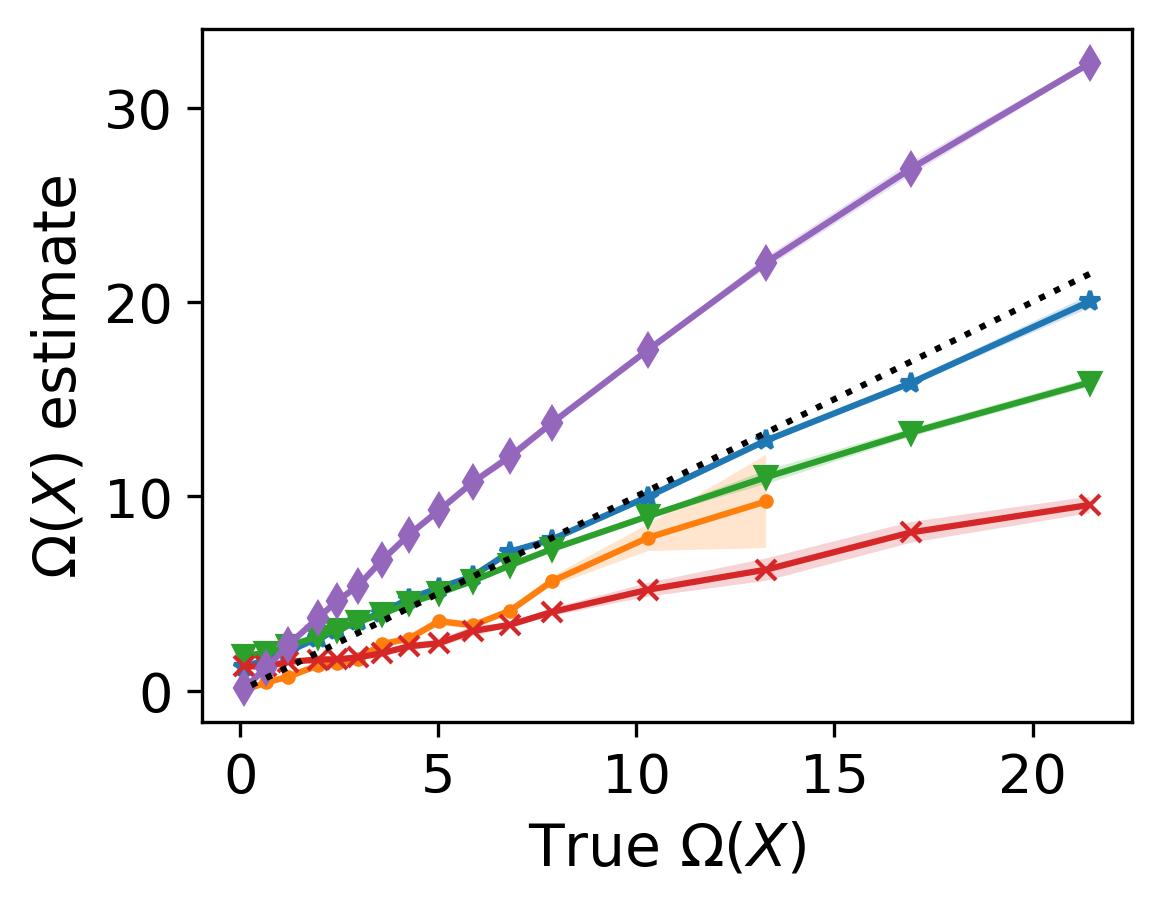}
         \caption{Dim=20}
     \end{subfigure}
      \caption{ Redundant system with 10 variables, organized into subsets of sizes $\{3,3,4\}$ and increasing interaction strength. A \textbf{CDF} transformation is applied on-top of the multi-normal distribution
      }
       \label{cdf_red}

\end{figure}

\begin{figure} [h]
\centering
\begin{subfigure}{0.3\textwidth}
         \centering
\includegraphics[page=1,width=\linewidth]{assets/figures/exp_red/legend.PNG}
     \end{subfigure}

     \begin{subfigure}{0.24\textwidth}
         \centering

         \includegraphics[page=1,width=\linewidth]{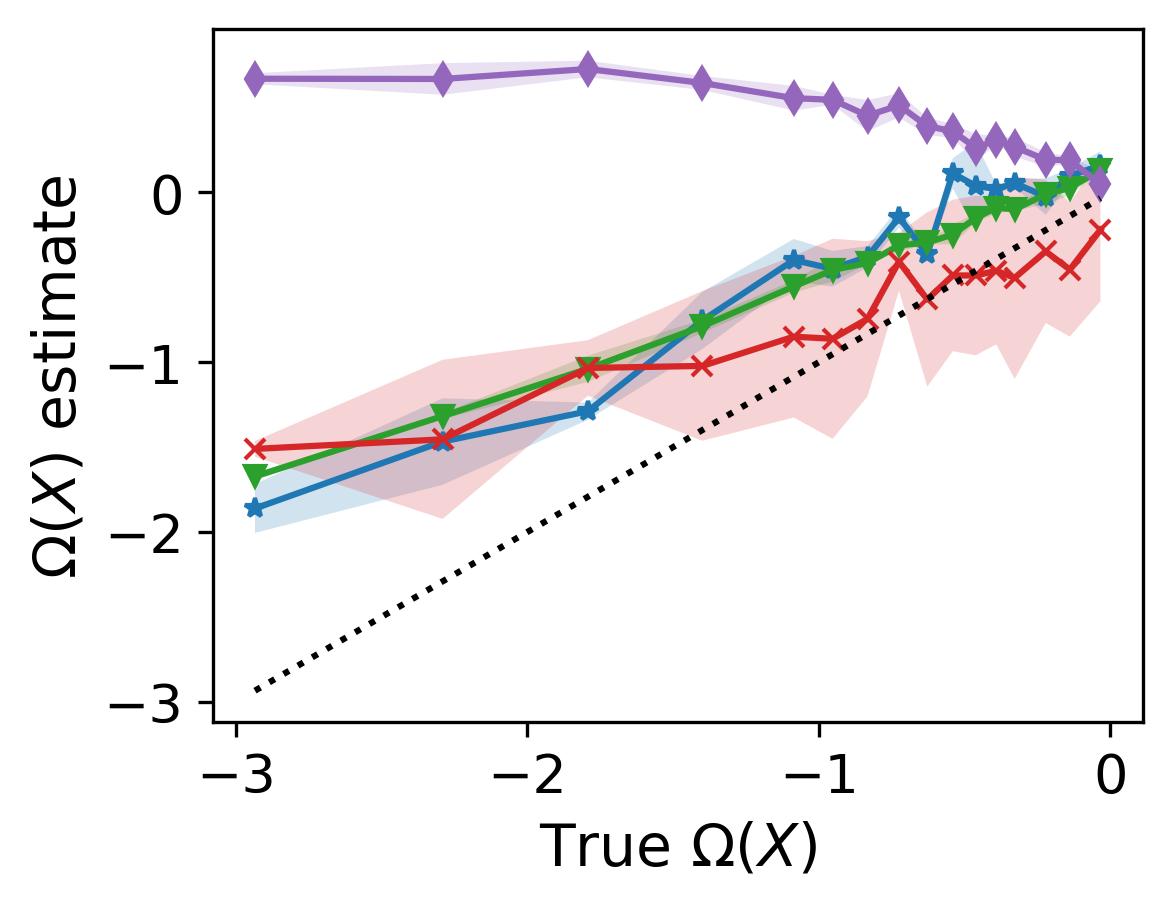}
         \caption{Dim=5}
     \end{subfigure}
      \begin{subfigure}{0.24\textwidth}
         \centering
  
         \includegraphics[page=1,width=\linewidth]{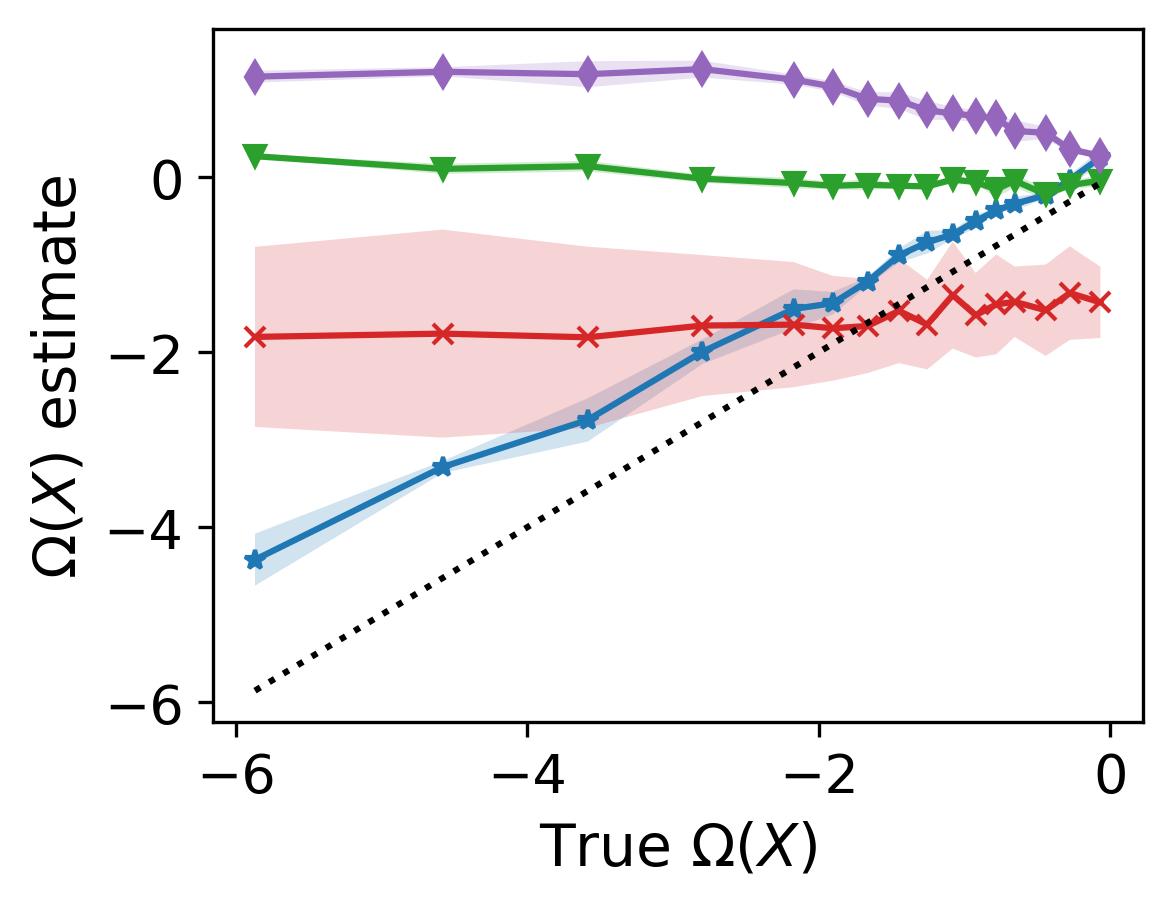}
         \caption{Dim=10}
     \end{subfigure}
     \begin{subfigure}{0.24\textwidth}
         \centering
    
         \includegraphics[page=1,width=\linewidth]{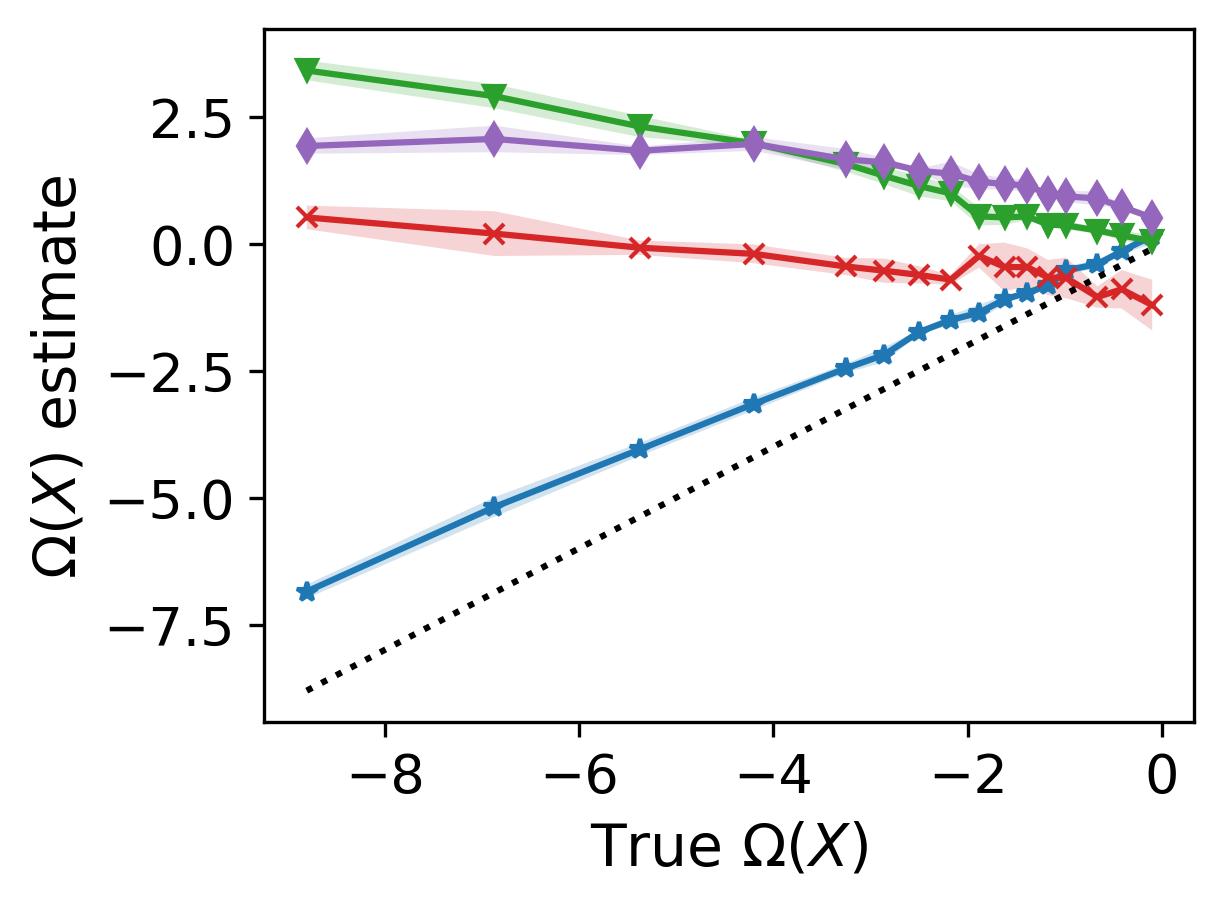}
         \caption{Dim=15}
     \end{subfigure}
         \begin{subfigure}{0.24\textwidth}
         \centering

         \includegraphics[page=1,width=\linewidth]{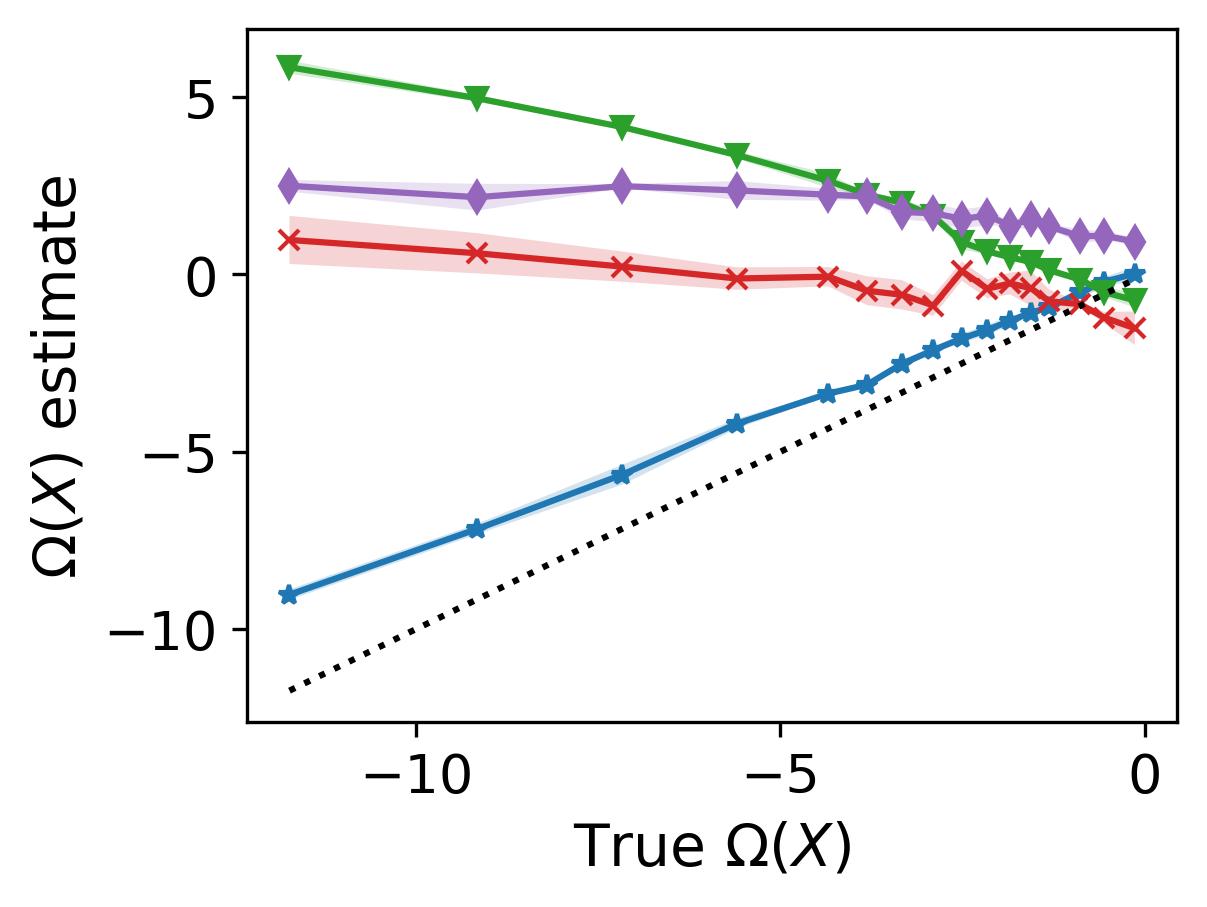}
         \caption{Dim=20}
     \end{subfigure}
      \caption{ Synergistic system with 10 variables, organized into subsets of sizes $\{3,3,4\}$ and increasing interaction strength. A \textbf{CDF} transformation is applied on-top of the multi-normal distribution
      }
      \label{cdf_syn}

\end{figure}

\begin{figure} [H]
\centering
\begin{subfigure}{0.3\textwidth}
         \centering
\includegraphics[page=1,width=\linewidth]{assets/figures/exp_red/legend.PNG}
     \end{subfigure}

     \begin{subfigure}{0.24\textwidth}
         \centering

         \includegraphics[page=1,width=\linewidth]{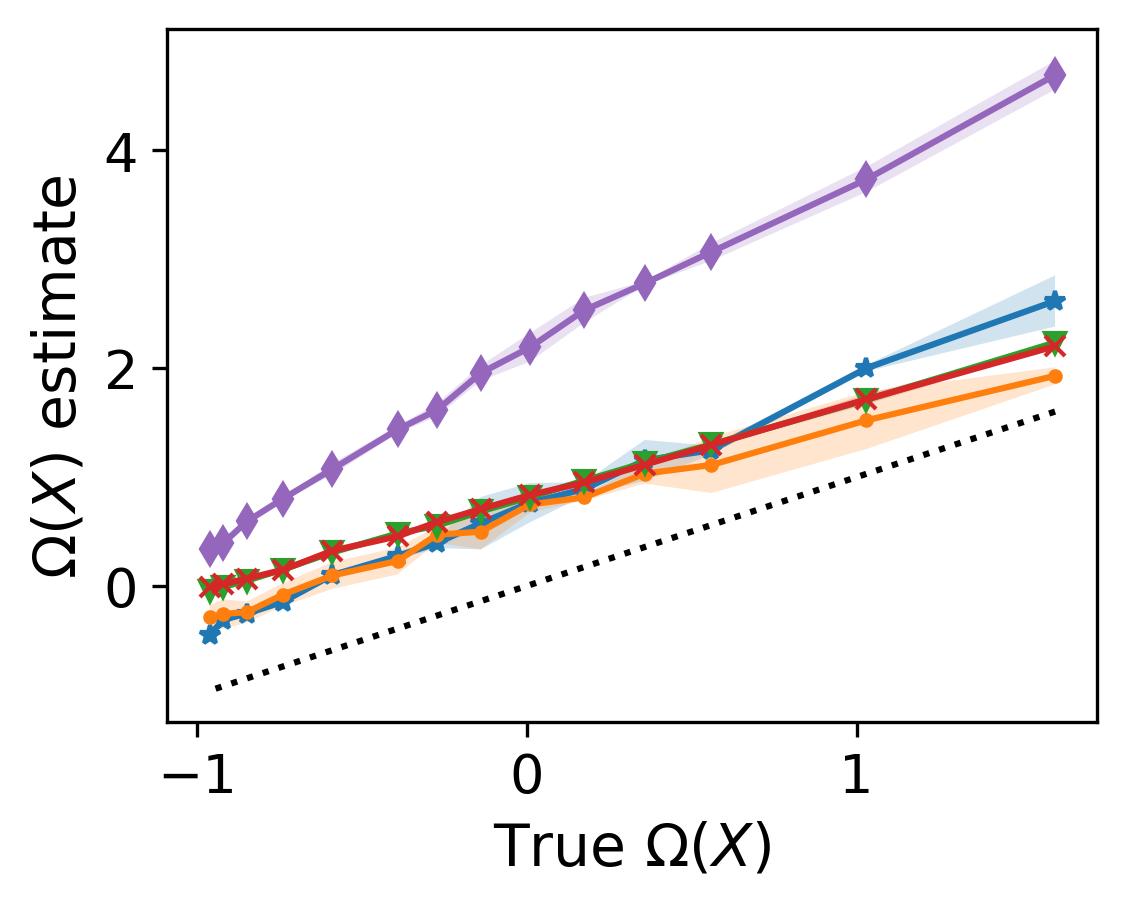}
         \caption{Dim=5}
     \end{subfigure}
      \begin{subfigure}{0.24\textwidth}
         \centering
  
         \includegraphics[page=1,width=\linewidth]{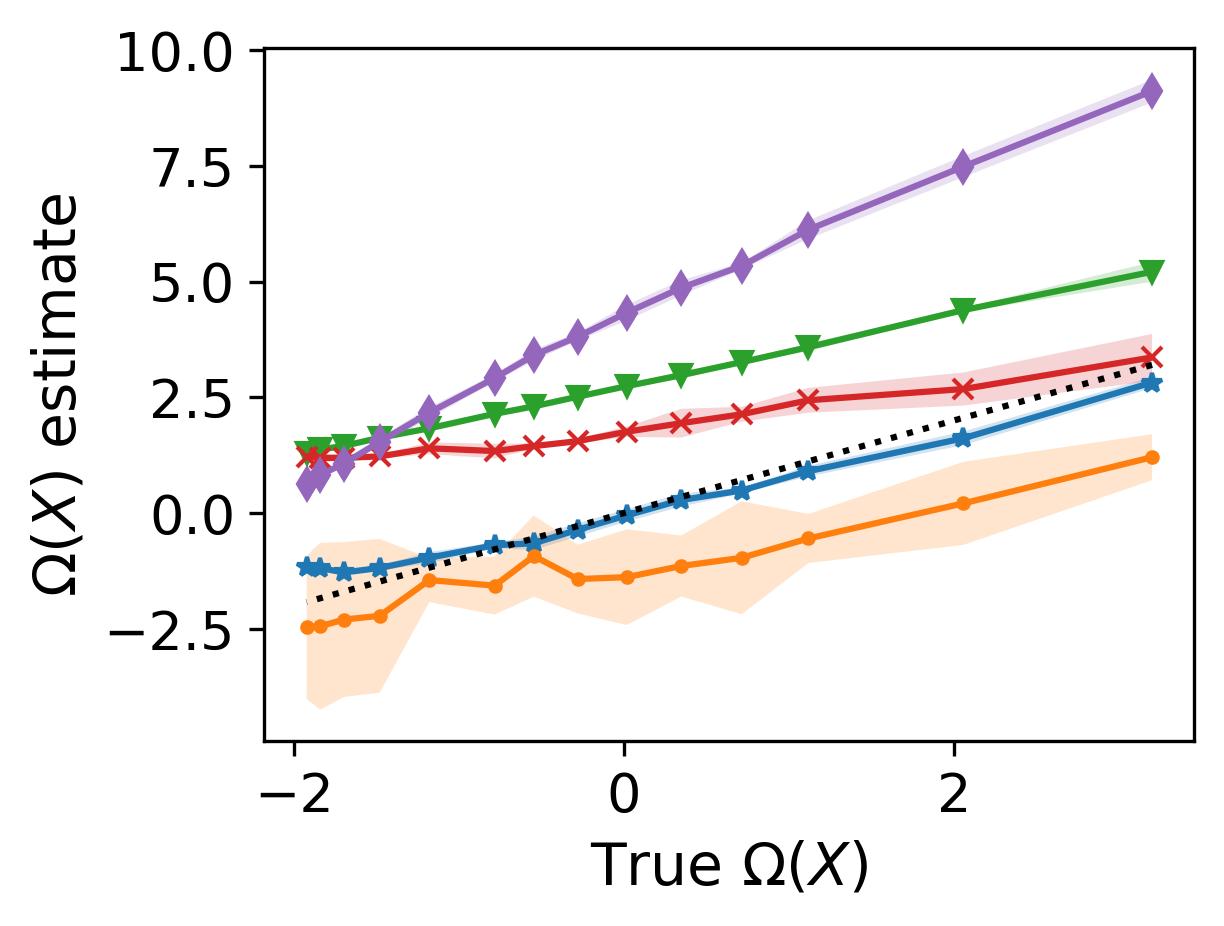}
         \caption{Dim=10}
     \end{subfigure}
     \begin{subfigure}{0.24\textwidth}
         \centering
    
         \includegraphics[page=1,width=\linewidth]{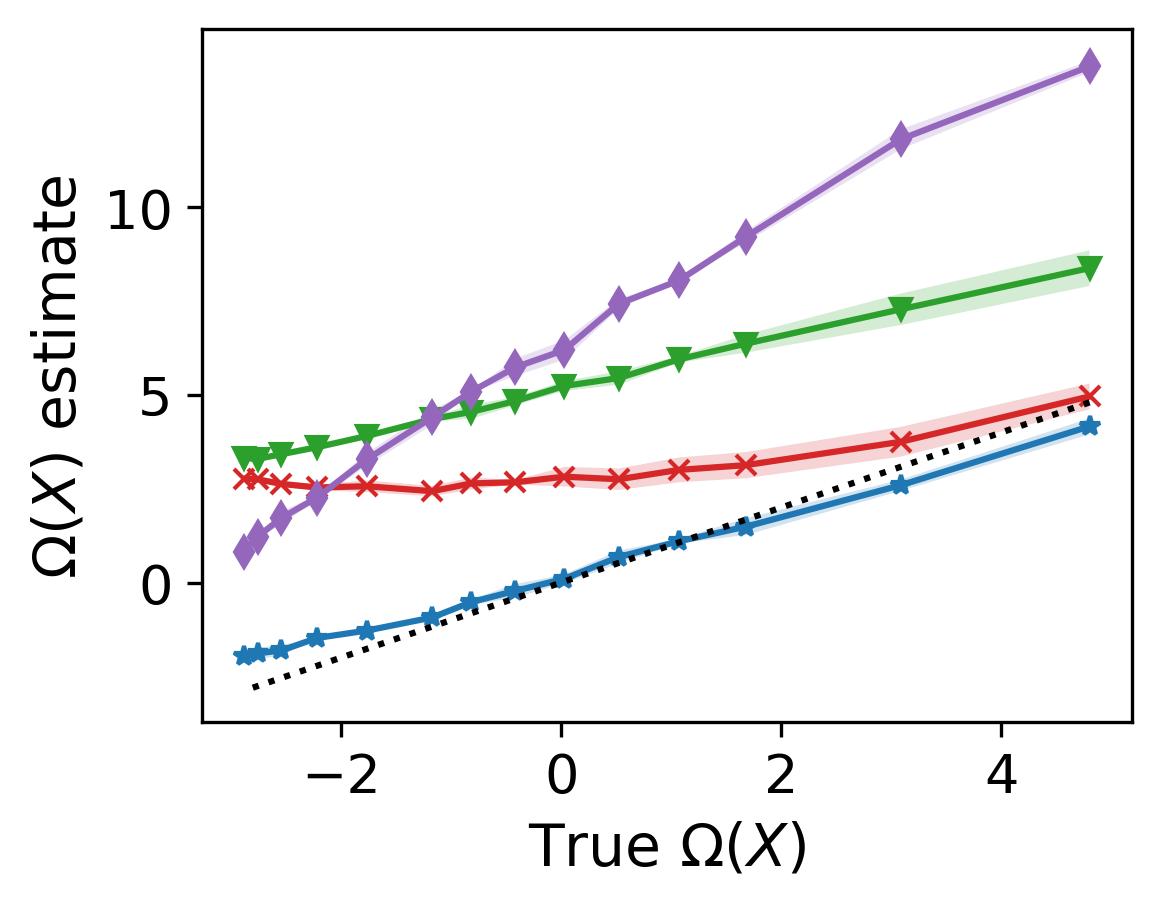}
         \caption{Dim=15}
     \end{subfigure}
         \begin{subfigure}{0.24\textwidth}
         \centering
         \includegraphics[page=1,width=\linewidth]{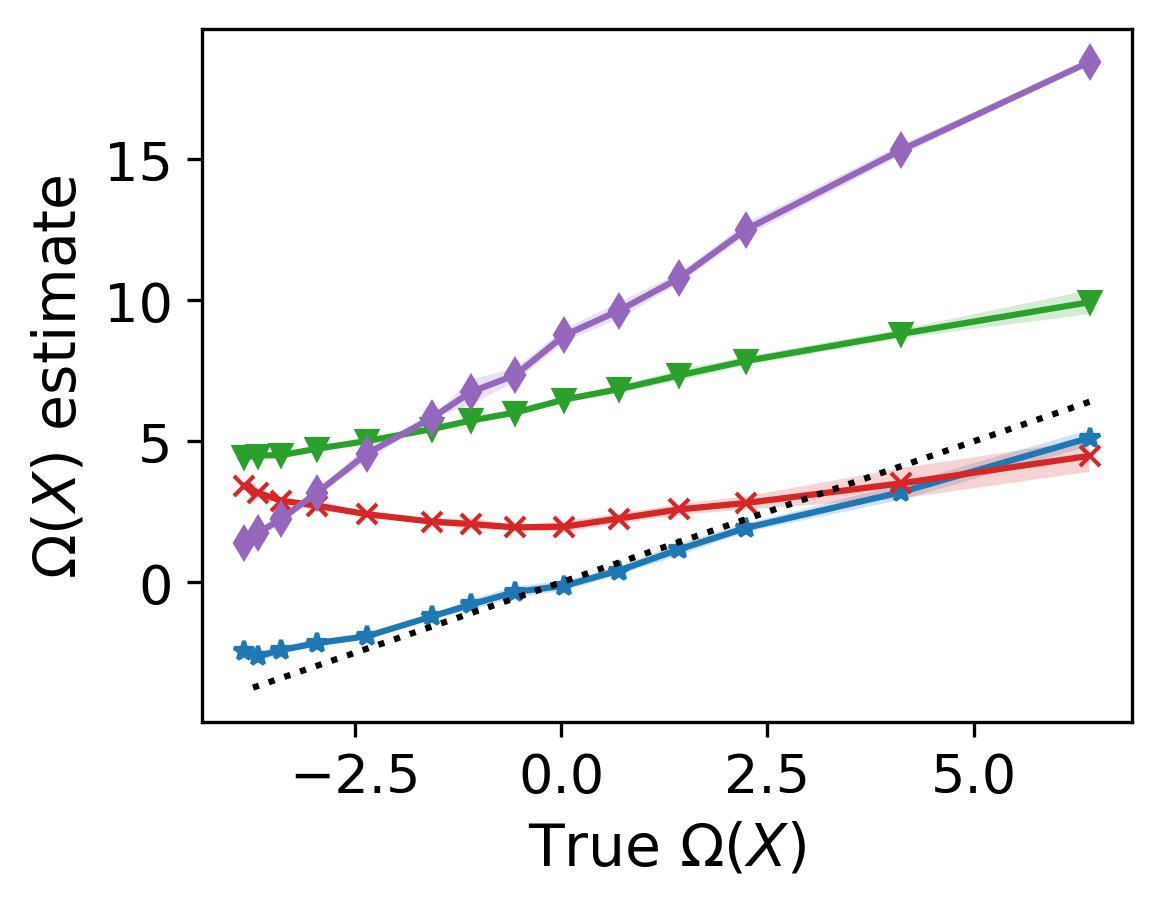}
         \caption{Dim=20}
     \end{subfigure}
      \caption{Mixed-interaction system with 10 variables, organized into 2 redundancy-dominant subsets of size $\{3,4\}$ variables and one synergy-dominant subset with $3$ variables. \acrshort{O-information} is modulated by fixing the synergy inter-dependency and increasing the redundancy. A \textbf{CDF} transformation is applied on-top of the multivariate-normal distribution.
      }
      \label{cdf_mix}

\end{figure}

\paragraph{CDF} 
The second transformation we consider is the application of a normal cumulative distribution function (CDF), which uniformizes the distribution margins (See \cite{czyz2023beyond}). In\Cref{cdf_red},\Cref{cdf_syn} and \Cref{cdf_mix}, we present the performance results of \gls{SOI} and alternatives on CDF-transformed benchmarks with a similar configuration used in \Cref{experiment}. Our method outperforms competitors, especially for high-dimensional variables. On the challenging synergistic benchmark, \gls{SOI} shows perfectible performance for very low \gls{O-information}, while competitors fail completely in this setting.

\section{Additional results }
\label{additional}
\subsection{Additional baseline}

\begin{figure} [h]

\centering
\begin{subfigure}{0.3\textwidth}
         \centering
    \includegraphics[page=1,width=\linewidth]{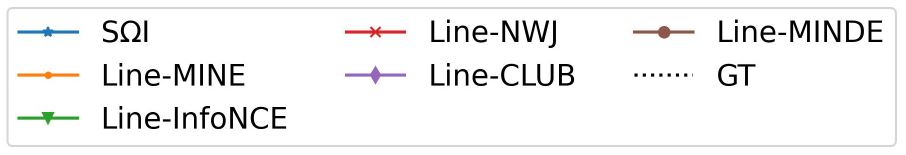}
    
     \end{subfigure}
     
     \begin{subfigure}{0.24\textwidth}
         \centering
  
         \includegraphics[page=1,width=\linewidth]{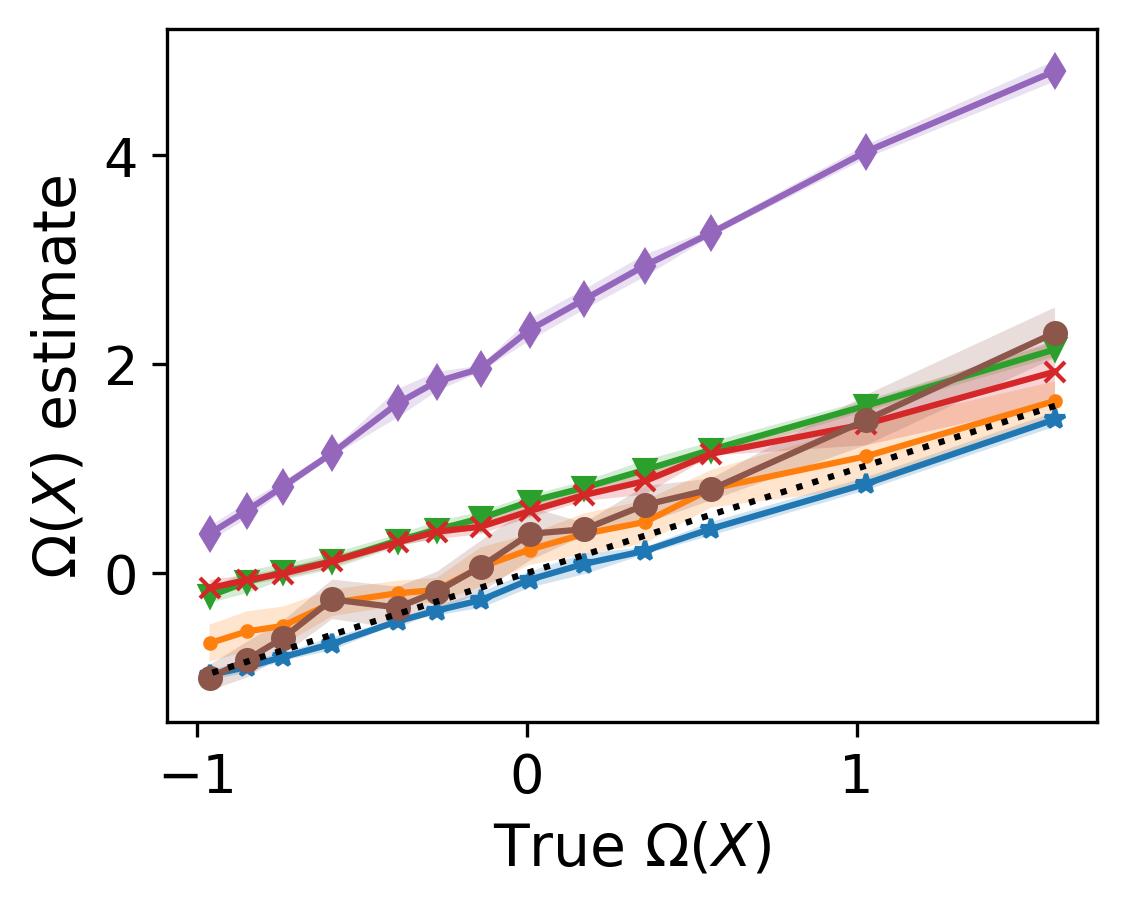}
         \caption{Dim=5}
     \end{subfigure}
      \begin{subfigure}{0.24\textwidth}
         \centering

         \includegraphics[page=1,width=\linewidth]{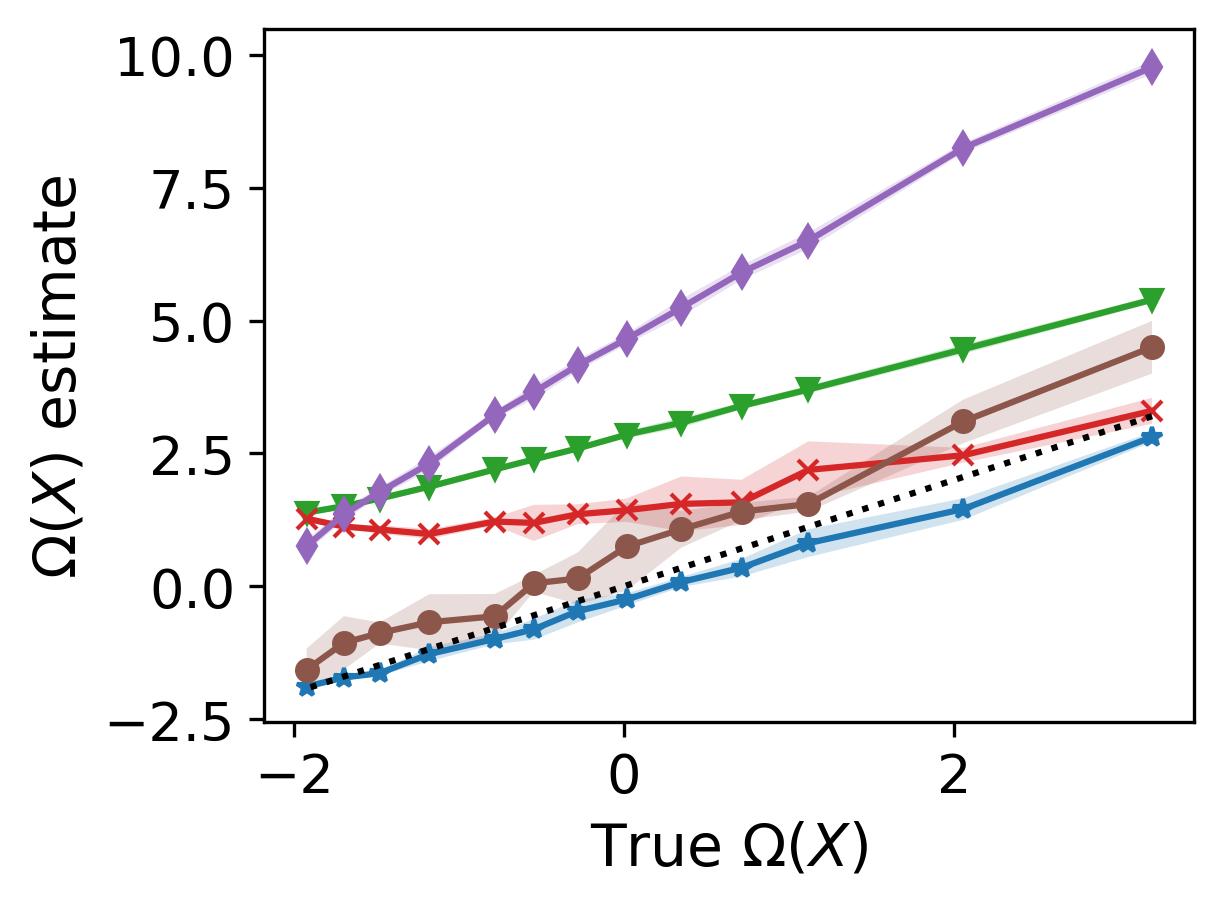}
         \caption{Dim=10}
     \end{subfigure}
     \begin{subfigure}{0.24\textwidth}
         \centering

         \includegraphics[page=1,width=\linewidth]{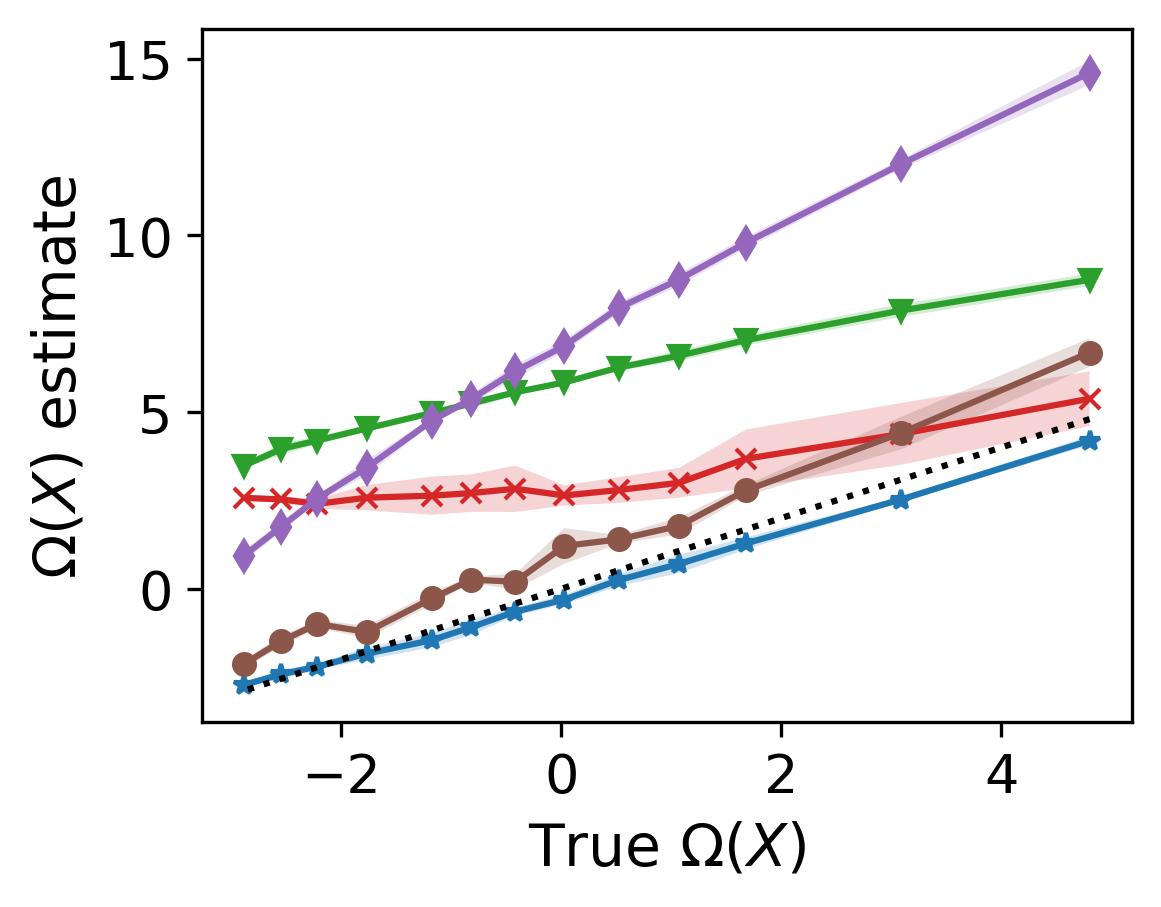}
         \caption{Dim=15}
     \end{subfigure}
         \begin{subfigure}{0.24\textwidth}
         \centering

         \includegraphics[page=1,width=\linewidth]{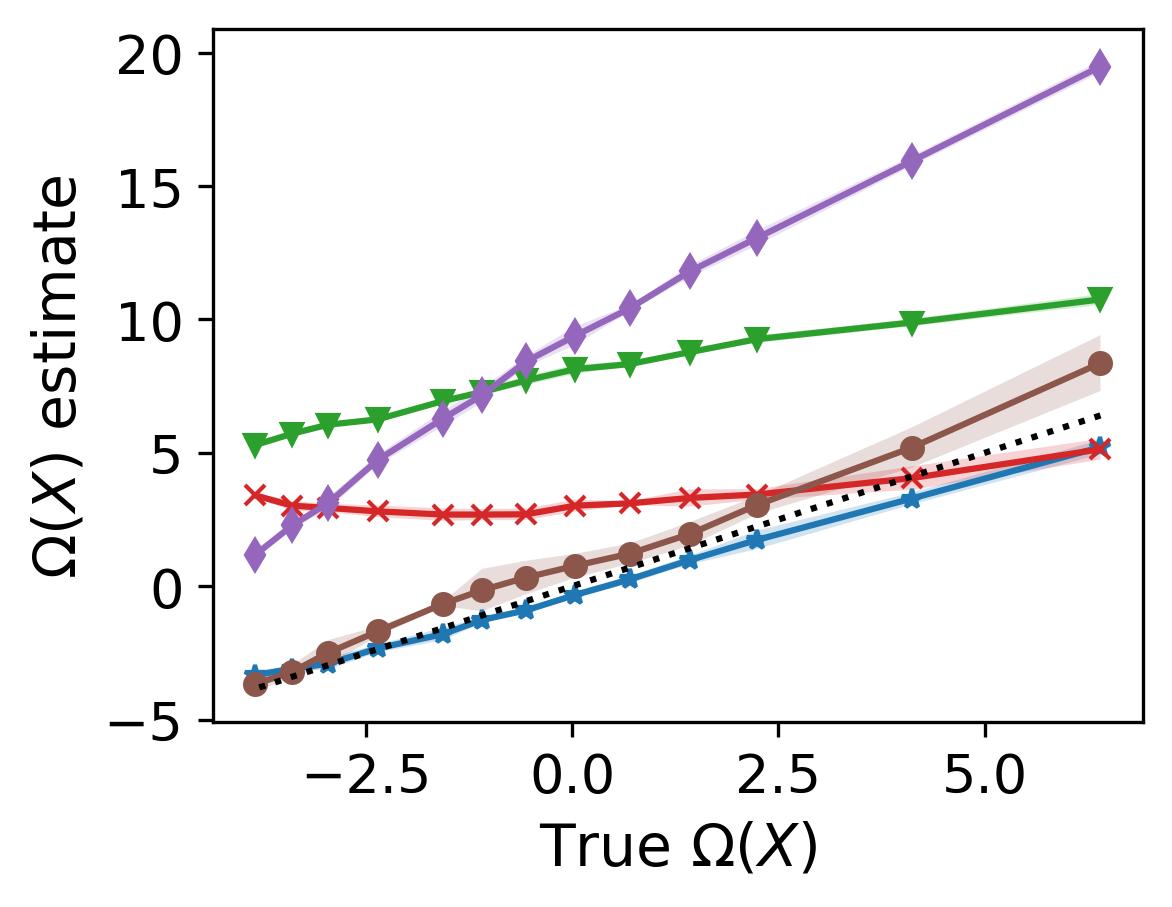}
         \caption{Dim=20}
     \end{subfigure}
      \caption{ \textbf{Additional Line-\acrshort{MINDE} \cite{franzese2023minde} baseline}.  Mixed-interaction system with 10 variables, organized into a redundancy-dominant subsets of size $3,4$ variables and one synergy-dominant subset with $3$ variables. \acrshort{O-information} is modulated by fixing the synergy inter-dependency and increasing the redundancy.
      }
      \label{fig:minde}

\end{figure}

 
\cite{franzese2023minde} have shown that the KL divergence between two distributions can be computed using the denoising score function enabling the proposition of an MI estimator.
In \Cref{fig:minde}, we present results on the mixed benchmark (redundancy and synergy) extended with the new baseline called Line-\acrshort{MINDE}, that computes O-information using the MI estimator from \cite{franzese2023minde}. Note that this approach requires learning a set of independent score models, one for each MI term: this increases the total number of parameters to learn,
resulting in a more computationally heavy training process compared to our proposed method. In these new experiments, we follow the authors hyper-parameters and score network architecture. We observe that while Line-\acrshort{MINDE} outperforms other pairwise MI based estimators, \gls{SOI} stands out with the best performance. Our findings indicate that the superiority of \gls{SOI} is due to efficiency of score based models in estimating information theoretic measures, which explains the superiority of \gls{SOI} and Line-\acrshort{MINDE} against other neural estimators. Secondly, the direct estimation of \gls{TC} and \gls{DTC} and the amortized training using a unique network is more efficient which explains why \gls{SOI} outperforms Line-\acrshort{MINDE}.

\subsection{Ablation study}

\subsubsection{Data size}
\begin{figure} [h]

\centering
\begin{subfigure}{0.3\textwidth}
         \centering
    \includegraphics[page=1,width=\linewidth]{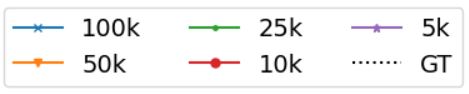}
    
     \end{subfigure}
     
     \begin{subfigure}{0.24\textwidth}
         \centering
  
         \includegraphics[page=1,width=\linewidth]{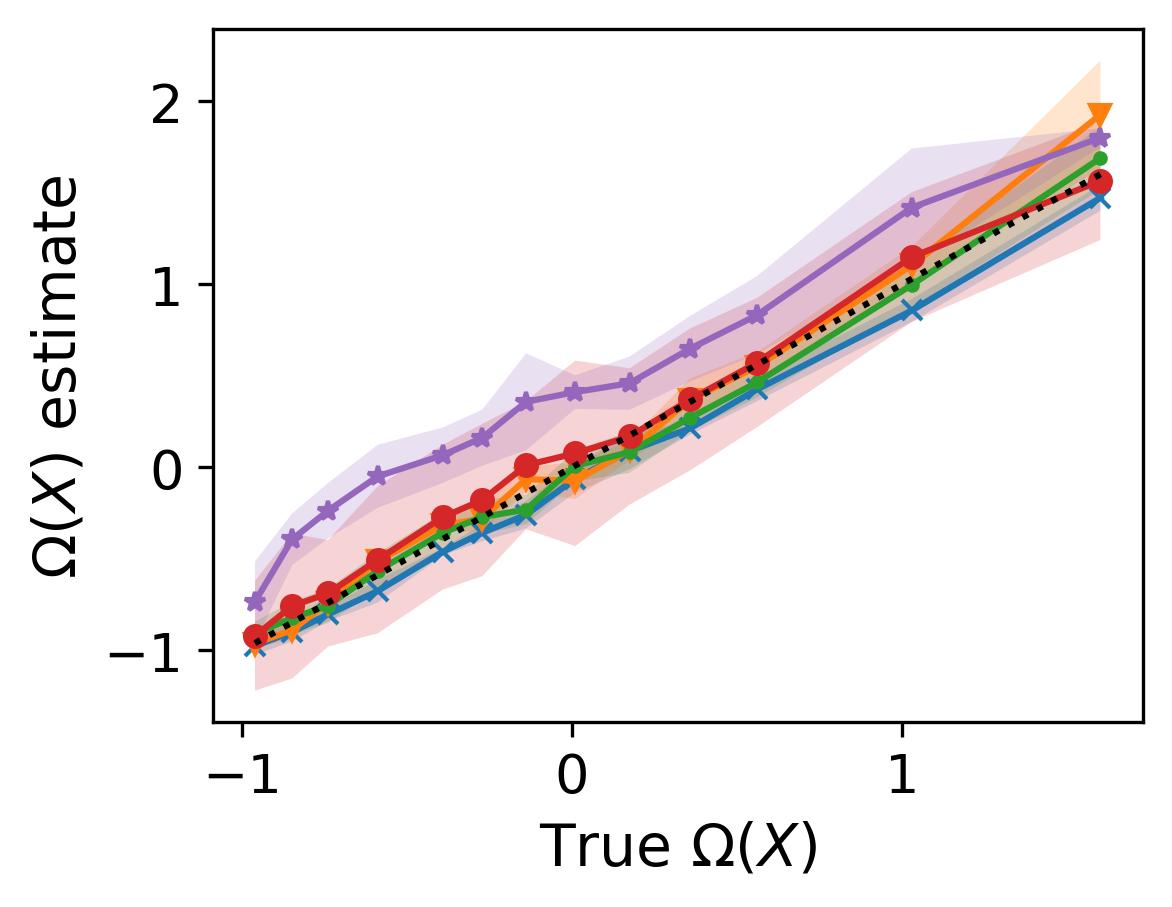}
         \caption{Dim=5}
     \end{subfigure}
      \begin{subfigure}{0.24\textwidth}
         \centering

         \includegraphics[page=1,width=\linewidth]{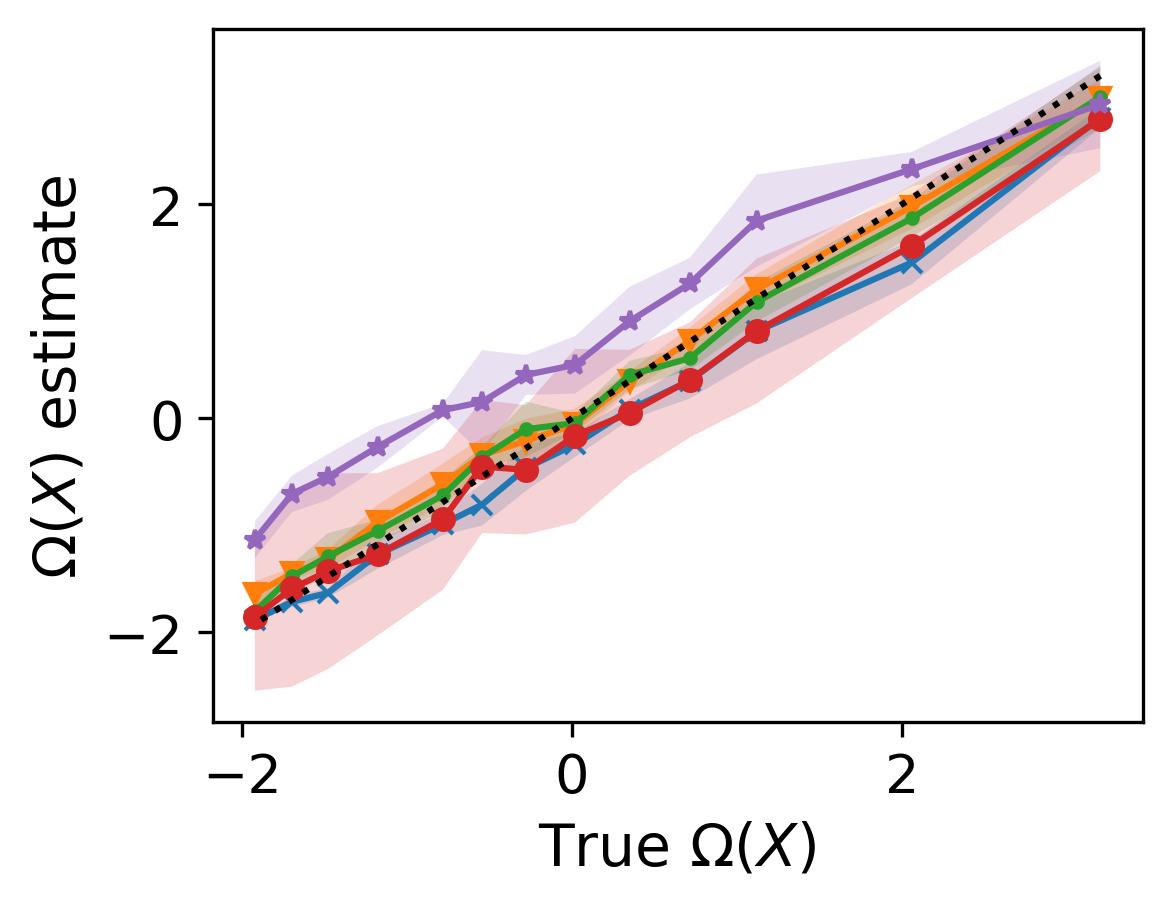}
         \caption{Dim=10}
     \end{subfigure}
     \begin{subfigure}{0.24\textwidth}
         \centering

         \includegraphics[page=1,width=\linewidth]{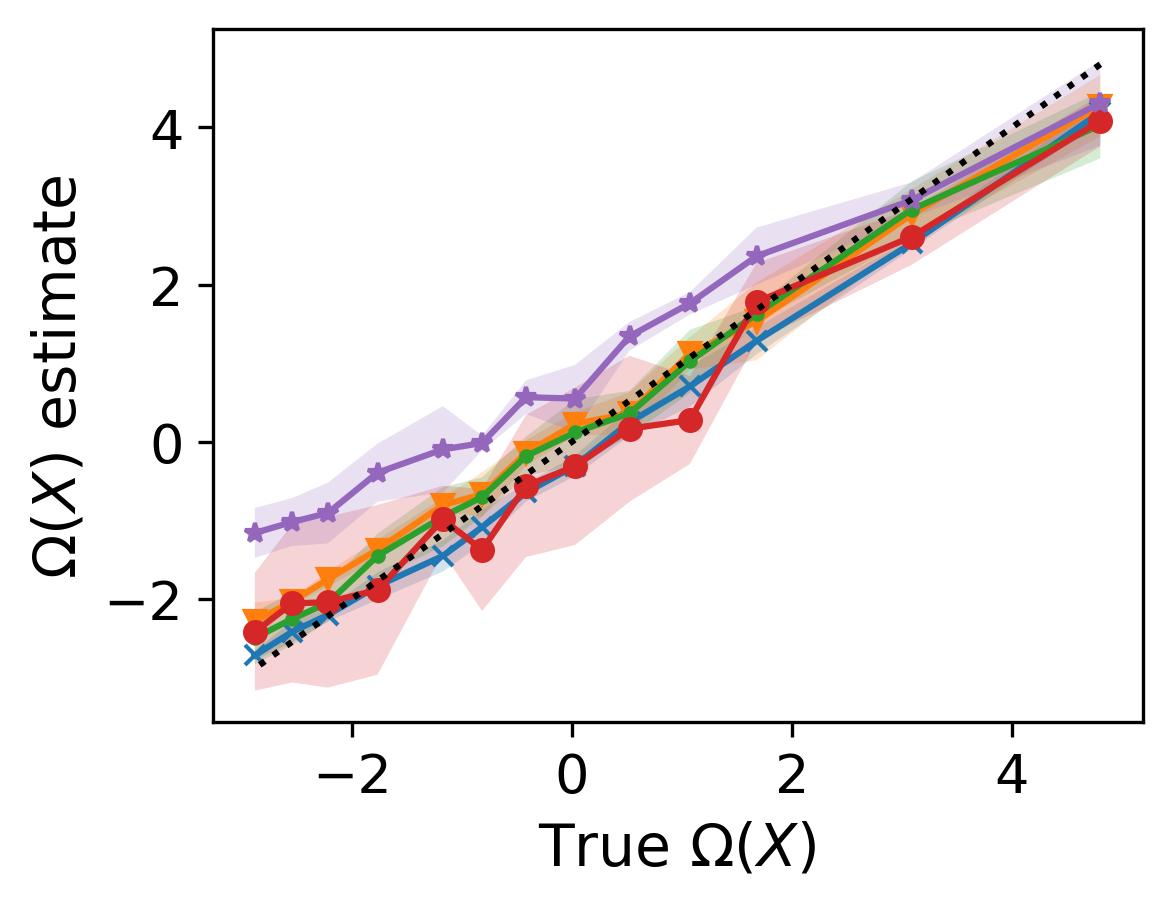}
         \caption{Dim=15}
     \end{subfigure}
         \begin{subfigure}{0.24\textwidth}
         \centering

         \includegraphics[page=1,width=\linewidth]{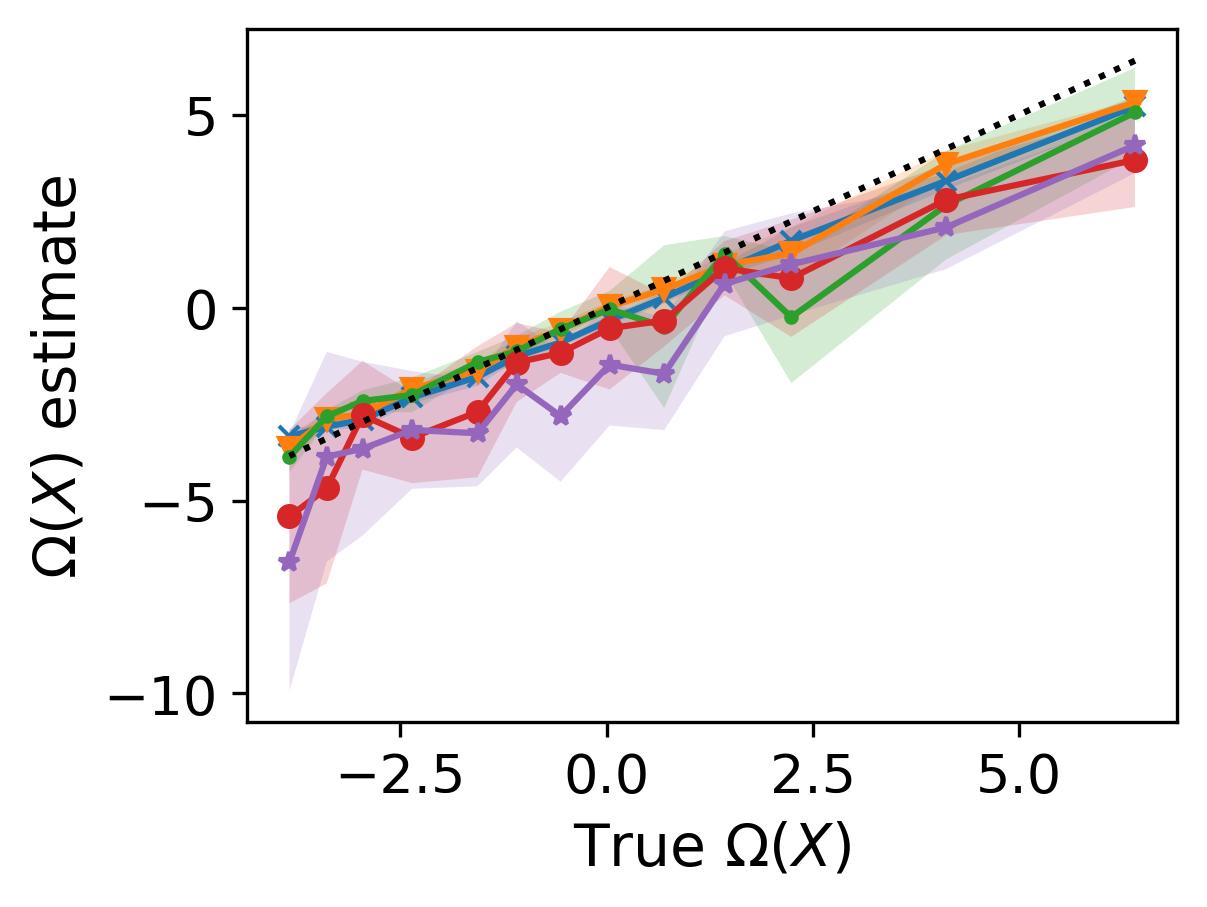}
         \caption{Dim=20}
     \end{subfigure}
      \caption{ \gls{SOI} training size ablation study : 100k,50k,25k,10k,5k. We use a test size of 10k for all the settings except when the train set size is equal to 5k where we use test size of similar size. The considered benchmark is a mixed-interaction system with 10 variables, organized into a redundancy-dominant subsets of size $3,4$ variables and one synergy-dominant subset with $3$ variables. \acrshort{O-information} is modulated by fixing the synergy inter-dependency and increasing the redundancy.
      }
\label{data_size}
\end{figure}

In \Cref{data_size}, we present a training size ablation study on the mixed benchmark. The considered number of training samples are of 5k,10k,25k,50k,100k samples. We fix the testset to 10k samples, except when the training size is 5k, for which we use 5k test samples. We observe that for data size  superior to 10k, \gls{SOI} obtains very good estimates in terms of bias and variance; when the training size has 10k samples, \gls{SOI} estimates have increased variance; when we use only 5k training samples, \gls{SOI} have increased bias. These results are to be expected, since neural estimators, in general, require sufficient training data to shine.

\subsubsection{Number of training iterations}
\begin{figure} [H]

\centering
     
     \begin{subfigure}{0.24\textwidth}
         \centering
  
         \includegraphics[page=1,width=\linewidth]{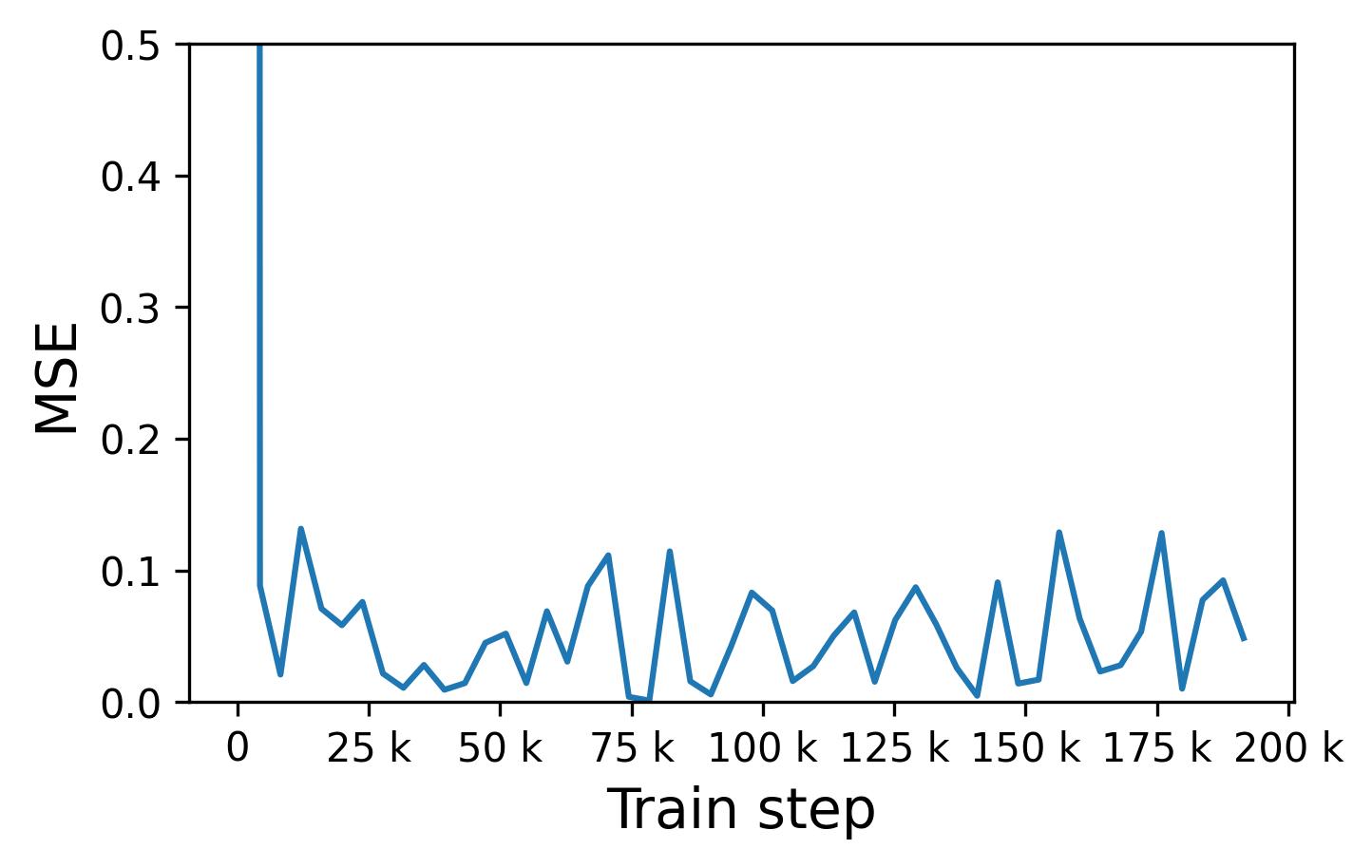}
         \caption{Dim=5}
     \end{subfigure}
      \begin{subfigure}{0.24\textwidth}
         \centering

         \includegraphics[page=1,width=\linewidth]{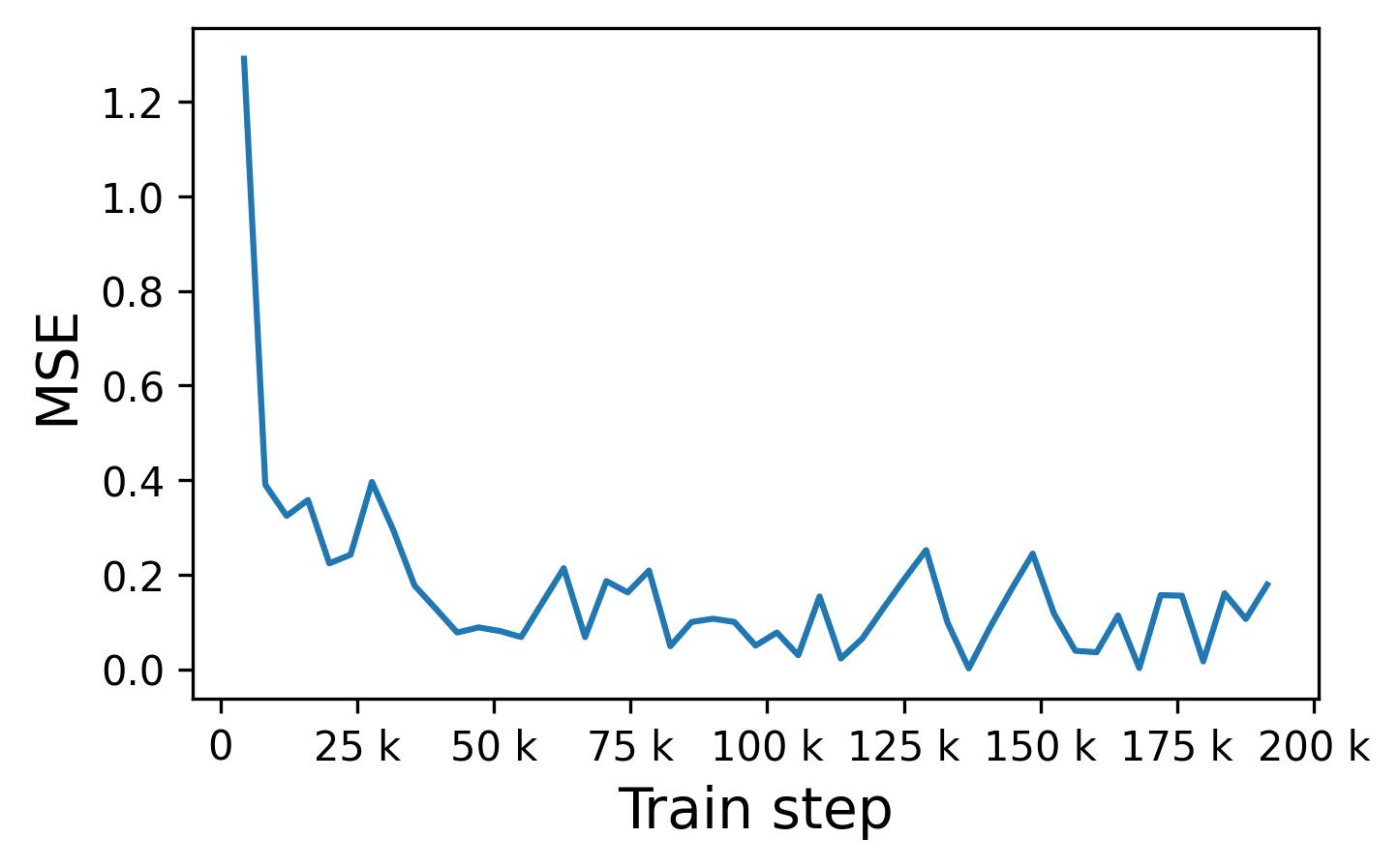}
         \caption{Dim=10}
     \end{subfigure}
     \begin{subfigure}{0.24\textwidth}
         \centering

         \includegraphics[page=1,width=\linewidth]{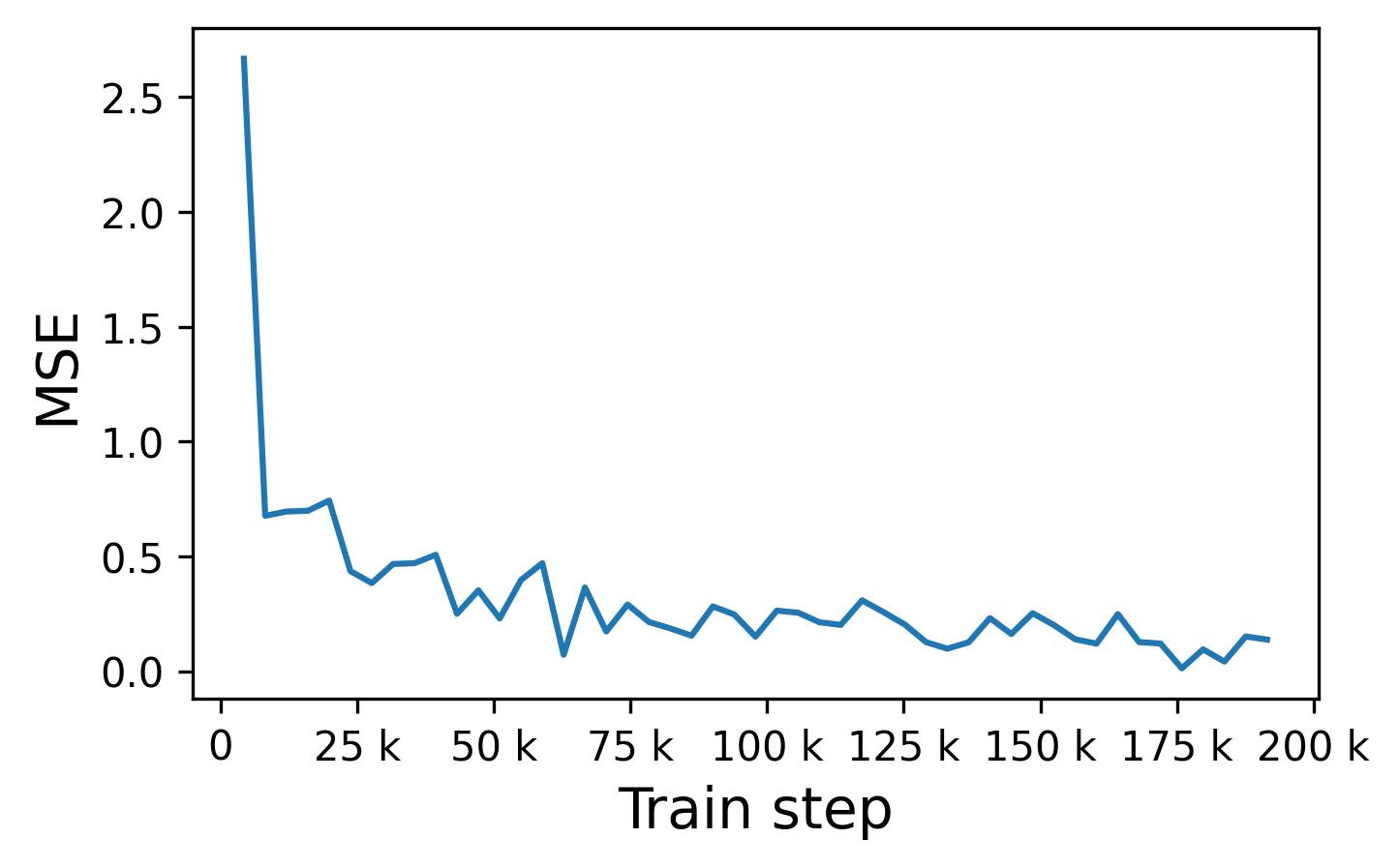}
         \caption{Dim=15}
     \end{subfigure}
         \begin{subfigure}{0.24\textwidth}
         \centering

         \includegraphics[page=1,width=\linewidth]{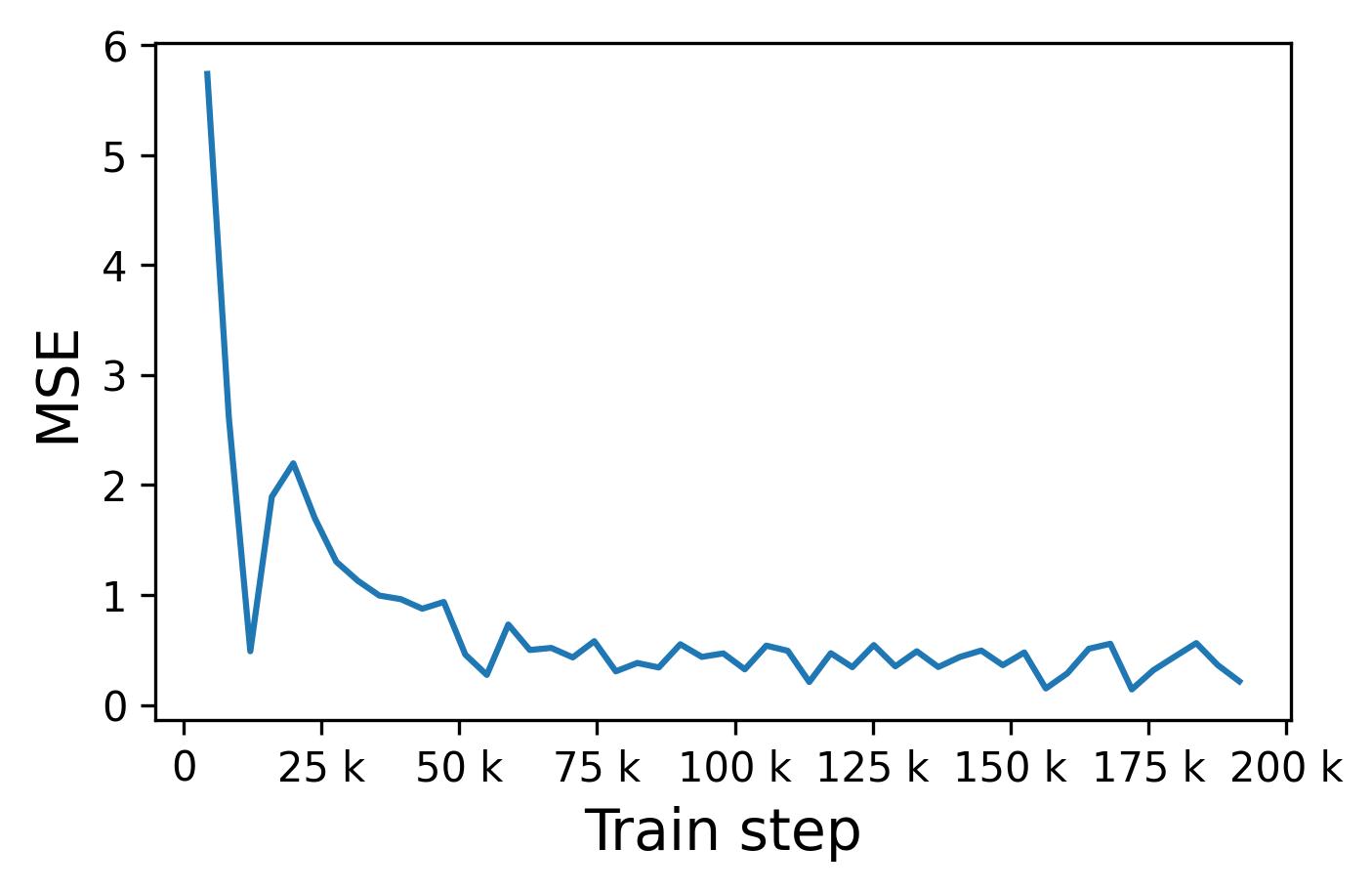}
         \caption{Dim=20}
     \end{subfigure}

        \begin{subfigure}{0.24\textwidth}
         \centering
  
         \includegraphics[page=1,width=\linewidth]{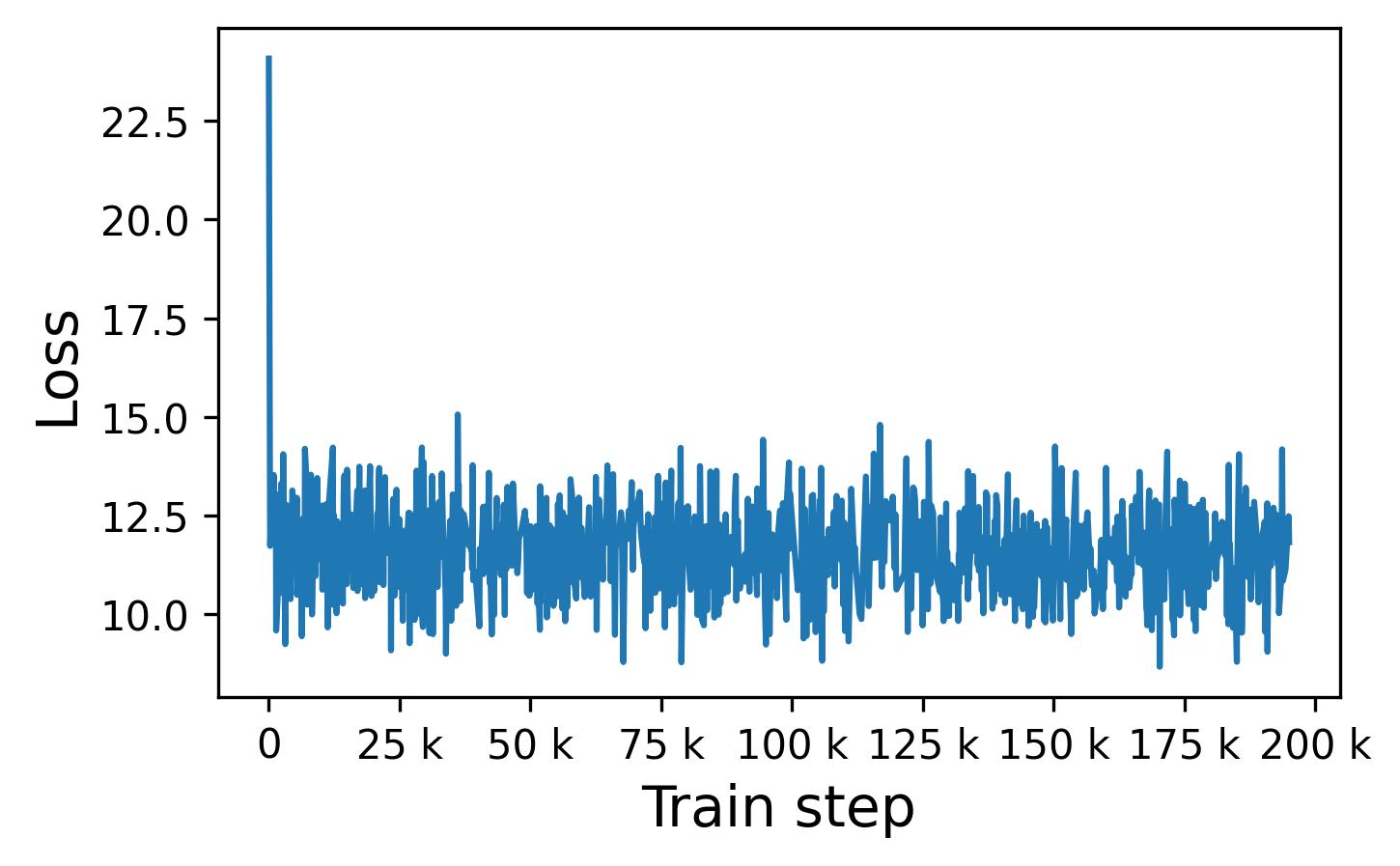}
         \caption{Dim=5}
     \end{subfigure}
      \begin{subfigure}{0.24\textwidth}
         \centering

         \includegraphics[page=1,width=\linewidth]{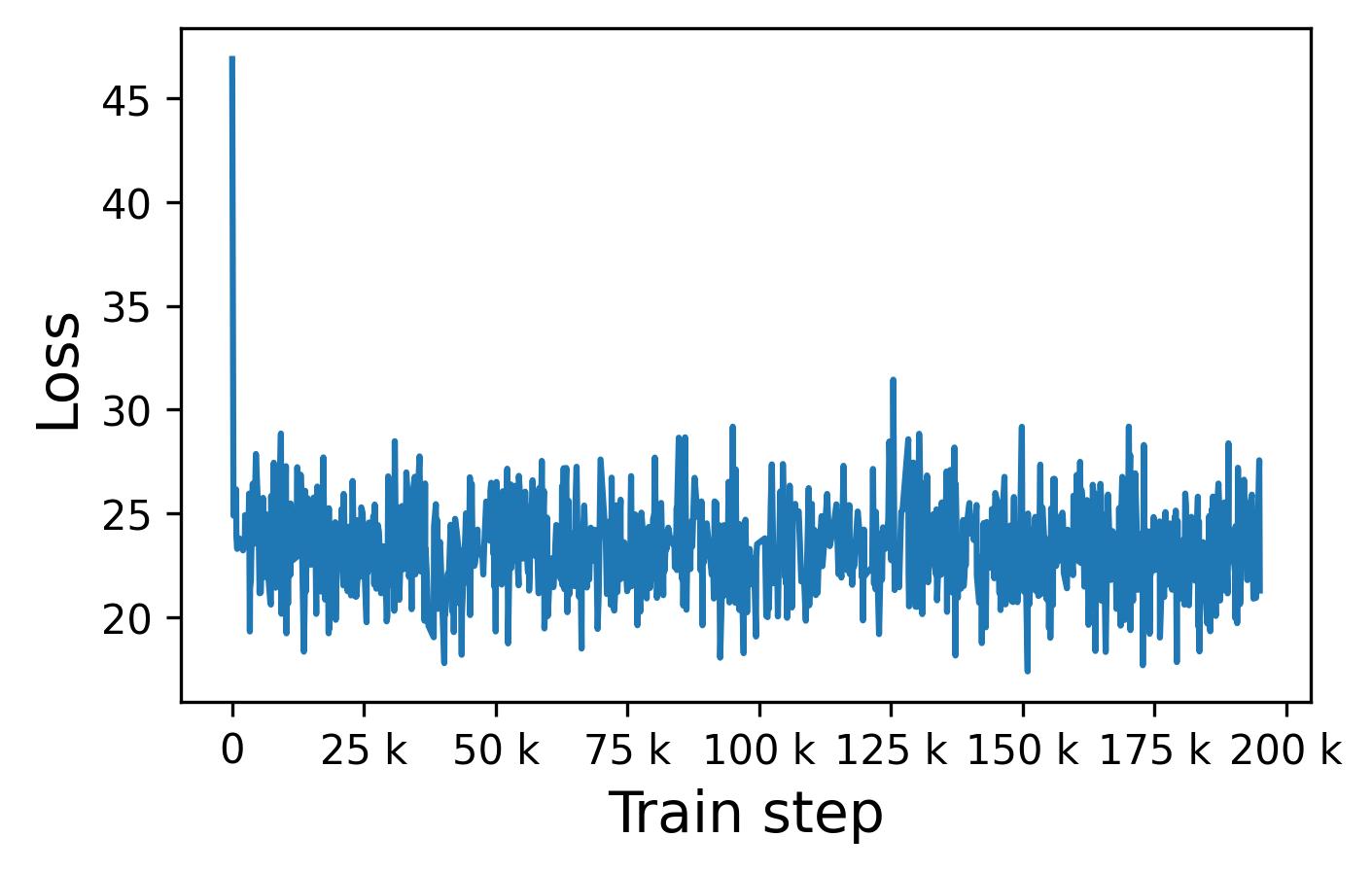}
         \caption{Dim=10}
     \end{subfigure}
     \begin{subfigure}{0.24\textwidth}
         \centering

         \includegraphics[page=1,width=\linewidth]{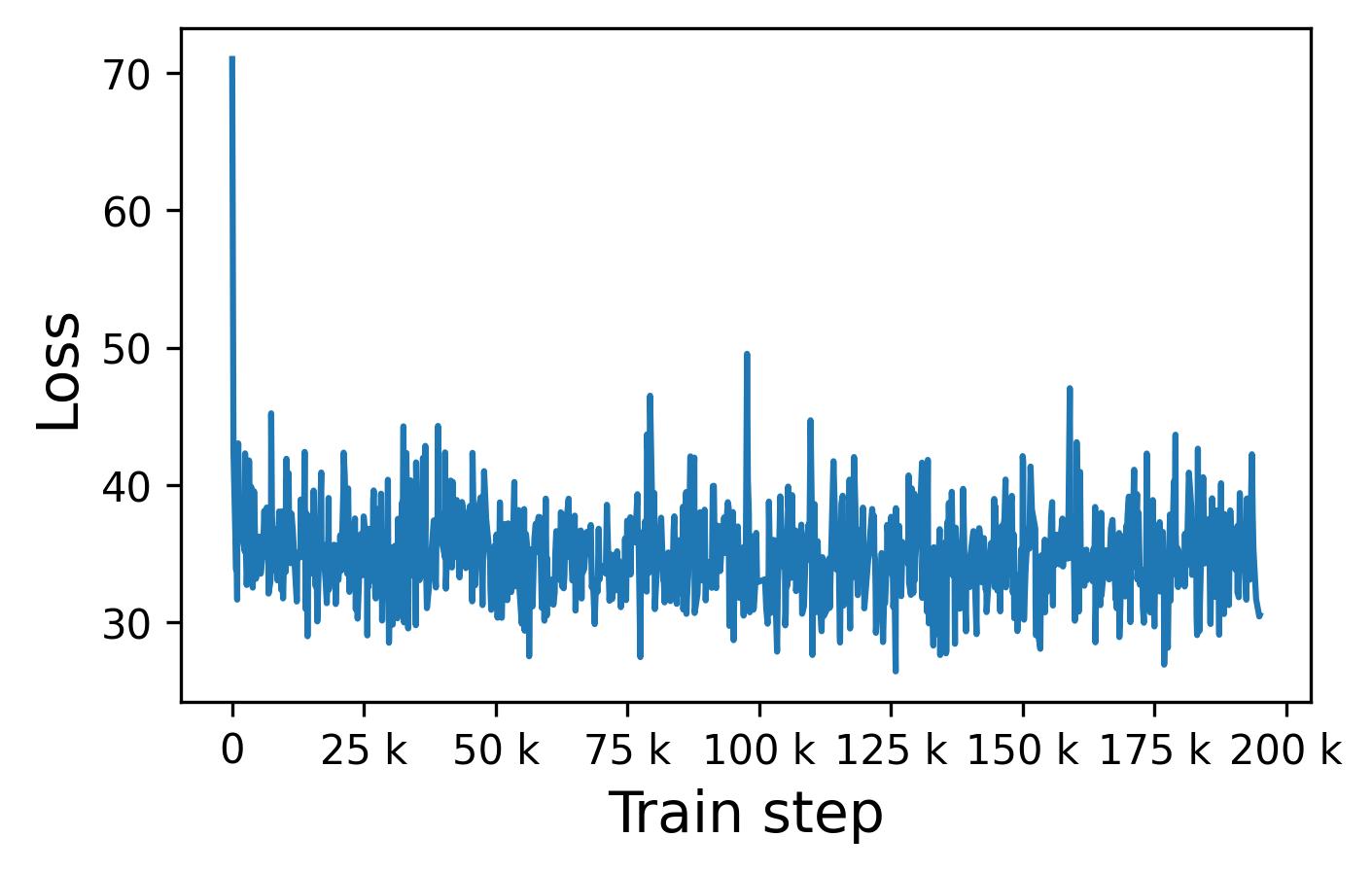}
         \caption{Dim=15}
     \end{subfigure}
         \begin{subfigure}{0.24\textwidth}
         \centering

         \includegraphics[page=1,width=\linewidth]{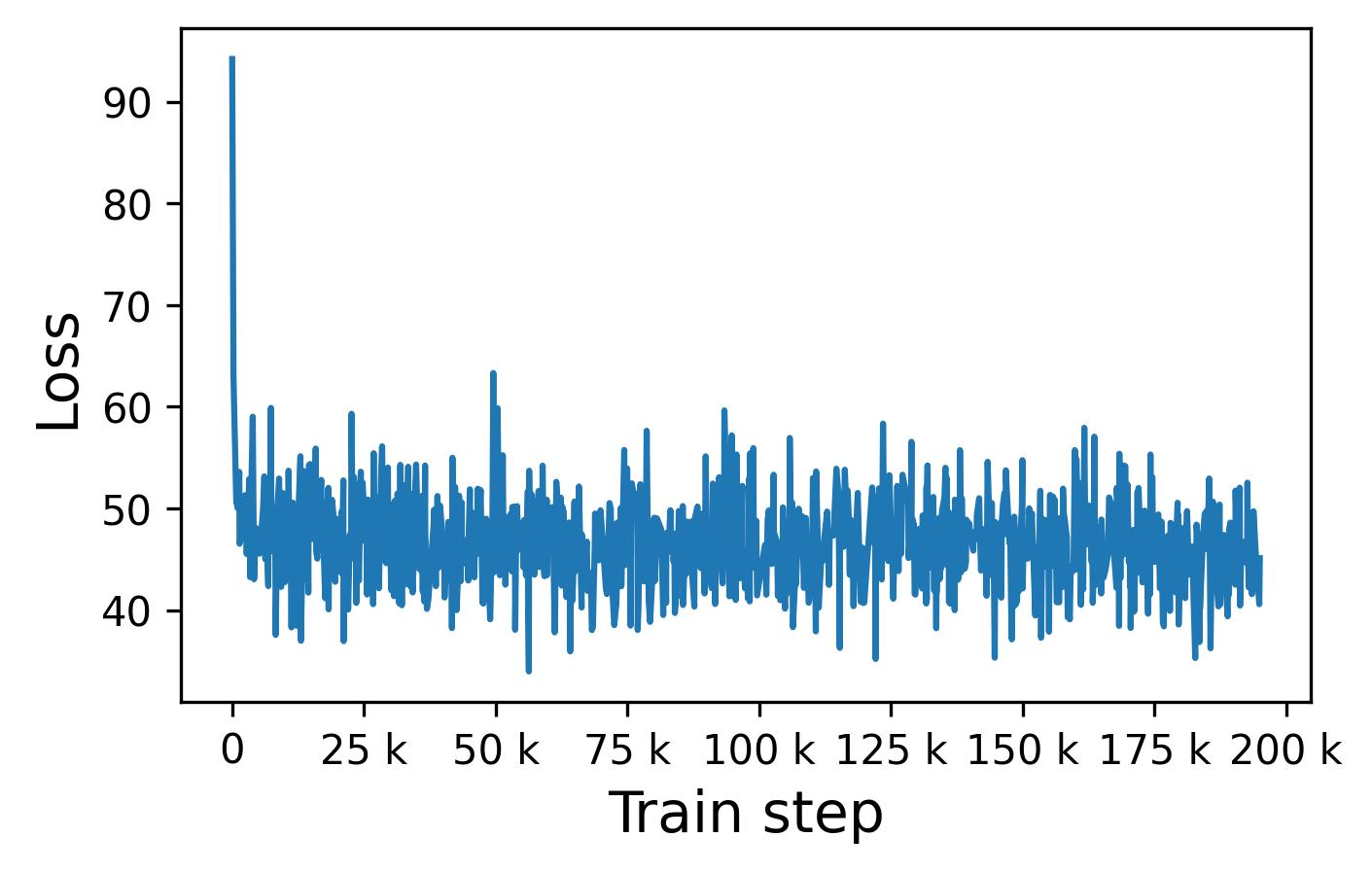}
         \caption{Dim=20}
     \end{subfigure}
      \caption{ \textbf{Training Loss curve Vs Estimation of \gls{O-information} MSE}.  Mixed-interaction system with 10 variables, organized into a redundancy-dominant subsets of size $3,4$ variables and one synergy-dominant subset with $3$ variables. For different benchmark dimensions, we report: \textbf{Top:} \gls{O-information}  estimation mean square error as a function of the training iterations. \textbf{Bottom:} Training loss curve.
      }
      \label{trainingloss}

\end{figure}

In \Cref{trainingloss}, we present the training curves contrasted with \gls{MI} estimate mean squared error. Clearly, the number of iterations required to achieve satisfactory results depends on the dataset complexity.

\subsubsection{Monte Carlo integration steps}
\begin{figure} [H]

\centering
     
     \begin{subfigure}{0.24\textwidth}
         \centering
  
         \includegraphics[page=1,width=\linewidth]{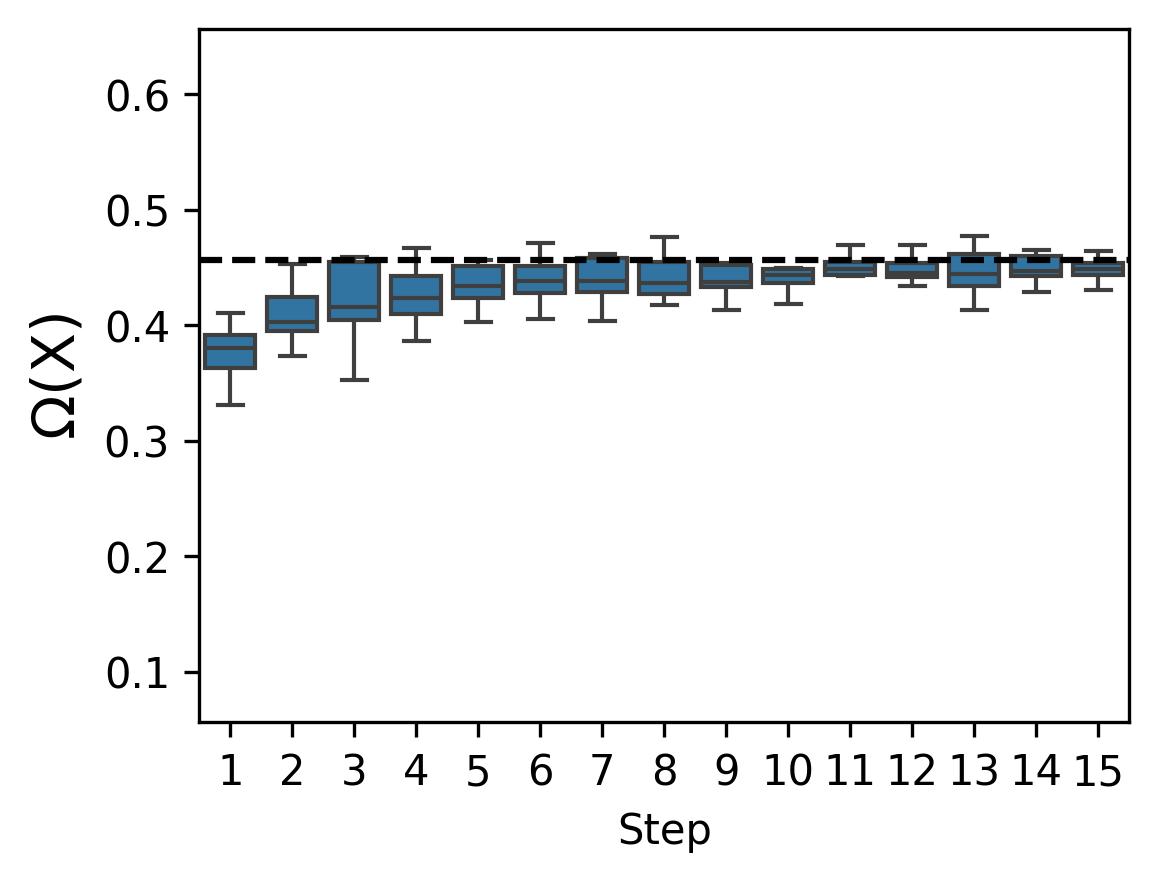}
         \caption{Dim=5}
     \end{subfigure}
      \begin{subfigure}{0.24\textwidth}
         \centering

         \includegraphics[page=1,width=\linewidth]{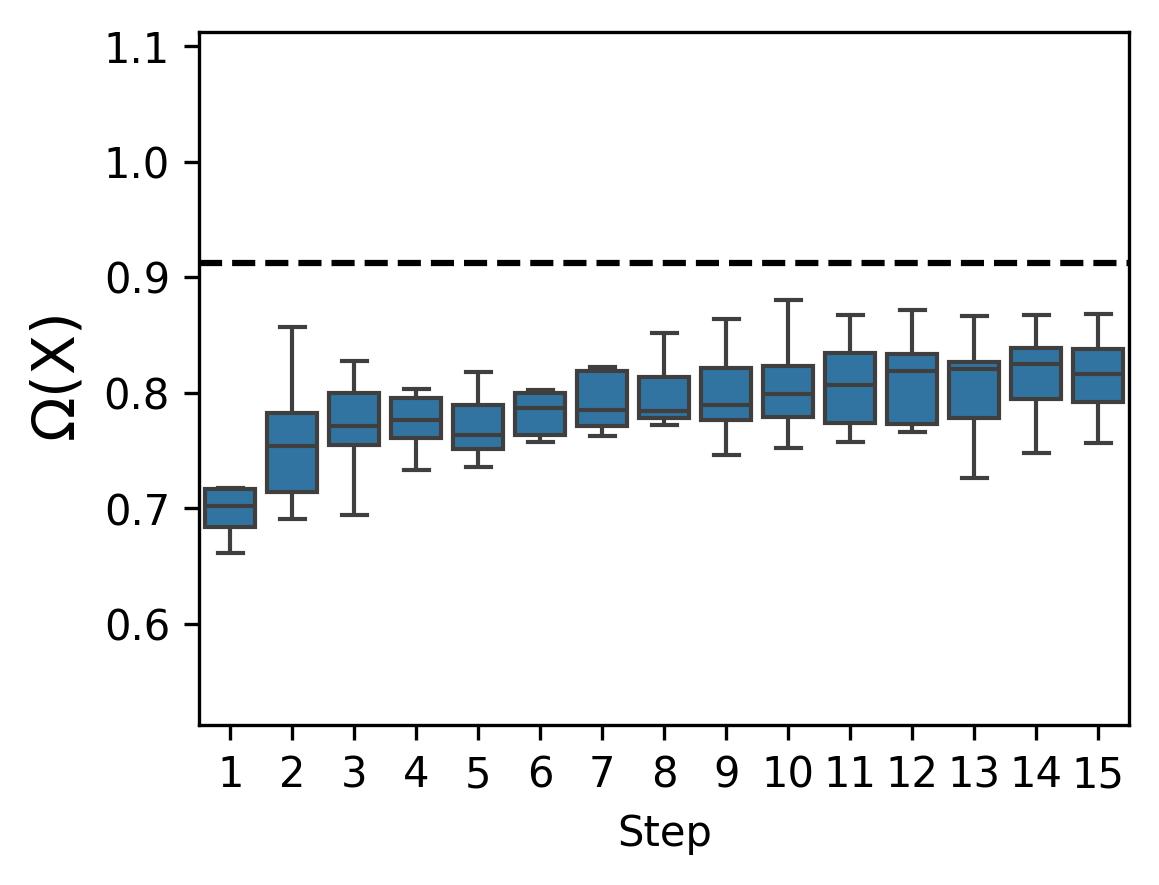}
         \caption{Dim=10}
     \end{subfigure}
     \begin{subfigure}{0.24\textwidth}
         \centering

         \includegraphics[page=1,width=\linewidth]{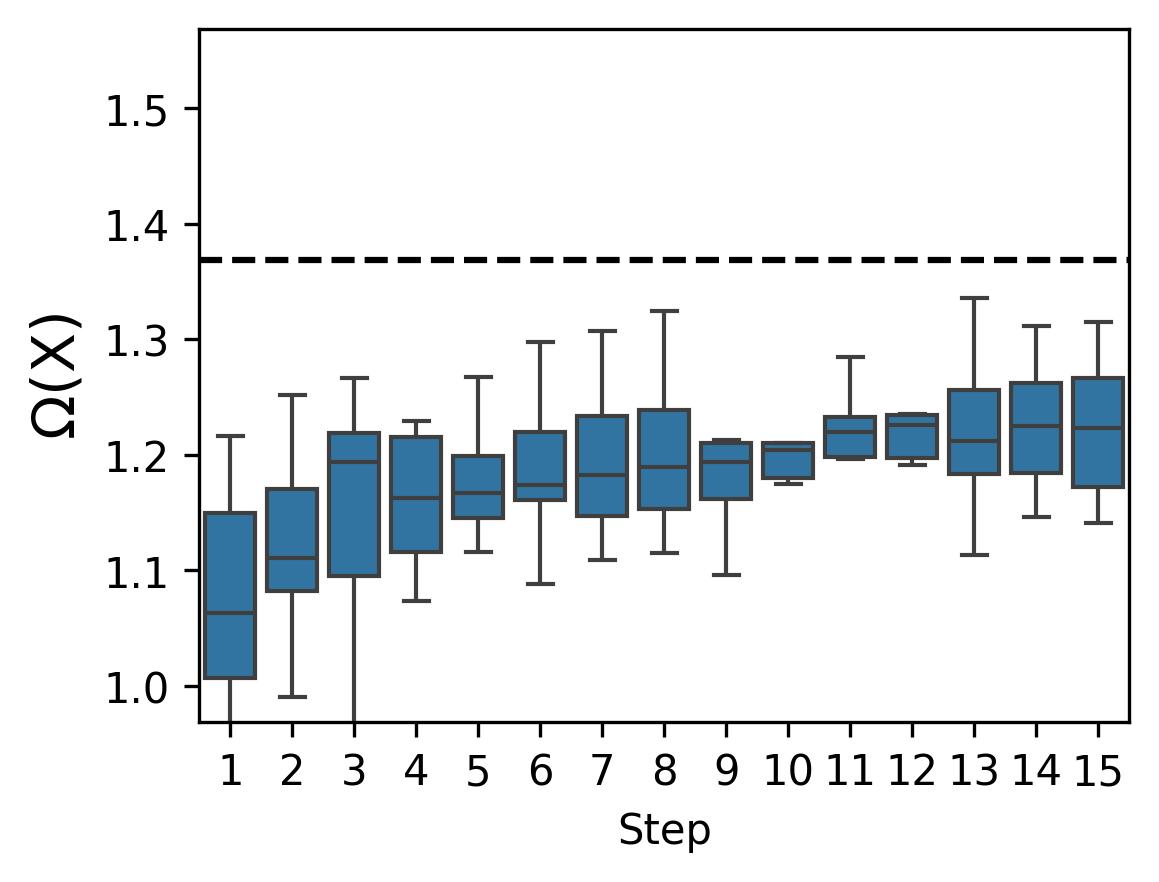}
         \caption{Dim=15}
     \end{subfigure}
         \begin{subfigure}{0.24\textwidth}
         \centering

         \includegraphics[page=1,width=\linewidth]{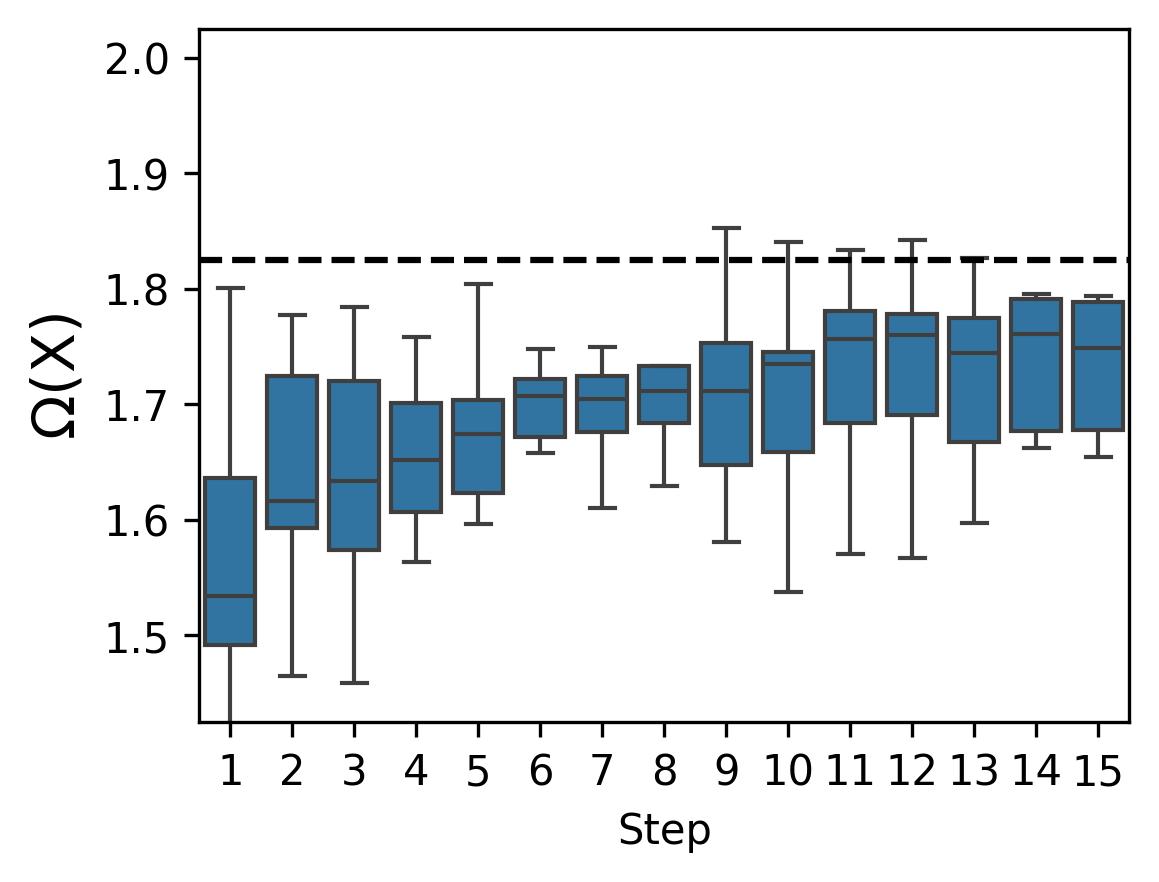}
         \caption{Dim=20}
     \end{subfigure}

      \caption{ \textbf{Estimation of \gls{O-information} as a function of Monte Carlo Averaging steps run over 10 seeds}.  Mixed-interaction system with 10 variables, organized into a redundancy-dominant subsets of size $3,4$ variables and one synergy-dominant subset with $3$ variables. Dashed line represents ground truth \gls{O-information}.
      }
      \label{mc_steps}

\end{figure}

In \Cref{mc_steps}, we present an ablation on the number of Monte Carlo steps, for the case of a mixed (redundancy and synergy) benchmark with $N=$ 10 random variables. We notice that an increased number of steps improves the estimation variance and bias. Naturally, this depends on the data dimension and complexity.

\subsection{Additional synthetic experiments}

\begin{figure} [h]
\centering
\begin{subfigure}{0.3\textwidth}
         \centering
\includegraphics[page=1,width=\linewidth]{assets/figures/exp_red/legend.PNG}
     \end{subfigure}

     \begin{subfigure}{0.24\textwidth}
         \centering

         \includegraphics[page=1,width=\linewidth]{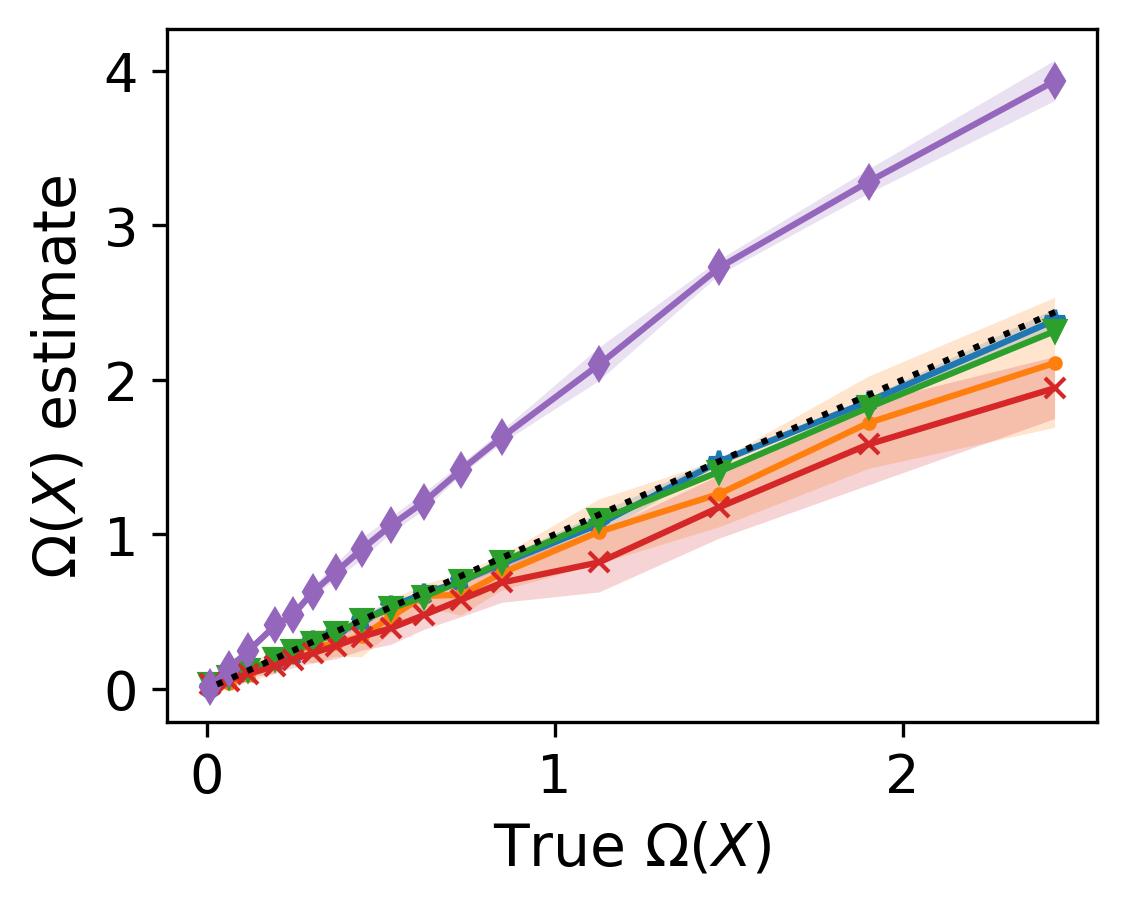}
         \caption{Dim=5}
     \end{subfigure}
      \begin{subfigure}{0.24\textwidth}
         \centering
  
         \includegraphics[page=1,width=\linewidth]{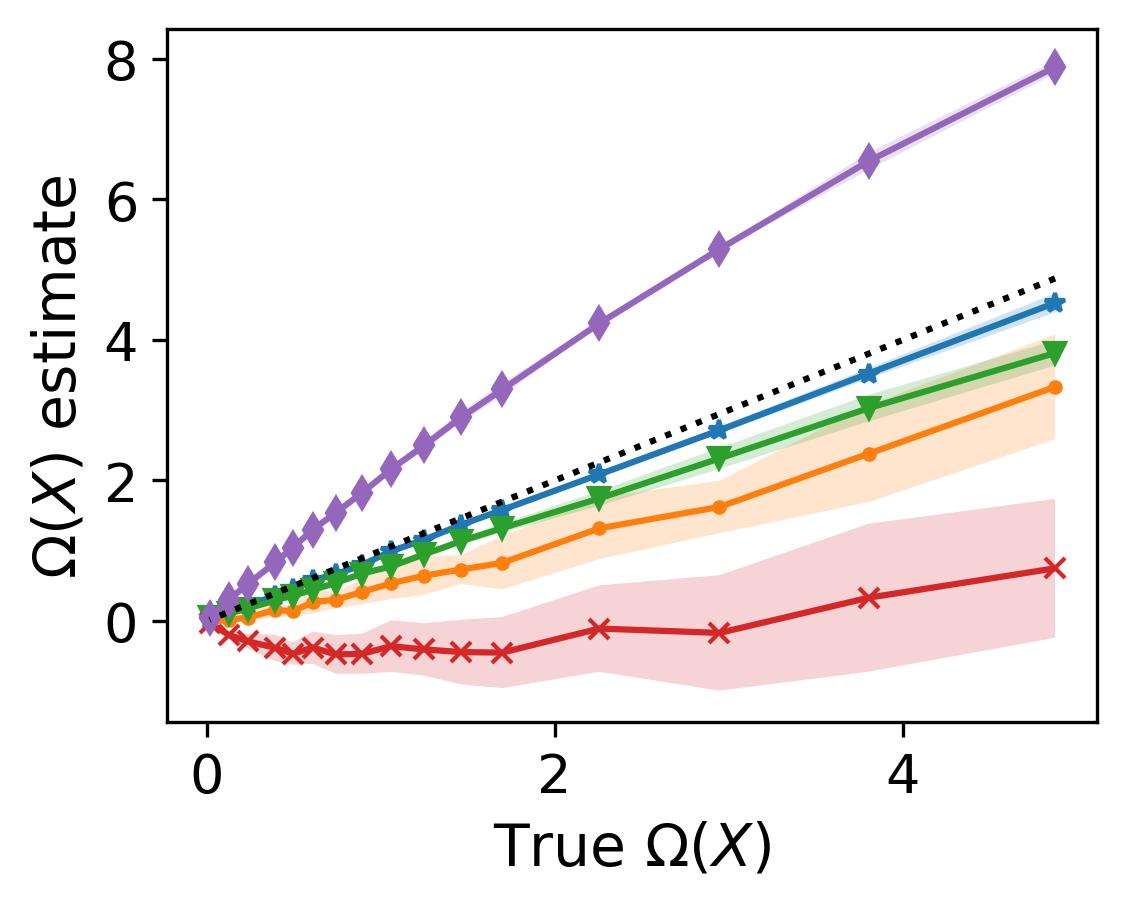}
         \caption{Dim=10}
     \end{subfigure}
     \begin{subfigure}{0.24\textwidth}
         \centering
    
         \includegraphics[page=1,width=\linewidth]{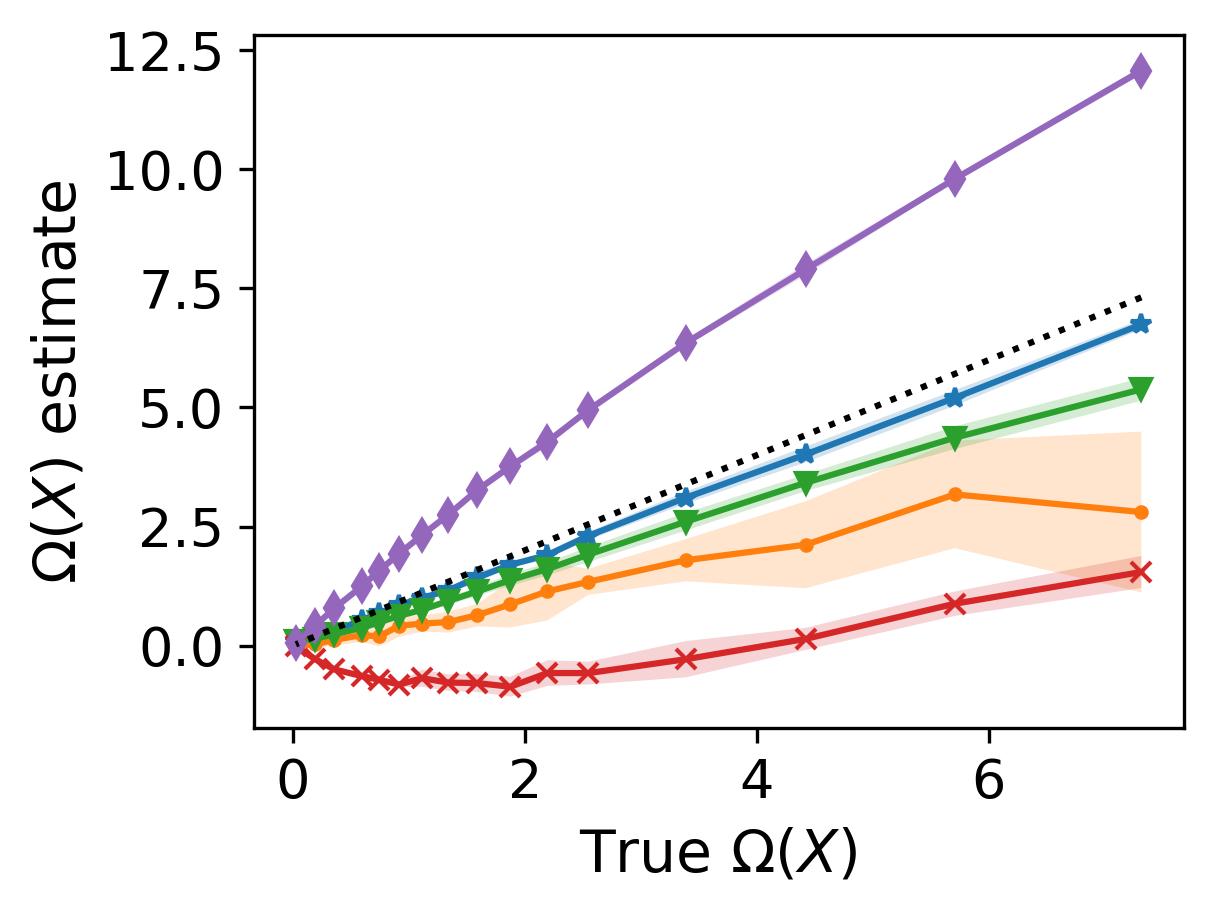}
         \caption{Dim=15}
     \end{subfigure}
         \begin{subfigure}{0.24\textwidth}
         \centering

         \includegraphics[page=1,width=\linewidth]{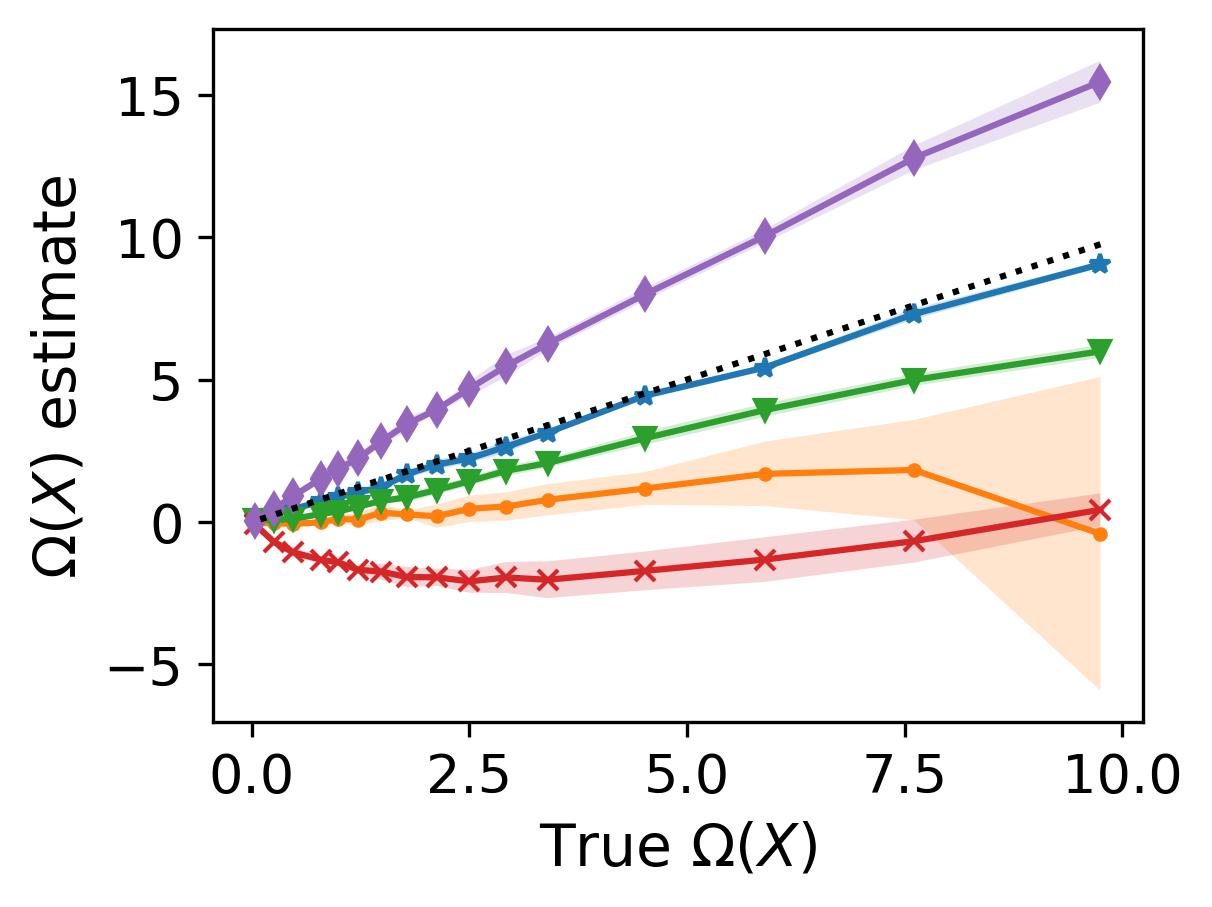}
         \caption{Dim=20}
     \end{subfigure}
      \caption{ Redundant system with 6 variables, organized into subsets of sizes $\{3,3\}$ and increasing interaction strength.
      }

\end{figure}

\begin{figure} [h]

\centering
\begin{subfigure}{0.3\textwidth}
         \centering
    \includegraphics[page=1,width=\linewidth]{assets/figures/exp_red/legend.PNG}
    
     \end{subfigure}
     
     \begin{subfigure}{0.24\textwidth}
         \centering
 
         \includegraphics[page=1,width=\linewidth]{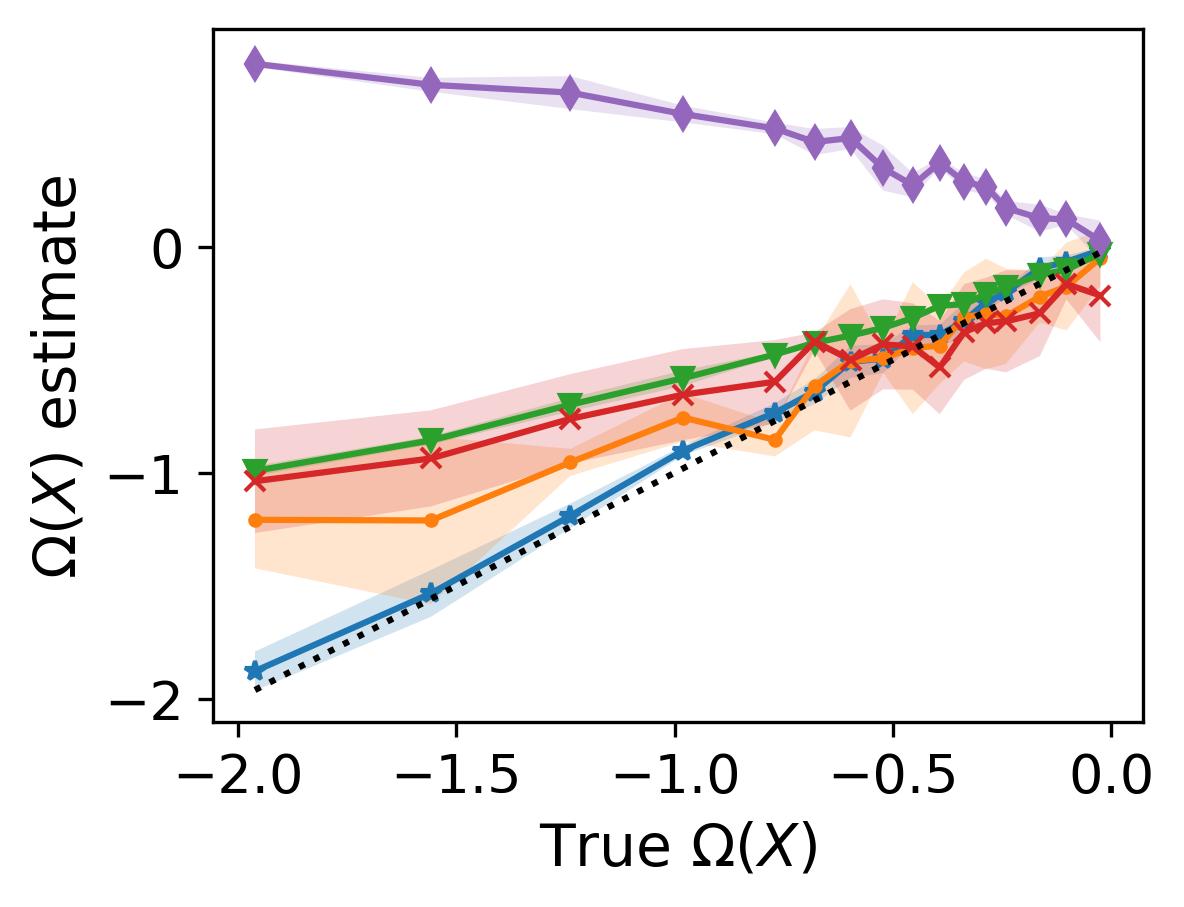}
         \caption{Dim=5}
     \end{subfigure}
      \begin{subfigure}{0.24\textwidth}
         \centering

         \includegraphics[page=1,width=\linewidth]{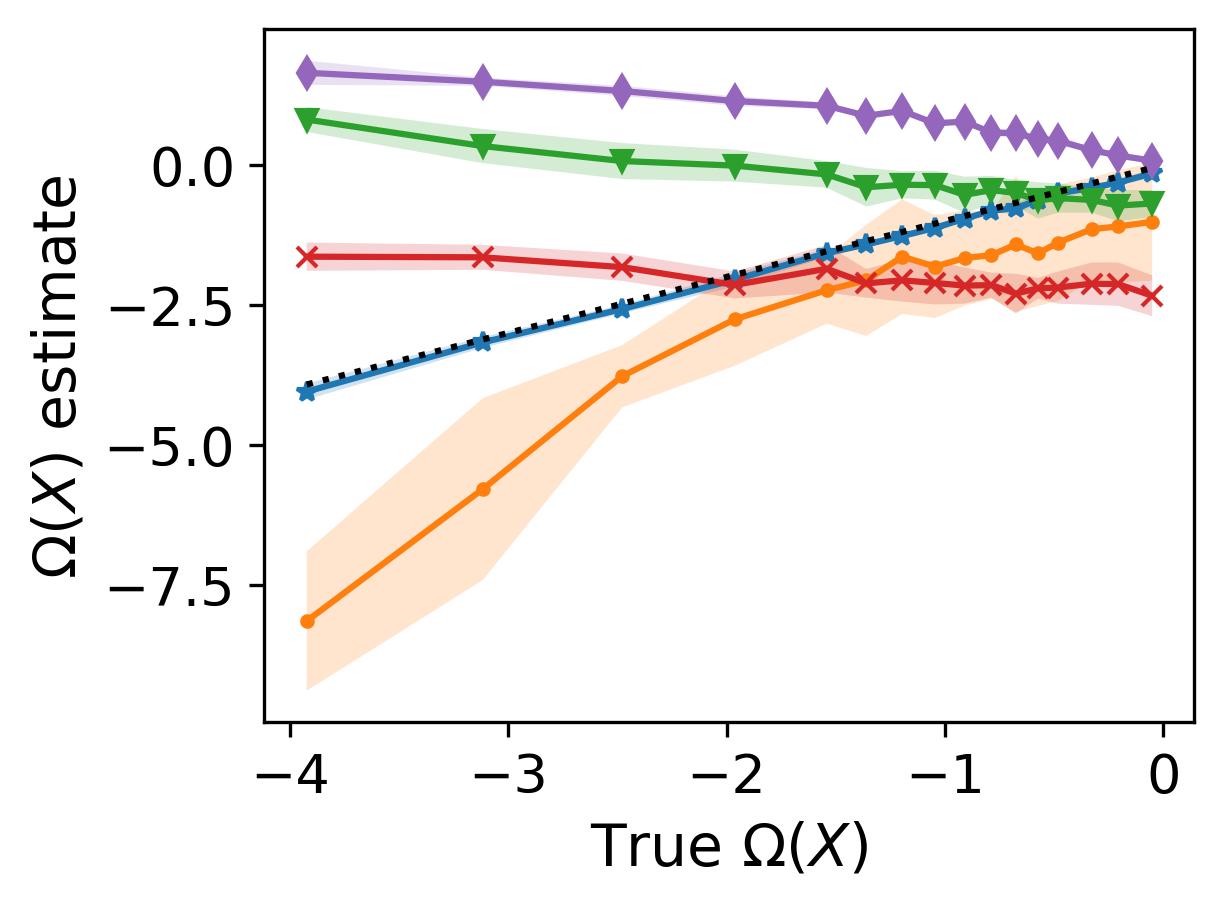}
         \caption{Dim=10}
     \end{subfigure}
     \begin{subfigure}{0.24\textwidth}
         \centering

         \includegraphics[page=1,width=\linewidth]{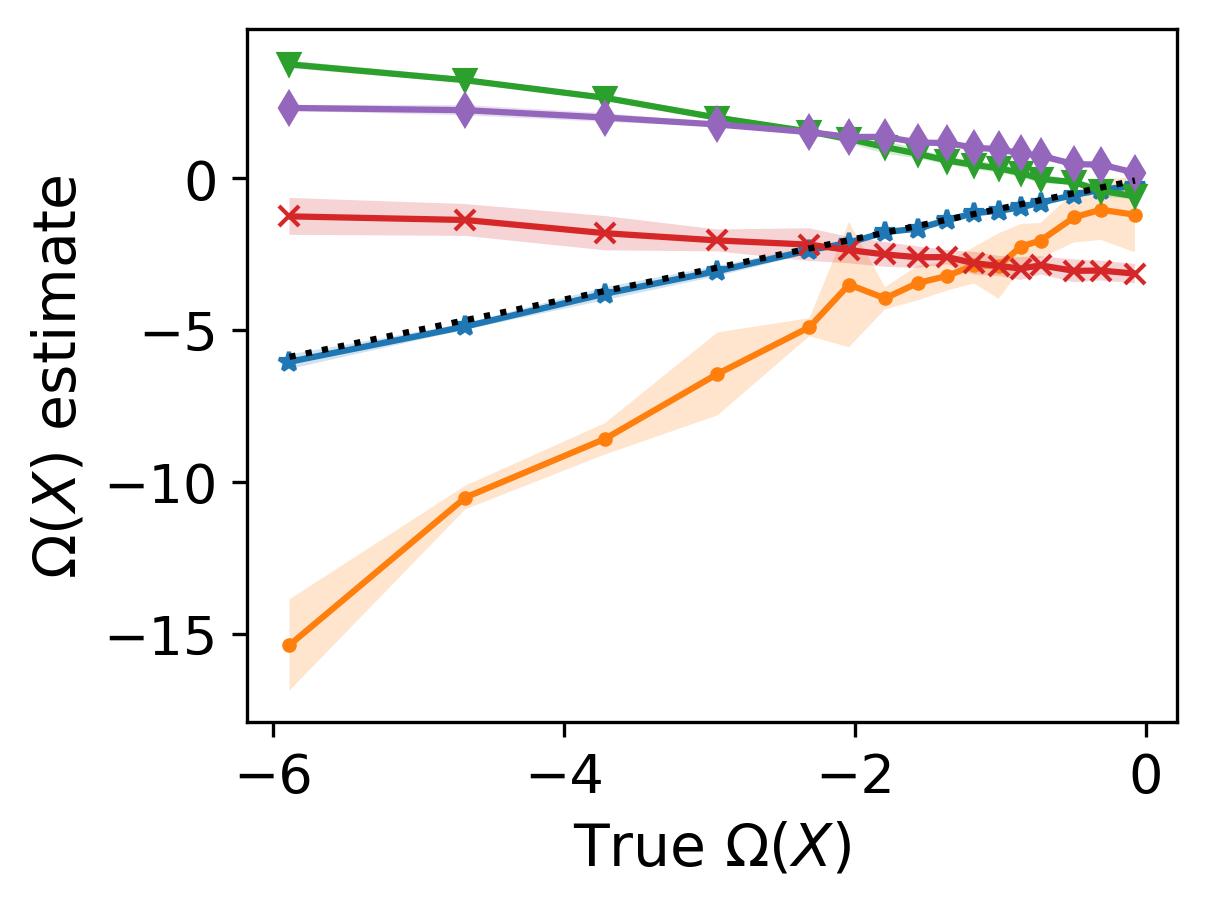}
         \caption{Dim=15}
     \end{subfigure}
         \begin{subfigure}{0.24\textwidth}
         \centering

         \includegraphics[page=1,width=\linewidth]{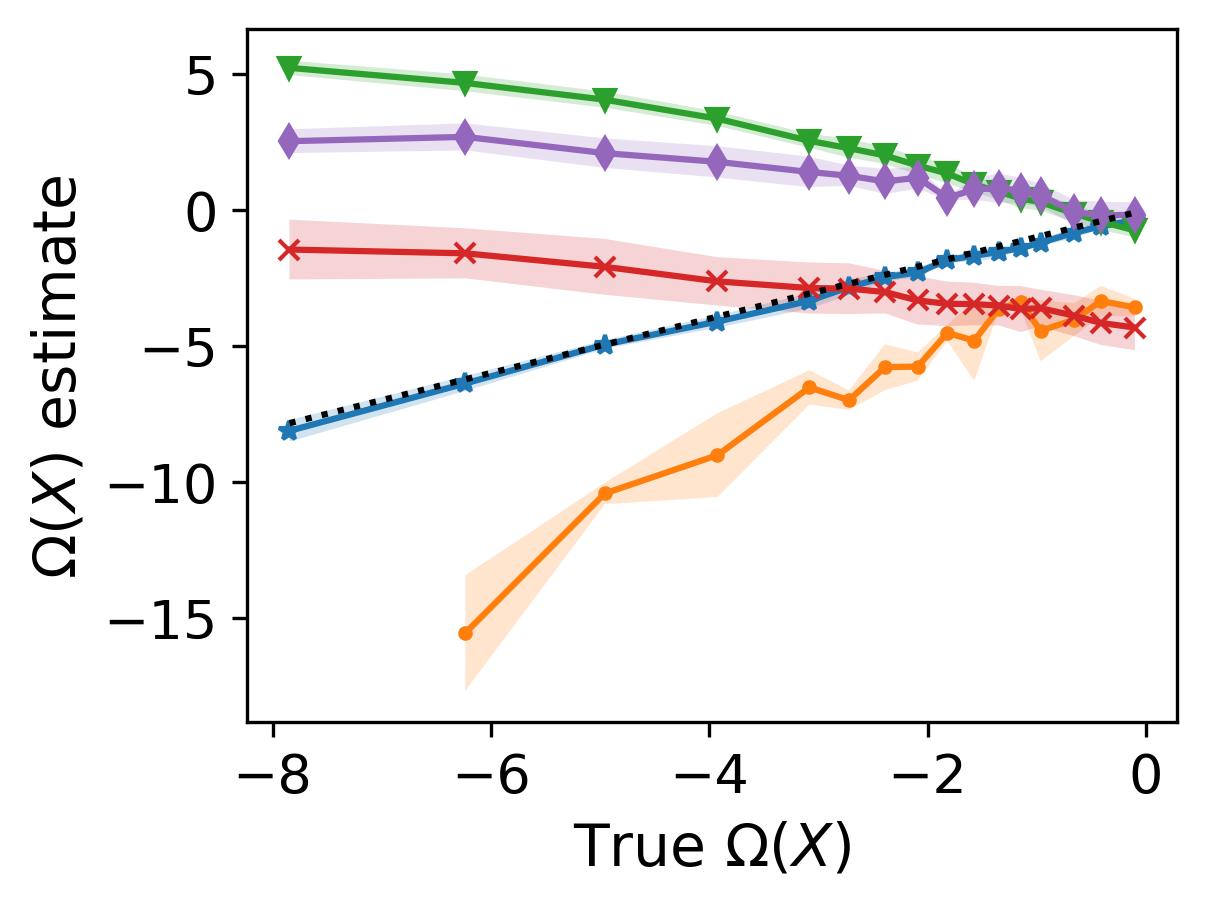}
         \caption{Dim=20}
     \end{subfigure}
      \caption{ Synergistic system with 6 variables, organized into subsets of sizes $\{3,3\}$ and increasing interaction strength.
      }

\end{figure}

\begin{figure} [h]

\centering
\begin{subfigure}{0.3\textwidth}
         \centering
    \includegraphics[page=1,width=\linewidth]{assets/figures/exp_red/legend.PNG}
    
     \end{subfigure}
     
     \begin{subfigure}{0.24\textwidth}
         \centering
  
         \includegraphics[page=1,width=\linewidth]{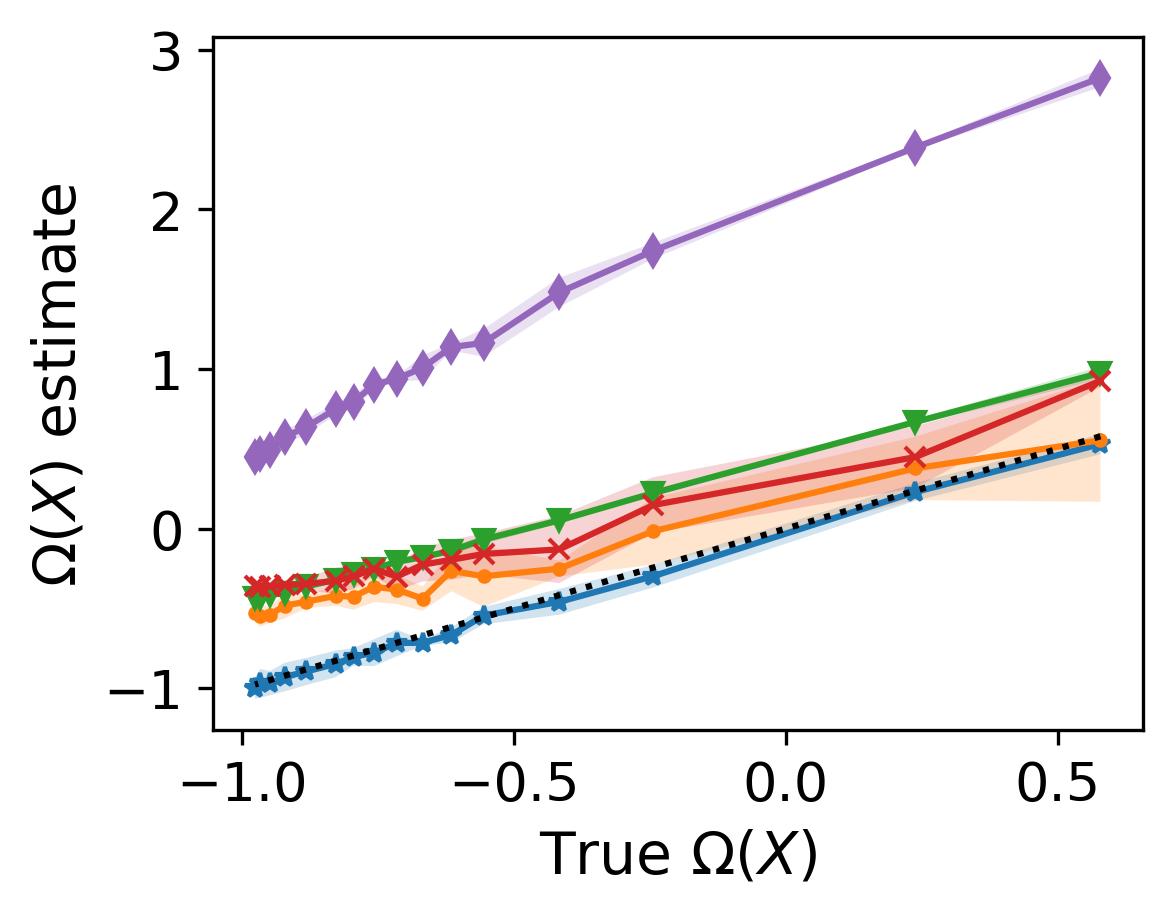}
         \caption{Dim=5}
     \end{subfigure}
      \begin{subfigure}{0.24\textwidth}
         \centering

         \includegraphics[page=1,width=\linewidth]{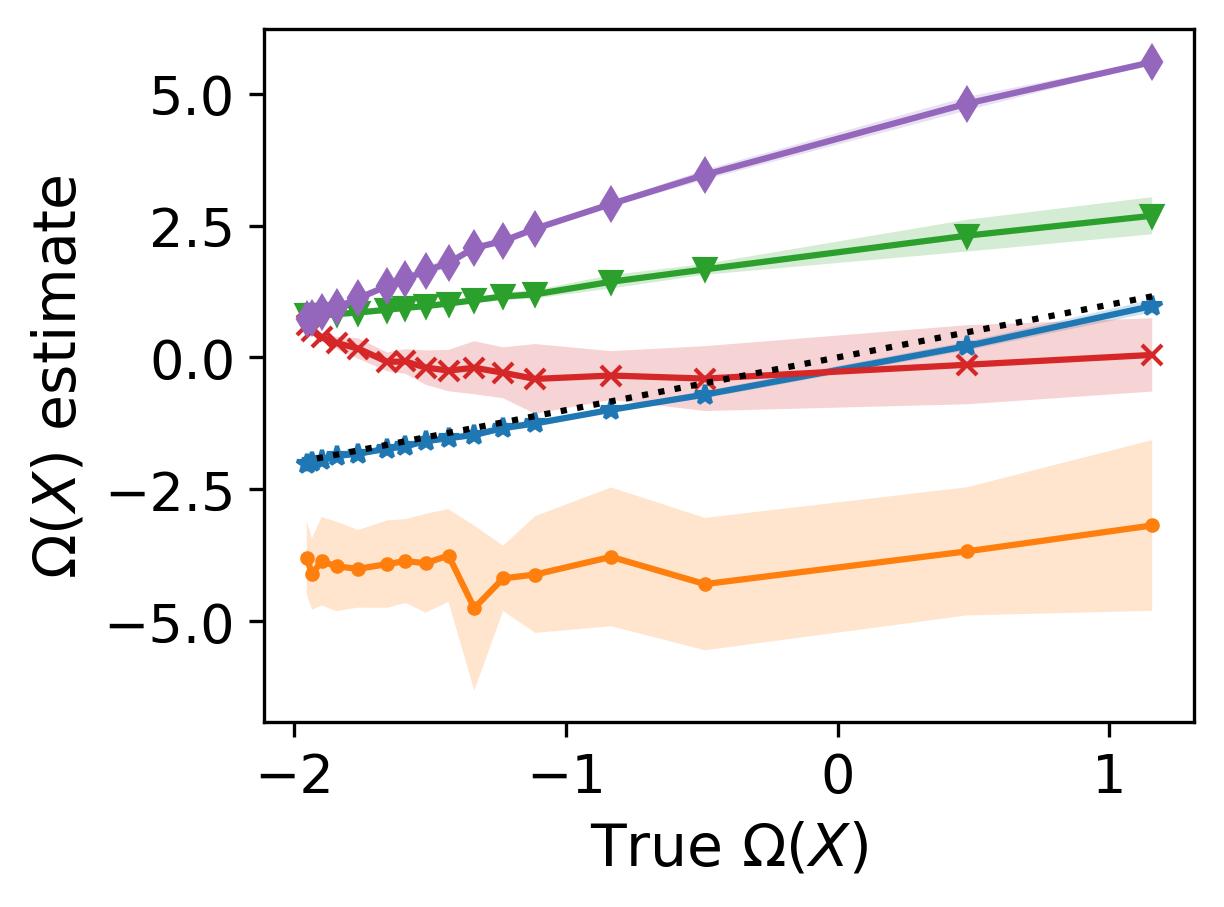}
         \caption{Dim=10}
     \end{subfigure}
     \begin{subfigure}{0.24\textwidth}
         \centering

         \includegraphics[page=1,width=\linewidth]{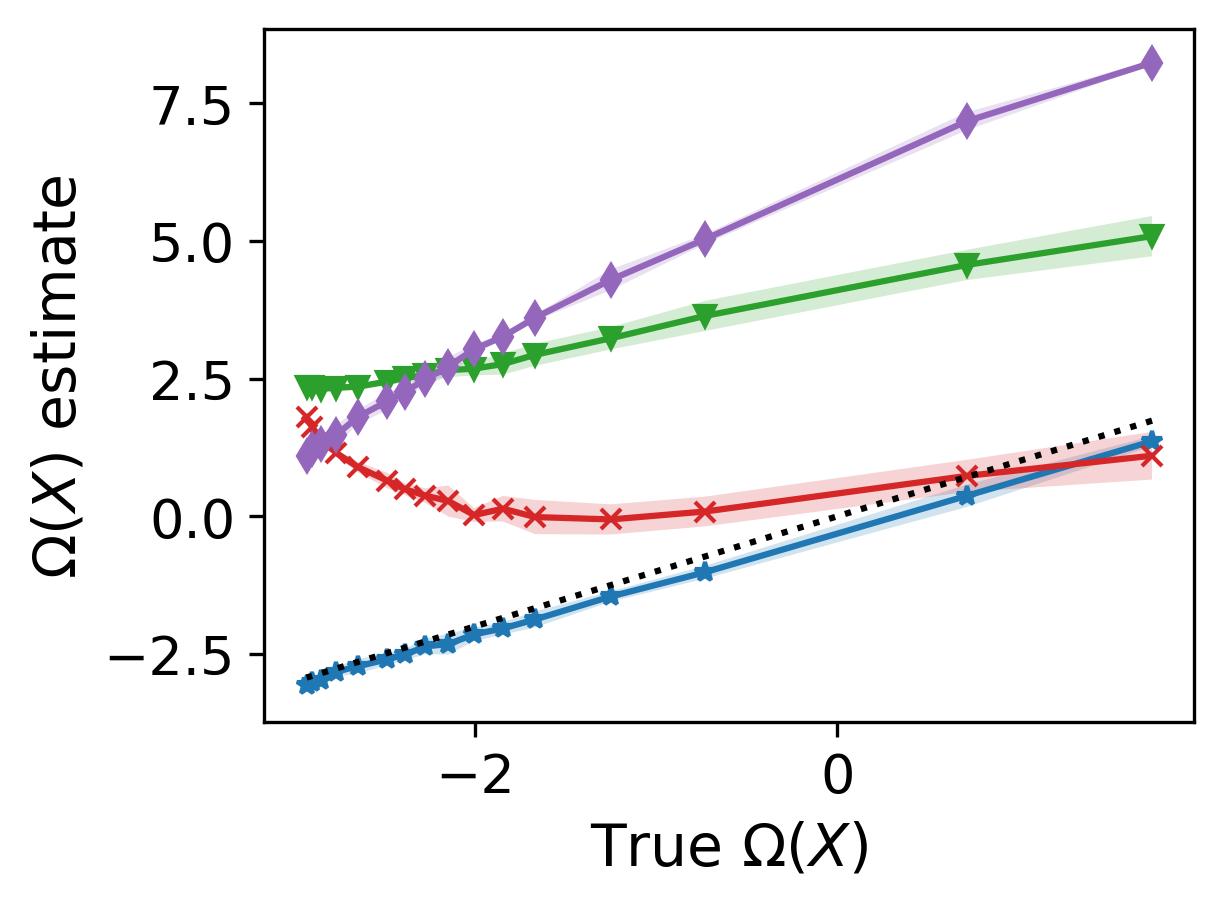}
         \caption{Dim=15}
     \end{subfigure}
         \begin{subfigure}{0.24\textwidth}
         \centering

         \includegraphics[page=1,width=\linewidth]{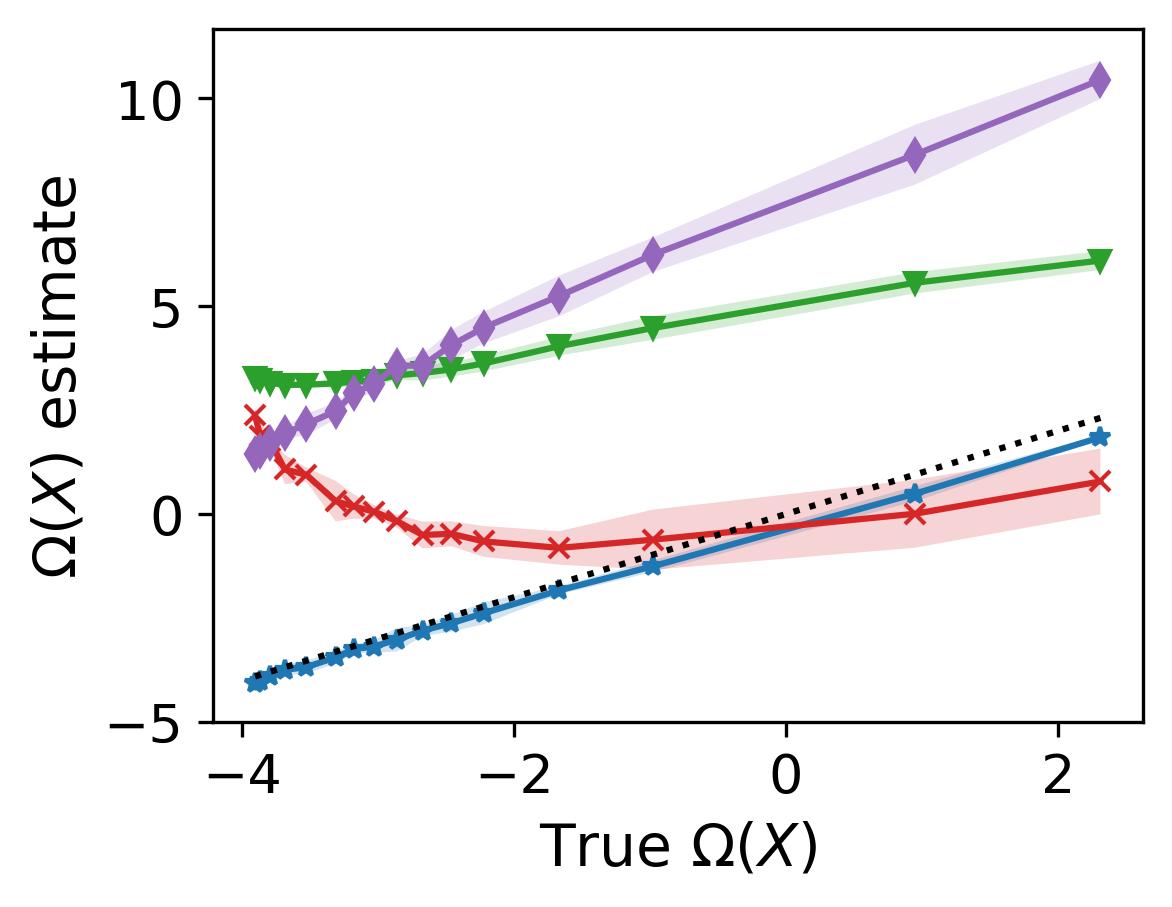}
         \caption{Dim=20}
     \end{subfigure}
      \caption{ Mixed-interaction system with 6 variables, organized into a redundancy-dominant subsets of size $3$ variables and one synergy-dominant subset with $3$ variables. \acrshort{O-information} is modulated by fixing the synergy inter-dependency and increasing the redundancy.
      }

\end{figure}

\subsection{The neural application additional experiments}

\end{document}